\PassOptionsToPackage{table,dvipsnames}{xcolor}
\PassOptionsToPackage{capitalize}{cleveref}
\documentclass{fairmeta}
\newcommand{\arxivversion}{}

\usepackage[T1]{fontenc}
\usepackage[utf8]{inputenc}

\usepackage{microtype}

\usepackage[table,dvipsnames]{xcolor}
\definecolor{darkgreen}{RGB}{80,200,80}

\usepackage{amsmath}
\usepackage{amssymb}
\usepackage{amsfonts}
\usepackage{amsthm}
\usepackage{mathtools}
\usepackage{bm}
\usepackage{bbm}
\usepackage{nicefrac}

\usepackage{graphicx}
\usepackage{caption}
\usepackage{subcaption}

\usepackage{booktabs}
\usepackage{array}
\usepackage{makecell}
\usepackage[inline]{enumitem}
\usepackage{url}

\usepackage{algorithm}
\usepackage{algpseudocode}

\usepackage[most]{tcolorbox}

\usepackage{natbib}

\usepackage{hyperref}
\usepackage[capitalize,noabbrev]{cleveref}

\definecolor{newblue}{HTML}{0064E0}
\definecolor{nblue}{HTML}{D7E5FF}
\definecolor{ngreen}{HTML}{E7FCF2}
\definecolor{nred}{HTML}{FFE8ED}

\hypersetup{
  colorlinks,
  linkcolor=newblue,
  citecolor=newblue,
  urlcolor=newblue
}

\crefname{section}{Sec.}{Secs.}
\crefname{subsection}{Subsec.}{Subsecs.}
\crefname{figure}{Fig.}{Figs.}
\crefname{table}{Tab.}{Tabs.}
\crefname{equation}{Eq.}{Eqs.}
\crefname{appendix}{App.}{Apps.}

\newtcolorbox[auto counter]{theorembox}[2][]{colframe=nblue, colback=nblue, breakable,
  before upper={{\textbf{Theorem \thetcbcounter.} #2}}, #1
}
\makeatletter
\def\tcb@cnt@theoremboxautorefname{Thm.}
\makeatother
\crefname{tcb@cnt@theorembox}{Thm.}{Thms.}

\newtcolorbox{qoutebox}[1][]{colframe=ngreen, colback=ngreen,
  #1
}

\newtcolorbox[auto counter]{definitionbox}[2][]{colframe=nred, colback=nred, breakable,
  before upper={{\textbf{Definition \thetcbcounter: } #2}}, #1
}
\makeatletter
\def\tcb@cnt@definitionboxautorefname{Def.}
\makeatother
\crefname{tcb@cnt@definitionbox}{Def.}{Defs.}

\newtcolorbox[auto counter]{algbox}[2][]{colframe=nred, colback=nred, left=1pt, right=1pt, top=0.5pt, breakable,
  before upper={{\textbf{Algorithm \thetcbcounter.} #2}}, #1
}
\makeatletter
\def\tcb@cnt@algboxautorefname{Alg.}
\makeatother
\crefname{tcb@cnt@algbox}{Alg.}{Algs.}

\newtcolorbox[auto counter]{mainbox}[2][]{colframe=ngreen,
  colback=ngreen,
  breakable,
  before upper={{\textbf{Main Contributions.} #2}},
  #1
}

\definecolor{takeawaybg}{RGB}{235,243,250}
\definecolor{takeawayborder}{RGB}{130,165,195}
\definecolor{takeawaytitle}{RGB}{190,215,235}

\newtcolorbox{takeawaybox}{
  enhanced,
  breakable,
  colback=takeawaybg,
  colframe=takeawayborder,
  boxrule=0.5pt,
  arc=2mm,
  left=3mm,
  right=3mm,
  top=2mm,
  bottom=2mm,
  before skip=4pt,
  after skip=4pt,
  title={Practical Takeaway},
  fonttitle=\small\bfseries,
  coltitle=black,
  attach boxed title to top left={
    xshift=4mm,
    yshift=-1.5mm
  },
  boxed title style={
    colback=takeawaytitle,
    colframe=takeawayborder,
    boxrule=0.5pt,
    arc=1.5mm,
    left=1.5mm,
    right=1.5mm,
    top=0.2mm,
    bottom=0.2mm
  }
}

\DeclareMathAlphabet{\mathsfit}{\encodingdefault}{\sfdefault}{m}{sl}
\SetMathAlphabet{\mathsfit}{bold}{\encodingdefault}{\sfdefault}{bx}{n}

\title{How to Loop Your Transformer: Loop Scaling, Specialization, and Parallel Inference}

\author[1,2,\dagger]{Chanakya Ekbote}
\author[1,*]{Artidoro Pagnoni}
\author[1]{Devendra Singh Sachan}
\author[1]{Shruti Bhosale}
\author[1]{Mike Lewis}
\author[1]{Maryam Fazel-Zarandi}
\author[1,*]{Shashank Rajput}
\author[1,*]{Sujan Kumar Gonugondla}
\affiliation[1]{FAIR at Meta}
\affiliation[2]{MIT}
\contribution[\dagger]{Work done at FAIR/Meta}
\contribution[*]{Equal contribution}

\abstract{Looping in transformer models reuses weights by repeatedly applying the same
transformer, or parts of it, to token representations, effectively making the
model deeper without increasing its parameter count. Whether looping is useful
depends on which constraint dominates: model size, training compute, or
autoregressive inference efficiency. We study these trade-offs jointly in dense
decoder-only transformers trained from scratch, varying which parts of the model
are looped, how many loops are applied, and how information is passed between
loops. Our experiments span models with up to \textbf{1B parameters}, trained on
up to \textbf{155B tokens} using up to $\mathbf{1.07 \times 10^{21}}$
\textbf{training FLOPs}, with up to $\mathbf{32}$ \textbf{loops}. We find that
looping the full transformer generally gives the strongest quality at a fixed
parameter budget, whereas looping a middle block of layers gives a better return
when training FLOPs are constrained. These trade-offs also depend on the number
of loops: adding loops is not uniformly beneficial, and gains eventually
saturate, reverse, or become unstable. This saturation motivates asking whether
forcing every loop to use exactly the same weights limits the benefit of deeper
looping. We address this by adding small loop-specific low-rank adapters that
improve quality, with larger gains as more loops are added. A separate practical
concern for looped transformers is the sequential dependency during
autoregressive generation, which causes a slowdown proportional to the number
of loops. Parallel Loop Transformers address this limitation by passing
information between loops across adjacent token positions via hidden states,
allowing loop stages to run concurrently during decoding. However, in our
experiments the original formulation does not consistently retain the quality
gains of ordinary looping. We find that preserving the current-token embedding
and separately normalizing it and the previous-token hidden state before
combining them recovers strong performance across model scales and training
budgets. Together, these results show that effective looping requires matching
the design to the resource constraint and inference setting.}

\date{October 2026}
\correspondence{Sujan Kumar Gonugondla at \email{sujan@meta.com}}

\begin{document}

\maketitle

\section{Introduction}
\label{sec:introduction}

\begin{figure*}[t]
    \centering
    \includegraphics[width=1\textwidth]{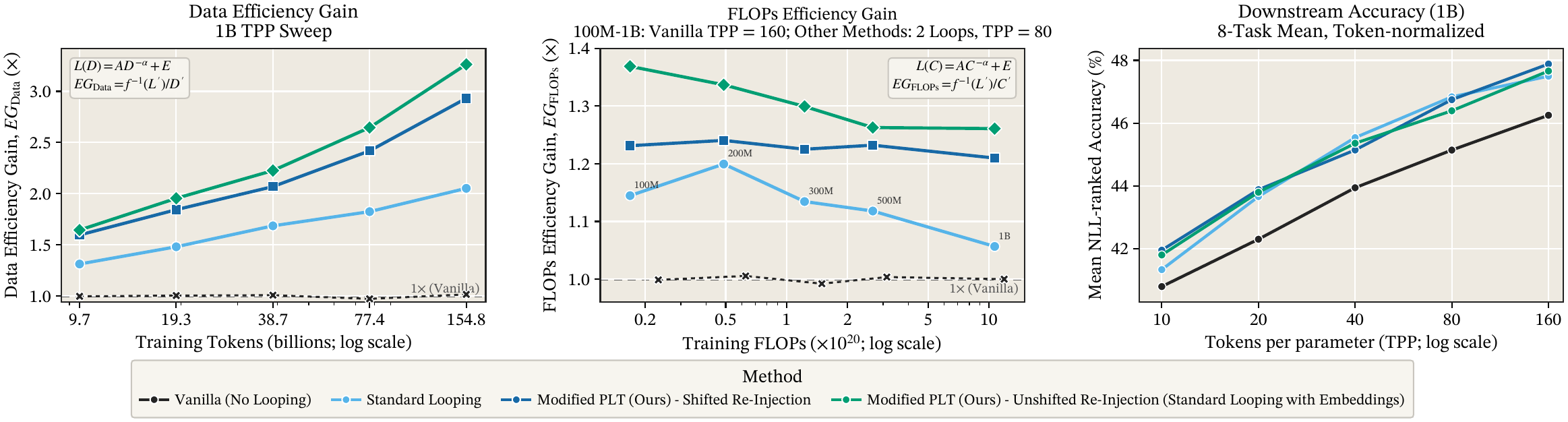
      }
\caption{\textbf{Looping can improve data and compute efficiency (in over-trained regimes) while retaining downstream quality.}
We compare Vanilla (no looping), Standard Looping (repeatedly applying the same Transformer sequentially), and our modified versions Parallel Loop Transformer (PLT) \citep{wu2025parallel} and Looped Transformers.
\emph{Left:} for 1B models trained across TPP 10--160, all looping methods achieve a given DCLM \citep{li2024datacomp} bits-per-byte (BPB) value with fewer training tokens than the fitted Vanilla baseline, with the modified PLT variants providing the largest relative data-efficiency gains.
\emph{Center:} across model sizes from 100M to 1B, we compare Vanilla models trained at TPP $=160$ with two-loop models trained at TPP $=80$. All looping methods achieve a given DCLM BPB with fewer training FLOPs than the fitted Vanilla baseline, and the modified PLT and Looped variants are more efficient than Standard Looping.
\emph{Right:} for 1B models across TPP 10--160, looping retains or improves mean NLL-ranked accuracy across eight reasoning and question-answering tasks; the modified PLT variants match or exceed Standard Looping.
See \autoref{app:headline-details} for the efficiency calculation and task-level results. }
    \label{fig:headline-looping-efficiency}
    \vspace{-1.5em}
\end{figure*}

Looping in transformer models reuses weights by repeatedly applying the same transformer, or selected parts of it, to token representations, allowing the model to behave like a deeper network without storing a new set of parameters at every `depth'  \citep{dehghani2018universal,saunshi2025reasoning,geiping2025scaling}. This makes looping attractive when model size is constrained, but the additional depth is not free: every extra loop must still be executed. As a result, a design that uses parameters efficiently may not use training compute efficiently.
Prior work reflects this trade-off, showing that repeated shared computation can recover some of the benefits of additional depth, while adding independently parameterized layers can be preferable when training compute is constrained \citep{schwethelm2026recurrenceworthisodepthscaling,prairie2026parcae}. The same weight reuse, however, raises another question as more loops are added. In ordinary looping, every loop applies the same weights, even though later loops operate on representations that have already been transformed several times. This suggests that exact weight sharing may 
give diminishing returns
with deeper looping. Prior work has relaxed this constraint by giving each loop iteration a different low-rank adapter while keeping the main transformer weights shared \citep{bae2025relaxed}.

Looped transformers also suffer from a separate issue during autoregressive generation: each loop for a token still depends on the output of the previous loop, so the model must execute them one after another, increasing the sequential computation required to generate each token. Parallel Loop Transformers (PLTs) reorganize this dependency by passing hidden states from one loop to the next across adjacent token positions, allowing multiple loop stages to be evaluated in parallel during decoding \citep{wu2025parallel}. This change alters the information available to each loop, and recent work reports that gains can decrease as more parallel loops are added \citep{yang2026loopcoder}. This leaves open what information the shifted transition must preserve for parallel looping to remain effective at greater loop depth. Together, these studies show that looping is not a single design choice but a set of coupled trade-offs. Repeating more of the model can make better use of a fixed parameter budget but requires more training compute; reusing the same weights across more loops can make exact sharing increasingly restrictive; and changing the dependency between loops can parallelize sequential looping while also changing the information available to each loop. These trade-offs may also change with model scale and training budget. Despite prior work studying these choices in different settings, it remains unclear which conclusions generalize and how the choices interact. We therefore study them within a common training from-scratch setting across model scales and training budgets, and ask:

\begin{qoutebox}{}
\centering
\textbf{How should transformer layers be looped under constraints on model size, training compute, and autoregressive inference?}
\end{qoutebox}

We address this question through a controlled study of dense decoder-only transformers trained from scratch. Our experiments span models with up to \textbf{1B parameters}, up to \textbf{155B training tokens}, up to $\mathbf{1.07 \times 10^{21}}$ \textbf{training FLOPs}, and up to $\mathbf{32}$ \textbf{loops}. We organize the study around three linked aspects of looping:

\begin{mainbox}{}
\begin{enumerate}[leftmargin=*]

\item \textbf{We characterize where and how much to loop.}
We compare different looping designs and vary the number of loops across each design. We find that looping the full transformer gives the strongest quality at a fixed parameter budget, whereas looping a contiguous block of transformers layers in the middle of the network, gives a better return when training FLOPs are constrained. We also find that adding more loops is not uniformly beneficial: gains eventually saturate, reverse, or become unstable (\autoref{sec:which-loops-work}).

\item \textbf{We show when loop-specific specialization becomes useful.}
We compare exact weight sharing across loops with small loop-specific low-rank adapters that allow each loop to learn a slightly different transformation. These adapters provide only modest gains when few loops are used, but become consistently more useful as the same weights are reused across more loops (\autoref{sec:loop-lora}).

\item \textbf{We present small modifications that improve stacked looping and PLTs}
We study Parallel Loop Transformers (PLTs) \citep{wu2025parallel}, which change the dependency between loops so that their computations can be parallelized during autoregressive decoding. Naively applying PLT in our setting does not preserve the gains from looping: quality can degrade as more loops are added. We find that preserving the current-token representation, separately normalizing it and the recurrent state, and controlling how they are combined substantially improves PLT. With these changes, parallel looping can outperform looping the full transformer while retaining a parallelizable decoding dependency. Applying similar changes to standard looping improves full-transformer looping further, showing that the modification is useful beyond the parallel setting (\autoref{sec:plt}).

\end{enumerate}
\end{mainbox}

\section{Background and Methods}
\label{sec:background}

\subsection{Different Ways to Loop a Transformer}
\label{sec:looping-designs}

We consider several ways to loop a transformer by varying which parts of the model are looped and the order in which the loops are applied. In \emph{Full-Stack Looping}, the output of the final transformer block is fed back to the first block, so the entire stack of transformer blocks is applied again at each loop. In \emph{Per-Layer Looping}, the output of each transformer block is fed back to the same block for all loops before moving to the next transformer block. We also consider looping only a contiguous set of transformer blocks in the middle of the model. In \emph{Middle-Stack Looping}, the output of the last block in this region is fed back to the first, so the middle blocks are looped as a group. In \emph{Middle-Layer Looping}, each transformer block in the middle region is looped before moving to the next transformer block. The transformer blocks outside this region are applied once. Finally, we vary which part of a transformer block is looped: the full block, attention only, the MLP only, or attention and the MLP independently. See \autoref{fig:looping-architectures} and \autoref{sec:background_looping} for details.

\subsection{Loop-Specific Specialization}
\label{sec:loop_specific_specialization}

In standard transformer looping, every loop uses the same transformer weights. This keeps parameter storage low, but may limit what later loops can learn, since they operate on representations already transformed by earlier loops. Giving each loop a separate set of weights would remove this constraint, but would also give up much of the parameter savings from looping. We instead use Low-Rank Adaptation (LoRA) \citep{hu2021lora} to give each loop a small loop-specific adapter while keeping the main transformer weights shared, as defined in \autoref{def:loop_specific_lora}. We refer to this as \emph{loop-specific specialization}.

\begin{definitionbox}[label=def:loop_specific_lora]
{\textbf{Loop-Specific Low-Rank Adapter. }}
For a weight matrix
\(\bm W\in\mathbb{R}^{d_{\mathrm{out}}\times d_{\mathrm{in}}}\)
shared across loops, loop \(\ell\) uses
\(\bm W^{(\ell)}
= \bm W+\frac{\alpha}{r}\bm B^{(\ell)}\bm A^{(\ell)}\),
where
\(\bm A^{(\ell)}\in\mathbb{R}^{r\times d_{\mathrm{in}}}\)
and
\(\bm B^{(\ell)}\in\mathbb{R}^{d_{\mathrm{out}}\times r}\)
are learned specifically for loop \(\ell\).
\end{definitionbox}

We also evaluate alternative ways of parameterizing and combining these loop-specific low-rank adapters; see \autoref{fig:loop-lora-architecture} and \autoref{sec:background_lora} for details.

\subsection{Parallelizing Loops During Inference}
\label{sec:parallel_looping}

In ordinary looping, each token is processed through each loop one after another: loop \(\ell\) receives the representation of that same token produced by loop \(\ell-1\). As a result, loop \(\ell\) cannot begin until loop \(\ell-1\) has finished. Parallel Loop Transformers (PLTs) \citep{wu2025parallel} change this dependency through \emph{Cross-Loop Parallelism} (CLP). Instead of passing the hidden state between loops at the same token position, CLP passes the state from the previous loop at the preceding token position. As these previous-token states are already available during autoregressive decoding, all the loop computations for the current token can be evaluated simultaneously. To make this 
dependency (CLP) precise, let \(\bm e_n\) denote the embedding of token \(n\), let \(\bm b_n^{(\ell)}\) and \(\bm h_n^{(\ell)}\) denote the input and output of loop \(\ell\), and let \(\mathcal{F}_{\theta}\) denote the transformer whose weights are shared across loops.

\begin{definitionbox}[label=def:plt_transition]
{\textbf{Cross-Loop Parallelism (CLP). }}
The first loop receives the current-token embedding,
\(\bm b_n^{(1)}=\bm e_n\). Each later loop receives the current-token embedding
together with the state produced by the previous loop for the preceding token:
\[
\bm b_n^{(\ell)}
=
\bm e_n+\bm h_{n-1}^{(\ell-1)},
\qquad
\bm h_n^{(\ell)}
=
\mathcal{F}_{\theta}\!\left(\bm b_n^{(\ell)}\right).
\]
\end{definitionbox}

The original PLT combines CLP with two mechanisms that reduce KV-cache memory: \emph{KV sharing} and \emph{(G-SWA)}.
\emph{KV sharing} lets later loops reuse the full KV cache produced by the first
loop instead of storing a separate full cache for every loop. Each later loop
also keeps a small local sliding-window cache, and \emph{gated sliding-window
attention} (G-SWA) combines the attention outputs from the shared and local
caches. In our experiments, we separate CLP, KV sharing, and G-SWA to study
their effects independently. See \autoref{fig:plt-diagonal} and
\autoref{sec:background_plt} for details.

\section{Experiments}
\label{sec:experiments}

\subsection{Experimental Setup}
\label{sec:loop-factor-setup}

We train models with 70M, 100M, 200M, 300M, 500M, and 1B parameters,
containing 4, 6, 8, 10, 12, and 16 stored transformer layers, respectively.
Training budgets range from 10 to 160 tokens per parameter (TPP). All models
use the same pretraining mixture of natural language, code, mathematical, and
scientific content, with the data sources, sampling distribution, preprocessing,
tokenizer, and training setup held fixed across architectures. We use a fixed subset of the DCLM Baseline 1.0 validation set
\citep{li2024datacomp}, containing approximately 41 million tokens, as our
primary measure of language-modeling quality and report bits per byte (BPB;
lower is better). To test whether the trends observed on DCLM hold across
datasets, we additionally evaluate on other language-modeling corpora and a
broader set of mathematical, code, reasoning, and question-answering
benchmarks. These results, together with the full model configurations,
training details, evaluation protocols, and task-level results, are provided
in the appendix (see \autoref{app:model-configurations-and-budgets}).

\subsection{Which Forms of Looping Work, and When?}
\label{sec:which-loops-work}

\begin{figure}[ht]
  \centering
  \includegraphics[width=\textwidth]{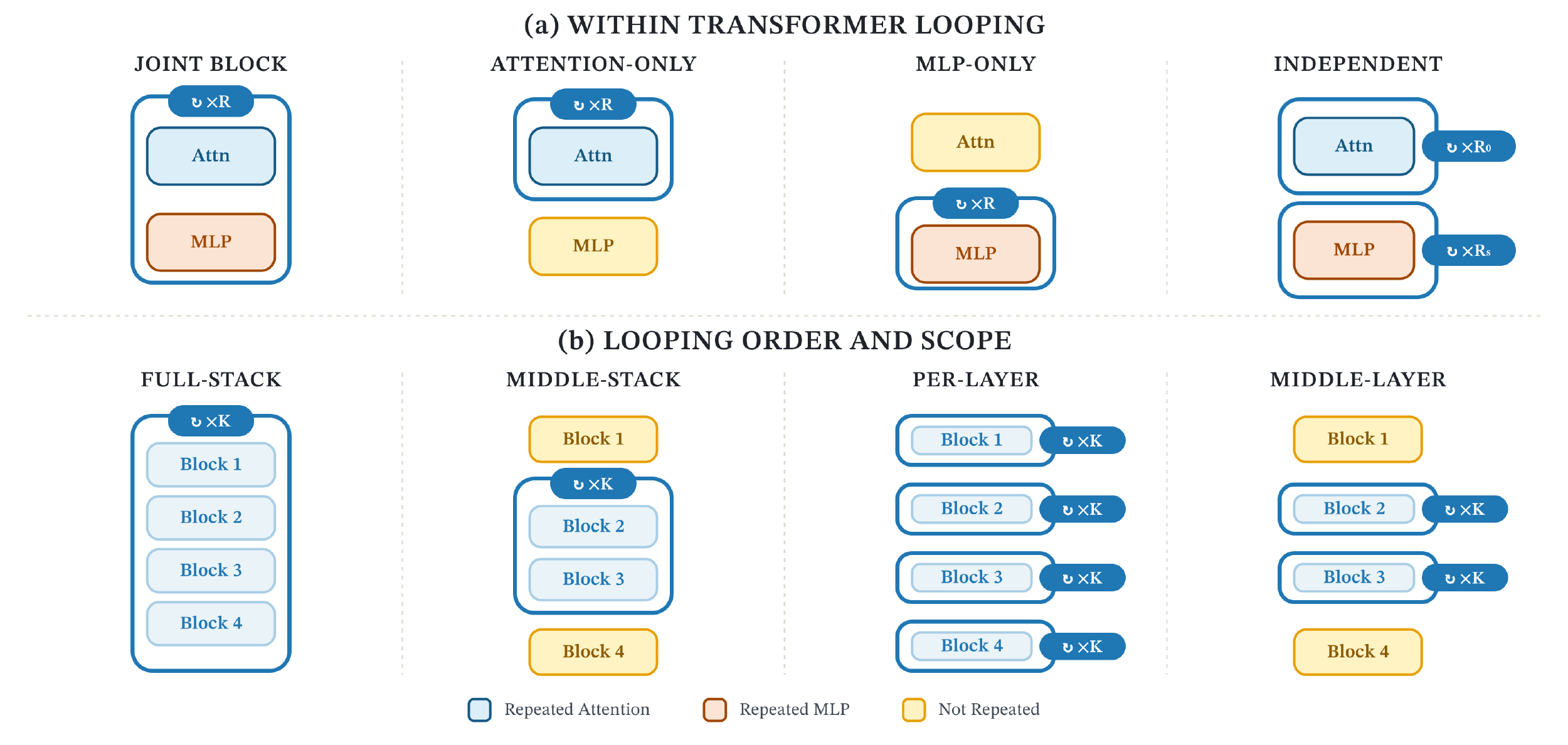}
\caption{\textbf{Looping designs: what is repeated, where, and in what order.}
Top: the looped component can be the complete transformer block,
attention alone, the MLP alone, or attention and the MLP separately.
Bottom: full-stack looping traverses the complete transformer for \(L\) loops,
whereas per-layer looping completes \(L\) loops of each block before
advancing to the next. Middle-stack and middle-layer looping restrict the
corresponding schedule to the central blocks. For a model containing
\(B\) blocks, the first and last \(\lfloor B/3\rfloor\) blocks are run once.
}
  \label{fig:looping-architectures}
  \vspace{-1.5em}
\end{figure}

\begin{figure}[t]
  \centering
  \includegraphics[width=\textwidth]{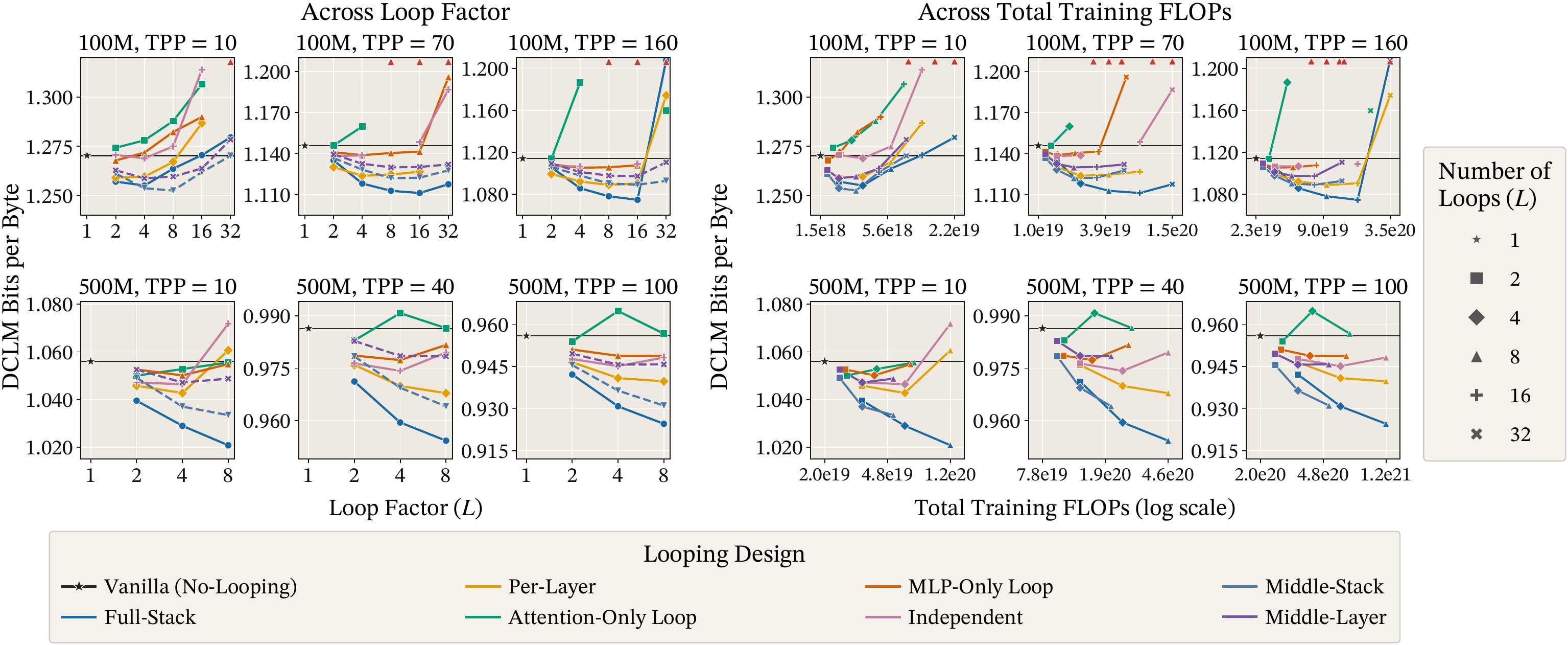}
\caption{\textbf{Full-Stack Looping is favored under fixed
parameter-count and training-data budgets, whereas Middle-Stack Looping is favored under FLOP constraints.}
DCLM validation BPB versus the number of loops or total training
  FLOPs for 100M and 500M models. Each panel holds the model size
  and training-data budget constant, so the differences reflect how
  each strategy uses the additional computation introduced by looping. The advantage of full-stack looping becomes clearer at larger model
  scales and training budgets. \emph{Middle-Stack Looping} with fewer loops
  achieves lower BPB when compute is limited, whereas \emph{Full-Stack Looping} can
  attain lower BPB when more compute is available. Increasing the number of
  loops does not improve quality indefinitely: the 100M results exhibit
  saturation, regression, and instability when more loops are used. Lower BPB
  is better; red triangles indicate unstable or clipped runs.
}
  \label{fig:loopfactor-dclm-main}
\end{figure}

  While keeping the model size fixed, there is a large design space for
  deciding which components of the transformer to loop, in what order, and
  how many times to loop them. We investigate which of these choices improve
  model quality and whether their effects depend on model scale and training
  budget. We compare seven designs (\autoref{fig:looping-architectures})
  across five dense model scales ranging from 70M to 500M parameters,
  token-to-parameter ratios (TPP) of up to 160, and loop factors
  \(L \in \{2,4,8,16,32\}\), where available. At each model size, we consider
  experiments that constrain either (a) the total amount of training data
  (i.e., fixed TPP) or (b) the total training compute (i.e., a fixed FLOP
  budget). Our primary metric is DCLM validation bits per byte (BPB; lower is
  better). \autoref{fig:loopfactor-dclm-main} shows the results at the 100M and 500M
  scales. Complete sweeps and targeted 1B comparisons appear in
  \autoref{sec:looping-additional-results}.

\paragraph{Benefits and limits of looping.}
  Using a moderate loop factor (\(L\leq 16\)), looping improves quality across the
  evaluated designs, model scales, and TPP settings: relative to the
  corresponding no-looping model, increasing the loop factor \(L\)
  reduces DCLM BPB (\autoref{fig:loopfactor-dclm-main}). However, the marginal
  gains diminish as \(L\) increases, and performance eventually degrades at very large \(L\). This effect is clearest in the 70M and 100M sweeps, which
  extend to \(L=32\). The point at which these gains saturate or reverse depends on both the looping
  strategy and TPP. At higher TPP, increasing \(L\) produces more
  consistent gains, whereas lower-TPP runs tend to saturate or regress sooner.
  The strategy that loops attention and the MLP independently degrades sharply as \(L\) increases,
  whereas \emph{Full-Stack Looping} and \emph{Middle-Stack Looping} avoid regression over
  a wider range of \(L\) (\autoref{fig:loopfactor-dclm-small}). Moreover, the relative
  performance of the looping strategies changes with model size: the benefits of
  looping persist as model size increases, while the strongest strategies become
  more consistent. At 100M and 200M, \emph{Per-Layer Looping} and \emph{Middle-Stack Looping}
   can be competitive with \emph{Full-Stack Looping} when a smaller
  \(L\leq 2\) is used, depending on TPP. At 300M and especially 500M,
  \emph{Full-Stack Looping} and \emph{Middle-Stack Looping} perform well across TPP
  settings and loop factors, with \emph{Full-Stack Looping} achieving the lowest BPB as TPP increases
  (\autoref{fig:loopfactor-dclm-large}). The
  200M--500M sweeps extend only to \(L=8\) and 1B sweeps up to \(L=4\), because the computational cost of
  looping grows linearly with \(L\). Within these ranges, we still see diminishing relative improvements with \(L\) (see \autoref{fig:loop-public-1b-dclm-factor} and
  \autoref{fig:loop-public-1b-dclm-flops} for 1B ablations). Comparing performance
  as a function of total training FLOPs shows that some looping strategies use compute more effectively than others.
  At lower training FLOP budgets, \emph{Middle-Stack Looping}
  provides the best return on additional compute, but when compute is not a constraint, \emph{Full-Stack Looping} can achieve lower absolute BPB, as shown in \autoref{fig:loopfactor-dclm-main} and
  \autoref{fig:loop-public-1b-dclm-flops}.

  Note that similar trends hold across additional model
  scales and evaluation datasets
  (\autoref{sec:looping-additional-results}). The diminishing returns from
  adding more loops raise a natural question: does sharing exactly the same
  weights across loops limit further improvements? We investigate this question
  in \autoref{sec:loop-lora}.

\subsection{When Should Loops Specialize?}
\label{sec:loop-lora}

\begin{figure}[t]
\centering
\includegraphics[width=\textwidth]{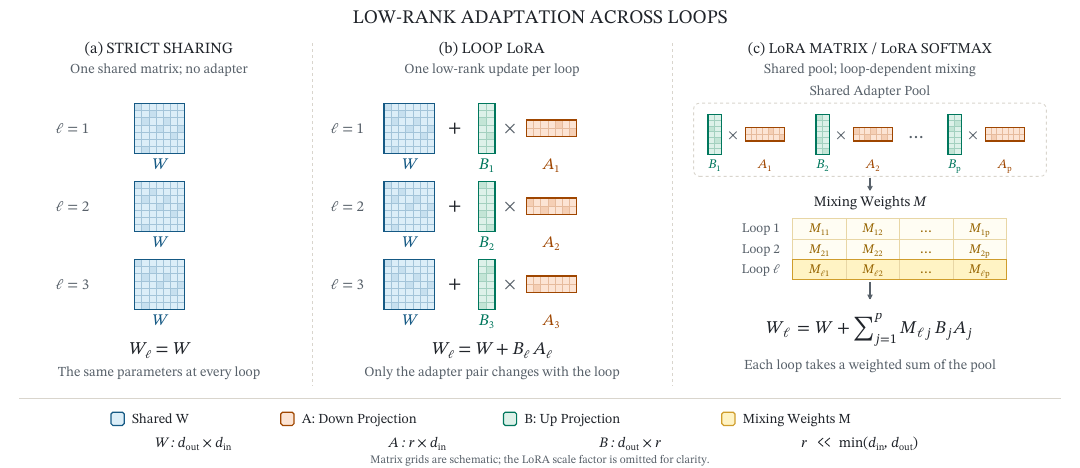}
\caption{\textbf{Three ways to vary weights across loops.} All designs retain the same base weights. Loop LoRA gives each loop a separate low-rank adapter. Loop Matrix lets each loop learn a combination of these adapters, while Loop Softmax restricts this combination to be nonnegative and sum to one.
}
\label{fig:loop-lora-architecture}
\vspace{-1.5em}
\end{figure}

\begin{figure}[ht]
\centering
\includegraphics[width=\textwidth]{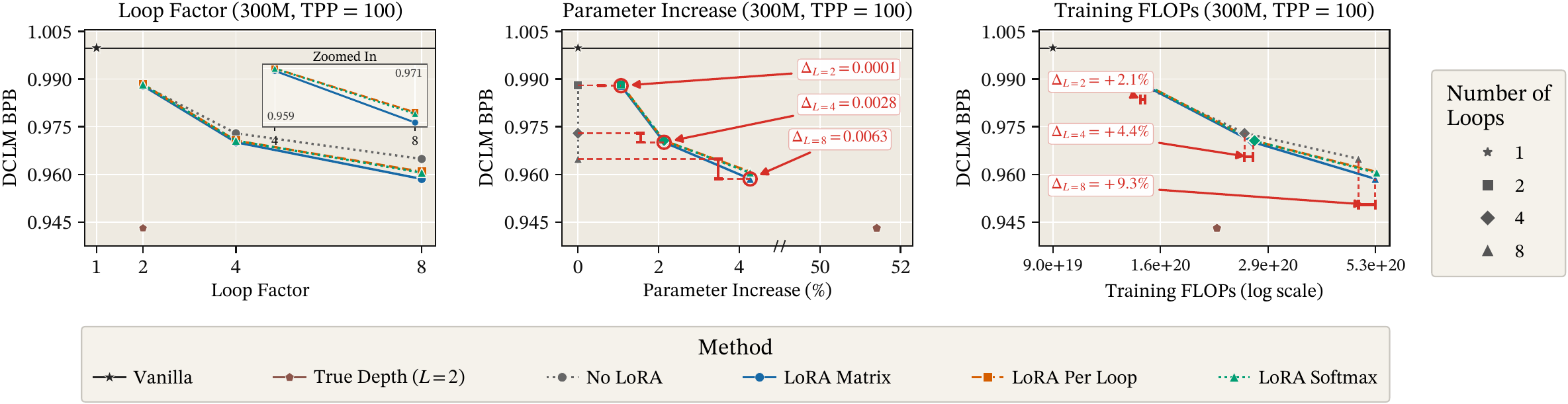}
\caption{\textbf{Loop-specific corrections become more useful with deeper looping while adding few parameters.} DCLM BPB for 300M models with \emph{Full-Stack Looping} at TPP 100. The left, middle, and right panels plot BPB against loop count, non-embedding parameter count, and total training FLOPs, respectively.
The break in the logarithmic parameter axis shows both parameter ranges. FLOPs include the corrections and the training tokens allocated at TPP 100. Lower BPB is better.}
 \label{fig:loop-lora-300m}
\end{figure}

  We now
  investigate whether requiring every loop to reuse the same weights limits
  model quality. To do so, we explore a middle ground between complete weight
  sharing in a standard looped model and using unique weights at every depth in
  a deeper model. For our primary experiments, we use \emph{Full-Stack Looping} because it generally
  provides the strongest absolute improvements. However, the proposed approach
  applies to any looping design, and we evaluate it with other designs in
  \autoref{app:loop-lora}. \autoref{fig:loop-lora-architecture} illustrates
  three ways to introduce loop-specific adapters. Given a weight matrix
  \(\bm W\) shared across \(L\) loops, loop
  \(\ell\in\{1,\ldots,L\}\) uses a loop-specific weight defined as
  \(\bm W^{(\ell)}=\bm W+\Delta\bm W^{(\ell)}\). We use different designs for \(\Delta\bm W^{(\ell)}\): \emph{Loop LoRA} assigns each loop its own low-rank adapter,
  \(\Delta\bm W^{(\ell)}=\bm B_\ell\bm A_\ell\).
  \emph{Loop Matrix} learns an unconstrained mixing matrix
  \(\bm M\in\mathbb{R}^{L\times L}\), where \(M_{\ell j}\) controls the
  contribution of low-rank component \(j\) to loop \(\ell\):
  \(\Delta\bm W^{(\ell)}=\sum_{j=1}^{L}M_{\ell j}\bm B_j\bm A_j\).
  \emph{Loop Softmax} parameterizes these coefficients using a learned score
  matrix \(\bm S\in\mathbb{R}^{L\times L}\), with
  \(M_{\ell j}=\operatorname{softmax}(\bm S_{\ell,:})_j\). This makes the
  coefficients nonnegative and normalizes them to sum to one. Loop Matrix and
  Loop Softmax can combine several low-rank components, increasing the
  effective rank and expressivity of each loop-specific adapter; see
  \autoref{app:loop-lora}. For these experiments, each low-rank component has rank at most 8. We use unit scaling and jointly train
  the shared weights and loop-specific adapters from initialization, and apply
  the updates to the attention and MLP weights in every transformer block. As
  in the preceding section, the main-text analysis focuses on DCLM.
  \autoref{fig:loop-lora-300m} shows the results for 300M models with
  \(L\in\{2,4,8\}\) at TPP 40 and 100.

\paragraph{Benefits and costs of loop-specific specialization.}
  The gains from loop-specific specialization are marginal at \(L=2\) but
  become clear at \(L\in\{4,8\}\)
  (\autoref{fig:loop-lora-300m}). Among the three specialization options, Loop Matrix provides the largest improvement, though the relative differences are smaller than the gain over the unspecialized looping baseline. We therefore use Loop Matrix by
  default in subsequent experiments involving loop-specific specialization,
  including those in \autoref{sec:plt}. At 300M, these specialization methods increase the number of non-embedding
  parameters by only 1.1--4.3\%, compared with 51.4\% for doubling the model
  depth. At comparable training FLOPs, the doubled-depth reference
  (the brown pentagon) achieves lower BPB than loop-specific models. Loop-specific specialization therefore offers a
  parameter-efficient, rather than compute-efficient, way to improve quality. We observe the same overall
  trends for 70M--200M models across TPP 10, 40, and 100, in targeted 500M and
  1B experiments, and on the downstream datasets introduced in
  \autoref{sec:loop-factor-setup}. Full results are provided in
  \autoref{app:loop-lora}; a more exhaustive study of specialization methods is
  left for future work. Loop-specific specialization improves quality but does not change the fact
  that the loops must be evaluated one after another for each generated token.

\subsection{Stabilizing PLT and Looping}
\label{sec:plt}

\ifdefined\arxivversion
\begin{figure}[t]
\else
\begin{wrapfigure}[22]{r}{0.7\textwidth}
\fi
    \centering
    \includegraphics[trim=0pt 5pt 25pt 5pt, clip, width=0.7\textwidth]{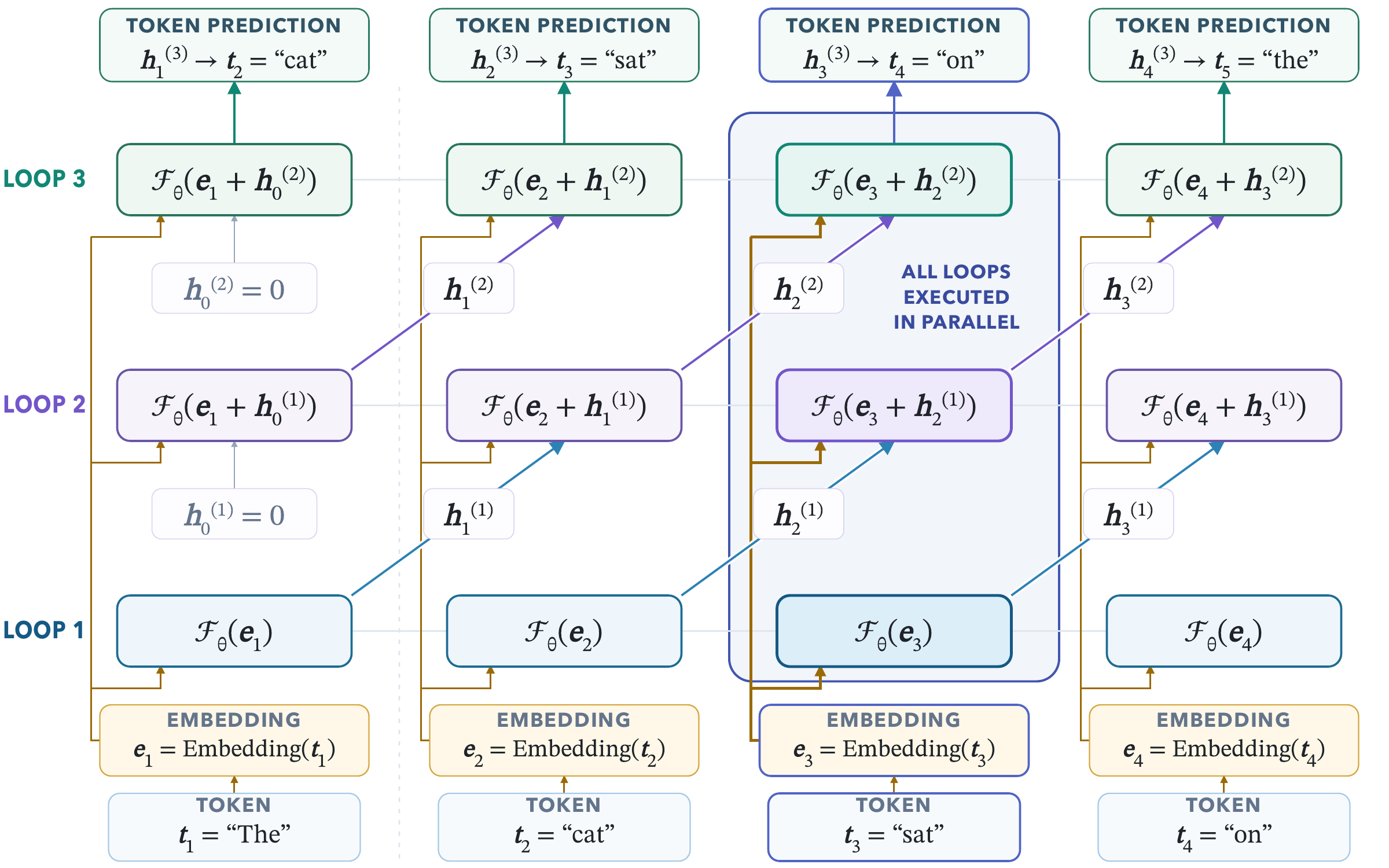}
  \caption{\textbf{PLT shifts the connection between loops across tokens.}
  Loop $\ell$ at token $n$ receives the current-token embedding and the state
  from loop $\ell-1$ at token $n-1$. KV sharing and G-SWA are
  omitted for clarity.}
  \label{fig:plt-diagonal}
\ifdefined\arxivversion
\end{figure}
\else
\end{wrapfigure}
\fi

\begin{figure*}[t]
  \centering
  \includegraphics[width=\textwidth]{
    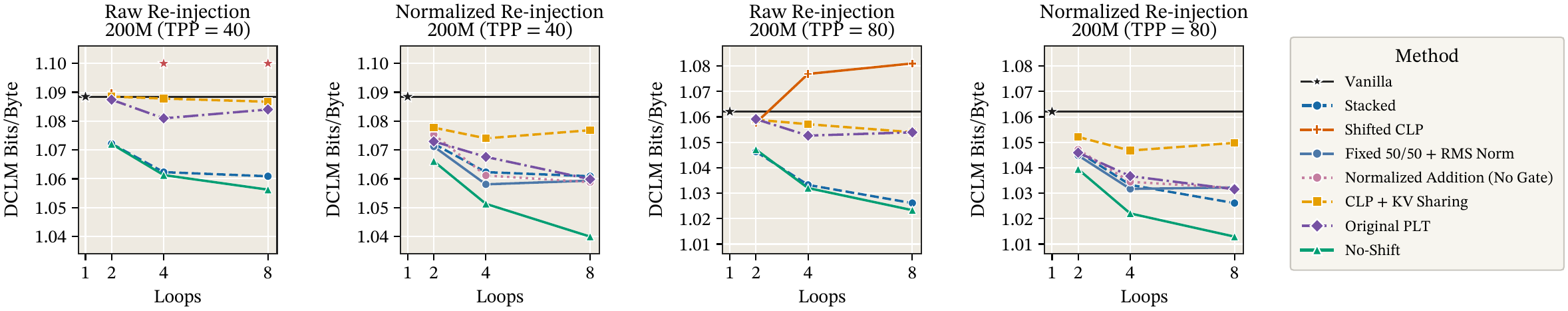
  }
  \caption{\textbf{Separate normalization recovers much of the quality lost
  by the shifted transition.} DCLM BPB for 200M models at TPP 40 and 80. Each
  raw panel is followed by its normalized counterpart. Except for Normalized
  Addition, normalized variants use separate RMS normalization and fixed
  50/50 mixing. Red stars mark unstable runs clipped for visualization. Lower
  BPB is better.}
  \label{fig:plt-design}
\end{figure*}

Looping improves quality but requires loops to run serially during
  autoregressive decoding. PLT (\autoref{sec:background_plt}) shifts the
  connections between loops across adjacent tokens, allowing different loops
  to run in parallel (\autoref{fig:plt-diagonal}). In our experiments, however,
  PLT underperforms \emph{Full-Stack Looping}
  (\autoref{fig:plt-design}). We investigate whether this quality loss is an
  inherent limitation of PLT or whether it can be mitigated by reconstructing PLT one
  component at a time.

\begin{figure*}[t]
  \centering
  \includegraphics[width=\textwidth]{
    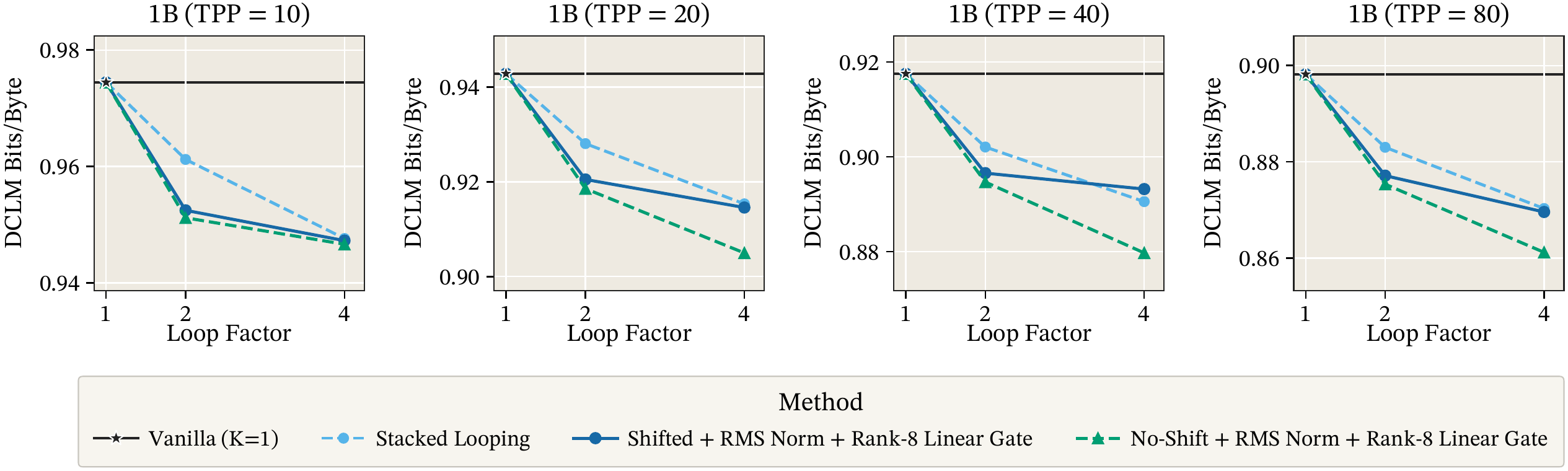
  }
  \caption{\textbf{Normalized shifted PLT retains the benefit of looping at
  1B.} DCLM BPB at TPP 10, 20, 40, and 80. Shifted and unshifted PLT use
  separate RMS normalization and a rank-$8$ linear gate. At $L=2$, shifted PLT
  closely tracks the unshifted control and improves upon \emph{Full-Stack Looping};
  at $L=4$, a clearer gap to the unshifted control remains. Lower BPB is better.
  }
  \label{fig:plt-1b-loops}
\end{figure*}

\textbf{Recovering and improving PLT quality.}
To isolate the source of the quality degradation, we reconstruct PLT one
  component at a time. As defined in \autoref{fig:plt-diagonal}, CLP passes
  \(\bm b_n^{(\ell)}=\bm e_n+\bm h_{n-1}^{(\ell-1)}\) to loop \(\ell\), while
  our unshifted control uses
  \(\bm b_n^{(\ell)}=\bm e_n+\bm h_n^{(\ell-1)}\). The unshifted control can
  therefore be viewed as \emph{Full-Stack Looping} with the current-token
  embedding added again at every loop. Starting from raw shifted CLP, we add KV
  sharing and normalized G-SWA to reconstruct \emph{Original PLT}
  (\autoref{sec:background_plt}). For the 200M model at both TPP 40 and 80,
  neither \emph{Original PLT} nor the intermediate variants formed by adding
  its components recover the performance of \emph{Full-Stack Looping}. These
  shifted variants perform near the \emph{Vanilla (No Looping)} baseline when
  fewer loops are used ($L\leq 2$), while raw shifted CLP can become even worse as the
  number of loops increases. In contrast, the unshifted control outperforms
  \emph{Full-Stack Looping} (\autoref{fig:plt-design}). Our experiments reveal a second source of degradation: the RMS magnitudes of
  the current-token embedding and the previous-token state differ substantially,
  with the magnitude of this mismatch varying across TPP settings and loop depths
  (\autoref{app:plt-normalization}). When the two inputs are added directly, the
  higher-magnitude representation can overwhelm the contribution of the other.
  To mitigate this mismatch, we define
  \(\hat{\bm e}_n=\operatorname{RMSNorm}(\bm e_n)\) and
  \(\hat{\bm h}_{n-1}^{(\ell-1)}
  =\operatorname{RMSNorm}(\bm h_{n-1}^{(\ell-1)})\), and combine them using a
  gate \(\bm g_n^{(\ell)}\in[0,1]^d\):
  \(\bm b_n^{(\ell)}=\bm g_n^{(\ell)}\odot\hat{\bm e}_n+
  (\bm 1-\bm g_n^{(\ell)})\odot\hat{\bm h}_{n-1}^{(\ell-1)}\).
  As seen in \autoref{fig:plt-design}, simply applying RMSNorm to both inputs and then adding them performs worse
  than controlling their mixture, showing that their relative weighting also
  matters. For \autoref{fig:plt-design}, we use the fixed equal mixture
  \(\bm g_n^{(\ell)}=\tfrac{1}{2}\bm 1\). Across the evaluated TPP settings,
  numbers of loops, and PLT component configurations, the resulting shifted,
  inference-efficient variants approach the quality of
  \emph{Full-Stack Looping}. The normalized and gated unshifted control also
  improves over the unnormalized unshifted control and again outperforms
  \emph{Full-Stack Looping}. Maintaining a separate KV cache for each loop
  provides the best quality across the evaluated designs; however, KV sharing
  combined with normalized G-SWA, as in \emph{Original PLT}, approaches this
  performance while requiring less cache storage. Additional ablations at 300M
  appear in \autoref{app:plt-design-progression}.

  \textbf{The current-token embedding matters more than the exact gate.}
  The analysis above uses the fixed gate
  \(\bm g_n^{(\ell)}=\tfrac{1}{2}\bm 1\). To test whether a more expressive
  gate improves quality, we compare four alternatives defined in
  \autoref{app:plt-gates}: fixed 50/50, scalar, rank-\(8\) linear, and
  rank-\(8\) MLP gates. They perform similarly across model scales, TPP
  settings, and numbers of loops, although the rank-\(8\) linear gate is marginally better. We use the rank-\(8\) linear gate in subsequent
  experiments. It is shared across loops and adds less than \(0.015\%\) to the
  non-embedding parameter count. We separately test the role of the
  current-token embedding in \autoref{app:plt-current-token}. Removing it
  causes shifted PLT to perform worse than \emph{Vanilla (No Looping)}, and
  several runs diverge. We hypothesize that, without \(\bm e_n\), each shifted loop must infer the
  current token solely from the previous-token state. As this state
  represents a contextual prediction rather than the token that actually
  occurred, prediction errors may propagate through subsequent loops and
  accumulate across token positions, causing quality degradation or divergence.
  Reintroducing \(\bm e_n\) anchors every loop to the observed current token and
  limits this accumulation.

  \textbf{Our normalized shifted PLT scales.}
  Hereafter, \emph{Normalized Shifted PLT} refers to our modified design that
  uses the previous token's hidden state and allows loops to run in parallel,
  whereas \emph{Normalized Unshifted PLT} uses the current token's hidden state
  and therefore remains sequential. Both variants separately normalize the
  current-token embedding and hidden state, combine them using a rank-\(8\)
  linear gate, and maintain a separate KV cache for each loop. Unlike
  \emph{Original PLT}, neither uses KV sharing nor G-SWA. We evaluate both variants from 100M to 1B. At 1B and \(L=2\),
  \emph{Normalized Shifted PLT} achieves lower DCLM BPB than
  \emph{Full-Stack Looping} at every displayed training budget and closely
  tracks \emph{Normalized Unshifted PLT}
  (\autoref{fig:plt-1b-loops}). At \(L=4\), it remains competitive with
  \emph{Full-Stack Looping}, although its gap to
  \emph{Normalized Unshifted PLT} widens. The same trends hold across six
  evaluation corpora at 1B (\autoref{app:plt-public-evals}). Additional sweeps
  across model sizes, numbers of loops, and TPP settings appear in
  \autoref{app:plt-loop-sweeps} and \autoref{app:plt-tpp-sweeps}. We next
  examine the training-data and training-FLOP efficiency of looping.

\textbf{Looping can improve data and FLOP efficiency.}
  As shown in \autoref{fig:headline-looping-efficiency}, looping can reach a
  given DCLM BPB using fewer training tokens than
  \emph{Vanilla (No Looping)}, suggesting that looping can partly
  substitute for additional training data. Data efficiency does not
  automatically imply FLOP efficiency because each loop adds computation per
  token. In the evaluated over-trained regime, however, the reduction in
  required training tokens outweighs this additional cost: across model sizes
  from 100M to 1B, two-loop models trained at TPP 80 reach a given BPB using
  fewer training FLOPs than \emph{Vanilla} models trained at TPP 160. This
  advantage does not hold uniformly across numbers of loops: although \(L=4\)
  can achieve lower absolute BPB, it provides a weaker improvement per training
  FLOP (\autoref{app:plt-compute}).

\section{Related Work}
\label{sec:related_work}

Prior work studies looping from several complementary angles. Scaling studies ask how much additional looping is worth under different parameter, data, and compute budgets, either through scaling laws or by comparing shared recurrence with added physical depth \citep{prairie2026parcae,schwethelm2026recurrenceworthisodepthscaling}. Other work explores particular choices for where and how recurrence is introduced in the transformer \citep{cai2026tmlr,zeitoun2026hyperlooptransformers}. As these studies use different looping designs and evaluate them under different resource constraints, it is difficult to separate the effect of the design from the effect of the budget. We instead vary where and how much to loop within a common dense model family and evaluate these choices under both parameter and training-compute constraints. Another line of work asks whether every loop should use exactly the same weights. Relaxed Recursive Transformers \citep{bae2025relaxed} give each loop a small set of additional parameters when converting a pretrained transformer into a looped model. We instead train the looped transformer and its loop-specific adapters together from scratch and test whether these adapters become more useful as the number of loops increases. Looping also raises a separate challenge during autoregressive inference. In ordinary looping, the model must finish one loop for a token before starting the next. Parallel Loop Transformers (PLTs) combat this by shifting information across adjacent token positions so that loop computations can be parallelized during decoding \citep{wu2025parallel}. This change alters the information available to each loop, and recent work reports that gains can diminish as more parallel loops are added \citep{yang2026loopcoder}. This leaves open what the shifted transition must preserve to retain quality as the number of loops increases. We find that preserving the current-token embedding and separately normalizing it and the state passed from the previous token substantially improves parallel looping at larger loop counts. Our work therefore connects three aspects of looping within a common setting: where and how much to loop, whether different loops should use exactly the same weights, and how loops can be parallelized during autoregressive inference. See \autoref{app:extended_related_work} for a more detailed discussion.

\section{Conclusion and Future Directions}

  We show that looping is a multidimensional design
  problem that must jointly account for stored parameters, training FLOPs, data
  efficiency, and inference efficiency. A promising future direction is to
  find ways of combining PLT with speculative decoding to further reduce inference latency.
  Another is to develop a theory of how normalization and current-token
  embedding reinjection affect stability in both the cross-token recurrence of
  PLT and the same-token recurrence of standard looping, and to identify which
  regimes favor each approach.

\section{AI Use Statement}

In this work, we used generative AI tools, specifically ChatGPT, Codex, Claude,
and Claude Code, for assistance with writing, code development and editing,
debugging, and auxiliary sanity checks. We did not use generative AI tools to
develop the core research methodology, formulate the main research hypotheses,
design the experiments, or interpret the experimental results. All AI-generated
or AI-suggested text and code were reviewed by the authors. AI-assisted code was
tested and validated for correctness before being used in the reported
experiments. The authors made all methodological and experimental decisions and
independently verified the analyses and reported results. We take full
responsibility for the final content of this work, including any text, code,
claims, or artifacts produced with the assistance of generative AI tools.

\bibliographystyle{assets/plainnat}
\bibliography{references}

@techreport{microsoft2025mai,
  title={Mai-thinking-1: Building a hill-climbing machine},
  author={The Microsoft AI Team},
  year={2026},
  institution={Microsoft AI}
}

@inproceedings{saunshi2025reasoning,
  title={Reasoning with latent thoughts: On the power of looped transformers},
  author={Saunshi, Nikunj and Dikkala, Nishanth and Li, Zhiyuan and Kumar, Sanjiv and J Reddi, Sashank},
  booktitle={International Conference on Learning Representations},
  volume={2025},
  pages={14855--14881},
  year={2025}
}

@inproceedings{
geiping2025scaling,
title={Scaling up Test-Time Compute with Latent Reasoning: A Recurrent Depth Approach},
author={Jonas Geiping and Sean Michael McLeish and Neel Jain and John Kirchenbauer and Siddharth Singh and Brian R. Bartoldson and Bhavya Kailkhura and Abhinav Bhatele and Tom Goldstein},
booktitle={The Thirty-ninth Annual Conference on Neural Information Processing Systems},
year={2025},
url={https://openreview.net/forum?id=S3GhJooWIC}
}

@article{dehghani2018universal,
  title={Universal transformers},
  author={Dehghani, Mostafa and Gouws, Stephan and Vinyals, Oriol and Uszkoreit, Jakob and Kaiser, {\L}ukasz},
  journal={arXiv preprint arXiv:1807.03819},
  year={2018}
}

@misc{zhu2026scalinglatentreasoninglooped,
      title={Scaling Latent Reasoning via Looped Language Models}, 
      author={Rui-Jie Zhu and Zixuan Wang and Kai Hua and Tianyu Zhang and Ziniu Li and Haoran Que and Boyi Wei and Zixin Wen and Fan Yin and He Xing and Lu Li and Jiajun Shi and Kaijing Ma and Shanda Li and Taylor Kergan and Andrew Smith and Xingwei Qu and Mude Hui and Bohong Wu and Qiyang Min and Hongzhi Huang and Xun Zhou and Wei Ye and Jiaheng Liu and Jian Yang and Yunfeng Shi and Chenghua Lin and Enduo Zhao and Tianle Cai and Ge Zhang and Wenhao Huang and Yoshua Bengio and Jason Eshraghian},
      year={2026},
      eprint={2510.25741},
      archivePrefix={arXiv},
      primaryClass={cs.CL},
      url={https://arxiv.org/abs/2510.25741}, 
}

@article{ju2022staircase,
  title={Staircase attention for recurrent processing of sequences},
  author={Ju, Da and Roller, Stephen and Sukhbaatar, Sainbayar and Weston, Jason E},
  journal={Advances in neural information processing systems},
  volume={35},
  pages={13203--13213},
  year={2022}
}

@inproceedings{
prairie2026parcae,
title={Parcae: Scaling Laws For Stable Looped Language Models},
author={Hayden Prairie and Zachary Novack and Taylor Berg-Kirkpatrick and Daniel Y Fu},
booktitle={Third Conference on Language Modeling},
year={2026},
url={https://openreview.net/forum?id=pIBqtrqFeP}
}

@misc{schwethelm2026recurrenceworthisodepthscaling,
      title={How Much Is One Recurrence Worth? Iso-Depth Scaling Laws for Looped Language Models}, 
      author={Kristian Schwethelm and Daniel Rueckert and Georgios Kaissis},
      year={2026},
      eprint={2604.21106},
      archivePrefix={arXiv},
      primaryClass={cs.LG},
      url={https://arxiv.org/abs/2604.21106}, 
}

@misc{zeitoun2026hyperlooptransformers,
      title={Hyperloop Transformers}, 
      author={Abbas Zeitoun and Lucas Torroba-Hennigen and Yoon Kim},
      year={2026},
      eprint={2604.21254},
      archivePrefix={arXiv},
      primaryClass={cs.LG},
      url={https://arxiv.org/abs/2604.21254}, 
}

@inproceedings{
cai2026tmlr,
title={T2{MLR}: Transformer with Temporal Middle-Layer Recurrence},
author={Ziyang Cai and Xingyu Zhu and Yihe Dong and Yinghui He and Sanjeev Arora},
booktitle={Workshop on Latent {\&} Implicit Thinking {\textendash} Going Beyond CoT Reasoning},
year={2026},
url={https://openreview.net/forum?id=fQbk1EQWBO}
}

@article{bae2026mixture,
  title={Mixture-of-recursions: Learning dynamic recursive depths for adaptive token-level computation},
  author={Bae, Sangmin and Kim, Yujin and Bayat, Reza and Kim, Sungnyun and Ha, Jiyoun and Schuster, Tal and Fisch, Adam and Harutyunyan, Hrayr and Ji, Ziwei and Courville, Aaron and others},
  journal={Advances in Neural Information Processing Systems},
  volume={38},
  pages={96572--96617},
  year={2026}
}

@misc{lee2026sparselayerscriticalscaling,
      title={Sparse Layers are Critical to Scaling Looped Language Models}, 
      author={Ryan Lee and Jacob Biloki and Edward J. Hu and Jonathan May},
      year={2026},
      eprint={2605.09165},
      archivePrefix={arXiv},
      primaryClass={cs.LG},
      url={https://arxiv.org/abs/2605.09165}, 
}

@inproceedings{bae2025relaxed,
  title={Relaxed recursive transformers: Effective parameter sharing with layer-wise lora},
  author={Bae, Sangmin and Fisch, Adam and Harutyunyan, Hrayr and Ji, Ziwei and Kim, Seungyeon and Schuster, Tal},
  booktitle={International Conference on Learning Representations},
  volume={2025},
  pages={34282--34327},
  year={2025}
}

@article{wu2025parallel,
  title={Parallel loop transformer for efficient test-time computation scaling},
  author={Wu, Bohong and Chen, Mengzhao and Luo, Xiang and Yan, Shen and Yu, Qifan and Xia, Fan and Zhang, Tianqi and Zhan, Hongrui and Zhong, Zheng and Zhou, Xun and others},
  journal={arXiv preprint arXiv:2510.24824},
  year={2025}
}

@article{yang2026loopcoder,
  title={LoopCoder-v2: Only Loop Once for Efficient Test-Time Computation Scaling},
  author={Yang, Jian and Guo, Shawn and Zhang, Wei and Zheng, Tianyu and Du, Yaxin and Li, Haau-Sing and Wu, Jiajun and Song, Yue and Xing, Yan and Cai, Qingsong and others},
  journal={arXiv preprint arXiv:2606.18023},
  year={2026}
}

@article{labovich2026stability,
  title={Stability and Generalization in Looped Transformers},
  author={Labovich, Asher},
  journal={arXiv preprint arXiv:2604.15259},
  year={2026}
}

@article{kim2026looped,
  title={Looped Transformers with Source-Centered State Evolution},
  author={Kim, Bum Jun and Hayashi, Kohei and Kamiya, Shunsuke and Koyama, Masanori and Iwasawa, Yusuke and Matsuo, Yutaka},
  journal={arXiv preprint arXiv:2607.27656},
  year={2026}
}

@article{hu2021lora,
  title={Lora: Low-rank adaptation of large language models},
  author={Hu, Edward J and Shen, Yelong and Wallis, Phillip and Allen-Zhu, Zeyuan and Li, Yuanzhi and Wang, Shean and Wang, Lu and Chen, Weizhu},
  journal={arXiv preprint arXiv:2106.09685},
  year={2021}
}

@article{li2024datacomp,
  title={Datacomp-lm: In search of the next generation of training sets for language models},
  author={Li, Jeffrey and Fang, Alex and Smyrnis, Georgios and Ivgi, Maor and Jordan, Matt and Gadre, Samir and Bansal, Hritik and Guha, Etash and Keh, Sedrick and Arora, Kushal and others},
  journal={Advances in Neural Information Processing Systems},
  volume={37},
  pages={14200--14282},
  year={2024}
}

@article{magnusson2024paloma,
  title={Paloma: A benchmark for evaluating language model fit},
  author={Magnusson, Ian and Bhagia, Akshita and Hofmann, Valentin and Soldaini, Luca and Jha, Ananya H and Tafjord, Oyvind and Schwenk, Dustin and Walsh, Evan and Elazar, Yanai and Lo, Kyle and others},
  journal={Advances in Neural Information Processing Systems},
  volume={37},
  pages={64338--64376},
  year={2024}
}

@article{cobbe2021training,
  title={Training verifiers to solve math word problems},
  author={Cobbe, Karl and Kosaraju, Vineet and Bavarian, Mohammad and Chen, Mark and Jun, Heewoo and Kaiser, Lukasz and Plappert, Matthias and Tworek, Jerry and Hilton, Jacob and Nakano, Reiichiro and others},
  journal={arXiv preprint arXiv:2110.14168},
  year={2021}
}

@article{hendrycks2021measuring,
  title={Measuring mathematical problem solving with the math dataset},
  author={Hendrycks, Dan and Burns, Collin and Kadavath, Saurav and Arora, Akul and Basart, Steven and Tang, Eric and Song, Dawn and Steinhardt, Jacob},
  journal={arXiv preprint arXiv:2103.03874},
  year={2021}
}

@article{chen2021evaluating,
  title={Evaluating large language models trained on code},
  author={Chen, Mark and Tworek, Jerry and Jun, Heewoo and Yuan, Qiming and Pinto, Henrique Ponde De Oliveira and Kaplan, Jared and Edwards, Harri and Burda, Yuri and Joseph, Nicholas and Brockman, Greg and others},
  journal={arXiv preprint arXiv:2107.03374},
  year={2021}
}

@misc{austin2021programsynthesislargelanguage,
      title={Program Synthesis with Large Language Models}, 
      author={Jacob Austin and Augustus Odena and Maxwell Nye and Maarten Bosma and Henryk Michalewski and David Dohan and Ellen Jiang and Carrie Cai and Michael Terry and Quoc Le and Charles Sutton},
      year={2021},
      eprint={2108.07732},
      archivePrefix={arXiv},
      primaryClass={cs.PL},
      url={https://arxiv.org/abs/2108.07732}, 
}

@inproceedings{soldaini2024dolma,
  title={Dolma: An open corpus of three trillion tokens for language model pretraining research},
  author={Soldaini, Luca and Kinney, Rodney and Bhagia, Akshita and Schwenk, Dustin and Atkinson, David and Authur, Russell and Bogin, Ben and Chandu, Khyathi and Dumas, Jennifer and Elazar, Yanai and others},
  booktitle={Proceedings of the 62nd Annual Meeting of the Association for Computational Linguistics (Volume 1: Long Papers)},
  pages={15725--15788},
  year={2024}
}

@article{raffel2020exploring,
  title={Exploring the limits of transfer learning with a unified text-to-text transformer},
  author={Raffel, Colin and Shazeer, Noam and Roberts, Adam and Lee, Katherine and Narang, Sharan and Matena, Michael and Zhou, Yanqi and Li, Wei and Liu, Peter J},
  journal={Journal of machine learning research},
  volume={21},
  number={140},
  pages={1--67},
  year={2020}
}

@misc{weber2024redpajama,
      title={RedPajama: an Open Dataset for Training Large Language Models}, 
      author={Maurice Weber and Daniel Fu and Quentin Anthony and Yonatan Oren and Shane Adams and Anton Alexandrov and Xiaozhong Lyu and Huu Nguyen and Xiaozhe Yao and Virginia Adams and Ben Athiwaratkun and Rahul Chalamala and Kezhen Chen and Max Ryabinin and Tri Dao and Percy Liang and Christopher Ré and Irina Rish and Ce Zhang},
      year={2024},
      eprint={2411.12372},
      archivePrefix={arXiv},
      primaryClass={cs.CL},
      url={https://arxiv.org/abs/2411.12372},}

@article{merity2016pointer,
  title={Pointer sentinel mixture models},
  author={Merity, Stephen and Xiong, Caiming and Bradbury, James and Socher, Richard},
  journal={arXiv preprint arXiv:1609.07843},
  year={2016}
}

@misc{sakaguchi2019winograndeadversarialwinogradschema,
      title={WinoGrande: An Adversarial Winograd Schema Challenge at Scale}, 
      author={Keisuke Sakaguchi and Ronan Le Bras and Chandra Bhagavatula and Yejin Choi},
      year={2019},
      eprint={1907.10641},
      archivePrefix={arXiv},
      primaryClass={cs.CL},
      url={https://arxiv.org/abs/1907.10641}, 
}

@article{clark2018think,
  title={Think you have solved question answering? try arc, the ai2 reasoning challenge},
  author={Clark, Peter and Cowhey, Isaac and Etzioni, Oren and Khot, Tushar and Sabharwal, Ashish and Schoenick, Carissa and Tafjord, Oyvind},
  journal={arXiv preprint arXiv:1803.05457},
  year={2018}
}

@inproceedings{zellers2019hellaswag,
  title={Hellaswag: Can a machine really finish your sentence?},
  author={Zellers, Rowan and Holtzman, Ari and Bisk, Yonatan and Farhadi, Ali and Choi, Yejin},
  booktitle={Proceedings of the 57th annual meeting of the association for computational linguistics},
  pages={4791--4800},
  year={2019}
}

@inproceedings{mihaylov2018can,
  title={Can a suit of armor conduct electricity? a new dataset for open book question answering},
  author={Mihaylov, Todor and Clark, Peter and Khot, Tushar and Sabharwal, Ashish},
  booktitle={Proceedings of the 2018 conference on empirical methods in natural language processing},
  pages={2381--2391},
  year={2018}
}

@misc{bisk2019piqareasoningphysicalcommonsense,
      title={PIQA: Reasoning about Physical Commonsense in Natural Language}, 
      author={Yonatan Bisk and Rowan Zellers and Ronan Le Bras and Jianfeng Gao and Yejin Choi},
      year={2019},
      eprint={1911.11641},
      archivePrefix={arXiv},
      primaryClass={cs.CL},
      url={https://arxiv.org/abs/1911.11641}, 
}

@misc{hendrycks2021measuringmassivemultitasklanguage,
      title={Measuring Massive Multitask Language Understanding}, 
      author={Dan Hendrycks and Collin Burns and Steven Basart and Andy Zou and Mantas Mazeika and Dawn Song and Jacob Steinhardt},
      year={2021},
      eprint={2009.03300},
      archivePrefix={arXiv},
      primaryClass={cs.CY},
      url={https://arxiv.org/abs/2009.03300}, 
}

@article{wang2024mmlu,
  title={Mmlu-pro: A more robust and challenging multi-task language understanding benchmark},
  author={Wang, Yubo and Ma, Xueguang and Zhang, Ge and Ni, Yuansheng and Chandra, Abhranil and Guo, Shiguang and Ren, Weiming and Arulraj, Aaran and He, Xuan and Jiang, Ziyan and others},
  journal={Advances in Neural Information Processing Systems},
  volume={37},
  pages={95266--95290},
  year={2024}
}

@misc{talmor2019commonsenseqaquestionansweringchallenge,
      title={CommonsenseQA: A Question Answering Challenge Targeting Commonsense Knowledge}, 
      author={Alon Talmor and Jonathan Herzig and Nicholas Lourie and Jonathan Berant},
      year={2019},
      eprint={1811.00937},
      archivePrefix={arXiv},
      primaryClass={cs.CL},
      url={https://arxiv.org/abs/1811.00937}, 
}

@misc{sap2019socialiqacommonsensereasoningsocial,
      title={SocialIQA: Commonsense Reasoning about Social Interactions}, 
      author={Maarten Sap and Hannah Rashkin and Derek Chen and Ronan LeBras and Yejin Choi},
      year={2019},
      eprint={1904.09728},
      archivePrefix={arXiv},
      primaryClass={cs.CL},
      url={https://arxiv.org/abs/1904.09728}, 
}

\clearpage
\appendix

\vspace{1em}
\begin{center}
\rule{\textwidth}{1pt}\\[0.75em]
{\LARGE\bfseries Appendix}\\[0.5em]
\rule{\textwidth}{1pt}
\end{center}
\vspace{1em}
\tableofcontents

\clearpage

\section{Related Work}
\label{app:extended_related_work}

\paragraph{Looped Transformers.}
Looped Transformers increase effective depth by repeatedly applying shared computation \citep{dehghani2018universal}. \citet{saunshi2025reasoning} show that shared looping can recover much of the benefit of unique depth on iterative reasoning problems. \citet{geiping2025scaling} instead study looping as a mechanism for scaling latent test-time computation, training a language model whose loop depth can vary at inference. More recently, \citet{zhu2026scalinglatentreasoninglooped} combine looped pretraining with learned depth allocation at substantially larger data scale. Relatedly, Staircase Attention \citep{ju2022staircase} combines looping across sequence time and depth, with its Ladder variant repeatedly applying a shared Transformer core. Together, these works establish that repeated computation can provide useful capability and reasoning benefits. Compared with prior looping work, we provide a more systematic account of the trade-offs among parameter efficiency, training FLOPs, specialization, and the architectural requirements for efficient autoregressive inference. Our analysis shows that looping is not uniformly beneficial: the preferred design depends on which resource is constrained.

\paragraph{Scaling and efficiency of weight reuse.}
Recent work studies looping from scaling, capacity, and efficiency perspectives. \citet{prairie2026parcae} develop a stable looped architecture by analyzing the looped residual stream as a dynamical system, and use it to derive scaling laws for loop depth and data under fixed parameter and compute budgets. \citet{schwethelm2026recurrenceworthisodepthscaling} instead quantify the effective capacity of repeated shared computation relative to unique depth, finding that each additional loop provides only partial equivalence to an additional untied block. Other work modifies the looping architecture itself. Hyperloop combines a looped middle block with hyper-connections to improve parameter efficiency \citep{zeitoun2026hyperlooptransformers}, while T$^2$MLR introduces a temporal pathway through intermediate layers \citep{cai2026tmlr}. Mixture-of-Recursions uses token-level routing to allocate different loop depths across tokens \citep{bae2026mixture}. Sparse looped models use expert routing to allow the same physical layers to realize different effective transformations across loop applications \citep{lee2026sparselayerscriticalscaling}. Unlike work that studies a single scaling relation or advocates a particular architectural modification, we hold the backbone dense and the training setup fixed while jointly varying loop placement, loop depth, loop-specific specialization, and state propagation during autoregressive decoding. We evaluate these choices under fixed parameter, training-data, and training-FLOP budgets. This design reveals trade-offs that are obscured when only one resource is considered. Our goal is therefore not to identify a universally optimal looping mechanism, but to characterize how the preferred design changes with the resource constraint.

\paragraph{Relaxing Exact Parameter Sharing.}
Strict looping requires the same parameters to serve every depth application. Relaxed Recursive Transformers (RRTs) relax this constraint by converting a pretrained untied Transformer into a compact looped model and adding depth-specific low-rank adapters \citep{bae2025relaxed}. This procedure substantially improves over strict tying and can recover much of the performance lost when originally distinct layers are collapsed into a shared block. Related work introduces flexibility through different mechanisms. Hyperloop augments a looped middle block with loop-level hyper-connections, allowing the residual pathway to vary across loop applications with little additional parameter cost \citep{zeitoun2026hyperlooptransformers}. Mixture-of-Recursions instead uses routers to assign individual tokens different loop depths, combining parameter sharing with adaptive computation \citep{bae2026mixture}. Sparse looped models select different experts at different loop applications, so weight sharing need not imply that every loop realizes the same effective transformation \citep{lee2026sparselayerscriticalscaling}. Our setting isolates a different question. Rather than converting a pretrained untied model or introducing routing and expert-specific capacity, we train a dense looped model from initialization and add small loop-specific corrections to the shared block. We use these corrections primarily as a diagnostic: does the value of depth-specific capacity increase as the same shared transformation is applied more times? This controlled comparison distinguishes the general cost of exact sharing from the benefits of a particular architectural mechanism, while preserving a common backbone, training setup, and parameter-sharing structure.

\paragraph{Parallel Looping and State Propagation.}
Conventional looping reduces the number of unique parameters, but places successive loop applications on the sequential path of autoregressive generation. Parallel Loop Transformers (PLTs) reorganize this dependency through Cross-Loop Parallelism (CLP): recurrent state is shifted across token positions, allowing loop stages to be pipelined during decoding. PLTs combine this schedule with shared KV caches and gated sliding-window attention to bound cache growth \citep{wu2025parallel}. LoopCoder-v2 studies this design at larger scale and finds a strongly non-monotonic dependence on loop count: a second loop improves performance, whereas three or more loops regress as the benefit of further refinement is outweighed by the positional mismatch introduced by CLP \citep{yang2026loopcoder}. Related work studies how looped dynamics can remain stable and input-dependent. \citet{labovich2026stability} analyze fixed-point stability in looped Transformers, showing that recall together with outer normalization can yield reachable, input-dependent, and trainable dynamics. \citet{kim2026looped} develop source-centered state evolution, which preserves an input-conditioned anchor while allowing deviations from that anchor to drive recurrent computation. Our work addresses the interface created specifically by cross-token parallelization. Under CLP, a loop stage must combine a representation of the current token with a shifted state produced at a different token position and loop depth. We find that the main limitation of naïve PLT is this shifted re-injection transition, rather than cache sharing itself. Our controlled ablations isolate the contributions of the cross-token shift, attention and cache mechanisms, normalization, recurrent-state mixing, and access to the current-token representation. This identifies separate normalization of the current-token and shifted-state streams, together with preservation of the current-token path, as the central repair. Thus, rather than proposing normalization or input preservation as general primitives, we characterize their role in making the parallel looping interface effective.

\clearpage

\section{Background}
\label{app:extended_background}

\paragraph{Notation.}
We use bold uppercase letters to denote matrices (e.g., $\bm X,\bm W$), bold lowercase letters to denote vectors (e.g., $\bm x$), and unbolded letters for scalars and indices. Calligraphic letters denote functions or transformations (e.g., $\mathcal{F}$), while Greek letters are used for scalar- or vector-valued parameters when convenient. For a positive integer $K$, we write $[K]\coloneqq\{1,\ldots,K\}$. We use $[\bm A,\bm B]$ to denote concatenation along the feature dimension and $[\bm A;\bm B]$ to denote stacking along the sequence or row dimension. The operator $\odot$ denotes element-wise multiplication, and $\sigma(\cdot)$ denotes the logistic sigmoid. For a Transformer with $L$ physical layers, we denote the parameters of layer $m\in[L]$ by $\theta_m$, and write $\Theta\coloneqq\{\theta_1,\ldots,\theta_L\}$ for the collection of all Transformer-layer parameters. When a layer or block is reused across multiple loop iterations, the same parameter set $\theta_m$ is used at every corresponding invocation unless explicitly stated otherwise. Architecture-specific notation is introduced alongside the corresponding construction.

\subsection{Looped Transformers}
\label{sec:background_looping}

\paragraph{Transformer blocks.}
We begin with the basic unit of computation that will be reused throughout the paper. Let $\mathcal{A}_{\theta_A}$ denote causal self-attention and $\mathcal{M}_{\theta_M}$ denote the position-wise MLP. We use a pre-normalized Transformer block, in which attention and the MLP each act on a normalized residual stream.

\begin{definitionbox}[label=def:transformer_block]
{\textbf{Pre-normalized Transformer block. }}
Given a sequence representation $\bm X\in\mathbb{R}^{N\times D}$, a Transformer block with parameters $\theta=(\theta_A,\theta_M)$ computes
\begin{align*}
    \bm U
    &\coloneqq
    \bm X
    + \mathcal{A}_{\theta_A}\!\left(\operatorname{LN}_A(\bm X)\right), \\
    \mathcal{T}_{\theta}(\bm X)
    &\coloneqq
    \bm U
    + \mathcal{M}_{\theta_M}\!\left(\operatorname{LN}_M(\bm U)\right).
\end{align*}
A standard depth-$B$ Transformer composes $B$ such blocks, each with its own parameters (Note that $H$ denotes hidden representations):
\begin{align*}
    \bm H_0
    &\coloneqq
    \bm X, \\
    \bm H_b
    &\coloneqq
    \mathcal{T}_{\theta_b}\!\left(\bm H_{b-1}\right),
    \qquad \forall b\in[B].
\end{align*}
\end{definitionbox}

In a standard Transformer, computational depth and parameter depth grow together: moving from $\bm H_{b-1}$ to $\bm H_b$ introduces both computation by another transformer block and a new parameter set $\theta_b$. Increasing depth therefore increases not only the amount of computation performed on each token, but also the number of parameters that must be stored. Looping breaks this coupling. Rather than introducing a new transformation at every additional depth, it repeatedly applies the same transformation, allowing computational depth to grow while keeping the  parameters fixed.

\paragraph{Weight reuse through looping.}
To formalize this distinction, let $\mathcal{F}_{\theta}$ denote any Transformer computation chosen for reuse. Depending on the construction, $\mathcal{F}_{\theta}$ may correspond to a single Transformer block, a contiguous group of blocks, or the entire Transformer stack. A looped Transformer applies this same parameterized function repeatedly.

\begin{definitionbox}[label=def:looped_transformer]
{\textbf{Looped Transformer} \citep{saunshi2025reasoning,geiping2025scaling, dehghani2018universal}. }
Given an input representation $\bm X$, the $L$-loop model induced by $\mathcal{F}_{\theta}$ is
\begin{align*}
    \bm H^{(0)}
    &\coloneqq
    \bm X, \\
    \bm H^{(\ell)}
    &\coloneqq
    \mathcal{F}_{\theta}\!\left(\bm H^{(\ell-1)}\right),
    \qquad \forall \ell\in[L].
\end{align*}
The parameter set $\theta$ is shared across all loop iterations. Thus, the superscript $(\ell)$ indexes repeated applications of the same transformation rather than distinct parameterized layers.
\end{definitionbox}

This separation between \emph{stored depth} and \emph{executed depth} is the central appeal of looping. Suppose a single application of $\mathcal{F}_{\theta}$ contains $P$ parameters and incurs compute cost $C$. Repeating it $L$ times requires approximately $LC$ compute while still storing only the same $P$ parameters:
$P_{\mathrm{loop}}\approx P, C_{\mathrm{loop}}
    \approx LC$. Looping therefore trades additional computation for parameter reuse: it can increase the effective computational depth of a model without requiring a proportional increase in parameter memory. The extra depth is not free, however, since the reused function must still be evaluated at every loop iteration. This definition deliberately leaves $\mathcal{F}_{\theta}$ unspecified. In a Transformer, there are several natural choices for what computation to reuse and in what order to reuse it. One may repeat the entire stack, repeatedly apply each layer before advancing to the next, restrict reuse to a middle portion of the network, or loop only particular subcomponents such as attention or the MLP. These choices induce different computational graphs despite sharing the same underlying principle of parameter reuse. We define the constructions considered in this work next.

\paragraph{Loop placement and computation order.}
A natural next question is \emph{where} this weight reuse can occur and \emph{in what order} the repeated operations are executed. These choices are distinct: two architectures may reuse exactly the same set of Transformer blocks the same number of times, yet induce different computational graphs because the repetitions are ordered differently. To make this distinction explicit, we write $\mathcal{G}^{\circ R}$ for $R$ successive applications of a transformation $\mathcal{G}$.

\begin{definitionbox}[label=def:whole_and_block_looping]
{\textbf{Full-Stack and Per-Layer Looping. }}
Let
$\mathcal{F}_{\theta}
=\mathcal{T}_{\theta_B}\circ\cdots\circ\mathcal{T}_{\theta_1}$
denote a stack of $B$ Transformer blocks.

\begin{enumerate}
    \item \emph{Full-Stack Looping} repeatedly applies the complete stack of the layers of a transformer model:
    \begin{align*}
        \bm H^{(L)}
        &=
        \mathcal{F}_{\theta}^{\circ L}(\bm X).
    \end{align*}

    \item \emph{Per-Layer Looping} instead completes the looping of each block before advancing to the next. If block $b$ is applied $R_b$ times, define
    \begin{align*}
        \bm Z_{1,0}
        &\coloneqq
        \bm X, \\
        \bm Z_{b,0}
        &\coloneqq
        \bm Z_{b-1,R_{b-1}},
        \qquad
        \forall b\in\{2,\ldots,B\}, \\
        \bm Z_{b,r}
        &\coloneqq
        \mathcal{T}_{\theta_b}\!\left(\bm Z_{b,r-1}\right),
        \qquad
        \forall b\in[B],\ \forall r\in[R_b].
    \end{align*}
\end{enumerate}
\end{definitionbox}

When $R_b=L$ for every block, Full-Stack and Per-Layer Looping use the same stored parameters and perform the same number of Transformer-block evaluations; they differ only in the ordering of those evaluations. Full-Stack Looping repeatedly traverses the full stack of transformer blocks, whereas Per-Layer Looping completes all loops of a block before advancing to the next. Since a Transformer block is itself composed of attention and MLP sublayers, we next consider whether reuse should operate on the entire block or on its individual components.

\paragraph{Independent attention and MLP looping.}
Per-Layer Looping can be refined further by choosing which subcomponents of a Transformer block are reused. To make these choices explicit, we define again the residual attention and MLP maps as follows.
\begin{align*}
    \widehat{\mathcal{A}}_{\theta_A}(\bm X)
    &\coloneqq
    \bm X
    +
    \mathcal{A}_{\theta_A}
    \!\left(\operatorname{LN}_A(\bm X)\right), \\
    \widehat{\mathcal{M}}_{\theta_M}(\bm X)
    &\coloneqq
    \bm X
    +
    \mathcal{M}_{\theta_M}
    \!\left(\operatorname{LN}_M(\bm X)\right),
\end{align*}
Then a standard Transformer block can be written as
\begin{align*}
    \mathcal{T}_{\theta}
    &=
    \widehat{\mathcal{M}}_{\theta_M}
    \circ
    \widehat{\mathcal{A}}_{\theta_A}.
\end{align*}
This decomposition lets us distinguish whether repeated computation is allocated to attention, to the MLP, or to the complete block.

\begin{definitionbox}[label=def:independent_sublayer_looping]
{\textbf{Joint and independent sublayer looping. }}
Let $R_A$ and $R_M$ denote the numbers of attention and MLP applications, respectively. We consider four designs:
\begin{enumerate}
    \item \emph{Joint Block Looping} repeats attention and the MLP together as a complete Transformer block. (Note that this is in principle similar to Per-Layer Looping):
    \begin{align*}
        \mathcal{T}_{\mathrm{joint}}^{(R)}
        &\coloneqq
        \left(
            \widehat{\mathcal{M}}_{\theta_M}
            \circ
            \widehat{\mathcal{A}}_{\theta_A}
        \right)^{\circ R}.
    \end{align*}

    \item \emph{Attention-Only Looping} repeats attention while applying the MLP once:
    \begin{align*}
        \mathcal{T}_{\mathrm{attn}}^{(R_A)}
        &\coloneqq
        \widehat{\mathcal{M}}_{\theta_M}
        \circ
        \widehat{\mathcal{A}}_{\theta_A}^{\circ R_A}.
    \end{align*}

    \item \emph{MLP-Only Looping} applies attention once and repeats the MLP:
    \begin{align*}
        \mathcal{T}_{\mathrm{mlp}}^{(R_M)}
        &\coloneqq
        \widehat{\mathcal{M}}_{\theta_M}^{\circ R_M}
        \circ
        \widehat{\mathcal{A}}_{\theta_A}.
    \end{align*}

    \item \emph{Independent Attention, MLP Looping} repeats the two sublayers separately:
    \begin{align*}
        \mathcal{T}_{\mathrm{ind}}^{(R_A,R_M)}
        &\coloneqq
        \widehat{\mathcal{M}}_{\theta_M}^{\circ R_M}
        \circ
        \widehat{\mathcal{A}}_{\theta_A}^{\circ R_A}.
    \end{align*}
\end{enumerate}
In every case, repeated applications of a sublayer reuse the same parameters.
\end{definitionbox}

These designs separate two aspects of looping. Joint looping repeatedly alternates token mixing and channel-wise transformation, whereas independent looping groups repeated applications of each operation. Attention-only and MLP-only looping further isolate whether additional computation is more useful for token mixing or for position-wise transformation. The constructions above determine \emph{what} computation is reused and \emph{how} repeated operations are ordered. A separate design choice is \emph{where} in the network this reuse is introduced. In particular, recent looped architectures often preserve distinct input and output layers while restricting recurrence to an intermediate portion of the network~\citep{prairie2026parcae}. We introduce this middle-looping construction next.

\paragraph{Middle-Stack Looping.}
The constructions above specify \emph{what} is repeated and \emph{how} repeated computations are ordered. A separate choice is \emph{where} recurrence is introduced. Instead of reusing every block in the network, one can preserve unique transformations at the beginning and end of the model and restrict recurrence to a contiguous middle region. This retains  input and output processing while concentrating repeated computation in the interior of the network. Within the looping region, the same ordering choices remain: the selected blocks can either be repeated individually or reused together as a complete stack.

\begin{definitionbox}[label=def:middle_looping]
{\textbf{Middle-Layer and Middle-Stack Looping.} }
Consider a depth-$B$ Transformer and choose indices $1\leq s\leq e\leq B$. We partition the model into a prefix, a recurrent middle region, and a suffix:
\begin{align*}
    \mathcal{F}_{\mathrm{pre}}
    &\coloneqq
    \mathcal{T}_{\theta_{s-1}}
    \circ\cdots\circ
    \mathcal{T}_{\theta_1}, \\
    \mathcal{F}_{\mathrm{mid}}
    &\coloneqq
    \mathcal{T}_{\theta_e}
    \circ\cdots\circ
    \mathcal{T}_{\theta_s}, \\
    \mathcal{F}_{\mathrm{post}}
    &\coloneqq
    \mathcal{T}_{\theta_B}
    \circ\cdots\circ
    \mathcal{T}_{\theta_{e+1}},
\end{align*}
where an empty prefix or suffix is interpreted as the identity map.

\begin{enumerate}
    \item \emph{Middle-Layer Looping} repeats each block in the selected middle region before advancing to the next block. If block $b$ is applied $R_b$ times, then
    \begin{align*}
        \bm Z_{s,0}
        &\coloneqq
        \mathcal{F}_{\mathrm{pre}}(\bm X), \\
        \bm Z_{b,0}
        &\coloneqq
        \bm Z_{b-1,R_{b-1}},
        \qquad
        \forall b\in\{s+1,\ldots,e\}, \\
        \bm Z_{b,r}
        &\coloneqq
        \mathcal{T}_{\theta_b}\!\left(\bm Z_{b,r-1}\right),
        \qquad
        \forall b\in\{s,\ldots,e\},\ \forall r\in[R_b], \\
        \mathcal{F}_{\mathrm{middle\text{-}layer}}(\bm X)
        &\coloneqq
        \mathcal{F}_{\mathrm{post}}\!\left(\bm Z_{e,R_e}\right).
    \end{align*}

    \item \emph{Middle-Stack Looping} treats the entire selected middle region as a single composite transformation and repeats that transformation $L$ times:
    \begin{align*}
        \bm Z^{(0)}
        &\coloneqq
        \mathcal{F}_{\mathrm{pre}}(\bm X), \\
        \bm Z^{(\ell)}
        &\coloneqq
        \mathcal{F}_{\mathrm{mid}}\!\left(\bm Z^{(\ell-1)}\right),
        \qquad
        \forall \ell\in[L], \\
        \mathcal{F}_{\mathrm{middle\text{-}stack}}(\bm X)
        &\coloneqq
        \mathcal{F}_{\mathrm{post}}\!\left(\bm Z^{(L)}\right).
    \end{align*}
\end{enumerate}
\end{definitionbox}

Middle-layer and Middle-Stack Looping therefore reuse the same region of the network but impose different computation orders. When every selected block is repeated the same number of times, $R_b=L$, Middle-Layer Looping applies all repetitions of one block before moving to the next, $\mathcal{T}_{\theta_e}^{\circ L}
    \circ\cdots\circ
    \mathcal{T}_{\theta_s}^{\circ L}$ whereas Middle-Stack Looping repeatedly traverses the entire recurrent region,
$\left(
        \mathcal{T}_{\theta_e}
        \circ\cdots\circ
        \mathcal{T}_{\theta_s}
    \right)^{\circ L}.$ In both cases, the prefix and suffix are evaluated only once and retain their own parameters. Middle looping therefore separates the question of \emph{where} additional computation is allocated from the question of \emph{how} that computation is ordered.

\paragraph{Why the form of looping matters.}
Looping is often described simply as a way to obtain more computation from a
fixed set of parameters. This description is correct, but incomplete. In
practice, the usefulness of weight reuse depends on \emph{which resource is
scarce}. Parameters, training FLOPs, memory movement, communication, and
sequential inference time impose different constraints, and different looping
constructions trade among these resources in different ways. Consequently,
two architectures that reuse the same number of parameters can behave very
differently as computational systems. The choice of what to loop, where to
loop it, and in what order to execute the repeated computation is therefore
not merely an architectural detail; it determines what kind of resource
efficiency the model is attempting to obtain.

Consider first the distinction between full-stack and middle looping.
Full-Stack Looping repeatedly evaluates the entire Transformer, and therefore
uses the additional compute budget everywhere in the network. If parameter
memory is the primary constraint while additional FLOPs are comparatively
cheap, this is a natural strategy: every extra loop provides another complete
pass through attention, MLP computation, and the hierarchy of learned
representations. In such a regime, there is little reason to leave part of the
network idle simply to save computation. The situation changes when training
FLOPs themselves become the limiting resource. Repeating the complete network
may then spend substantial compute on layers for which additional evaluation
provides relatively little marginal benefit. Middle looping offers a different
trade-off: the beginning and end of the network are evaluated once, while the
additional compute is concentrated in a smaller recurrent core. Thus, even
when full-stack and middle looping are motivated by the same goal of parameter
reuse, they answer different resource-allocation questions. The former asks
how much quality can be extracted from a fixed parameter budget when more
compute is available, while the latter asks whether that additional compute
can be spent more selectively.

Computation order introduces a different systems consideration. Suppose the
same collection of layers is evaluated the same number of times. Stack looping
cycles through the entire collection before returning to the first layer,
whereas Per-Layer Looping repeatedly evaluates one layer before moving on to
the next. Their arithmetic FLOPs can therefore be identical, yet their
execution patterns are different. This distinction can matter in regimes where
the cost of moving parameters is comparable to, or larger than, the cost of
performing arithmetic with parameters that are already resident on the
accelerator. A per-layer schedule repeatedly uses the same weights in
consecutive operations, potentially allowing more computation to be performed
while those weights remain locally available. Stack looping instead revisits a
layer only after the intervening layers have been executed. We do not assume
that one design is universally faster; the answer depends on the hardware,
batch size, and whether execution is compute- or memory-bound, but the two
designs expose different opportunities for parameter reuse at the systems
level even when their nominal FLOP counts are matched. The distinction between attention and MLP looping is similarly motivated by
the fact that Transformer components need not have comparable costs. In a
standard dense Transformer, repeating attention spends additional computation
on token-to-token interaction, whereas repeating the MLP spends it on
position-wise feature transformation. In other model families, however, the
systems consequences can be much larger. Consider a mixture-of-experts (MoE)
Transformer in which the MLP is replaced by routed experts distributed across
different accelerators or nodes. Repeating the MLP may then require not only
additional matrix multiplications, but also repeated token routing,
all-to-all communication, and expert-weight access. Attention-only looping
would increase effective computational depth without repeatedly incurring
those expert-related costs. Conversely, for very long sequences, attention
itself can become an increasingly expensive part of the model because its
compute and memory traffic depend on sequence length, making additional MLP
computation comparatively attractive. These examples are not claims about our
dense experiments as we do not evaluate MoE models here, but they illustrate
why deciding whether to loop attention, the MLP, or the complete block can
become an important systems choice rather than a cosmetic architectural
ablation. Finally, reuse depth determines how aggressively the model makes this
parameter--compute trade. Increasing the number of loops allows the same
parameters to support progressively greater effective depth, which is the
central appeal of looping. At the same time, every additional application
must still be executed. Training FLOPs therefore increase with loop count, and
under conventional looping the loops are also executed sequentially, leading to slower autoregressive generation during inference.
A model can consequently become more parameter-efficient while becoming less
attractive from the standpoint of training compute or autoregressive latency.
This tension becomes particularly important at large loop factors: the
question is not simply whether more loops continue to improve loss, but
whether their marginal improvement justifies the additional computation and
sequential depth.

Taken together, these examples show why there is no single notion of the
``best'' looping architecture independent of the resource regime. Full-stack
looping may be preferable when parameters are expensive and additional FLOPs
are available; middle looping may be preferable when compute must be allocated
more selectively; Per-Layer Looping changes the pattern with which shared
weights are reused by the hardware; and attention- versus MLP-level looping
can become important when the two components have different computational or
communication costs. The architectural choices introduced above therefore
correspond to different ways of allocating a fixed parameter budget across
computation, memory movement, communication, and sequential depth.

\paragraph{Our architectural ablations.} Motivated by these trade-offs, we systematically vary loop placement, computation order, sublayer choice, and reuse depth across model scales and training budgets. We compare these constructions both at matched data and as a function of executed FLOPs, since the two views answer different questions: the former measures how much quality can be extracted from a fixed parameter budget when additional computation is available, while the latter measures how effectively each construction uses a constrained compute budget. Individual points in this design space have precedents. Prior work repeats complete Transformer stacks \citep{saunshi2025reasoning}, while other approaches retain distinct outer layers and apply looping to a middle stack \citep{zeitoun2026hyperlooptransformers, prairie2026parcae}. We therefore do not treat any individual looping primitive as the contribution of this work. Rather, our goal is to compare these choices within a common dense, from-scratch setting and determine how their relative behavior changes with model scale, training budget, loop depth, and the resource under which they are compared.

\paragraph{Sequential inference.} The discussion above concerns \emph{how much} repeated computation should be performed and \emph{where} that computation should be allocated. It leaves open a separate issue that becomes especially important at inference time: in conventional looping, the additional effective depth is realized through a sequence of dependent computations. Since $\bm H^{(\ell)}=\mathcal{F}_{\theta}\!\left(\bm H^{(\ell-1)}\right)$, loop $\ell$ cannot be evaluated before loop $\ell-1$ has completed. Increasing the loop factor therefore increases not only executed computation, but also the amount of sequential computation on the inference critical path. This creates a tension that is not captured by training FLOPs alone. A looping construction may be attractive because it achieves better quality for a fixed parameter budget, or even because it provides a favorable quality--FLOP trade-off, yet still be unattractive for autoregressive decoding if obtaining that gain requires proportionally more sequential model evaluations for every generated token. In other words, reducing the \emph{amount} of repeated computation, for example by looping only a middle region, does not resolve the fact that the remaining loop iterations are still ordered sequentially. Parameter efficiency, compute efficiency, and inference latency are therefore distinct objectives. This observation motivates the next question we study: whether the benefits of repeated computation can be retained while relaxing the serial dependency imposed by conventional looping.

\subsection{Low-Rank Adaptation and Relaxed Weight Sharing}
\label{sec:background_lora}

A looped Transformer repeats the same layer, or group of layers, \(L\) times.
Normally, every loop uses the same attention and MLP weights. Sharing the
weights in this way keeps the model small, since adding loops does not require
storing new Transformer layers. However, the loops do not all receive the same
kind of input. The first loop processes the initial representation, while later
loops process representations that have already passed through the shared
layers one or more times. It may therefore be useful for different loops to
learn slightly different transformations.

One option would be to give every loop its own attention and MLP weights. This
would provide complete specialization, but it would also make the loops
equivalent to separately parameterized Transformer layers. The number of
stored parameters would then grow with the number of loops, removing much of
the motivation for weight reuse. We instead use Low-Rank Adaptation (LoRA)
\citep{hu2021lora} to place a small loop-specific correction around a larger
shared weight matrix.

\begin{definitionbox}[label=def:lora]
{\textbf{Loop-specific LoRA.} }
Consider a weight matrix
\(\bm W\in\mathbb{R}^{d_{\mathrm{out}}\times d_{\mathrm{in}}}\)
that is shared across \(L\) loops. In loop
\(\ell\in\{1,\ldots,L\}\), we use
\[\bm W^{(\ell)}
=\bm W+\frac{\alpha}{r}\bm B^{(\ell)}\bm A^{(\ell)}\],
where
\(\bm A^{(\ell)}\in\mathbb{R}^{r\times d_{\mathrm{in}}}\) and
\(\bm B^{(\ell)}\in\mathbb{R}^{d_{\mathrm{out}}\times r}\) are learned
separately for that loop. The rank \(r\) controls the size of the correction,
and \(\alpha/r\) controls its scale. Each loop-specific correction adds
\(r(d_{\mathrm{in}}+d_{\mathrm{out}})\) parameters, rather than the
\(d_{\mathrm{in}}d_{\mathrm{out}}\) parameters required to store another
complete version of \(\bm W\).
\end{definitionbox}

For one adapted weight matrix, using the same weights in every loop requires
\(d_{\mathrm{in}}d_{\mathrm{out}}\) parameters. Loop-specific LoRA requires
\(d_{\mathrm{in}}d_{\mathrm{out}}
+Lr(d_{\mathrm{in}}+d_{\mathrm{out}})\) parameters, while giving every loop
a different weight matrix requires
\(Ld_{\mathrm{in}}d_{\mathrm{out}}\) parameters. Loop-specific LoRA therefore
lies between these two choices: most parameters remain shared, but each loop
has a small amount of capacity with which to specialize.

Prior work explores a similar middle ground in Relaxed Recursive Transformers
(RRTs) \citep{bae2025relaxed}. RRT begins with a pretrained Transformer whose
layers have different weights and converts it into a smaller looped model by
replacing groups of those layers with repeated uses of shared weights. Because
this conversion removes differences among the original layers, RRT gives each
loop a low-rank adapter initialized from a truncated SVD of the difference
between the original weight and its shared replacement. The converted model is
then trained further. Our setting differs because we do not begin with a
larger pretrained model or attempt to preserve its layer-specific weights.
Instead, the Transformer is looped from the beginning, and the shared weights
and loop-specific adapters are trained jointly from initialization. We use this
setting to ask whether loop-specific specialization is useful during looped
pretraining itself and whether its value increases as the same weights are
reused across more loops.

\subsection{Parallel Loop Transformers}
\label{sec:background_plt}

In conventional looping, loop \(\ell\) for token \(n\) receives the output of
loop \(\ell-1\) for that same token. The model must therefore finish loop
\(1\) before starting loop \(2\), finish loop \(2\) before starting loop
\(3\), and continue in this way through all \(L\) loops. Producing one token
consequently requires \(L\) consecutive evaluations of the shared
Transformer. Parallel Loop Transformers (PLTs) \citep{wu2025parallel} remove this dependency.
For loop \(\ell\), PLT combines the current token representation with the
output of loop \(\ell-1\) from the preceding token \(n-1\). When decoding
token \(n\), the model has already processed token \(n-1\), so all of these
preceding-token outputs are available. The inputs to the \(L\) loops can
therefore be constructed at the same time and evaluated together, rather than
waiting for one loop to finish before beginning the next. PLT still performs
\(L\) evaluations and therefore does not reduce total FLOPs, but parallel
hardware can execute them concurrently, reducing the time required to produce
each token. Formally, let $\bm t=(t_1,\ldots,t_N)$ denote a token sequence. At position $n\in[N]$, let $\bm e_n\in\mathbb{R}^{D}$ denote the token embedding, $\bm b_n^{(\ell)}\in\mathbb{R}^{D}$ the representation supplied to loop $\ell$, and $\bm h_n^{(\ell)}\in\mathbb{R}^{D}$ the corresponding output state. We use the boundary condition $\bm h_0^{(\ell)}=\bm 0$ and initialize the first loop with the token embedding, $\bm b_n^{(1)}=\bm e_n$. The key departure from conventional looping appears in every subsequent loop.

\begin{definitionbox}[label=def:app-plt-transition]
{\textbf{Cross-loop PLT transition (CLP). }}
For every position $n$ and loop $\ell\geq 2$, the input to the next loop combines the current token embedding with the previous token's state from the preceding loop:
\begin{align*}
    \bm b_n^{(1)}
    &\coloneqq
    \bm e_n, \\
    \bm b_n^{(\ell)}
    &\coloneqq
    \bm e_n
    +
    \bm h_{n-1}^{(\ell-1)},
    \qquad
    \forall n\in[N],\
    \forall \ell\in\{2,\ldots,L\}, \\
    \bm h_n^{(\ell)}
    &\coloneqq
    \mathcal{F}_{\theta}^{(\ell)}
    \!\left(\bm b_n^{(\ell)}\right),
    \qquad
    \forall n\in[N],\
    \forall \ell\in[L].
\end{align*}
Here, $\mathcal{F}_{\theta}^{(\ell)}$ denotes the $\ell$-th application of the same shared Transformer weights. The superscript therefore indexes the computation stage rather than a distinct parameter set. The essential distinction from conventional looping is the recurrent source: PLT uses $\bm h_{n-1}^{(\ell-1)}$, whereas ordinary recurrence would feed the same-token state $\bm h_n^{(\ell-1)}$ into the next loop.
\end{definitionbox}

The goal of PLT is not to reduce the total computation introduced by looping.
Training still evaluates the loops sequentially, and inference still applies
the shared Transformer \(L\) times. Instead, PLT targets the way this
computation is executed during autoregressive decoding. Conventional looping
requires \(L\) consecutive evaluations for every generated token. These
single-token evaluations expose little parallel work and repeatedly access the
same shared weights. PLT reorganizes the dependencies so that the \(L\) loop
inputs become available together and can be processed as a batch. This can
reduce decoding latency and increase arithmetic intensity by performing more
computation per weight load. We first describe the training algorithm below (as seen in \cite{wu2025parallel}).

\paragraph{PLT Training.}
Let $\bm E\coloneqq[\bm e_1;\ldots;\bm e_N]$, $\bm B^{(\ell)}\coloneqq[\bm b_1^{(\ell)};\ldots;\bm b_N^{(\ell)}]$, and $\bm H^{(\ell)}\coloneqq[\bm h_1^{(\ell)};\ldots;\bm h_N^{(\ell)}]$. We define the one-position causal shift by
$\operatorname{Shift}([\bm h_1;\ldots;\bm h_N])\coloneqq[\bm 0;\bm h_1;\ldots;\bm h_{N-1}]$.
The first loop receives the token embeddings directly, while every subsequent loop receives the embeddings together with the shifted state of the preceding loop:
\begin{align*}
    \bm B^{(1)}
    &\coloneqq
    \bm E, \\
    \bm B^{(\ell)}
    &\coloneqq
    \bm E
    +
    \operatorname{Shift}\!\left(\bm H^{(\ell-1)}\right),
    \qquad
    \forall \ell\in\{2,\ldots,L\}.
\end{align*}
Note that at the token level we use the boundary convention $\bm h_0^{(\ell)}\coloneqq\bm 0$ for every $\ell\in[L]$, so that
$\bm b_1^{(\ell)}=\bm e_1$ at the left boundary, while
$\bm b_n^{(\ell)}=\bm e_n+\bm h_{n-1}^{(\ell-1)}$ for $n>1$ and $\ell>1$. To reiterate, this shift changes the dependency structure in an important way. Conventional looping connects successive loop iterations at the same token position, so the next loop cannot begin until the preceding loop for that token has completed. PLT instead makes the input to loop $\ell$ at the current position depend on a state that was produced at the preceding position. During training, the full sequence is available, so the model is evaluated one loop at a time while retaining parallelism over token positions. Below we describe the training algorithm for PLT.

\begin{algbox}[label=alg:plt_training]{Training a Parallel Loop Transformer}
\begin{algorithmic}[1]
\Require tokens $\bm t=(t_1,\ldots,t_N)$, loop count $L$, shared Transformer $\mathcal{F}_{\theta}$, prediction head $g$
\State $\bm E \gets \operatorname{Embed}(\bm t)$
\State $\bm B^{(1)} \gets \bm E$
\State $\bm H^{(1)} \gets \mathcal{F}_{\theta}(\bm B^{(1)})$
\For{$\ell=2,\ldots,L$}
    \State $\bm B^{(\ell)}
           \gets
           \bm E+\operatorname{Shift}\!\left(\bm H^{(\ell-1)}\right)$
    \State $\bm H^{(\ell)}
           \gets
           \mathcal{F}_{\theta}(\bm B^{(\ell)})$
\EndFor
\State $\bm Z\gets g(\bm H^{(L)})$
\State \Return $-\sum_{n=1}^{N-1}\log\!\left(\operatorname{softmax}(\bm Z_n)_{t_{n+1}}\right)$
\end{algorithmic}
\end{algbox}

On the other hand, the cost of looping appears differently during training and inference. During
training, the entire token sequence is available, so all token positions can
be processed in parallel within each loop. The \(L\) loops must still be
evaluated one after another, and we accept this additional training cost.
During autoregressive decoding, however, tokens are generated one at a time.
If the \(L\) loops are also evaluated consecutively, every additional loop
directly increases the time required to generate each token.

CLP is designed to improve this inference-time behavior rather than reduce
training compute. Before token \(n\) is processed, token \(n-1\) has already
produced and cached the states
\(\{\bm h_{n-1}^{(\ell)}\}_{\ell=1}^{L-1}\). The \(L\) inputs for token \(n\)
can therefore be constructed together, stacked along the loop dimension, and
passed through the shared Transformer as a single batch. The model performs
the same \(L\) loop computations, but parallel hardware can execute them
together rather than consecutively. Because every loop uses the same
Transformer weights, this batching can also improve arithmetic intensity by
applying each loaded weight to multiple loop inputs. Note that, at the beginning of a sequence, some preceding-token states are unavailable
and are replaced by the zero boundary state. After the prompt or initial
tokens populate the cache, decoding reaches a steady state in which every loop
receives a state computed at the preceding token position. The algorithm below
describes one decoding step in this steady-state regime.

\begin{algbox}[label=alg:plt_inference]{Steady-state autoregressive decoding with PLT}
\begin{algorithmic}[1]
\Require current token $t_n$, loop count $L$, cached predecessor states $\{\bm h_{n-1}^{(\ell)}\}_{\ell=1}^{L-1}$, shared Transformer $\mathcal{F}_{\theta}$, prediction head $g$
\State $\bm e_n\gets\operatorname{Embed}(t_n)$
\State $\bm b_n^{(1)}\gets\bm e_n$
\For{$\ell=2,\ldots,L$}
    \State $\bm b_n^{(\ell)}
           \gets
           \bm e_n+\bm h_{n-1}^{(\ell-1)}$
\EndFor
\State $\bm B_n
       \gets
       \operatorname{Batch}\!\left(
       \bm b_n^{(1)},\ldots,\bm b_n^{(L)}
       \right)$
\State $\bm H_n
       \gets
       \mathcal{F}_{\theta}(\bm B_n)$
       \Comment{loop stages evaluated along the batch dimension}
\For{$\ell=1,\ldots,L$}
    \State $\bm h_n^{(\ell)}
           \gets
           [\bm H_n]_{\ell}$
\EndFor
\State $p(t_{n+1}\mid t_{\leq n})
       \gets
       \operatorname{softmax}\!\left(
       g(\bm h_n^{(L)})
       \right)$
\State sample or select $t_{n+1}$ and cache $\{\bm h_n^{(\ell)}\}_{\ell=1}^{L-1}$
\end{algorithmic}
\end{algbox}

Here, \(\operatorname{Batch}(\cdot)\) stacks the \(L\) loop inputs along an
independent batch dimension rather than concatenating them along the token
dimension. The shared Transformer \(\mathcal{F}_{\theta}\) processes this
batch in one invocation, and \([\bm H_n]_{\ell}\) denotes the output
corresponding to loop \(\ell\). The algorithm therefore maps the loop
dimension onto the batch dimension during steady-state decoding.

\paragraph{Shared global and loop-local context.}
Evaluating the \(L\) loops together reduces the delay caused by running them
one after another, but it does not reduce the memory needed for attention. If
each loop stores its own sequence-wide KV cache, then a model with sequence
length \(N\), hidden dimension \(D\), and \(L\) loops requires
\(\mathcal{O}(LND)\) cache memory. PLT avoids this cost by storing one global
cache, produced by the first loop, and sharing it with all later loops. This
gives every loop access to the full preceding sequence without storing another
complete set of keys and values.

Using only the shared cache, however, would prevent later loops from retaining
attention states produced by their own computations. PLT therefore gives each
later loop an additional private cache containing only its most recent \(w\)
positions. The shared cache provides global context from the first loop, while
the private window provides recent context specific to the current loop.
Together, these caches require
\(\mathcal{O}(ND+(L-1)wD)\) memory.

PLT combines the two sources of context using
\emph{gated sliding-window attention} (G-SWA). For loop \(\ell\geq 2\), let
\(\bm Q^{(\ell)}\) denote its queries,
\((\bm K_{\mathrm{share}},\bm V_{\mathrm{share}})\) the global KV cache from
the first loop, and
\((\bm K_{\mathrm{local}}^{(\ell)},
\bm V_{\mathrm{local}}^{(\ell)})\) its private KV cache over the most recent
\(w\) positions. G-SWA computes attention over the shared and local caches
separately and uses a learned head-wise gate to combine the two outputs. Below we describe this concretely.

\begin{definitionbox}[label=def:plt_representation_enhancement]
{\textbf{Gated Sliding Window Attention (G-SWA). }}
For every loop $\ell\geq 2$, define
\begin{align*}
    \bm Y_{\mathrm{global}}^{(\ell)}
    &\coloneqq
    \operatorname{Attn}\!\left(
        \bm Q^{(\ell)},
        \bm K_{\mathrm{share}},
        \bm V_{\mathrm{share}}
    \right), \\
    \bm Y_{\mathrm{local}}^{(\ell)}
    &\coloneqq
    \operatorname{SWA}\!\left(
        \bm Q^{(\ell)},
        \bm K_{\mathrm{local}}^{(\ell)},
        \bm V_{\mathrm{local}}^{(\ell)};
        w
    \right), \\
    \bm g^{(\ell)}
    &\coloneqq
    \sigma\!\left(
        f_{\mathrm{gate}}\!\left(\bm Q^{(\ell)}\right)
    \right), \\
    \widetilde{\bm Y}^{(\ell)}
    &\coloneqq
    \bm g^{(\ell)}
    \odot
    \bm Y_{\mathrm{local}}^{(\ell)}
    +
    \left(\bm 1-\bm g^{(\ell)}\right)
    \odot
    \bm Y_{\mathrm{global}}^{(\ell)}.
\end{align*}
Here, $f_{\mathrm{gate}}$ produces one scalar gate for each attention head,
which is broadcast across the corresponding head representation. Each head
can therefore determine how much of its output should come from loop-local
recent context versus the global context shared from the first loop.
\end{definitionbox}

The shared first-loop cache requires $\mathcal{O}(ND)$ memory, while the
private sliding-window caches contribute
$\mathcal{O}((L-1)wD)$. The resulting KV-cache  memory complexity is therefore
$\mathcal{O}(ND+(L-1)wD)$ rather than $\mathcal{O}(LND)$. When
$w\ll N$, G-SWA allows later loops to retain their own recent attention state
without requiring a separate sequence-length cache for every recurrent stage. The two main components of PLT therefore address different costs of recurrent
computation. The cross-loop transition in \autoref{def:app-plt-transition}
changes the dependency structure, allowing recurrent stages to be pipelined
across token positions during autoregressive decoding. KV sharing and G-SWA
address the separate memory cost of maintaining attention state for those
stages. The former supplies a common global context, while the latter restores
a small amount of loop-specific context that would otherwise disappear under
complete KV sharing.

\paragraph{Relation to our study.}
Our study examines whether the change in information flow introduced by CLP
affects model quality independently of its inference benefits. In the original
PLT transition, each later loop directly adds the current-token embedding
\(\bm e_n\) to the shifted state
\(\bm h_{n-1}^{(\ell-1)}\). These representations play different roles and may
have different scales, so how they are combined can determine whether parallel
looping preserves the benefits of ordinary looping. We isolate this effect by
reconstructing PLT component by component, separating the shifted transition
from KV sharing and G-SWA. We then vary access to the current-token
representation, normalization, and the mechanism used to combine it with the
shifted state. These experiments distinguish limitations of CLP itself from
limitations of the transition and cache mechanisms used to implement it.

\clearpage

\section{Model Sizes and Training Budgets}
\label{app:model-configurations-and-budgets}

We train six model sizes, ranging from 70M to 1B, and vary the amount of
training data from 10 to 160 tokens per parameter (TPP). This section specifies
the architecture used at each model size, explains how the model-size labels
are assigned, and describes how TPP determines the training-token budget.

\begin{table*}[ht]
  \centering
  \small
  \setlength{\tabcolsep}{6pt}

  \caption{\textbf{Model configurations used in our experiments.}
  Transformer layers give the number of Transformer blocks. Model width is
  the dimension of the representation passed between layers, and MLP width is
  the intermediate dimension of the SwiGLU MLP. Q/KV gives the numbers of
  query heads and key--value heads, respectively. The first parameter column
  excludes the input embedding and is used to assign the model label and set
  the training-token budget. The final column includes the input embedding
  and gives the total number of trainable parameters. Parameter counts are
  reported in millions. All models use a vocabulary of 128,256 tokens, a
  context length of 8,192, and an attention-head dimension of 128.}
  \label{tab:model-ladder}

  \vspace{0.75em}

  \begin{tabular}{@{}ccccccc@{}}
    \toprule
    & \multicolumn{4}{c}{Architecture}
    & \multicolumn{2}{c}{Parameter count (M)} \\
    \cmidrule(lr){2-5}
    \cmidrule(l){6-7}
    \shortstack{Model\\label}
      & \shortstack{Transformer\\layers}
      & \shortstack{Model\\width}
      & \shortstack{MLP\\width}
      & \shortstack{Attention heads\\(Q/KV)}
      & \shortstack{Excluding input\\embedding}
      & \shortstack{Including input\\embedding} \\
    \midrule
    70M  & 4  & 512   & 1,536 & 4/1  & 77.726  & 143.397 \\
    100M & 6  & 768   & 2,048 & 6/1  & 135.070 & 233.580 \\
    200M & 8  & 1,024 & 2,816 & 8/1  & 219.415 & 350.766 \\
    300M & 10 & 1,280 & 3,584 & 10/1 & 337.838 & 502.033 \\
    500M & 12 & 1,536 & 4,096 & 12/2 & 489.554 & 686.594 \\
    1B   & 16 & 2,048 & 5,632 & 16/2 & 967.311 & 1,230.047 \\
    \bottomrule
  \end{tabular}
\end{table*}

\paragraph{Model-size labels.}
We assign each model the label closest to its number of parameters excluding
the input embedding, rounding up or down to the nearest scale in our model
ladder. For example, the model with 489.554M such parameters is labeled 500M,
while the model with 967.311M is labeled 1B. These labels are therefore
convenient rounded names rather than exact parameter counts. The total parameter count is larger because it also includes the input
embedding. This difference is largest for the smaller models. For example,
the model labeled 70M has 77.726M parameters when the input embedding is
excluded and 143.397M parameters when it is included. We retain the same
model-size label for looped variants so that it continues to identify the
underlying model dimensions, even when weight sharing changes the number of
parameters that must be stored.

\paragraph{Training-token budgets.}
TPP specifies the number of training tokens allocated per parameter,
excluding the input embedding. We use TPP values of
\(10\), \(20\), \(40\), \(80\), and \(160\). The target number of training
tokens is \(D=\mathrm{TPP}\times P_{\mathrm{non\text{-}embed}}\), where
\(P_{\mathrm{non\text{-}embed}}\) is the parameter count in the
second-to-last column of \autoref{tab:model-ladder}. For example, the 100M
model has 135.070M non-embedding parameters and is trained on approximately
5.403 billion tokens at TPP 40. Across all model sizes and TPP values, the
training budgets range from 0.777 billion to 154.770 billion tokens. For looped models, parameters reused across multiple loops are counted once,
and any loop-specific parameters are also included. Adding loops also increases training FLOPs even when it does not
    increase the parameter count or token budget. We therefore report training
    tokens and training FLOPs separately when comparing models. We note that in \autoref{tab:model-ladder}, the non-embedding parameter count used to determine the token budget omits the RMSNorm parameters. These account for less than 0.008\% of every model
and do not meaningfully change either the reported model size or its token
budget.

\paragraph{Models, training, data, and evaluation.}
We train models with 70M, 100M, 200M, 300M, 500M, and 1B parameters,
containing 4, 6, 8, 10, 12, and 16 stored transformer layers, respectively.
All models use grouped-query attention, SwiGLU MLPs, RoPE with
$\theta=10{,}000$, document-causal attention, a context length of 8,192, and
an attention head dimension of 128. We use a Llama-3-compatible
\texttt{tiktoken} tokenizer with a vocabulary of 128,256 and untied input and
output embeddings. We train in bfloat16 with AdamW
($\beta_1=0.9$, $\beta_2=0.95$, $\epsilon=10^{-8}$), gradient clipping at
$1.0$, and a peak learning rate of $3\times10^{-4}$. The learning rate is
warmed up over the first $10\%$ of training and then cosine-decayed. Training
budgets range from 10 to 160 tokens per parameter (TPP), and all models are
trained with the same random seeds. Additional model configurations and
training budgets are given in
\autoref{app:model-configurations-and-budgets}. All models use the same
pretraining mixture of natural language, code, mathematical, and scientific
content, with the data sources, sampling distribution, preprocessing, and
tokenizer held fixed across architectures. We use a fixed subset of the DCLM Baseline 1.0 validation set
\citep{li2024datacomp}, comprising approximately 41 million scored tokens, as
our primary language-modeling evaluation corpus. We use DCLM as our primary
metric because it is the largest corpus in our evaluation suite and provides
the broadest estimate of language-modeling quality. To test whether the
observed trends generalize across domains, we additionally evaluate on the
Paloma benchmark \citep{magnusson2024paloma} versions of C4, WikiText-103,
Dolma v1.5, RedPajama, and Dolma Programming Languages
\citep{raffel2020exploring,merity2016pointer,soldaini2024dolma,weber2024redpajama}.
We also evaluate likelihood on the mathematical and code datasets GSM8K
\citep{cobbe2021training}, MATH \citep{hendrycks2021measuring}, HumanEval
\citep{chen2021evaluating}, and MBPP \citep{austin2021programsynthesislargelanguage}. For these
evaluations, we teacher-force the reference sequence and report bits per byte
(BPB; lower is better). We separately evaluate downstream multiple-choice performance on ARC-Easy and
ARC-Challenge \citep{clark2018think}, HellaSwag
\citep{zellers2019hellaswag}, PIQA \citep{bisk2019piqareasoningphysicalcommonsense}, WinoGrande
\citep{sakaguchi2019winograndeadversarialwinogradschema}, OpenBookQA \citep{mihaylov2018can},
CommonsenseQA \citep{talmor2019commonsenseqaquestionansweringchallenge}, and
SocialIQA \citep{sap2019socialiqacommonsensereasoningsocial}. For each
question, we score every candidate answer using its average negative
log-likelihood per tokenizer token and predict the candidate with the lowest
score. We report NLL-ranked accuracy as well as the normalized NLL assigned to
the correct answer. The headline downstream result is the unweighted mean of
NLL-ranked accuracy across these eight tasks. We report MMLU
\citep{hendrycks2021measuringmassivemultitasklanguage} and MMLU-Pro
\citep{wang2024mmlu} separately as knowledge diagnostics. These
multiple-choice evaluations measure discrimination among the provided answer
choices rather than freely generated answers.

\clearpage

\section{Details Behind the Headline Figure}
\label{app:headline-details}

The headline figure combines two complementary evaluations of looping. The
relative-efficiency panels measure how much training data or total training
compute a candidate model requires to attain a given DCLM BPB,
relative to a fitted Vanilla (no looping) baseline. Here, we provide the raw
scaling curves, fitting procedure, and task-level results underlying these
comparisons.

\subsection{Relative Data and Compute Efficiency}
\label{app:relative-efficiency}

\begin{figure*}[ht]
  \centering
  \includegraphics[width=\textwidth]{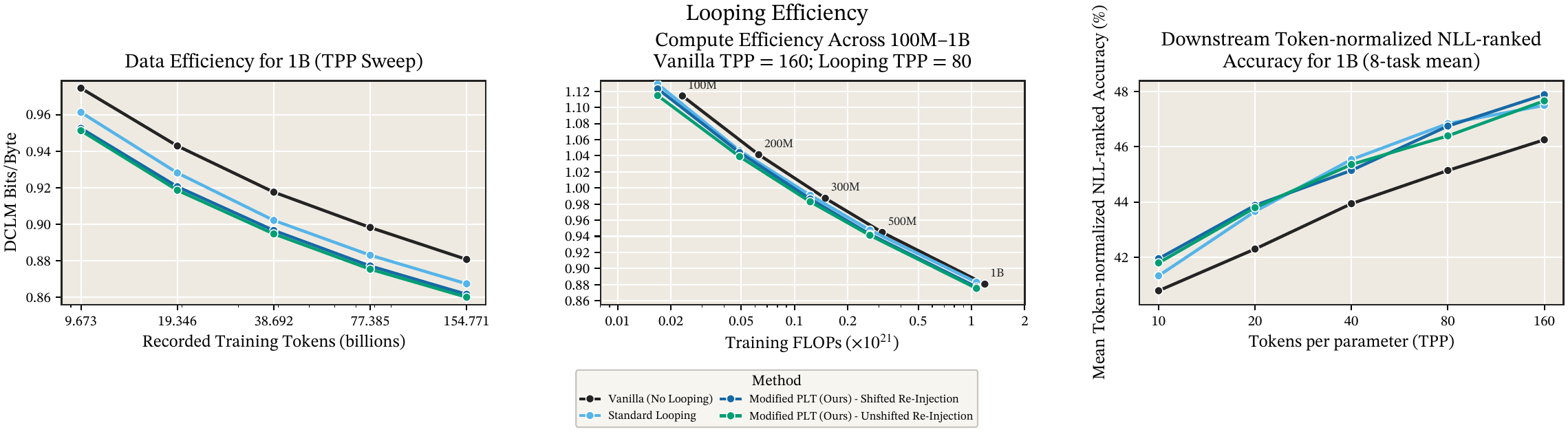}
  \caption{\textbf{Raw scaling curves underlying the relative-efficiency
  analysis.}
  We plot observed DCLM validation BPB together with unconstrained Vanilla
  scaling-law fits. \emph{Left:} BPB as a function of actual training tokens
  for fixed 1B models across training-token budgets. \emph{Middle:} BPB as a
  function of total training FLOPs across model families from 100M to 1B.
  Vanilla models use TPP \(=160\), whereas looping models use two loops and
  TPP \(=80\). Points show observed runs, connecting lines show empirical
  trajectories, and dashed black curves show fitted Vanilla scaling laws.
  Lower BPB is better.}
  \label{fig:headline-raw}
\end{figure*}

Looping increases the computation applied to each token but may reduce the
  amount of training needed to reach a given quality. We measure
  \emph{data efficiency} by the number of training tokens required by a model to reach a
  target BPB, and \emph{FLOP efficiency} by the total training FLOPs required by a model
  to reach the same BPB. Following \citet{microsoft2025mai}, we fit two curves to the BPB achieved by
  \emph{Vanilla (No Looping)} models: one as a function of training tokens and
  one as a function of total training FLOPs. Both use
  \(L(x)=Ax^{-\alpha}+E\), with \(A\), \(\alpha\), and \(E\) estimated by
  unweighted, unconstrained nonlinear least squares minimizing
  \(\sum_i[L_i-(Ax_i^{-\alpha}+E)]^2\). Only non-looping models are used to fit
  these curves. For a looped model achieving BPB \(L'\) at cost \(x'\), we estimate the cost
  that a non-looping model would require to achieve the same BPB as
  \(f^{-1}(L')=[A/(L'-E)]^{1/\alpha}\). We define the relative efficiency gain
  as \(EG_x=f^{-1}(L')/x'\), where \(EG_x>1\) means that the looped model uses
  fewer tokens or FLOPs than the fitted non-looping baseline. For data efficiency, we fit 1B non-looping models across training-token
  budgets. The looping models reach a given BPB using fewer tokens, with the
  modified PLT variants providing the largest relative gains. For FLOP
  efficiency, we fit non-looping models from 100M to 1B using their total
  training FLOPs. In the evaluated over-trained regime, two-loop models trained
  at TPP 80 reach a given BPB using fewer FLOPs than non-looping models trained
  at TPP 160. The key conclusion is that looping is not only parameter-efficient. In the
  evaluated over-trained regime, its additional computation per token reduces
  the required training data enough to lower total training FLOPs, making
 looping both data- and FLOP-efficient. This benefit is not uniform,
  however, and weakens as the number of loops increases.

 \subsection{Task-Level Results Behind the Headline Comparison}
  \label{app:plt-downstream}

  The right panel of \autoref{fig:headline-looping-efficiency} reports the
  unweighted mean across eight reasoning and question-answering tasks. Here, we
  provide task-level results and report MMLU and MMLU-Pro separately to evaluate
  performance across a broad range of academic subjects. We compare four 1B
  model families at \(\mathrm{TPP}\in\{10,20,40,80,160\}\):
  \emph{Vanilla} (\(L=1\)), \emph{Full-Stack Looping} (\(L=2\), labeled
  Standard Looping in the headline figure), \emph{Normalized Shifted PLT}
  (\(L=2\)), and \emph{Normalized Unshifted PLT} (\(L=2\)). Both PLT variants
  use the rank-\(8\) linear gate.

\begin{figure*}[t]
  \centering
  \includegraphics[width=\textwidth]{
    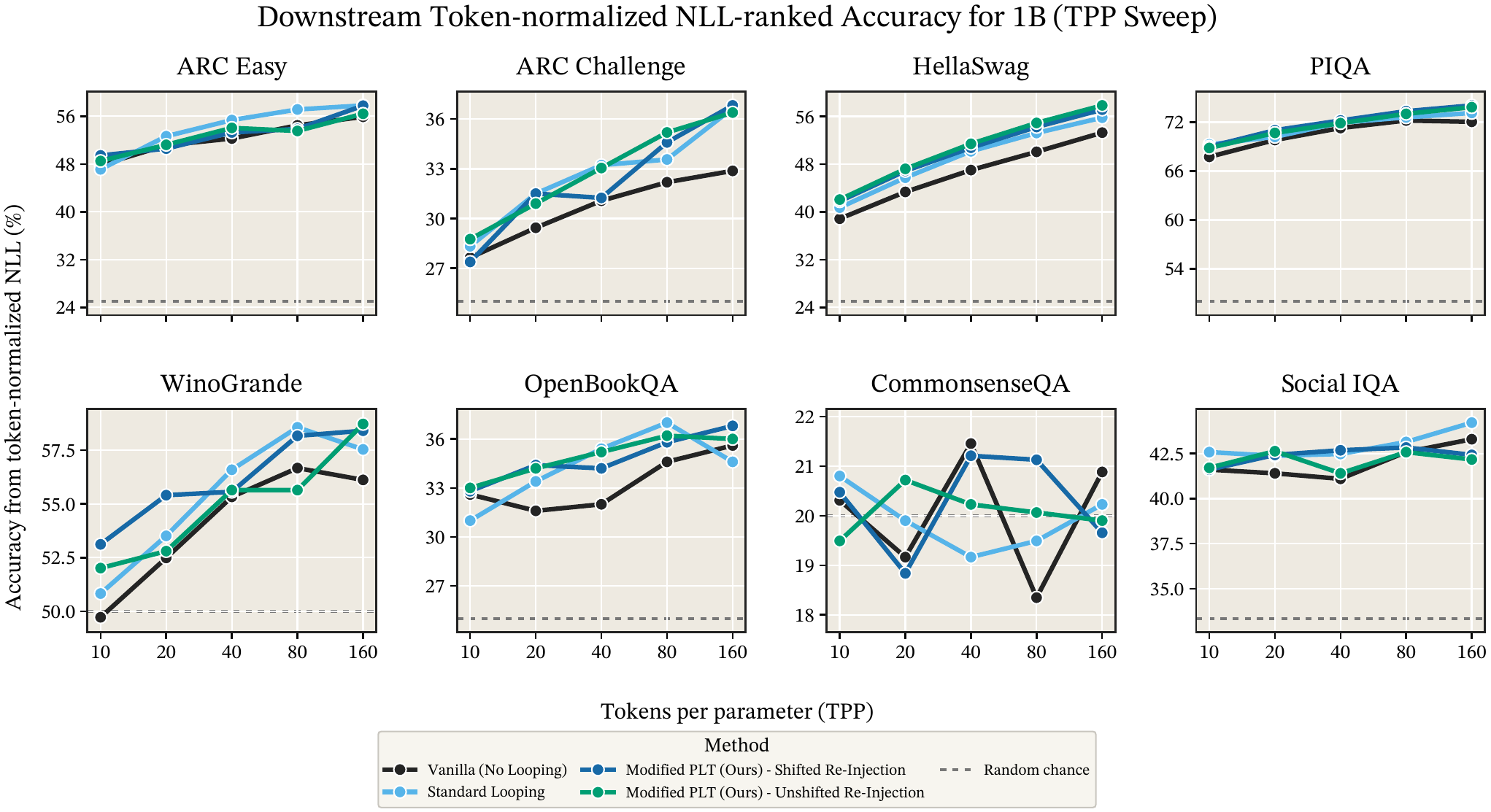}
  \caption{\textbf{Task-level NLL-ranked accuracy for the 1B models.}
  Results are shown for ARC Easy, ARC Challenge, HellaSwag, PIQA, WinoGrande,
  OpenBookQA, CommonsenseQA, and Social IQA. For each question, the model
  predicts the candidate with the lowest token-normalized NLL. Dashed
  horizontal lines denote random-choice accuracy. Lines connect measured
  checkpoints and are not fitted scaling laws. Higher is better.}
  \label{fig:plt-downstream-accuracy}
\end{figure*}

\begin{figure*}[ht]
  \centering
  \includegraphics[width=\textwidth]{
    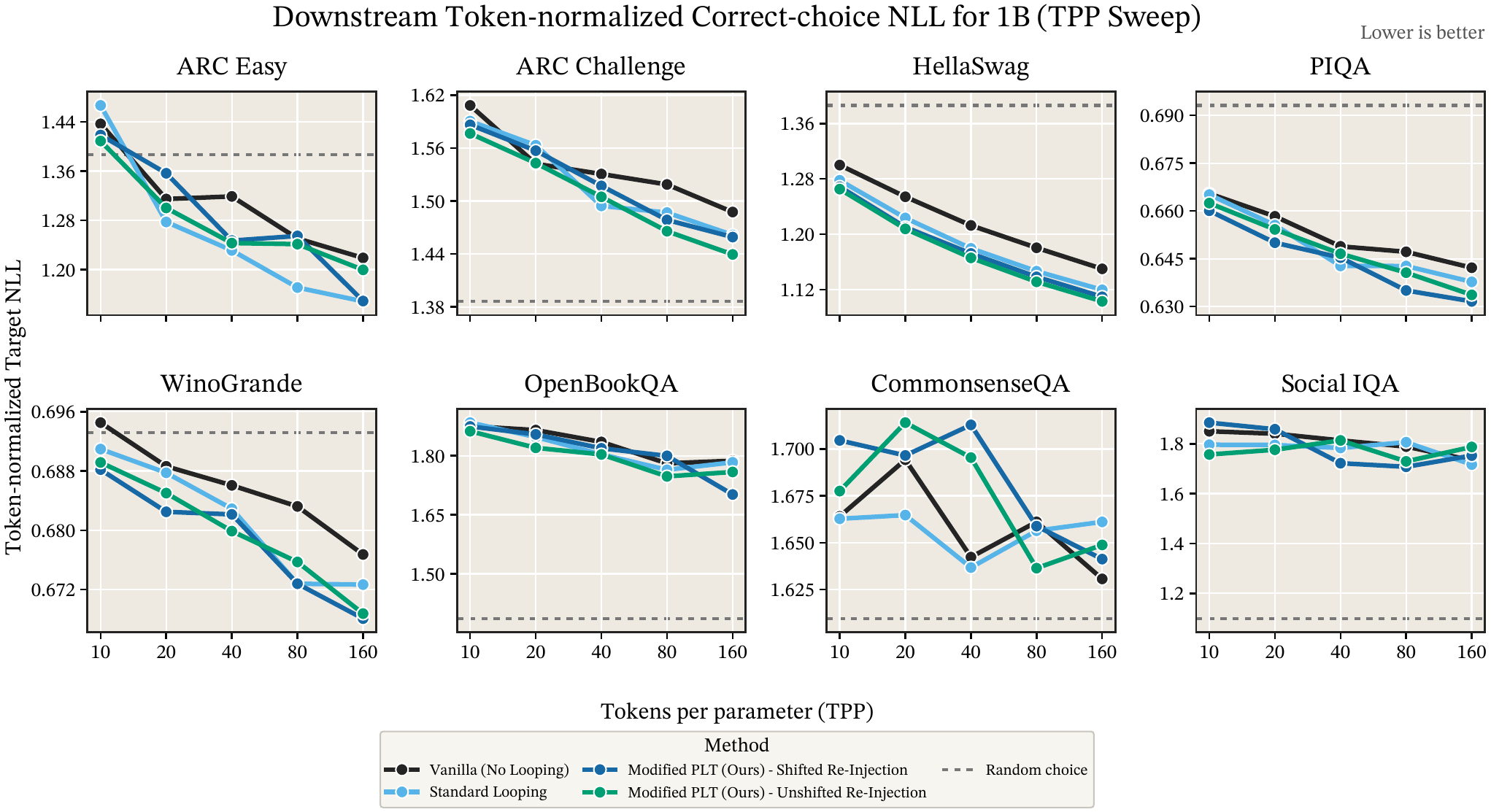}
  \caption{\textbf{Candidate-normalized correct-answer NLL for the same eight
  tasks.}
  Candidate scores are computed from token-normalized NLL and normalized
  across the available answers. This continuous metric measures how strongly
  the model prefers the correct answer relative to the alternatives, including
  changes that do not alter the top-ranked answer. Dashed lines show the
  random-choice value \(\log C\) for a task with \(C\) candidates. Because
  choice counts and prompt formats differ, values should be compared within,
  rather than across, datasets. Lower is better.}
  \label{fig:plt-downstream-nll}
\end{figure*}

\begin{figure*}[ht]
  \centering
  \includegraphics[width=\textwidth]{
    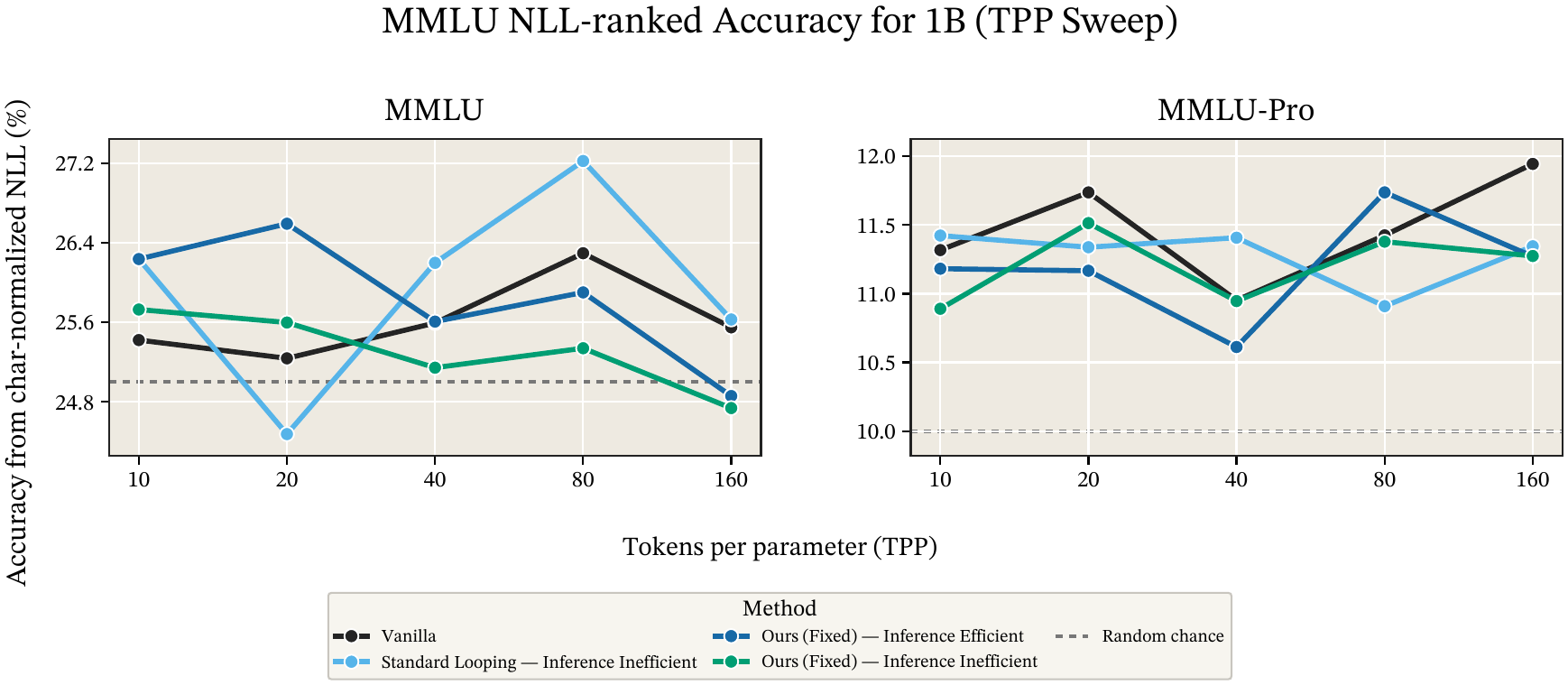}

  \vspace{0.5em}

  \includegraphics[width=\textwidth]{
    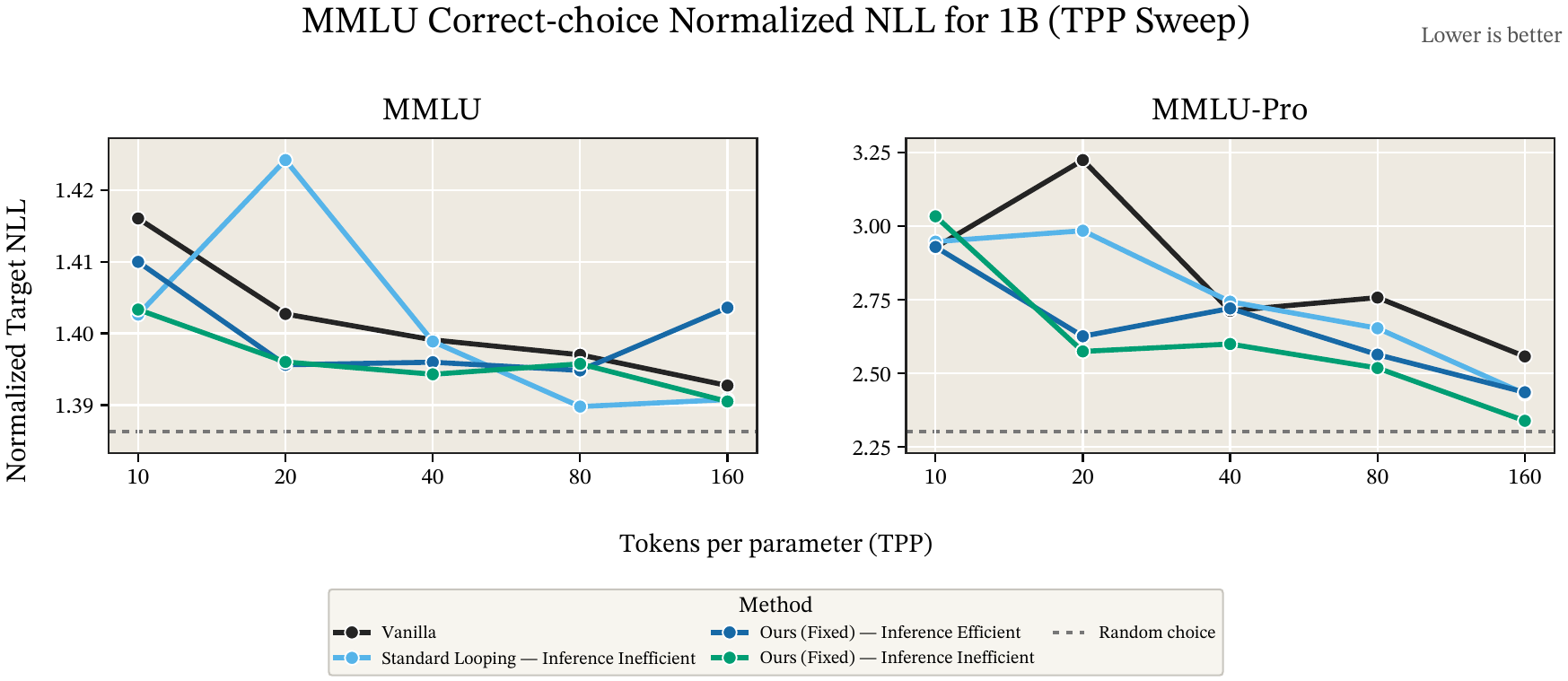}
  \caption{\textbf{MMLU and MMLU-Pro results for the 1B models.}
  Top: macro-averaged NLL-ranked accuracy, where the predicted answer is the
  candidate with the lowest token-normalized NLL. Bottom:
  candidate-normalized correct-answer NLL. Accuracy remains near the
  respective random-choice baselines, while candidate-normalized
  correct-answer NLL remains at or above the random-choice reference,
  particularly on MMLU-Pro. Dashed lines denote random-choice performance.
  Higher accuracy and lower NLL are better.}
  \label{fig:plt-downstream-mmlu}
\end{figure*}

\paragraph{Downstream metrics.}
  For each question, we score every answer option \(c\) by its average negative
  log-likelihood (NLL) per tokenizer token,
  \(q_c=-\frac{1}{T_c}\sum_{t=1}^{T_c}\log
  p_\theta(x_{c,t}\mid\text{prompt},x_{c,<t})\), where \(T_c\) is the number
  of tokens in option \(c\), \(x_{c,t}\) is its \(t\)-th token, and
  \(x_{c,<t}\) denotes its preceding tokens. We predict the option with the
  lowest \(q_c\). \autoref{fig:plt-downstream-accuracy} reports
  \emph{NLL-ranked accuracy} on ARC Easy, ARC Challenge, HellaSwag, PIQA,
  WinoGrande, OpenBookQA, CommonsenseQA, and Social IQA. The headline figure
  reports the unweighted mean across these eight tasks. The upper panel of
  \autoref{fig:plt-downstream-mmlu} reports macro-averaged accuracy for MMLU
  and MMLU-Pro. NLL-ranked accuracy records only whether the correct answer option has the
  best score, not how strongly it is preferred. For a question with \(C\)
  answer options, we therefore normalize the scores as
  \(\pi(c)=\exp(-q_c)/\sum_{j=1}^{C}\exp(-q_j)\), and report the target NLL
  \(-\log\pi(c^\star)\), where \(c^\star\) is the correct option. A lower value
  means that the model assigns greater relative preference to the correct
  option, even when the predicted answer remains unchanged. These results
  appear in \autoref{fig:plt-downstream-nll} and the lower panel of
  \autoref{fig:plt-downstream-mmlu}. Dashed lines show the random-choice
  baselines \(1/C\) for accuracy and \(\log C\) for target NLL. Both metrics
  evaluate the provided answer options rather than free-form generation.

\paragraph{Looping improves performance on most downstream tasks.}
  At TPP \(=160\), \emph{Full-Stack Looping},
  \emph{Normalized Shifted PLT}, and \emph{Normalized Unshifted PLT} outperform
  \emph{Vanilla} on six of the eight tasks in
  \autoref{fig:plt-downstream-accuracy}. The clearest improvements occur on ARC
  Challenge and HellaSwag, with additional gains on ARC Easy, PIQA, WinoGrande,
  and OpenBookQA. CommonsenseQA remains close to the random-choice baseline and
  varies non-monotonically across training budgets, while Social IQA shows no
  consistent improvement. No looping method dominates every task, but
  \emph{Normalized Shifted PLT} generally tracks the sequential looping
  methods, suggesting that parallel loop execution does not impose a systematic
  downstream-accuracy penalty.

\paragraph{Looping lowers correct-answer NLL on most downstream tasks.}
  The target-NLL results in \autoref{fig:plt-downstream-nll} show the same
  general pattern as accuracy. At TPP \(=160\), every looping method achieves
  lower choice-normalized target NLL than \emph{Vanilla} on the same six tasks.
  CommonsenseQA remains variable and close to its random-choice baseline, while
  Social IQA shows smaller, method-dependent differences. Thus, the aggregate
  improvement is distributed across several tasks rather than driven by a
  single benchmark, although it is not universal.

\paragraph{MMLU and MMLU-Pro remain near chance.}
We exclude these benchmarks from the eight-task headline mean and report them
separately in \autoref{fig:plt-downstream-mmlu}. None of the architectures
produces a consistent NLL-ranked accuracy advantage across the TPP sweep.
Although the candidate-normalized correct-answer NLL on MMLU-Pro generally
improves with additional training, it remains above the random-choice
reference. These results provide a useful knowledge diagnostic, but do not
show that looping improves MMLU-style performance at this scale.

\clearpage

\section{Additional Results on Looping Designs and Regimes}
\label{sec:looping-additional-results}

This appendix section provides the additional results supporting
\autoref{sec:which-loops-work}. The 100M and 500M results shown in
\autoref{fig:loopfactor-dclm-main}  are complemented by experiments at
70M, 200M, and 300M, a separate 1B study, and evaluations on additional
datasets. We first summarize the models, loop factors, and evaluation
setup used in these experiments, and then present the full DCLM results,
other dataset comparisons, the 1B extension, and mathematical and code
likelihood evaluations.

\subsection{Experimental Setup for the Additional Results}
\label{sec:looping-appendix-coverage}

\paragraph{Models and loop factors.}
For 70M and 100M models, we evaluate TPP values
10, 20, 40, 70, 100, 120, 140, and 160, with loop factors
\(L\in\{1,2,4,8,16,32\}\). For 200M, 300M, and 500M models, we evaluate
TPP values 10, 20, 40, 70, and 100, with loop factors
\(L\in\{1,2,4,8\}\). Not every configuration produces a stable run.
The 1B experiments use a smaller set of loop factors and are described
separately in \autoref{sec:looping-public-1b}.

\paragraph{Looping designs and training compute.}
We use the same looping designs introduced in
\autoref{sec:background_looping} and illustrated in
\autoref{fig:looping-architectures}. In the figure legends,
Stack Loop denotes Full-Stack Looping, Layer Loop denotes per-layer
looping, and Both Loop denotes separately looping the attention and MLP
sublayers. All designs use exact parameter sharing across loops. For a
model with \(B\) Transformer Blocks, Full-Stack Looping repeats the entire
sequence of Transformer Blocks \(L\) times, while Per-Layer Looping repeats
each Transformer Block \(L\) times before proceeding to the next; both
therefore have an effective depth of \(LB\). The middle variants restrict
this repeated computation to the middle \(M\) Transformer Blocks of the
network. Middle-Stack Looping repeats these \(M\) Transformer Blocks
together as a stack, whereas Middle-Layer Looping repeats each of these
Transformer Blocks individually. The Transformer Blocks before and after
the middle region are used only once, giving both middle variants an
effective depth of \(B+(L-1)M\). Thus, at the same loop factor \(L\), the
middle variants require less training compute because only the middle
portion of the network is repeated.

\paragraph{Evaluation.}
Each panel fixes model scale and TPP, where TPP denotes training tokens per
non-embedding parameter. We report teacher-forced bits per byte (BPB), with
lower values indicating better likelihood. The \(L=1\) model is the
standard non-looped baseline and is labeled Vanilla. In figures with
training FLOPs on the horizontal axis, marker shape denotes the loop factor.
These comparisons capture training compute only; they do not account for
system-level effects such as communication overhead or realized hardware
throughput.

\paragraph{Unstable runs.}
Some configurations become unstable during training and are excluded by our
shared stability criterion. We mark these runs with red stars in the figures.
If the corresponding BPB falls outside the displayed y-axis range, the star is
shown at the edge of the panel only to indicate that the run was unstable; its
vertical position should not be interpreted as the measured BPB. These points
are therefore different from configurations for which an evaluation is simply
missing. For example, in the small-model mathematical and code experiments,
11 of the 80 Both Loop configurations are unstable, mostly at \(L=32\) and
at several high-TPP 100M settings with \(L=8\).

\subsection{BPB Results on DCLM}
\label{sec:looping-appendix-dclm}

\begin{figure}[htbp]
  \centering
  \includegraphics[width=0.75\textwidth,height=0.75\textheight,keepaspectratio]{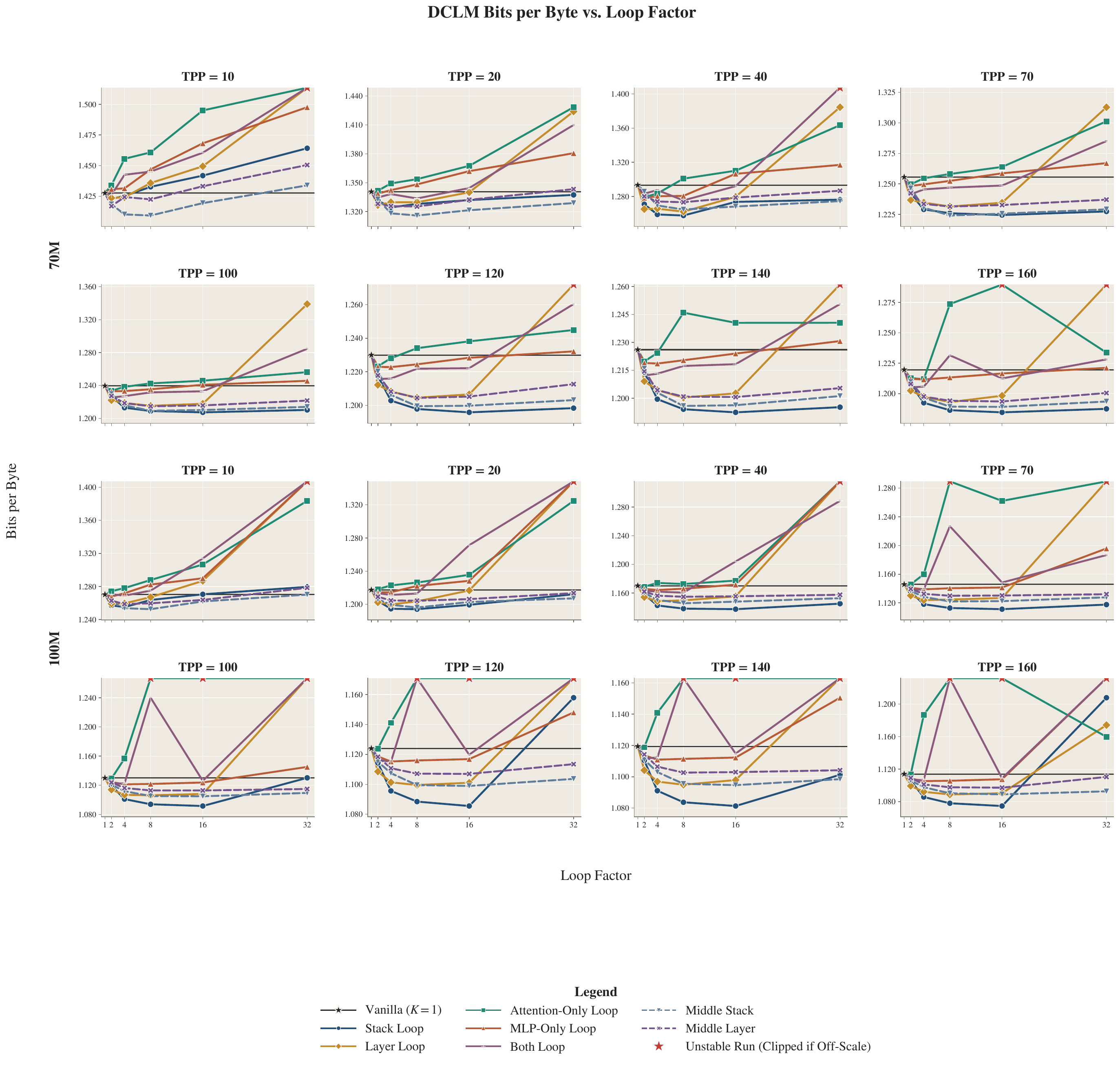}
  \caption{\textbf{DCLM BPB versus loop factor \(L\): 70M and 100M models.}
  Results span TPP 10 to 160 and loop factors \(L\in\{1,2,4,8,16,32\}\).
  Each panel fixes model scale and TPP, allowing the effect of increasing
  the loop factor to be compared across looping designs. Lower BPB is better;
  red stars denote unstable runs as described in
  \autoref{sec:looping-appendix-coverage}.}
  \label{fig:loopfactor-dclm-small}
\end{figure}

\begin{figure}[htbp]
  \centering
  \includegraphics[width=0.75\textwidth,height=0.75\textheight,keepaspectratio]{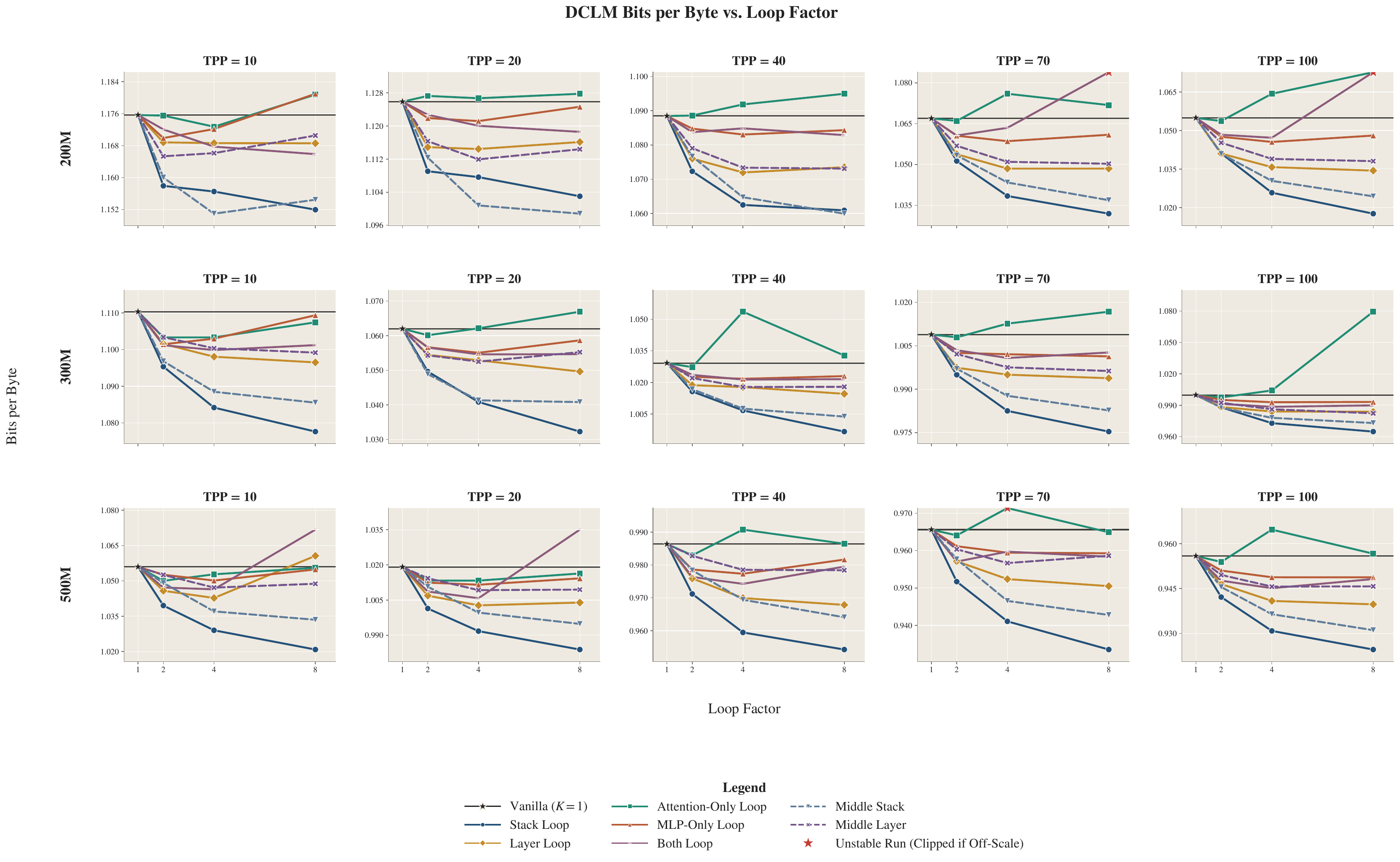}
  \caption{\textbf{DCLM BPB versus loop factor \(L\): 200M, 300M, and 500M models.}
  Results span TPP 10 to 100 and loop factors \(L\in\{1,2,4,8\}\).
  Each panel fixes model scale and TPP, allowing the effect of increasing
  the loop factor to be compared across looping designs. Lower BPB is better;
  red stars denote unstable runs as described in
  \autoref{sec:looping-appendix-coverage}.}
  \label{fig:loopfactor-dclm-large}
\end{figure}

\paragraph{How looping behavior changes with scale and loop factor.}
\autoref{fig:loopfactor-dclm-small} and
\autoref{fig:loopfactor-dclm-large} show how the preferred looping design
changes across model scales and as the loop factor \(L\) increases. At 70M
and 100M, several designs are competitive at small \(L\), and the strongest
choice depends on TPP. The 200M results remain mixed: Middle-Stack Looping
can match or outperform Full-Stack Looping at lower TPP, while full-stack
looping becomes stronger as TPP increases. At 300M and 500M, full-stack
looping more consistently reaches the lowest BPB. At the same time,
increasing \(L\) shows diminishing returns. In the 70M and 100M sweeps,
which extend to \(L=32\), the reduction in BPB typically becomes smaller as
\(L\) grows and can eventually saturate or reverse, especially for designs
that loop the attention and MLP sublayers separately. Full-stack and
Middle-Stack Looping remain more stable over a wider range of \(L\), but
their gains also diminish as looping increases. The 200M--500M experiments
extend only to \(L=8\), so they do not directly establish where these gains
eventually saturate or reverse. However, the marginal reduction in BPB
already becomes smaller as \(L\) increases, suggesting that further looping
would eventually reach a saturation regime rather than continue providing
comparable gains at each increase in loop factor.

\begin{figure}[htbp]
  \centering
  \includegraphics[width=0.60\textwidth,height=0.7\textheight,keepaspectratio]{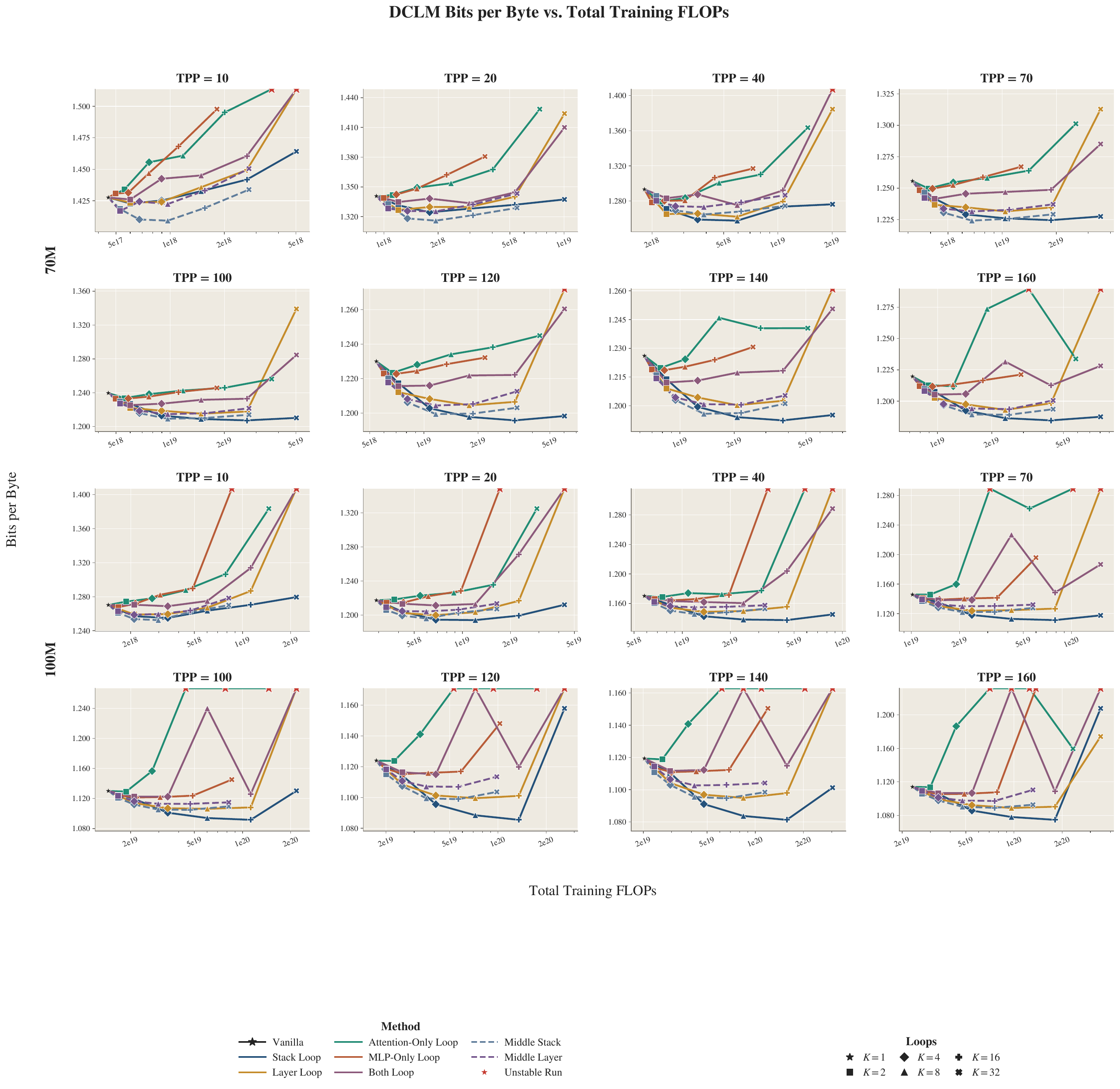}
  \caption{\textbf{DCLM BPB versus total training FLOPs: 70M and 100M models.}
  Results span TPP 10 to 160 and loop factors
  \(L\in\{1,2,4,8,16,32\}\).}
  \label{fig:loop-flops-dclm-small}
\end{figure}

\begin{figure}[htbp]
  \centering
  \includegraphics[width=0.60\textwidth,height=0.7\textheight,keepaspectratio]{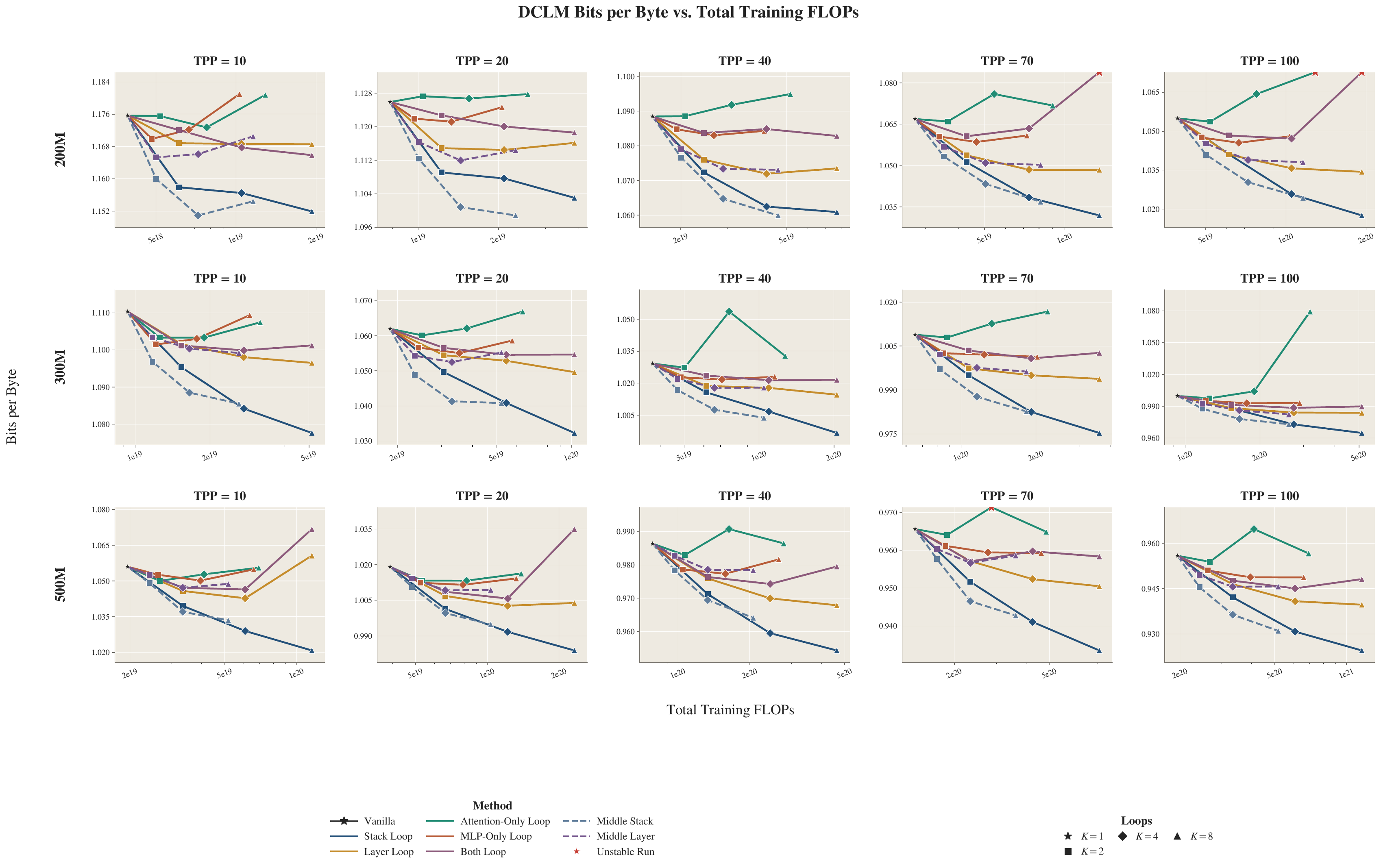}
  \caption{\textbf{DCLM BPB versus total training FLOPs: 200M, 300M, and 500M models.}
  Results span TPP 10 to 100 and loop factors \(L\in\{1,2,4,8\}\).
   Middle-Stack Looping often achieves strong gains for a smaller
  increase in compute, while Full-Stack Looping can attain lower BPB when
  more training compute is available.}
  \label{fig:loop-flops-dclm-large}
\end{figure}

\paragraph{How the compute budget changes the preferred looping design.}
\autoref{fig:loop-flops-dclm-small} and
\autoref{fig:loop-flops-dclm-large} compare DCLM BPB against total
training FLOPs across all model scales. When only a modest increase in
training compute is available, Middle-Stack Looping often provides a
larger improvement for the additional FLOPs because only the middle
portion of the network is repeated. With a larger compute budget,
Full-Stack Looping can attain lower absolute BPB, particularly at 300M
and 500M, where it more consistently reaches the strongest quality among
the evaluated looping designs. The 70M and 100M results also show that
spending progressively more compute by increasing the loop factor \(L\)
eventually gives diminishing returns: the BPB reductions become smaller
and can ultimately saturate or reverse. The 200M--500M sweeps extend only
to \(L=8\), but their marginal gains also tend to shrink as \(L\)
increases. Together, these results support the same trade-off observed in
the main text: Middle-Stack Looping is often more attractive when
additional training compute is limited, whereas Full-Stack Looping can
achieve better quality when more compute is available.

\clearpage

\clearpage

\subsection{Do the Looping Trends Hold Across Other Datasets?}
\label{sec:looping-cross-corpus}

We next ask whether the trends observed on DCLM persist across other
datasets. We first evaluate additional pretraining corpora: C4
(\autoref{sec:c4eng_looping_ablations}), WikiText-103
(\autoref{sec:wikitext_looping_ablations}), Dolma v1.5
(\autoref{sec:dolma_looping_ablations}), RedPajama
(\autoref{sec:redpajama_looping_ablations}), and Dolma Programming
Languages (\autoref{sec:dolma_programming_looping_ablations}). We then
consider more specialized mathematical and code datasets: GSM8K
(\autoref{sec:gsm8k_looping_ablations}), MATH
(\autoref{sec:math_looping_ablations}), HumanEval
(\autoref{sec:humaneval_looping_ablations}), and MBPP
(\autoref{sec:mbpp_looping_ablations}). Across the additional pretraining corpora, the same broad pattern observed
on DCLM generally persists. At smaller model scales and lower TPP,
full-stack and Middle-Stack Looping are often competitive, while full-stack
looping becomes more consistently favorable as model scale and TPP increase.
When training compute is considered, shallow Middle-Stack Looping remains an
attractive alternative because it can achieve substantial quality gains
while repeating only part of the network. The mathematical and code
datasets exhibit more variation across looping designs, but broadly retain
the tendency for Full-Stack Looping to become more favorable at larger
scales and higher TPP. Across both groups of datasets, the gains from
increasing the loop factor \(L\) also tend to diminish as \(L\) grows.

\clearpage

\subsubsection{C4 English}
\label{sec:c4eng_looping_ablations}
\begin{figure}[htbp]
  \centering
  \includegraphics[width=0.70\textwidth,height=0.70\textheight,keepaspectratio]{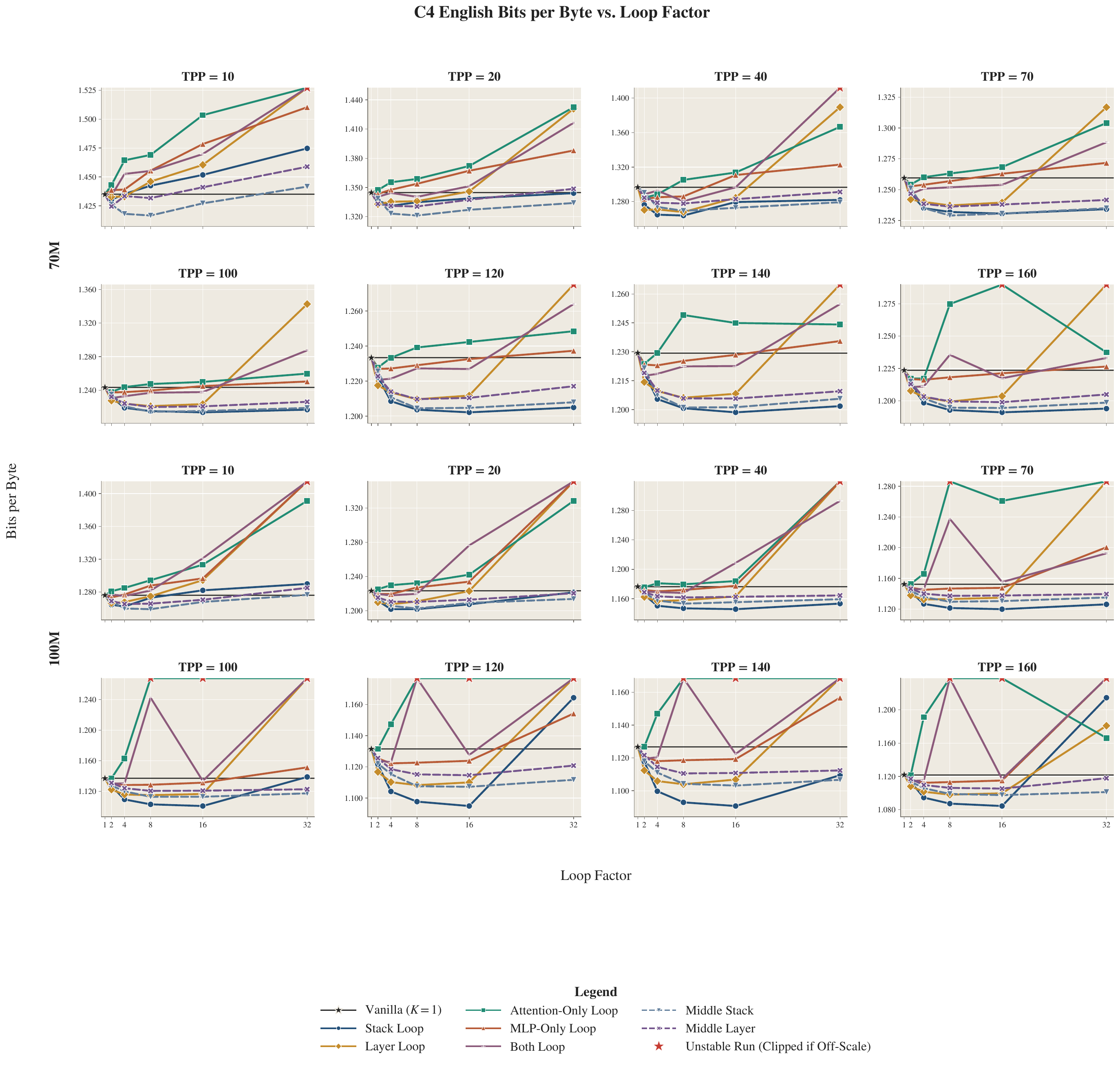}
  \caption{\textbf{C4 English BPB versus loop factor \(L\): 70M and 100M models.}
  Results span TPP 10 to 160 and loop factors
  \(L\in\{1,2,4,8,16,32\}\). At these smaller scales, several looping
  designs remain competitive, while the improvement from increasing \(L\)
  generally diminishes at larger loop factors. Each panel fixes model scale
  and TPP. Lower BPB is better; red stars denote unstable runs as described
  in \autoref{sec:looping-appendix-coverage}.}
  \label{fig:loop-public-c4-factor-small}
\end{figure}

\begin{figure}[htbp]
  \centering
  \includegraphics[width=0.70\textwidth,height=0.70\textheight,keepaspectratio]{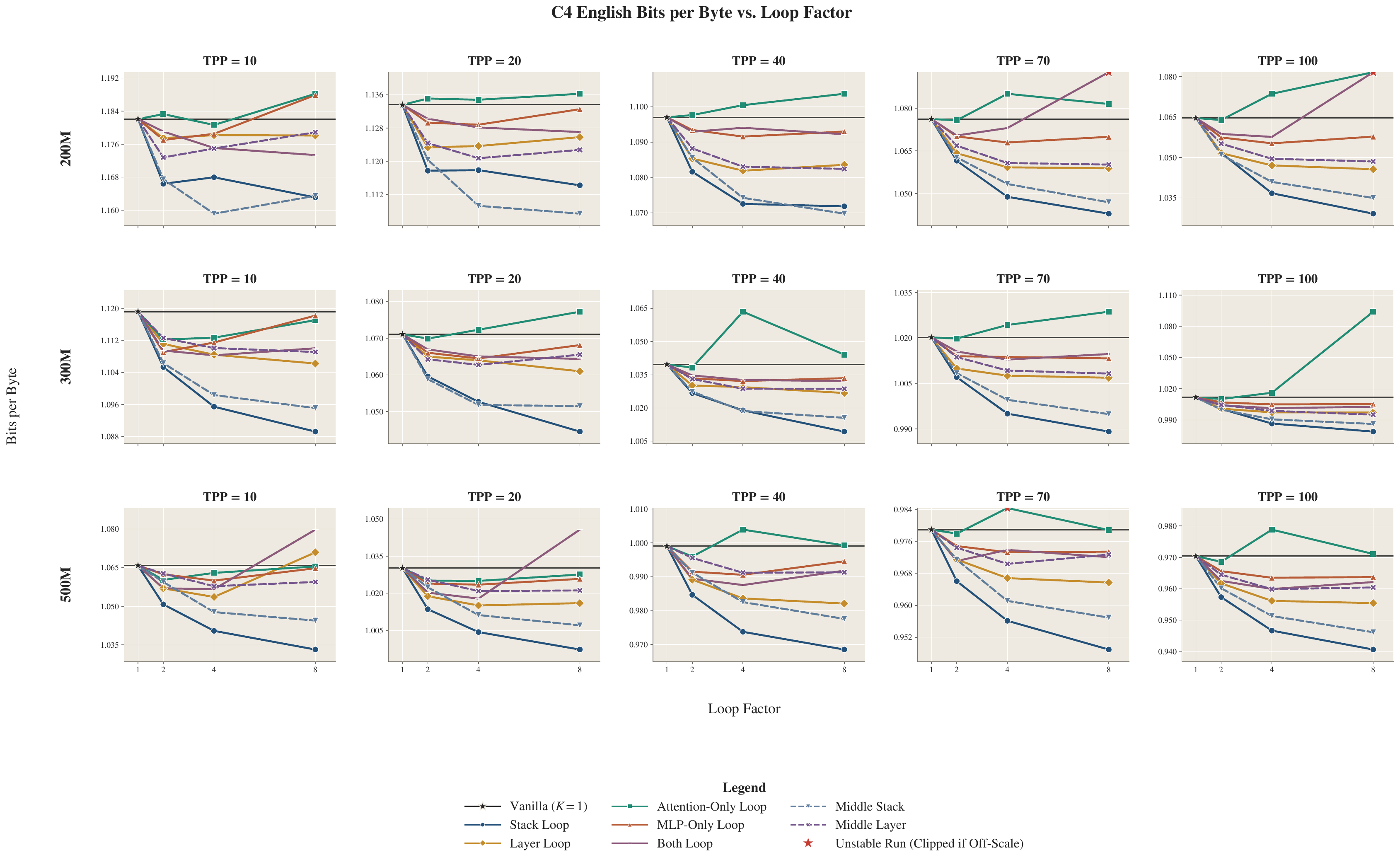}
  \caption{\textbf{C4 English BPB versus loop factor \(L\): 200M, 300M, and 500M models.}
  Results span TPP 10 to 100 and loop factors
  \(L\in\{1,2,4,8\}\). As model scale and TPP increase, Full-Stack Looping
  more consistently attains lower BPB, while the marginal improvement from
  increasing \(L\) becomes smaller. Each panel fixes model scale and TPP.
  Lower BPB is better; red stars denote unstable runs as described in
  \autoref{sec:looping-appendix-coverage}.}
  \label{fig:loop-public-c4-factor-large}
\end{figure}

\begin{figure}[htbp]
  \centering
  \includegraphics[width=0.70\textwidth,height=0.70\textheight,keepaspectratio]{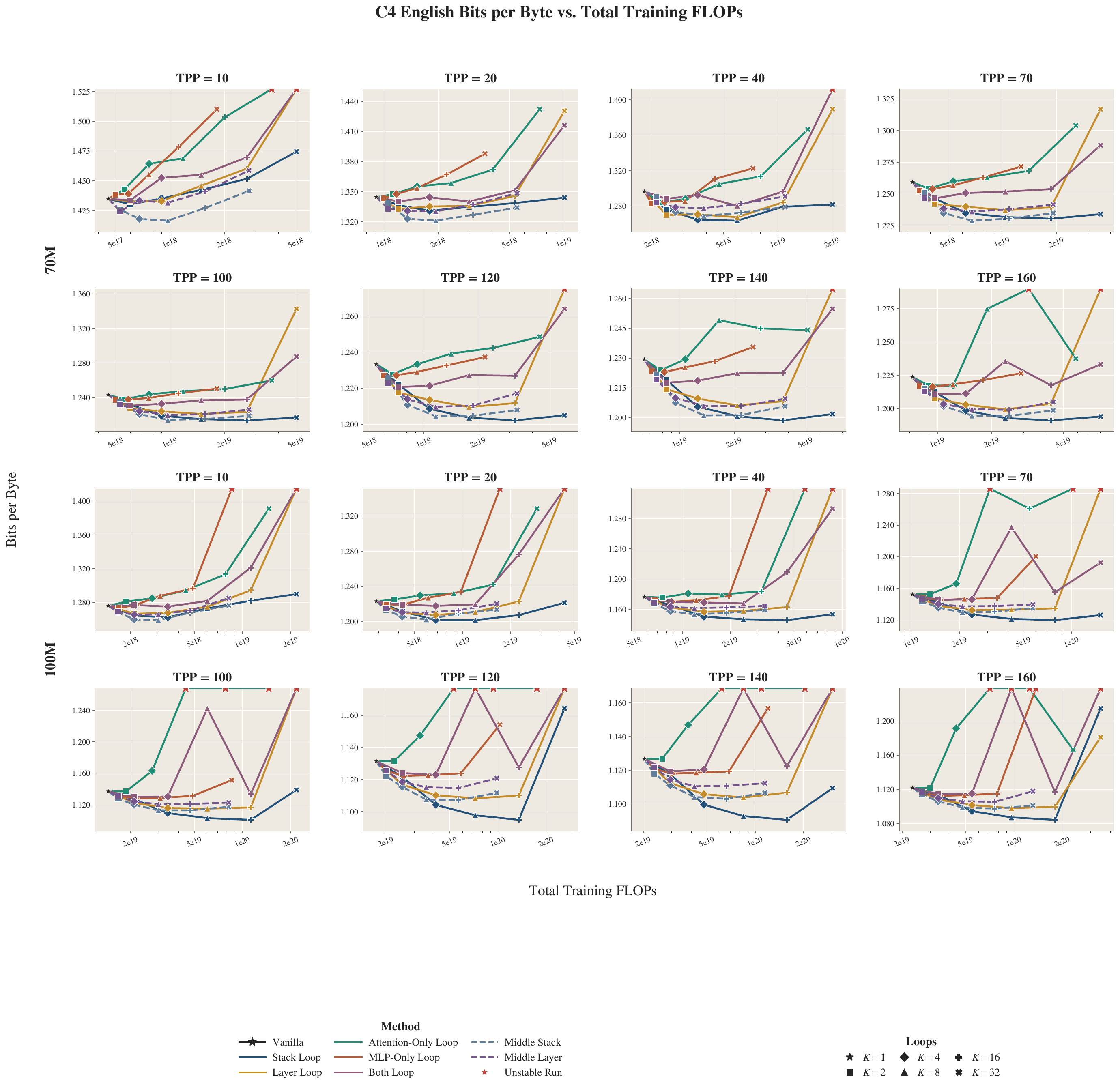}
  \caption{\textbf{C4 English BPB versus total training FLOPs: 70M and 100M models.}
  Results span TPP 10 to 160 and loop factors
  \(L\in\{1,2,4,8,16,32\}\). Middle-Stack Looping often captures much of
  the quality improvement with a smaller increase in training compute,
  while spending additional FLOPs through larger loop factors eventually
  shows diminishing returns. Each panel fixes model scale and TPP.
  Lower BPB is better; red stars denote unstable runs as described in
  \autoref{sec:looping-appendix-coverage}.}
  \label{fig:loop-public-c4-flops-small}
\end{figure}

\begin{figure}[htbp]
  \centering
  \includegraphics[width=0.70\textwidth,height=0.70\textheight,keepaspectratio]{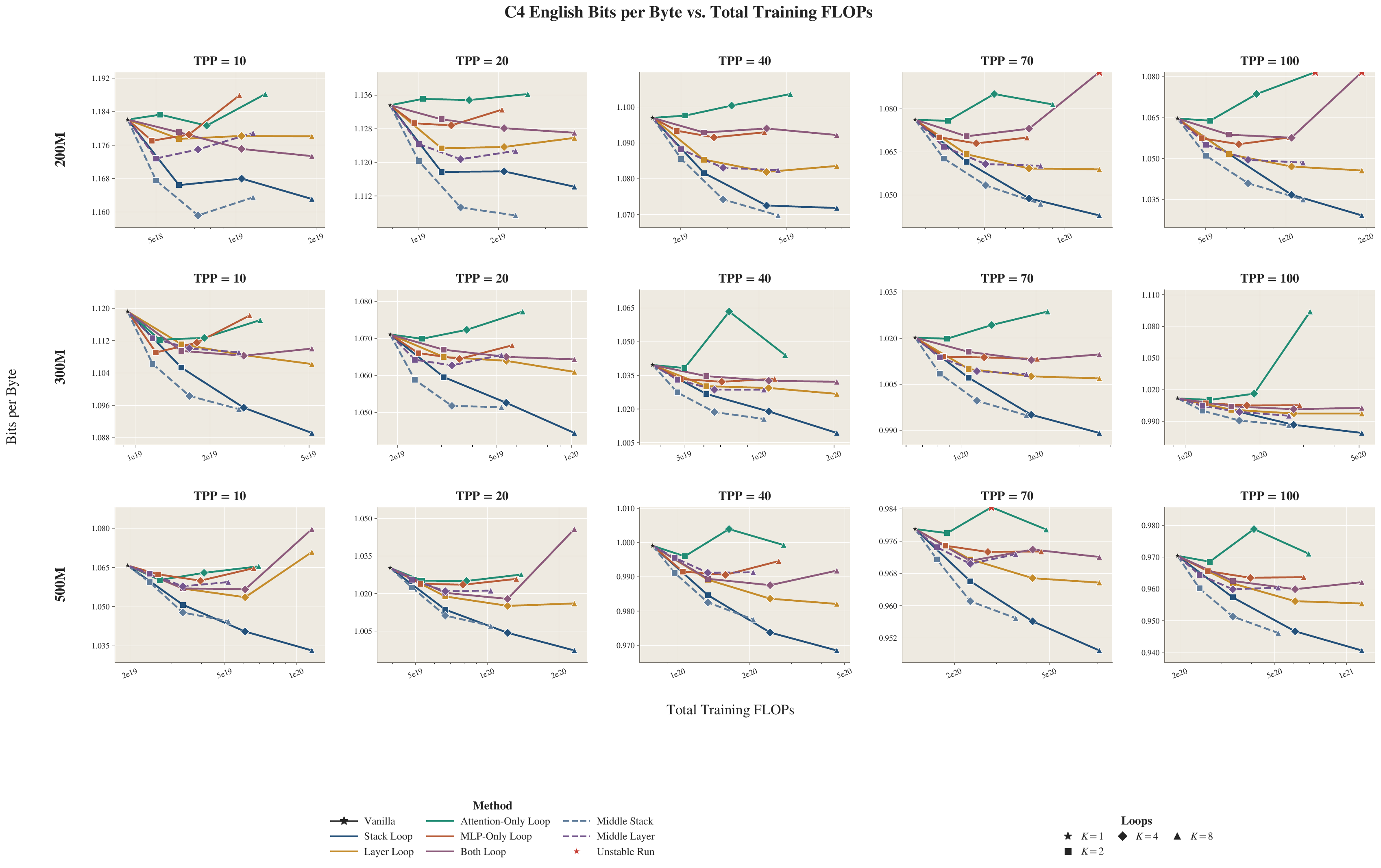}
  \caption{\textbf{C4 English BPB versus total training FLOPs: 200M, 300M, and 500M models.}
  Results span TPP 10 to 100 and loop factors
  \(L\in\{1,2,4,8\}\). Middle-Stack Looping often provides strong gains
  for a smaller increase in training compute, while Full-Stack Looping can
  attain lower BPB when more compute is available, particularly at the
  larger model scales. Each panel fixes model scale and TPP.
  Lower BPB is better; red stars denote unstable runs as described in
  \autoref{sec:looping-appendix-coverage}.}
  \label{fig:loop-public-c4-flops-large}
\end{figure}

The C4 sweeps in \autoref{fig:loop-public-c4-factor-small},
\autoref{fig:loop-public-c4-factor-large},
\autoref{fig:loop-public-c4-flops-small}, and
\autoref{fig:loop-public-c4-flops-large} closely follow the DCLM
pattern: rankings are more mixed at smaller scales, while
Full-Stack Looping becomes more favorable at larger scales and
middle-stack remains attractive for modest compute increments.

  \clearpage

\subsubsection{WikiText-103}
\label{sec:wikitext_looping_ablations}

\begin{figure}[htbp]
  \centering
  \includegraphics[width=0.61\textwidth,height=0.70\textheight,keepaspectratio]{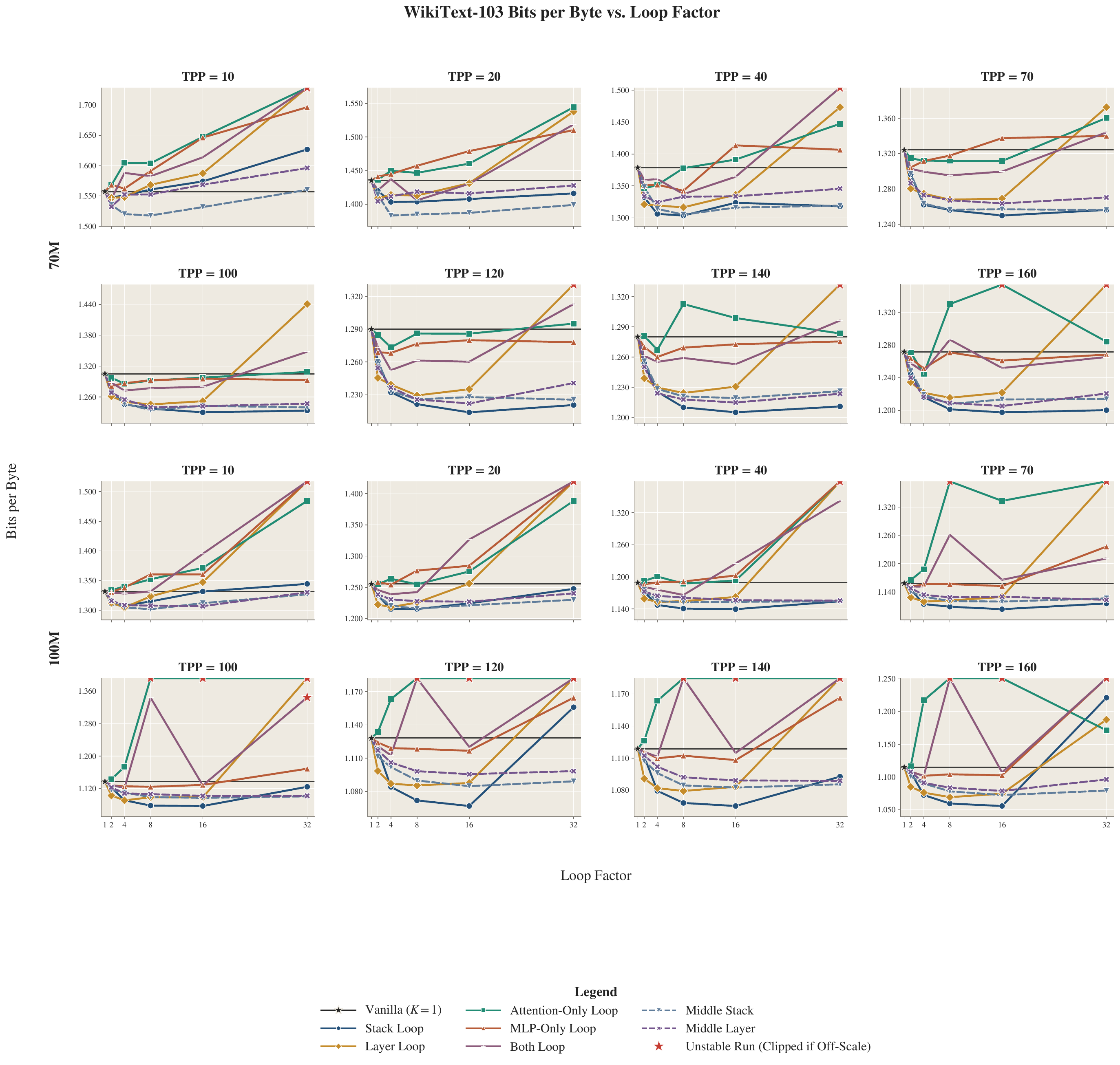}
  \caption{\textbf{WikiText-103 BPB versus loop factor \(L\): 70M and 100M models.}
  Results span TPP 10 to 160 and loop factors
  \(L\in\{1,2,4,8,16,32\}\). At these smaller scales, several looping
  designs remain competitive, while the improvement from increasing \(L\)
  generally diminishes at larger loop factors. Each panel fixes model scale
  and TPP. Lower BPB is better; red stars denote unstable runs as described
  in \autoref{sec:looping-appendix-coverage}.}
  \label{fig:loop-public-wikitext-factor-small}
\end{figure}

\begin{figure}[htbp]
  \centering
  \includegraphics[width=0.61\textwidth,height=0.70\textheight,keepaspectratio]{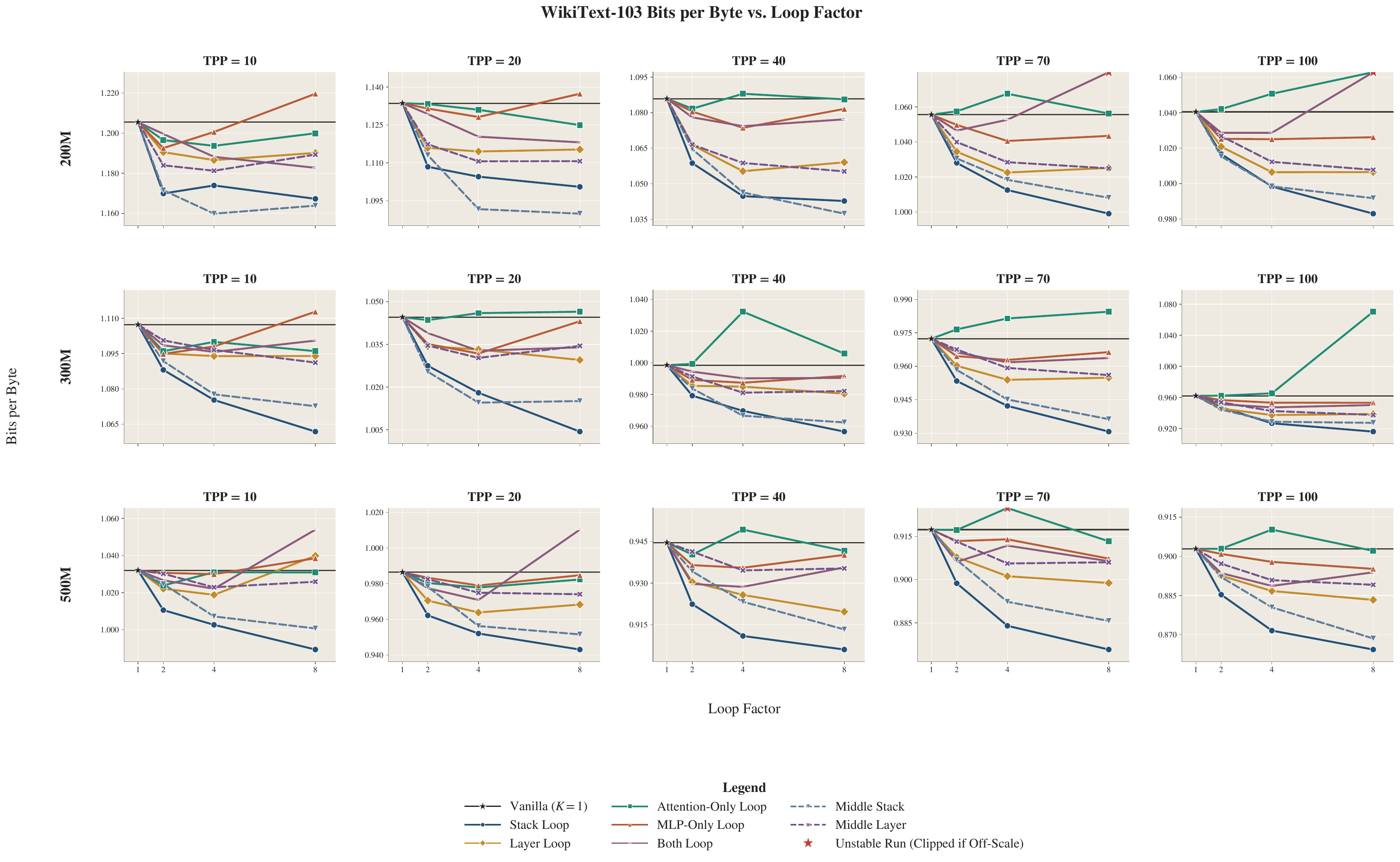}
  \caption{\textbf{WikiText-103 BPB versus loop factor \(L\): 200M, 300M, and 500M models.}
  Results span TPP 10 to 100 and loop factors
  \(L\in\{1,2,4,8\}\). As model scale and TPP increase, Full-Stack Looping
  more consistently attains lower BPB, while the marginal improvement from
  increasing \(L\) becomes smaller. Each panel fixes model scale and TPP.
  Lower BPB is better; red stars denote unstable runs as described in
  \autoref{sec:looping-appendix-coverage}.}
  \label{fig:loop-public-wikitext-factor-large}
\end{figure}

\begin{figure}[htbp]
  \centering
  \includegraphics[width=0.70\textwidth,height=0.70\textheight,keepaspectratio]{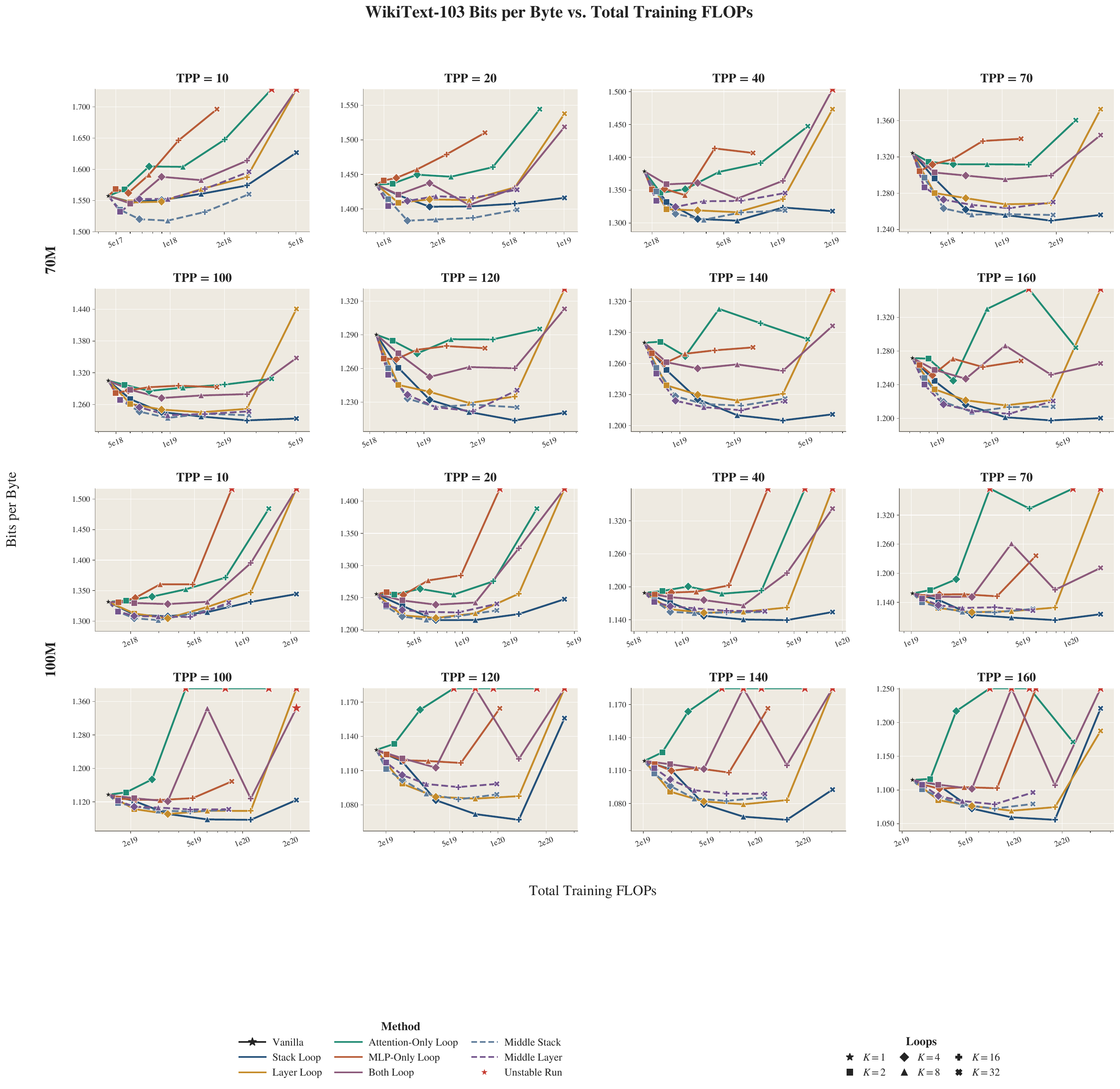}
  \caption{\textbf{WikiText-103 BPB versus total training FLOPs: 70M and 100M models.}
  Results span TPP 10 to 160 and loop factors
  \(L\in\{1,2,4,8,16,32\}\). Middle-Stack Looping often captures much of
  the quality improvement with a smaller increase in training compute,
  while increasing the loop factor further eventually gives diminishing
  returns. Each panel fixes model scale and TPP. Lower BPB is better; red
  stars denote unstable runs as described in
  \autoref{sec:looping-appendix-coverage}.}
  \label{fig:loop-public-wikitext-flops-small}
\end{figure}

\begin{figure}[htbp]
  \centering
  \includegraphics[width=0.70\textwidth,height=0.70\textheight,keepaspectratio]{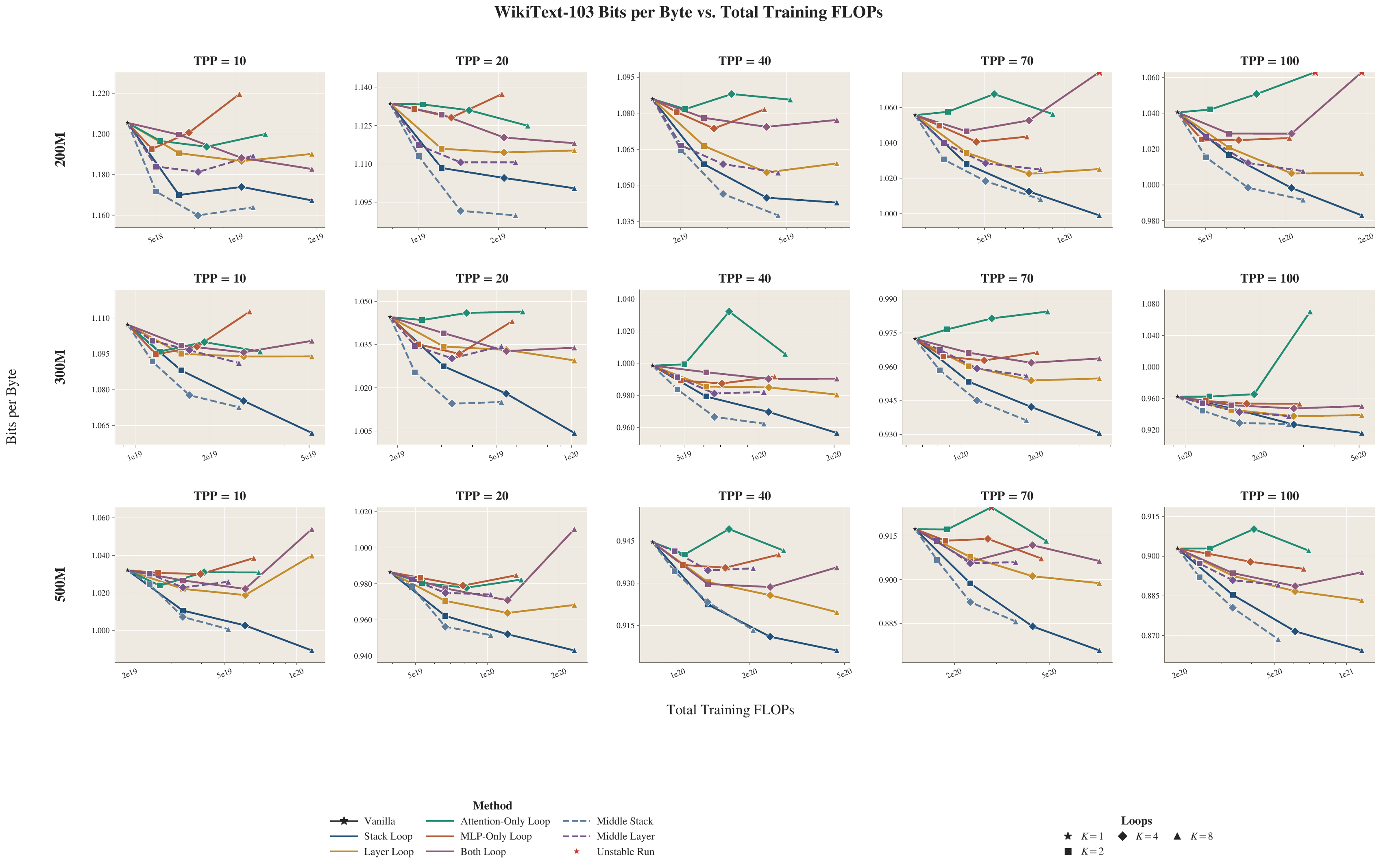}
  \caption{\textbf{WikiText-103 BPB versus total training FLOPs: 200M, 300M, and 500M models.}
  Results span TPP 10 to 100 and loop factors
  \(L\in\{1,2,4,8\}\). Middle-Stack Looping often provides strong gains
  for a smaller increase in training compute, while Full-Stack Looping can
  attain lower BPB when more compute is available, particularly at larger
  model scales. Each panel fixes model scale and TPP. Lower BPB is better;
  red stars denote unstable runs as described in
  \autoref{sec:looping-appendix-coverage}.}
  \label{fig:loop-public-wikitext-flops-large}
\end{figure}

The WikiText-103 results in
\autoref{fig:loop-public-wikitext-factor-small},
\autoref{fig:loop-public-wikitext-factor-large},
\autoref{fig:loop-public-wikitext-flops-small},
\autoref{fig:loop-public-wikitext-flops-large}
show the same broad trends observed on DCLM and C4. At smaller model scales
and lower TPP, several looping designs remain competitive, and the preferred
design depends on the training regime and loop factor \(L\). As model scale
and TPP increase, Full-Stack Looping more consistently attains lower BPB.
When training compute is considered, however, shallow Middle-Stack Looping
remains attractive because it can achieve substantial quality improvements
with a smaller increase in FLOPs. The gains from increasing \(L\) also tend
to diminish at larger loop factors.

\clearpage

\subsubsection{Dolma v1.5}
\label{sec:dolma_looping_ablations}

\begin{figure}[htbp]
  \centering
  \includegraphics[width=0.60\textwidth,height=0.7\textheight,keepaspectratio]{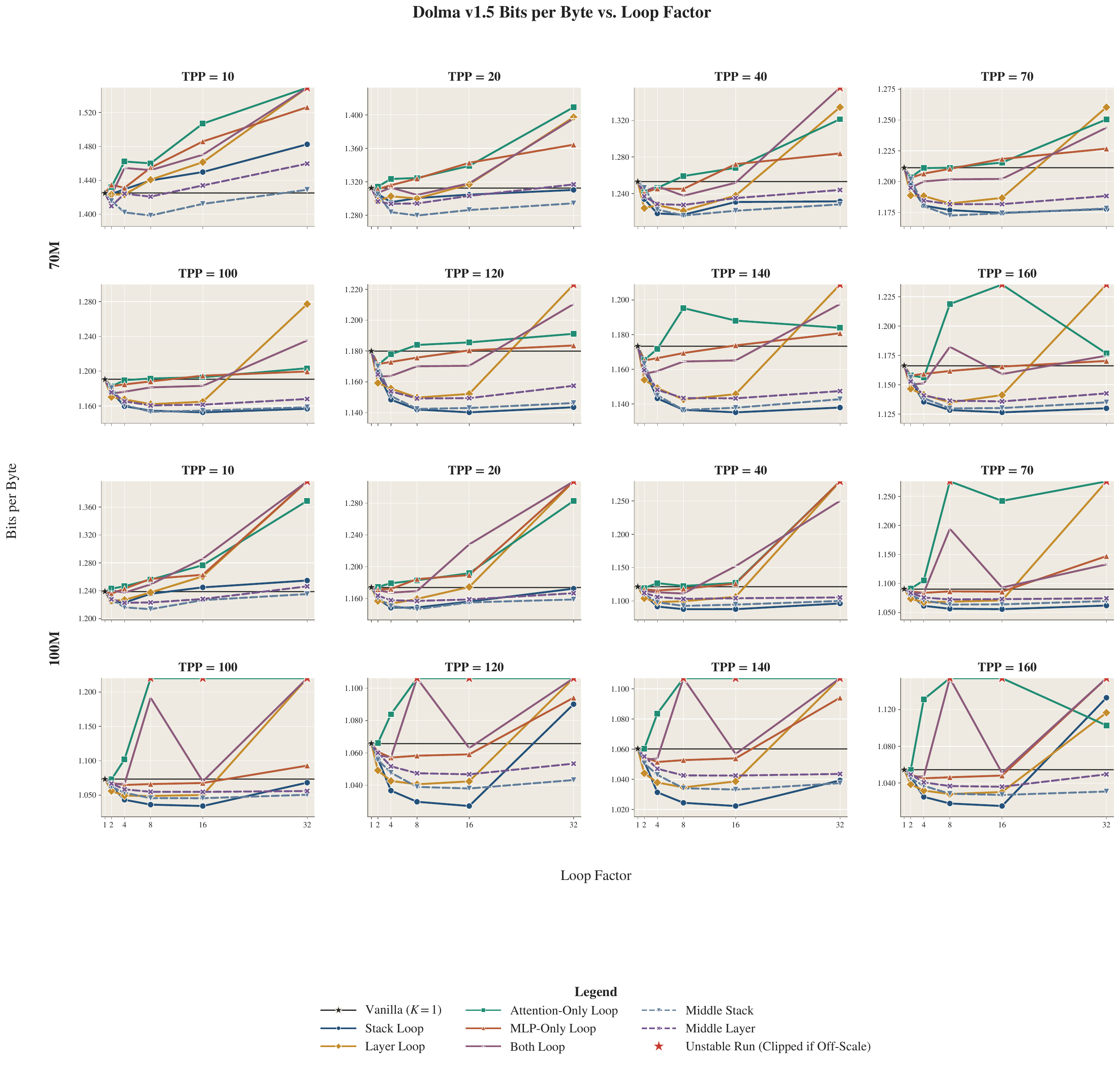}
  \caption{\textbf{Dolma v1.5 BPB versus loop factor \(L\): 70M and 100M models.}
  Results span TPP 10 to 160 and loop factors
  \(L\in\{1,2,4,8,16,32\}\). At these smaller scales, several looping
  designs remain competitive, while the improvement from increasing \(L\)
  generally diminishes at larger loop factors. Each panel fixes model scale
  and TPP. Lower BPB is better; red stars denote unstable runs as described
  in \autoref{sec:looping-appendix-coverage}.}
  \label{fig:loop-public-dolma-factor-small}
\end{figure}

\begin{figure}[htbp]
  \centering
  \includegraphics[width=0.60\textwidth,height=0.7\textheight,keepaspectratio]{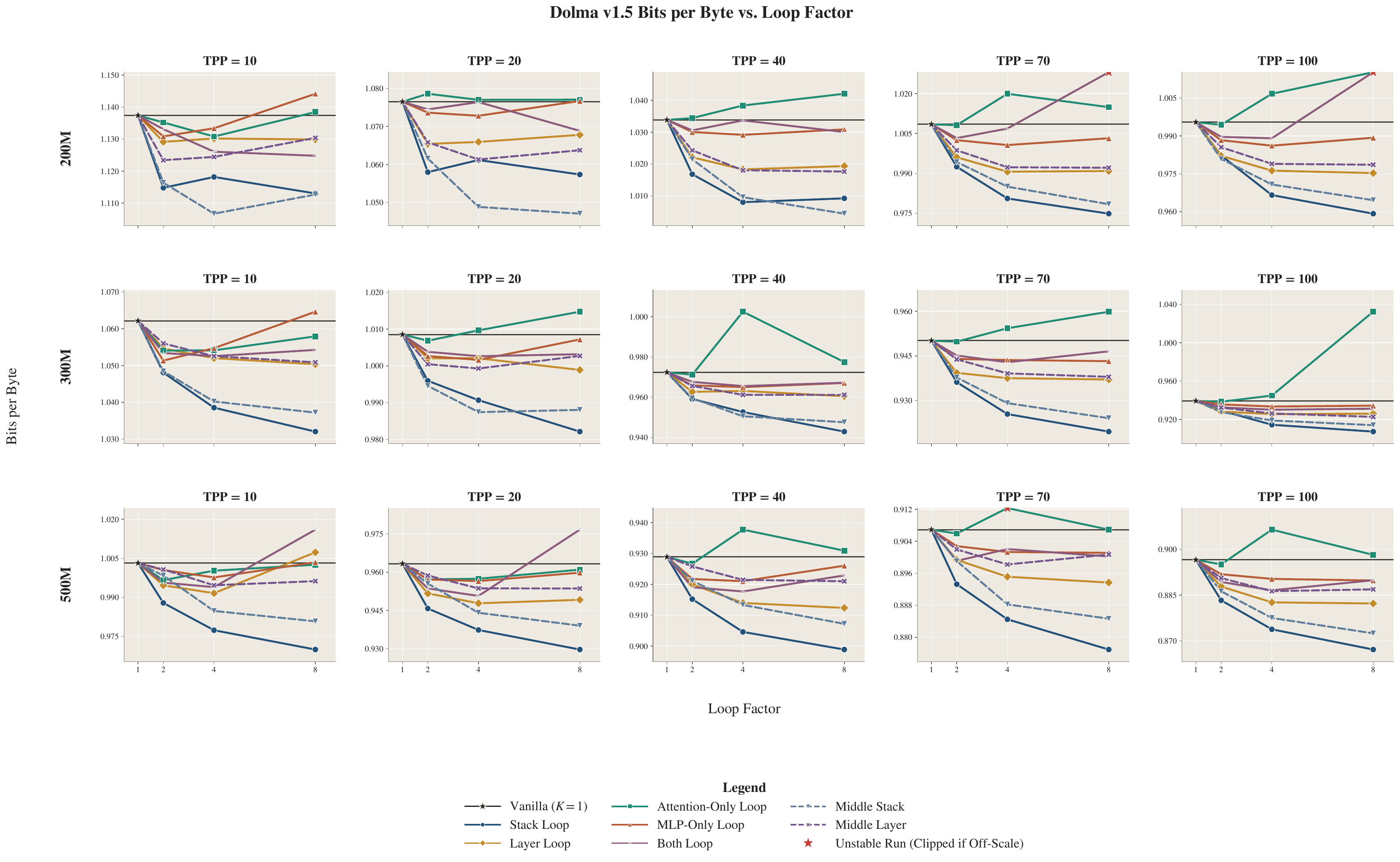}
  \caption{\textbf{Dolma v1.5 BPB versus loop factor \(L\): 200M, 300M, and 500M models.}
  Results span TPP 10 to 100 and loop factors
  \(L\in\{1,2,4,8\}\). As model scale and TPP increase, Full-Stack Looping
  more consistently attains lower BPB, while the marginal improvement from
  increasing \(L\) becomes smaller. Each panel fixes model scale and TPP.
  Lower BPB is better; red stars denote unstable runs as described in
  \autoref{sec:looping-appendix-coverage}.}
  \label{fig:loop-public-dolma-factor-large}
\end{figure}

\begin{figure}[htbp]
  \centering
  \includegraphics[width=0.70\textwidth,height=0.70\textheight,keepaspectratio]{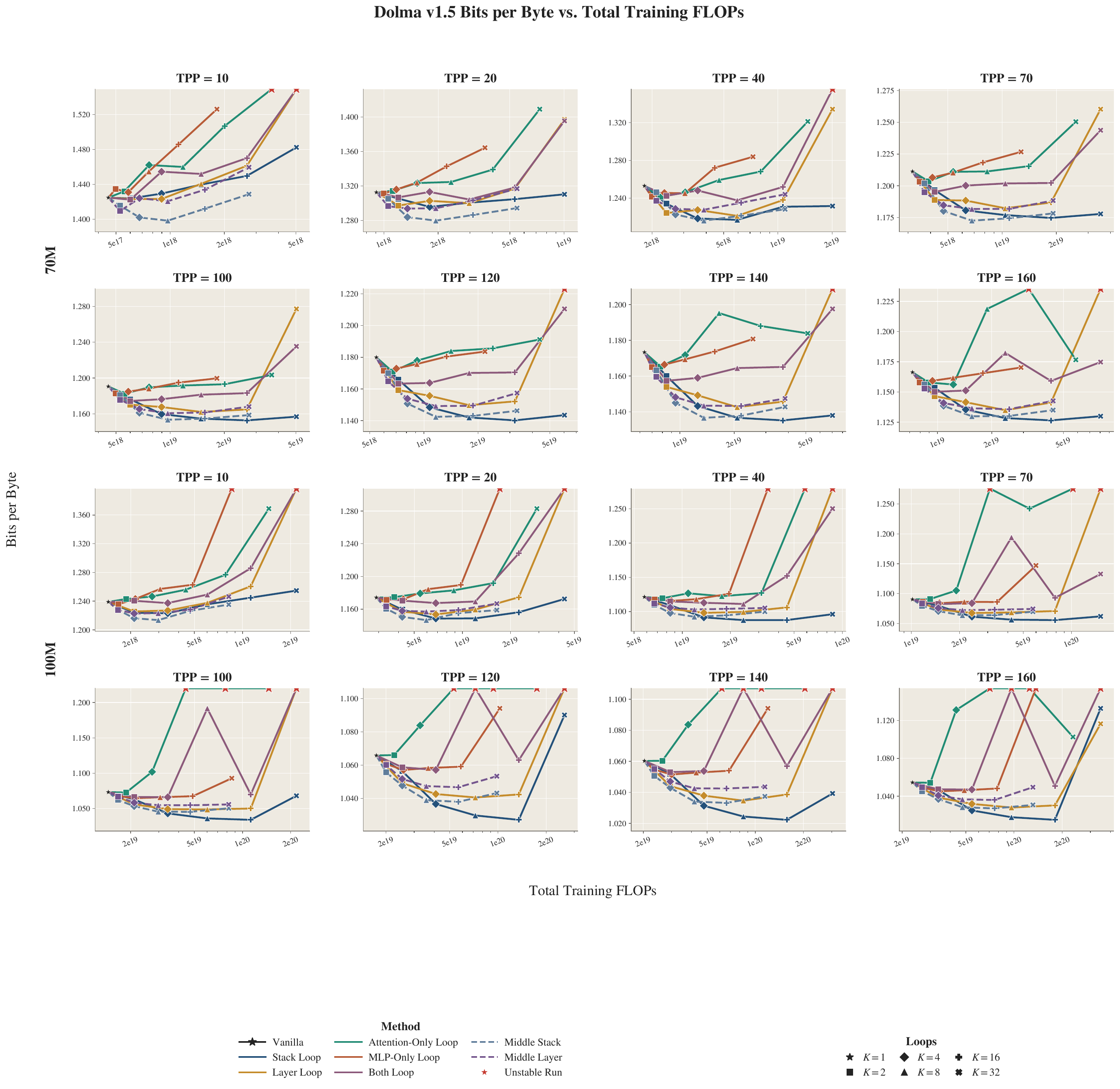}
  \caption{\textbf{Dolma v1.5 BPB versus total training FLOPs: 70M and 100M models.}
  Results span TPP 10 to 160 and loop factors
  \(L\in\{1,2,4,8,16,32\}\). Middle-Stack Looping often captures much of
  the quality improvement with a smaller increase in training compute,
  while increasing the loop factor further eventually gives diminishing
  returns. Each panel fixes model scale and TPP. Lower BPB is better; red
  stars denote unstable runs as described in
  \autoref{sec:looping-appendix-coverage}.}
  \label{fig:loop-public-dolma-flops-small}
\end{figure}

\begin{figure}[htbp]
  \centering
  \includegraphics[width=0.70\textwidth,height=0.70\textheight,keepaspectratio]{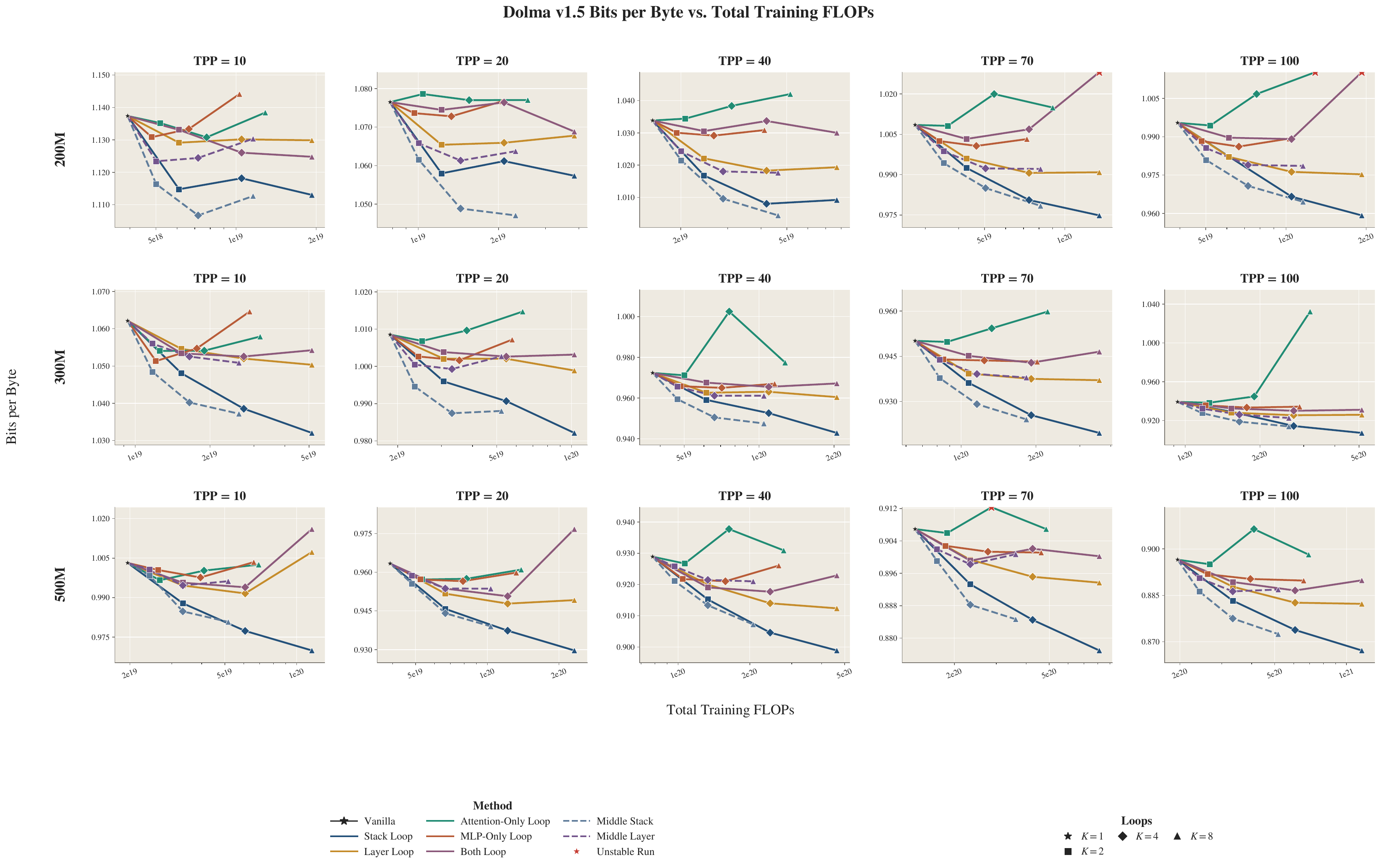}
  \caption{\textbf{Dolma v1.5 BPB versus total training FLOPs: 200M, 300M, and 500M models.}
  Results span TPP 10 to 100 and loop factors
  \(L\in\{1,2,4,8\}\). Middle-Stack Looping often provides strong gains
  for a smaller increase in training compute, while Full-Stack Looping can
  attain lower BPB when more compute is available, particularly at larger
  model scales. Each panel fixes model scale and TPP. Lower BPB is better;
  red stars denote unstable runs as described in
  \autoref{sec:looping-appendix-coverage}.}
  \label{fig:loop-public-dolma-flops-large}
\end{figure}

The Dolma v1.5 results in
\autoref{fig:loop-public-dolma-factor-small},
\autoref{fig:loop-public-dolma-factor-large},
\autoref{fig:loop-public-dolma-flops-small},
\autoref{fig:loop-public-dolma-flops-large}
show the same broad trends observed on DCLM, C4, and WikiText-103. At
smaller model scales and lower TPP, several looping designs remain
competitive, and the preferred design depends on the training regime and
loop factor \(L\). As model scale and TPP increase, Full-Stack Looping more
consistently attains lower BPB. When training compute is considered,
however, shallow Middle-Stack Looping remains attractive because it can
provide substantial quality gains with a smaller increase in FLOPs. The
benefit from increasing \(L\) also tends to diminish at larger loop factors.

\clearpage

\subsubsection{RedPajama}
\label{sec:redpajama_looping_ablations}

\begin{figure}[htbp]
  \centering
  \includegraphics[width=0.60\textwidth,height=0.7\textheight,keepaspectratio]{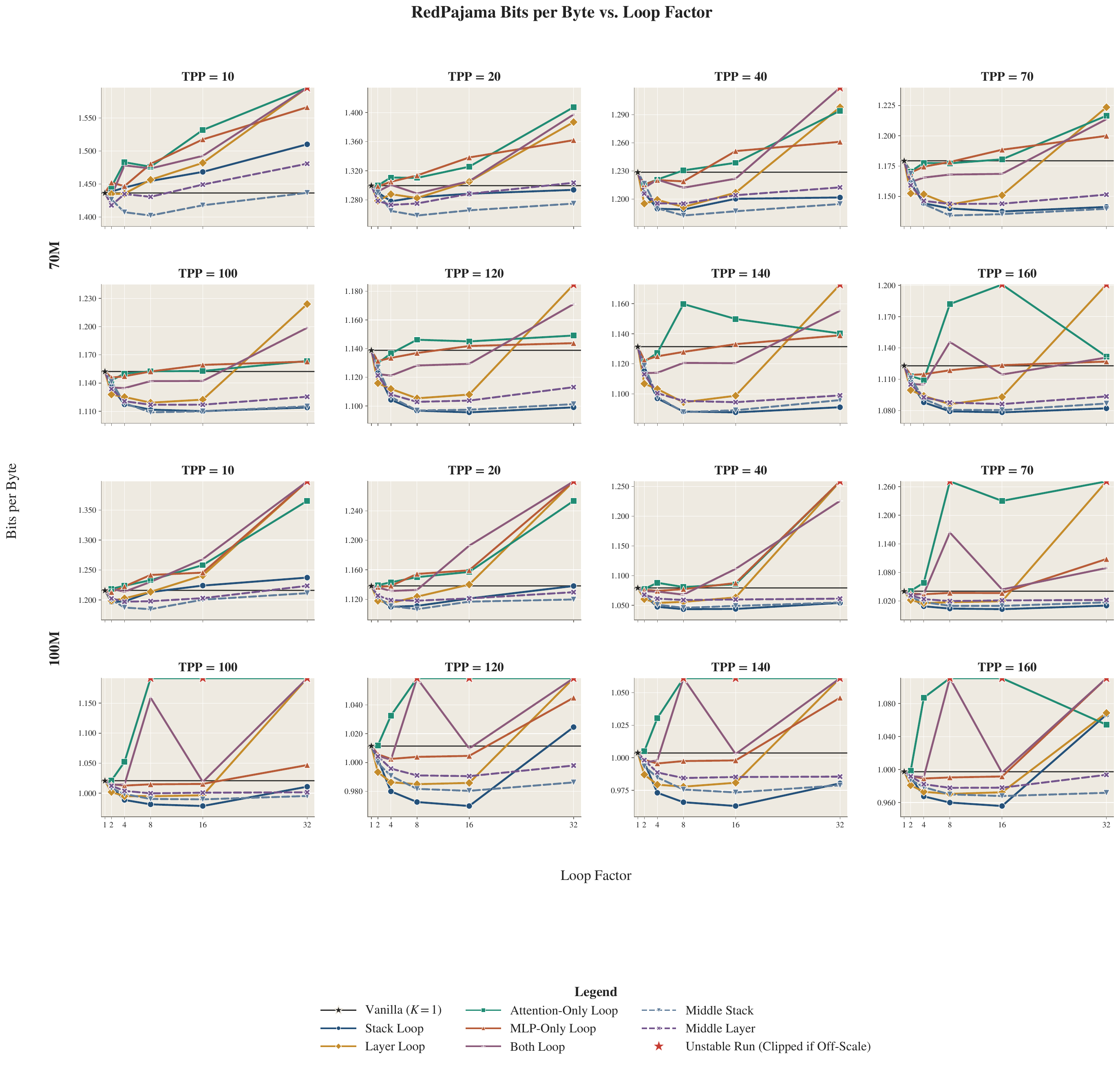}
  \caption{\textbf{RedPajama BPB versus loop factor \(L\): 70M and 100M models.}
  Results span TPP 10--160 and loop factors
  \(L\in\{1,2,4,8,16,32\}\). At these smaller scales, several looping
  designs remain competitive, while the improvement from increasing \(L\)
  generally diminishes at larger loop factors. Each panel fixes model scale
  and TPP. Lower BPB is better; red stars denote unstable runs as described
  in \autoref{sec:looping-appendix-coverage}.}
  \label{fig:loop-public-redpajama-factor-small}
\end{figure}

\begin{figure}[htbp]
  \centering
  \includegraphics[width=0.60\textwidth,height=0.7\textheight,keepaspectratio]{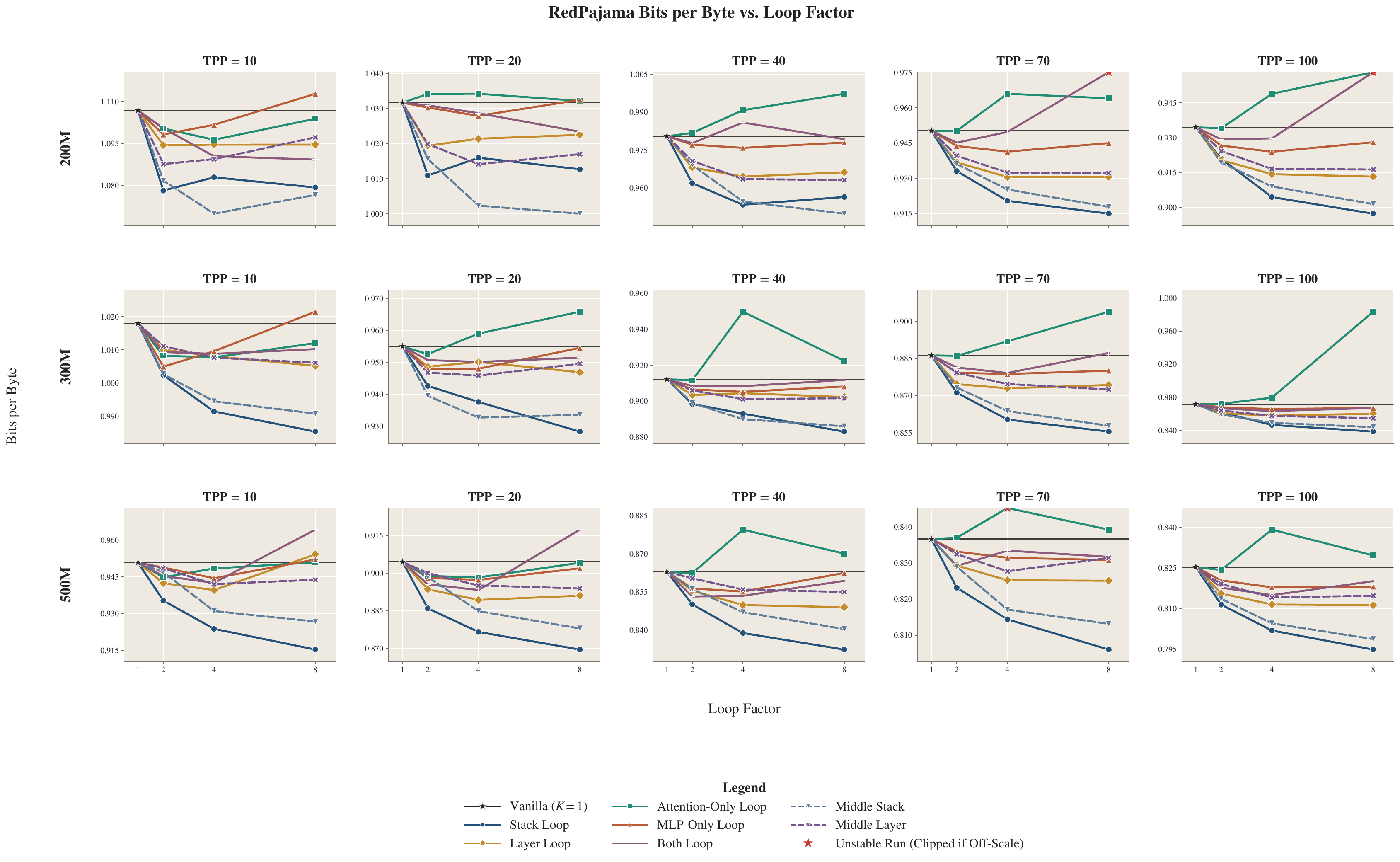}
  \caption{\textbf{RedPajama BPB versus loop factor \(L\): 200M, 300M, and 500M models.}
  Results span TPP 10--100 and loop factors
  \(L\in\{1,2,4,8\}\). As model scale and TPP increase, Full-Stack Looping
  more consistently attains lower BPB, while the marginal improvement from
  increasing \(L\) becomes smaller. Each panel fixes model scale and TPP.
  Lower BPB is better; red stars denote unstable runs as described in
  \autoref{sec:looping-appendix-coverage}.}
  \label{fig:loop-public-redpajama-factor-large}
\end{figure}

\begin{figure}[htbp]
  \centering
  \includegraphics[width=0.70\textwidth,height=0.70\textheight,keepaspectratio]{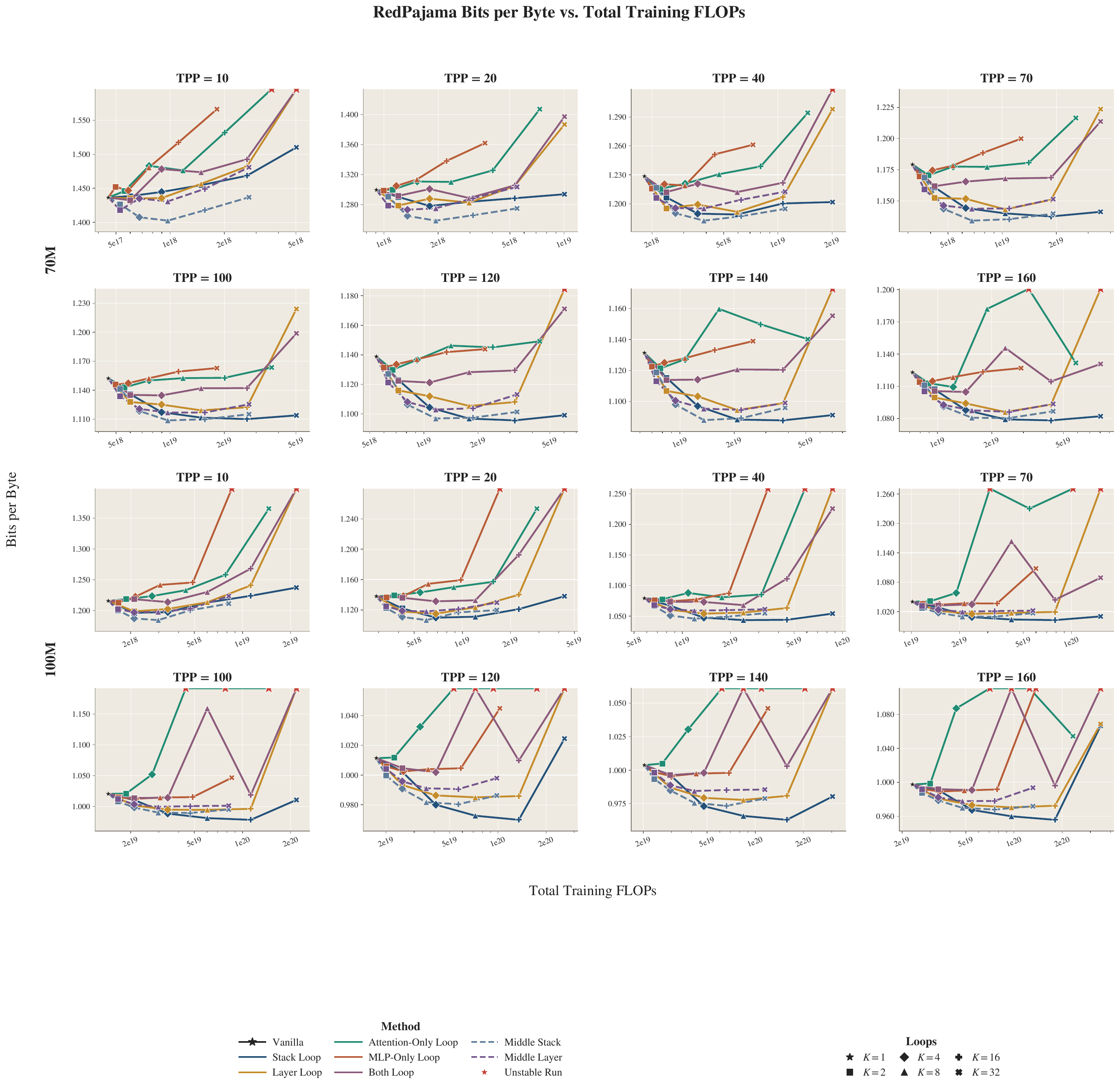}
  \caption{\textbf{RedPajama BPB versus total training FLOPs: 70M and 100M models.}
  Results span TPP 10--160 and loop factors
  \(L\in\{1,2,4,8,16,32\}\). Middle-Stack Looping often captures much of
  the quality improvement with a smaller increase in training compute,
  while increasing the loop factor further eventually gives diminishing
  returns. Each panel fixes model scale and TPP. Lower BPB is better; red
  stars denote unstable runs as described in
  \autoref{sec:looping-appendix-coverage}.}
  \label{fig:loop-public-redpajama-flops-small}
\end{figure}

\begin{figure}[htbp]
  \centering
  \includegraphics[width=0.70\textwidth,height=0.70\textheight,keepaspectratio]{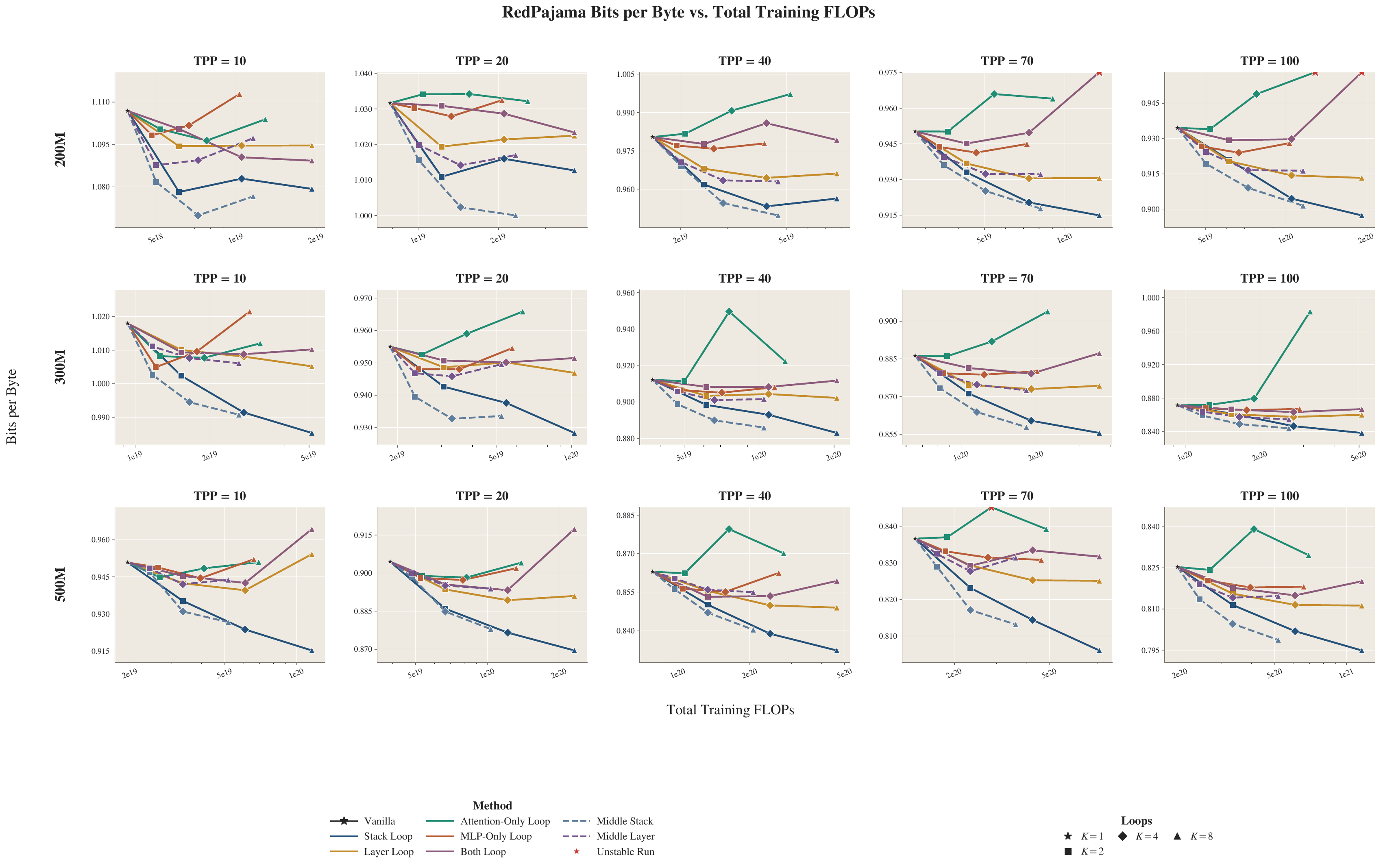}
  \caption{\textbf{RedPajama BPB versus total training FLOPs: 200M, 300M, and 500M models.}
  Results span TPP 10--100 and loop factors
  \(L\in\{1,2,4,8\}\). Middle-Stack Looping often provides strong gains
  for a smaller increase in training compute, while Full-Stack Looping can
  attain lower BPB when more compute is available, particularly at larger
  model scales. Each panel fixes model scale and TPP. Lower BPB is better;
  red stars denote unstable runs as described in
  \autoref{sec:looping-appendix-coverage}.}
  \label{fig:loop-public-redpajama-flops-large}
\end{figure}

The RedPajama results in
\autoref{fig:loop-public-redpajama-factor-small},
\autoref{fig:loop-public-redpajama-factor-large},
\autoref{fig:loop-public-redpajama-flops-small},
\autoref{fig:loop-public-redpajama-flops-large}
show the same broad trends observed across the other pretraining corpora.
At smaller model scales and lower TPP, several looping designs remain
competitive, and the preferred design depends on the training regime and
loop factor \(L\). As model scale and TPP increase, Full-Stack Looping more
consistently attains lower BPB. When training compute is considered,
however, shallow Middle-Stack Looping remains attractive because it can
provide substantial quality gains with a smaller increase in FLOPs. The
benefit from increasing \(L\) also tends to diminish at larger loop factors.

\clearpage

\subsubsection{Dolma Programming Languages}
\label{sec:dolma_programming_looping_ablations}

\begin{figure}[htbp]
  \centering
  \includegraphics[width=0.60\textwidth,height=0.7\textheight,keepaspectratio]{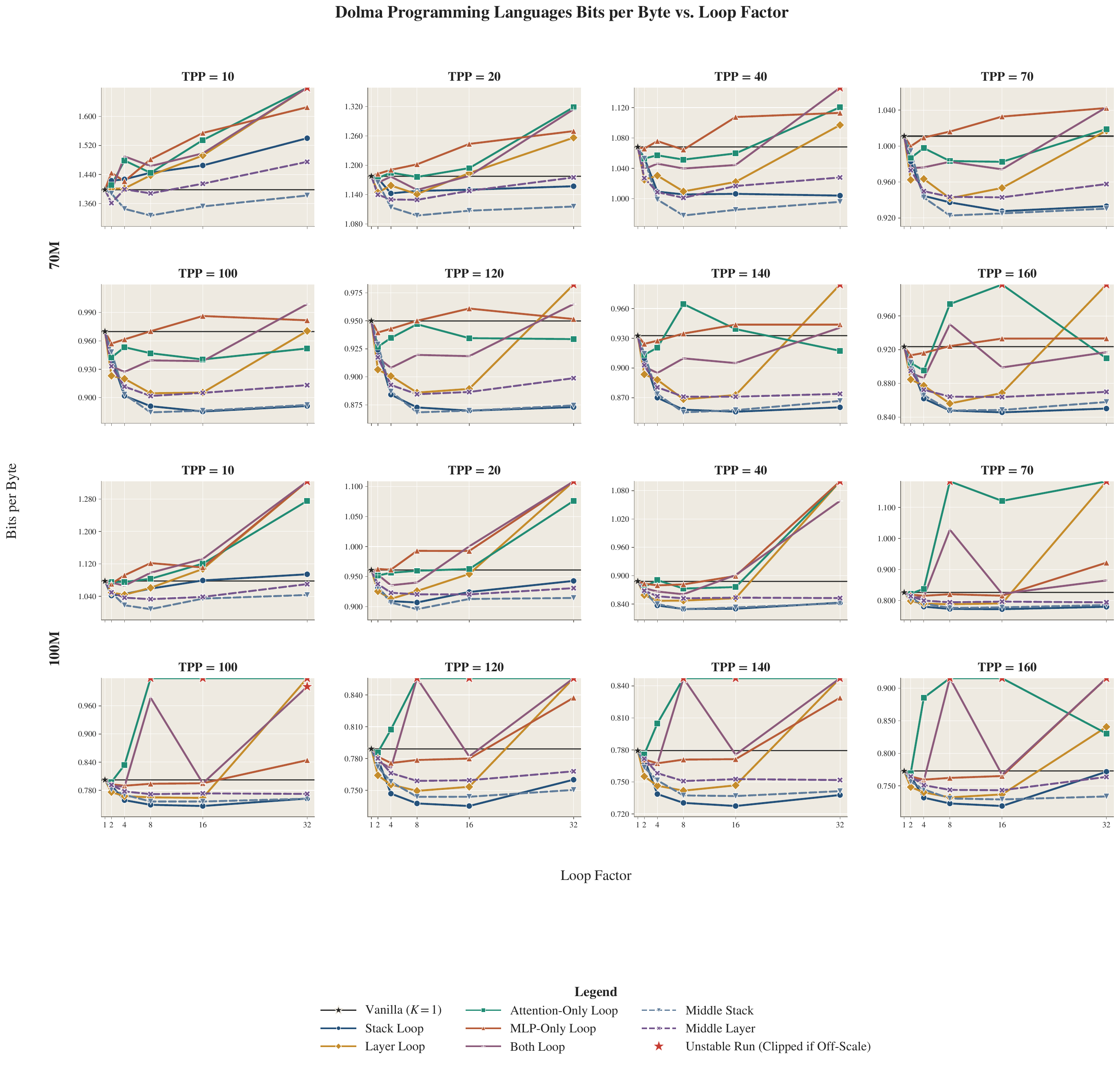}
  \caption{\textbf{Dolma programming-language BPB versus loop factor \(L\): 70M and 100M models.}
  Results span TPP 10--160 and loop factors
  \(L\in\{1,2,4,8,16,32\}\). At these smaller scales, Middle-Stack Looping
  is often competitive with Full-Stack Looping, while the improvement from
  increasing \(L\) generally diminishes at larger loop factors. Each panel
  fixes model scale and TPP. Lower BPB is better; red stars denote unstable
  runs as described in \autoref{sec:looping-appendix-coverage}.}
  \label{fig:loop-public-programming-factor-small}
\end{figure}

\begin{figure}[htbp]
  \centering
  \includegraphics[width=0.60\textwidth,height=0.7\textheight,keepaspectratio]{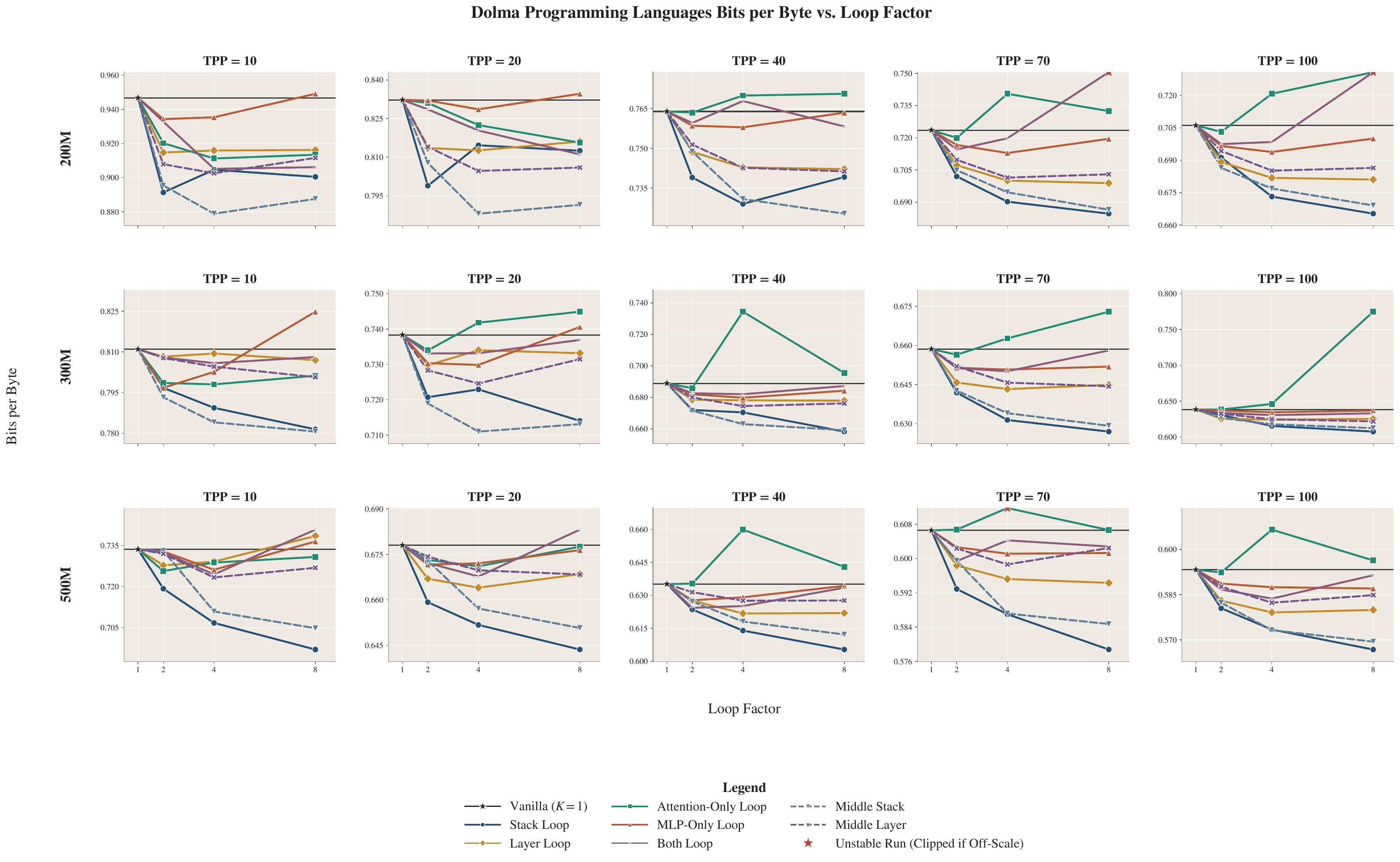}
  \caption{\textbf{Dolma programming-language BPB versus loop factor \(L\): 200M, 300M, and 500M models.}
  Results span TPP 10--100 and loop factors
  \(L\in\{1,2,4,8\}\). As model scale and TPP increase, Full-Stack Looping
  becomes more consistently favorable, while the marginal improvement from
  increasing \(L\) becomes smaller. Each panel fixes model scale and TPP.
  Lower BPB is better; red stars denote unstable runs as described in
  \autoref{sec:looping-appendix-coverage}.}
  \label{fig:loop-public-programming-factor-large}
\end{figure}

\begin{figure}[htbp]
  \centering
  \includegraphics[width=0.70\textwidth,height=0.70\textheight,keepaspectratio]{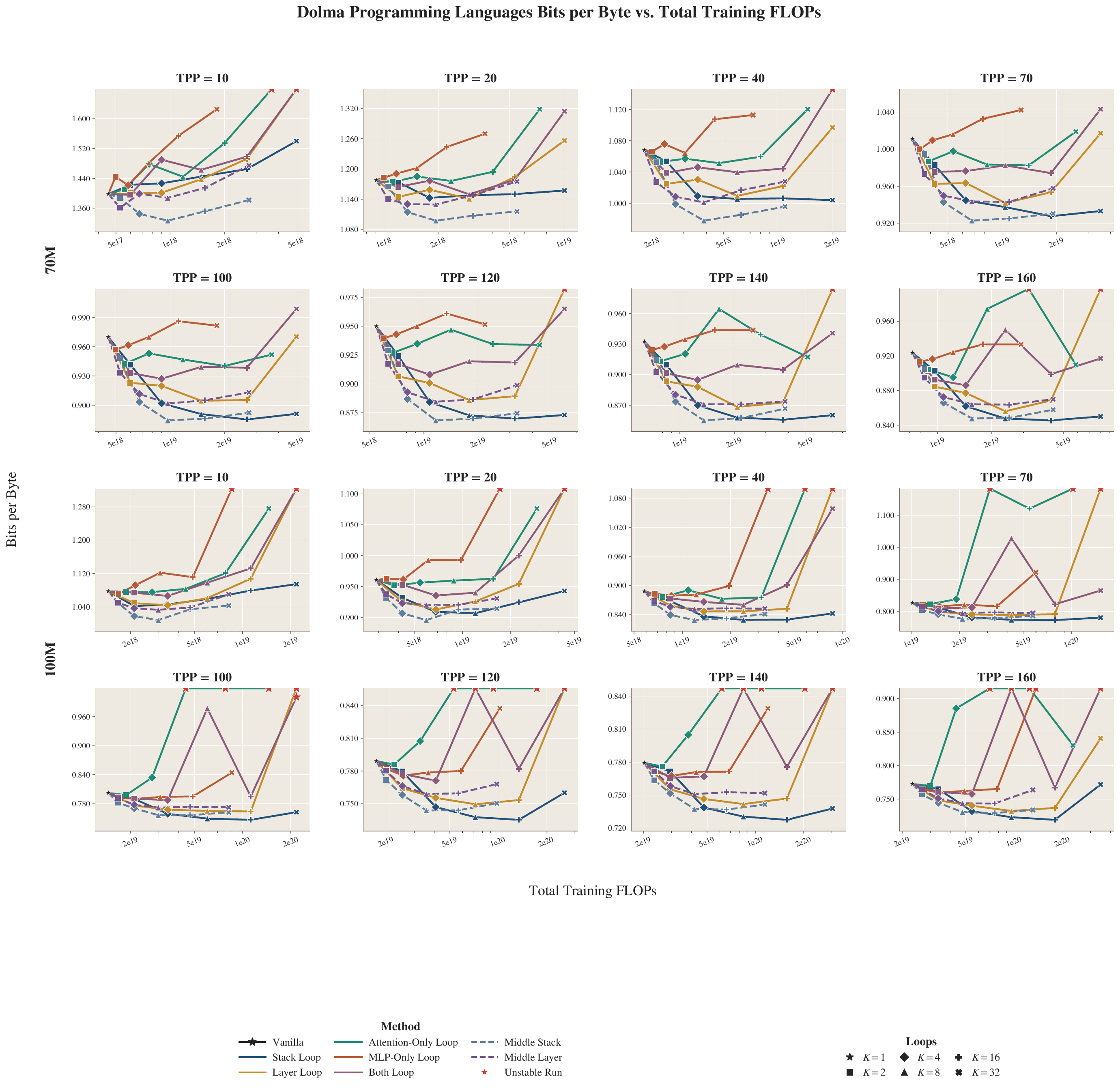}
  \caption{\textbf{Dolma programming-language BPB versus total training FLOPs: 70M and 100M models.}
  Results span TPP 10--160 and loop factors
  \(L\in\{1,2,4,8,16,32\}\). Middle-Stack Looping often provides strong
  quality improvements for a smaller increase in training compute, while
  spending additional FLOPs through larger loop factors eventually shows
  diminishing returns. Each panel fixes model scale and TPP. Lower BPB is
  better; red stars denote unstable runs as described in
  \autoref{sec:looping-appendix-coverage}.}
  \label{fig:loop-public-programming-flops-small}
\end{figure}

\begin{figure}[htbp]
  \centering
  \includegraphics[width=0.70\textwidth,height=0.70\textheight,keepaspectratio]{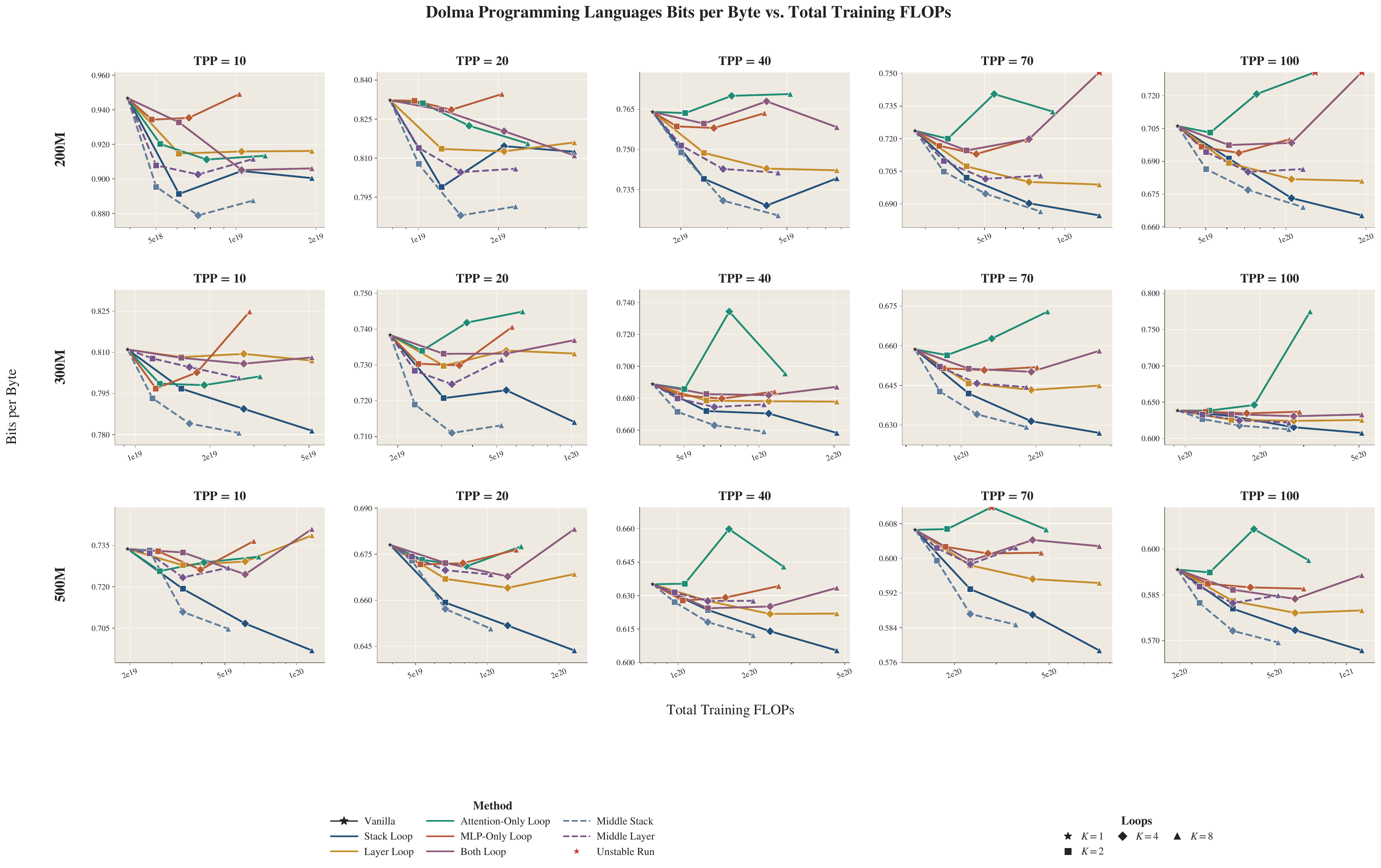}
  \caption{\textbf{Dolma programming-language BPB versus total training FLOPs: 200M, 300M, and 500M models.}
  Results span TPP 10--100 and loop factors
  \(L\in\{1,2,4,8\}\). Middle-Stack Looping remains competitive for smaller
  increases in training compute, while Full-Stack Looping can attain lower
  BPB when more compute is available, particularly at the larger model
  scales. Each panel fixes model scale and TPP. Lower BPB is better; red
  stars denote unstable runs as described in
  \autoref{sec:looping-appendix-coverage}.}
  \label{fig:loop-public-programming-flops-large}
\end{figure}

The Dolma programming-language results in
\autoref{fig:loop-public-programming-factor-small},
\autoref{fig:loop-public-programming-factor-large},
\autoref{fig:loop-public-programming-flops-small}, and
\autoref{fig:loop-public-programming-flops-large}
show a similar overall pattern, although the differences between looping
designs are more pronounced at lower training budgets. At smaller model
scales and lower TPP, Middle-Stack Looping is often especially competitive
with Full-Stack Looping. As model scale and TPP increase, Full-Stack Looping
more consistently attains lower BPB, particularly at 500M. When training
compute is considered, shallow Middle-Stack Looping remains attractive
because it can provide strong quality gains with a smaller increase in FLOPs.
As on the other corpora, the improvement from increasing the loop factor
\(L\) also tends to diminish at larger loop factors.

\clearpage

\subsubsection{GSM8K}
\label{sec:gsm8k_looping_ablations}

\begin{figure}[htbp]
  \centering
  \includegraphics[width=0.70\textwidth,height=0.70\textheight,keepaspectratio]{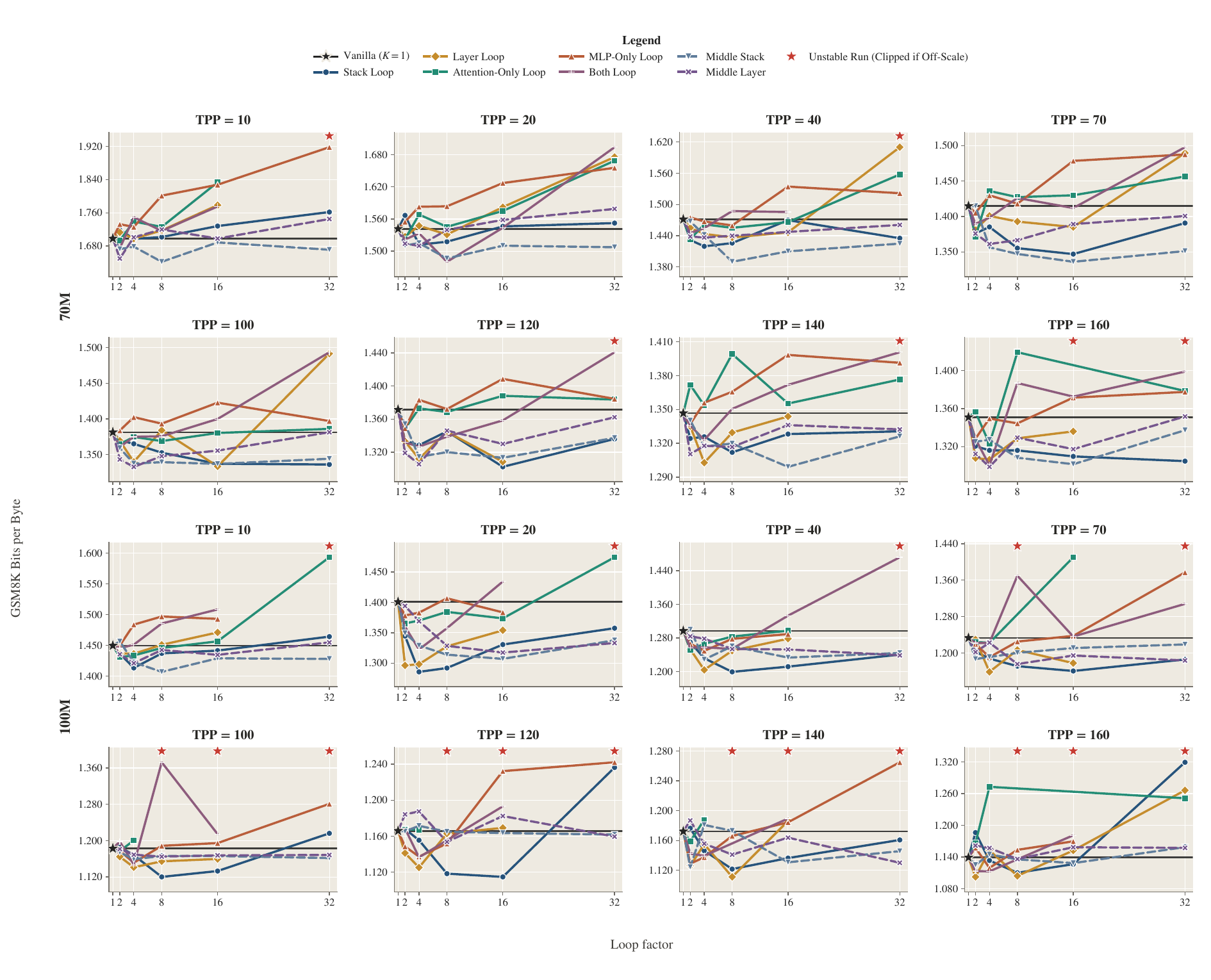}
  \caption{\textbf{GSM8K BPB versus loop factor \(L\): 70M and 100M models.}
  Results span TPP 10--160 and loop factors
  \(L\in\{1,2,4,8,16,32\}\). Several looping designs remain competitive,
  while the improvement from increasing \(L\) generally diminishes at larger
  loop factors. Each panel fixes model scale and TPP. BPB measures
  teacher-forced likelihood rather than answer accuracy; lower BPB is better.
  Red stars denote unstable runs as described in
  \autoref{sec:looping-appendix-coverage}.}
  \label{fig:loop-public-gsm8k-small}
\end{figure}

\begin{figure}[htbp]
  \centering
  \includegraphics[width=0.70\textwidth,height=0.70\textheight,keepaspectratio]{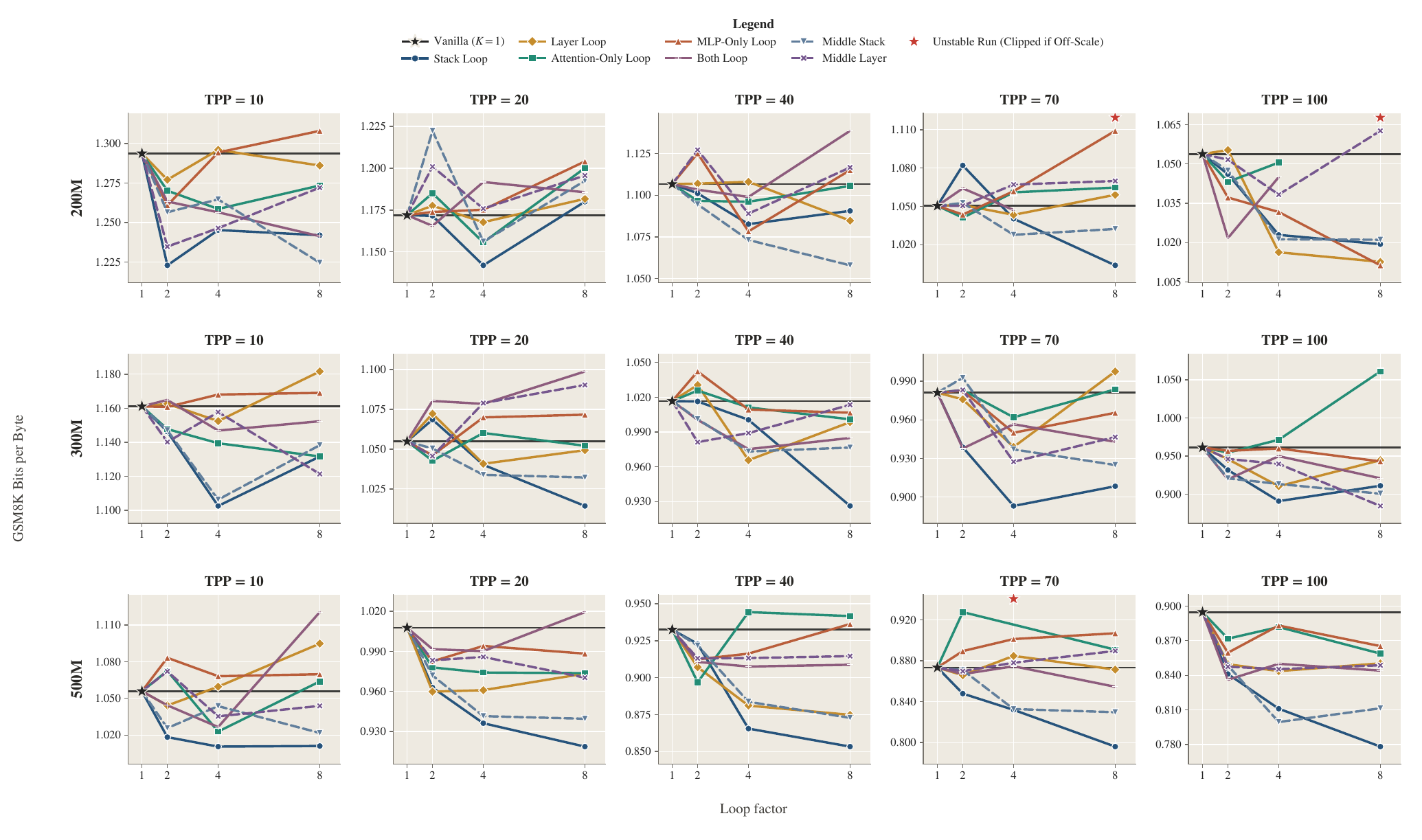}
  \caption{\textbf{GSM8K BPB versus loop factor \(L\): 200M, 300M, and 500M models.}
  Results span TPP 10--100 and loop factors
  \(L\in\{1,2,4,8\}\). Full-Stack Looping becomes more favorable as model
  scale and TPP increase, although the differences across looping designs
  remain less uniform than on the general pretraining corpora. Each panel
  fixes model scale and TPP. BPB measures teacher-forced likelihood rather
  than answer accuracy; lower BPB is better.}
  \label{fig:loop-public-gsm8k-large}
\end{figure}

The GSM8K results in
\autoref{fig:loop-public-gsm8k-small} and
\autoref{fig:loop-public-gsm8k-large}
show more variation across looping designs than the general pretraining
corpora. At smaller model scales and lower TPP, several designs remain
competitive, with no single design consistently attaining the lowest BPB.
As model scale and TPP increase, Full-Stack Looping becomes more favorable,
although the trend is less uniform than on DCLM. The smaller-model sweeps
also show diminishing returns as the loop factor \(L\) increases: BPB
typically improves initially, but the reduction becomes smaller at larger
\(L\) and can eventually saturate or reverse. These results measure
teacher-forced likelihood on GSM8K rather than generated-answer accuracy.

\clearpage

\subsubsection{MATH}
\label{sec:math_looping_ablations}

\begin{figure}[htbp]
  \centering
  \includegraphics[width=0.70\textwidth,height=0.70\textheight,keepaspectratio]{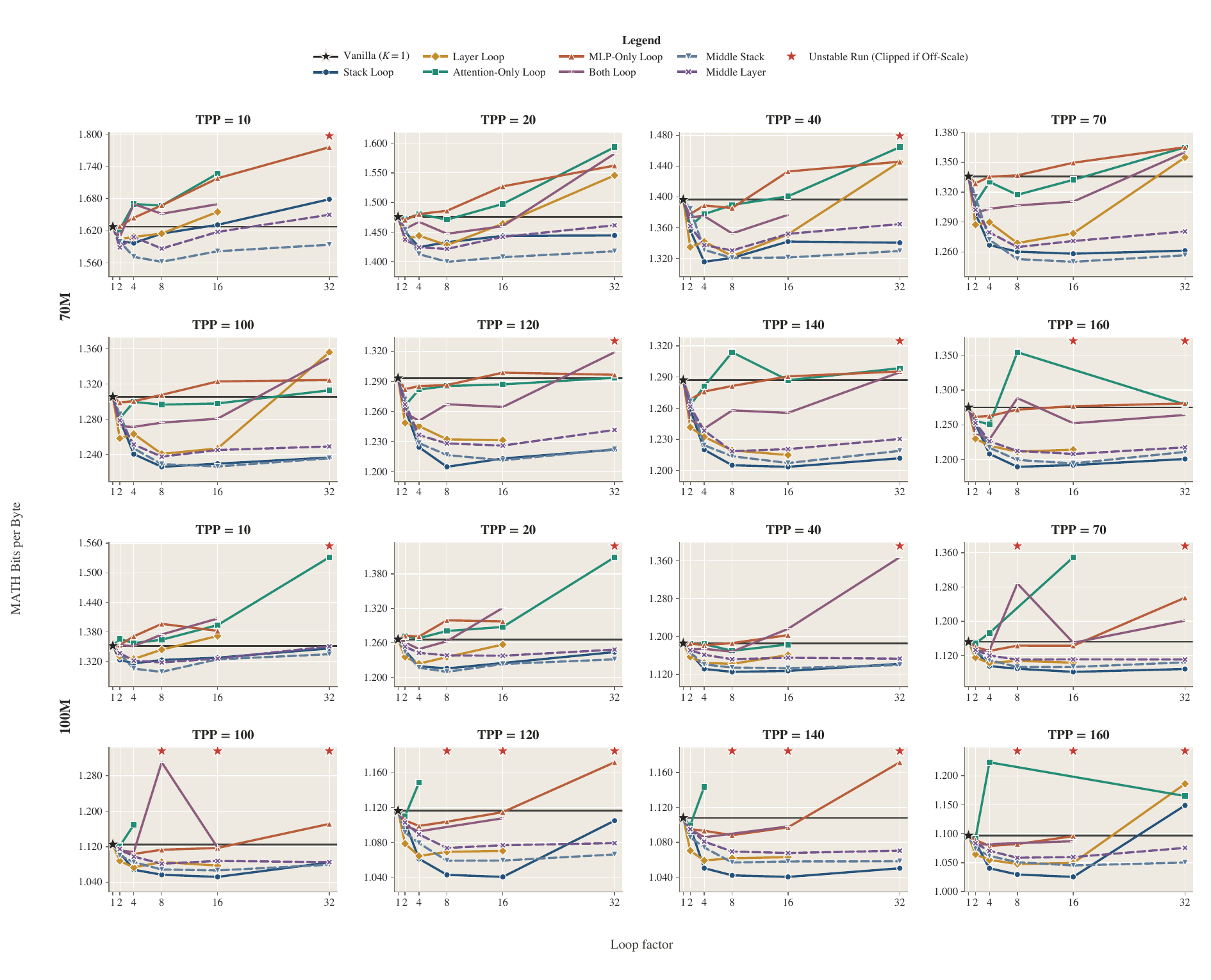}
  \caption{\textbf{MATH BPB versus loop factor \(L\): 70M and 100M models.}
  Results span TPP 10--160 and loop factors
  \(L\in\{1,2,4,8,16,32\}\). Several looping designs remain competitive at
  smaller training budgets, while the improvement from increasing \(L\)
  generally diminishes at larger loop factors. Each panel fixes model scale
  and TPP. BPB measures teacher-forced likelihood rather than answer accuracy;
  lower BPB is better. Red stars denote unstable runs as described in
  \autoref{sec:looping-appendix-coverage}.}
  \label{fig:loop-public-math-small}
\end{figure}

\begin{figure}[htbp]
  \centering
  \includegraphics[width=0.70\textwidth,height=0.70\textheight,keepaspectratio]{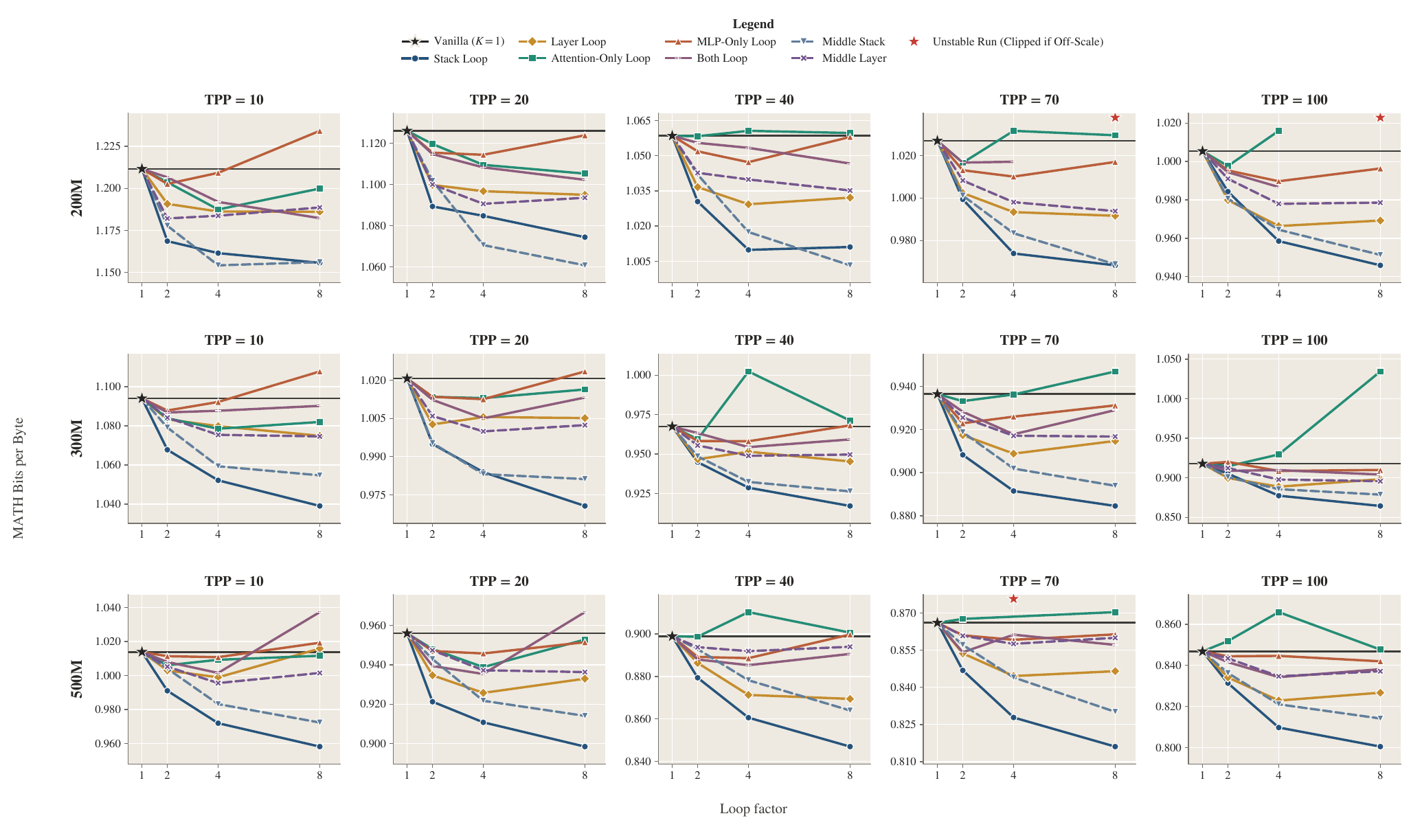}
  \caption{\textbf{MATH BPB versus loop factor \(L\): 200M, 300M, and 500M models.}
  Results span TPP 10--100 and loop factors
  \(L\in\{1,2,4,8\}\). Full-Stack Looping becomes more consistently favorable
  as model scale and TPP increase. Each panel fixes model scale and TPP.
  BPB measures teacher-forced likelihood rather than answer accuracy; lower
  BPB is better.}
  \label{fig:loop-public-math-large}
\end{figure}

The MATH results in
\autoref{fig:loop-public-math-small} and
\autoref{fig:loop-public-math-large}
show a clearer preference for Full-Stack Looping than GSM8K, particularly as
model scale and TPP increase. At smaller scales and lower TPP, several looping
designs remain competitive, while at larger scales and higher TPP full-stack
looping more consistently attains lower BPB. The small-model sweeps also show
diminishing returns as the loop factor \(L\) increases: the reduction in BPB
generally becomes smaller at larger \(L\), and deeper looping can eventually
cease to improve quality. These results measure teacher-forced likelihood on
MATH rather than mathematical answer accuracy.

\clearpage

\subsubsection{HumanEval}
\label{sec:humaneval_looping_ablations}

\begin{figure}[htbp]
  \centering
  \includegraphics[width=0.70\textwidth,height=0.70\textheight,keepaspectratio]{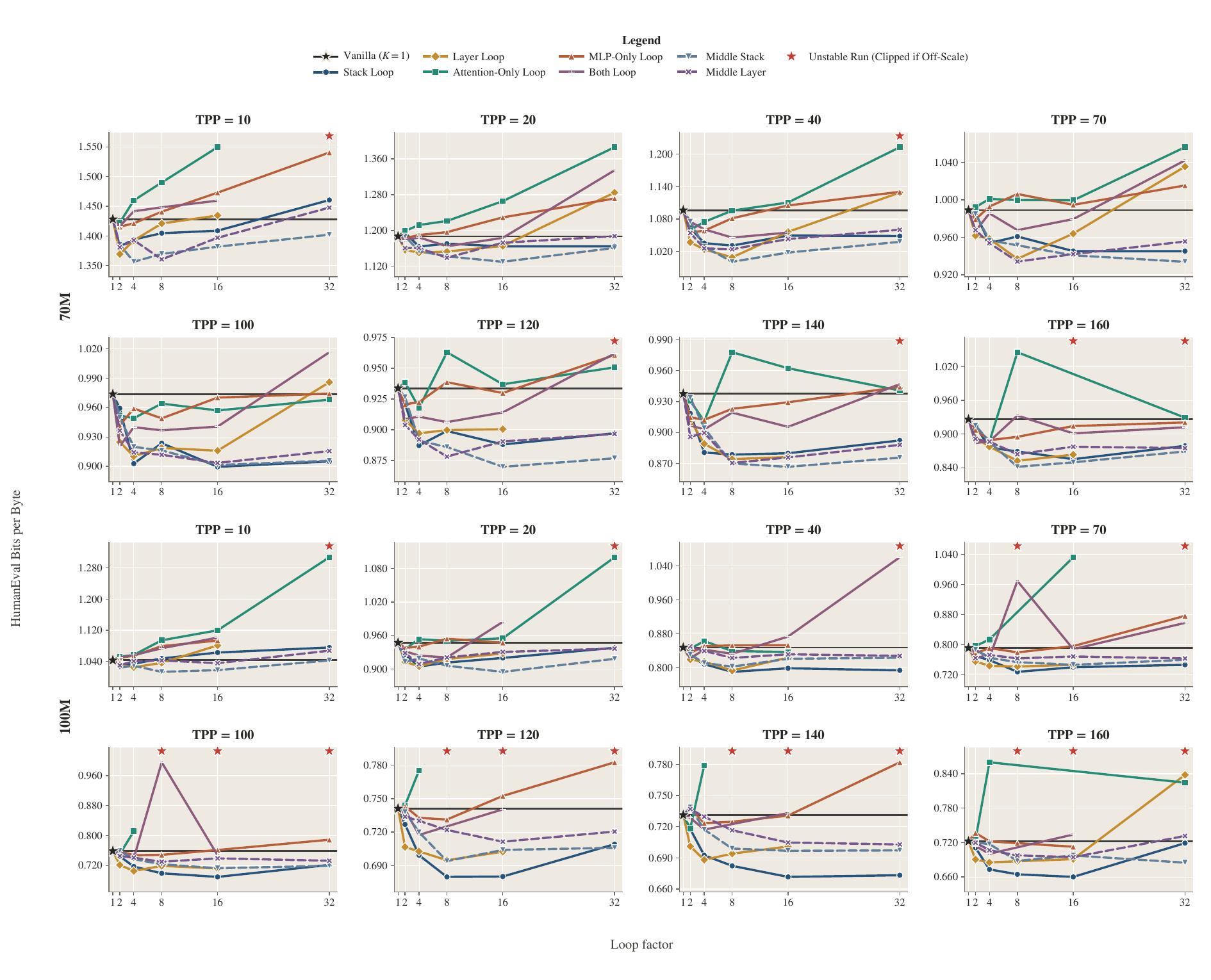}
  \caption{\textbf{HumanEval BPB versus loop factor \(L\): 70M and 100M models.}
  Results span TPP 10--160 and loop factors
  \(L\in\{1,2,4,8,16,32\}\). The preferred design varies across training
  budgets, and increasing \(L\) does not always improve BPB. Each panel
  fixes model scale and TPP. BPB measures teacher-forced likelihood rather
  than execution accuracy; lower BPB is better. Red stars denote unstable
  runs as described in \autoref{sec:looping-appendix-coverage}.}
  \label{fig:loop-public-humaneval-small}
\end{figure}

\begin{figure}[htbp]
  \centering
  \includegraphics[width=0.70\textwidth,height=0.70\textheight,keepaspectratio]{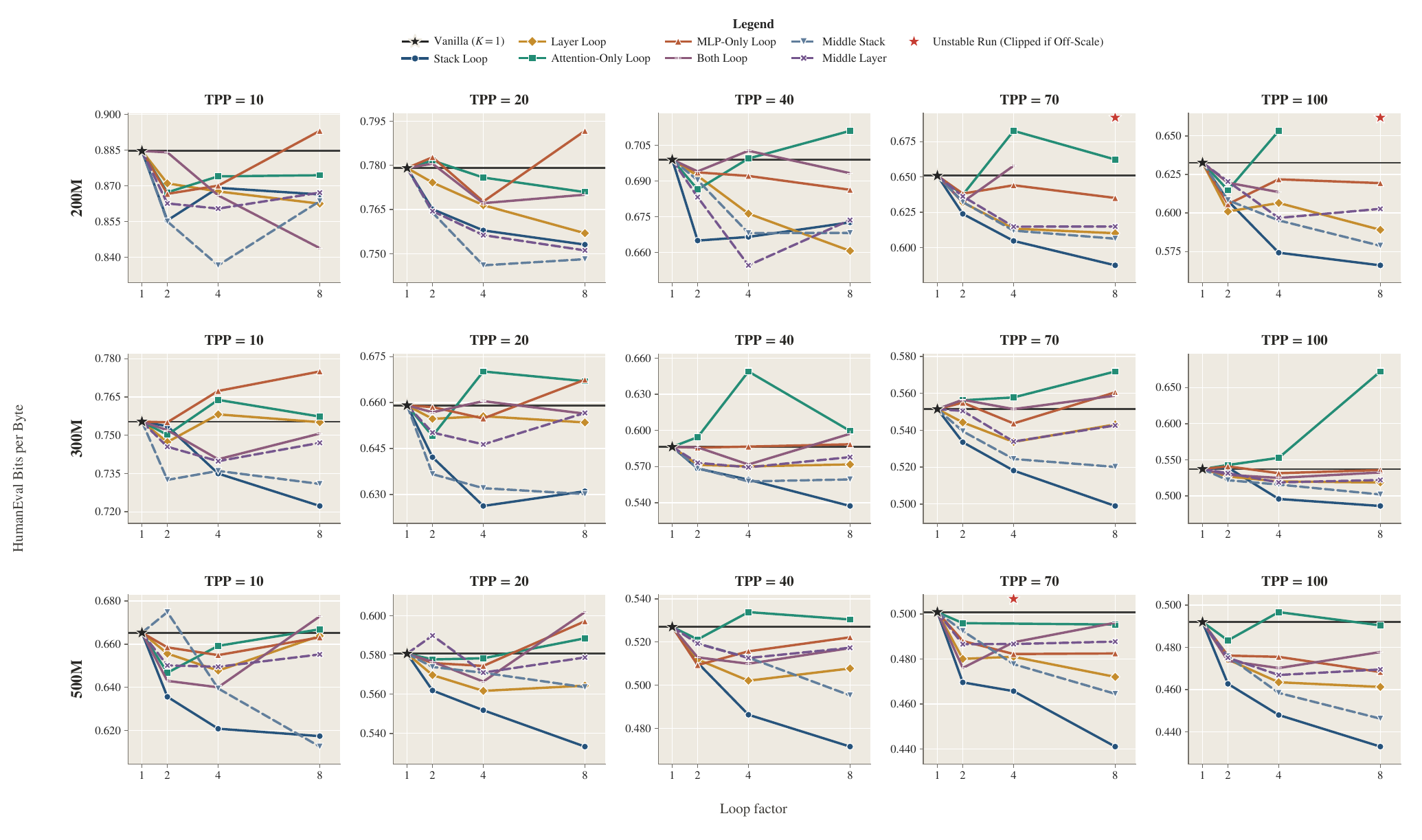}
  \caption{\textbf{HumanEval BPB versus loop factor \(L\): 200M, 300M, and 500M models.}
  Results span TPP 10--100 and loop factors
  \(L\in\{1,2,4,8\}\). Full-Stack Looping becomes more favorable at larger
  model scales and higher TPP, although it does not dominate every setting.
  Each panel fixes model scale and TPP. Lower BPB is better.}
  \label{fig:loop-public-humaneval-large}
\end{figure}

The HumanEval results in
\autoref{fig:loop-public-humaneval-small} and
\autoref{fig:loop-public-humaneval-large}
show that the preferred looping design changes with model scale and TPP.
At smaller scales and lower TPP, Full-Stack Looping is not consistently
better than the alternatives; it becomes more favorable as scale and TPP
increase. The deeper small-model sweeps also show diminishing returns:
additional loops can eventually stop reducing BPB or reverse earlier gains.
These results measure teacher-forced likelihood on reference code, not
whether generated programs pass the HumanEval tests.

\clearpage

\subsubsection{MBPP}
\label{sec:mbpp_looping_ablations}

\begin{figure}[htbp]
  \centering
  \includegraphics[width=0.70\textwidth,height=0.70\textheight,keepaspectratio]{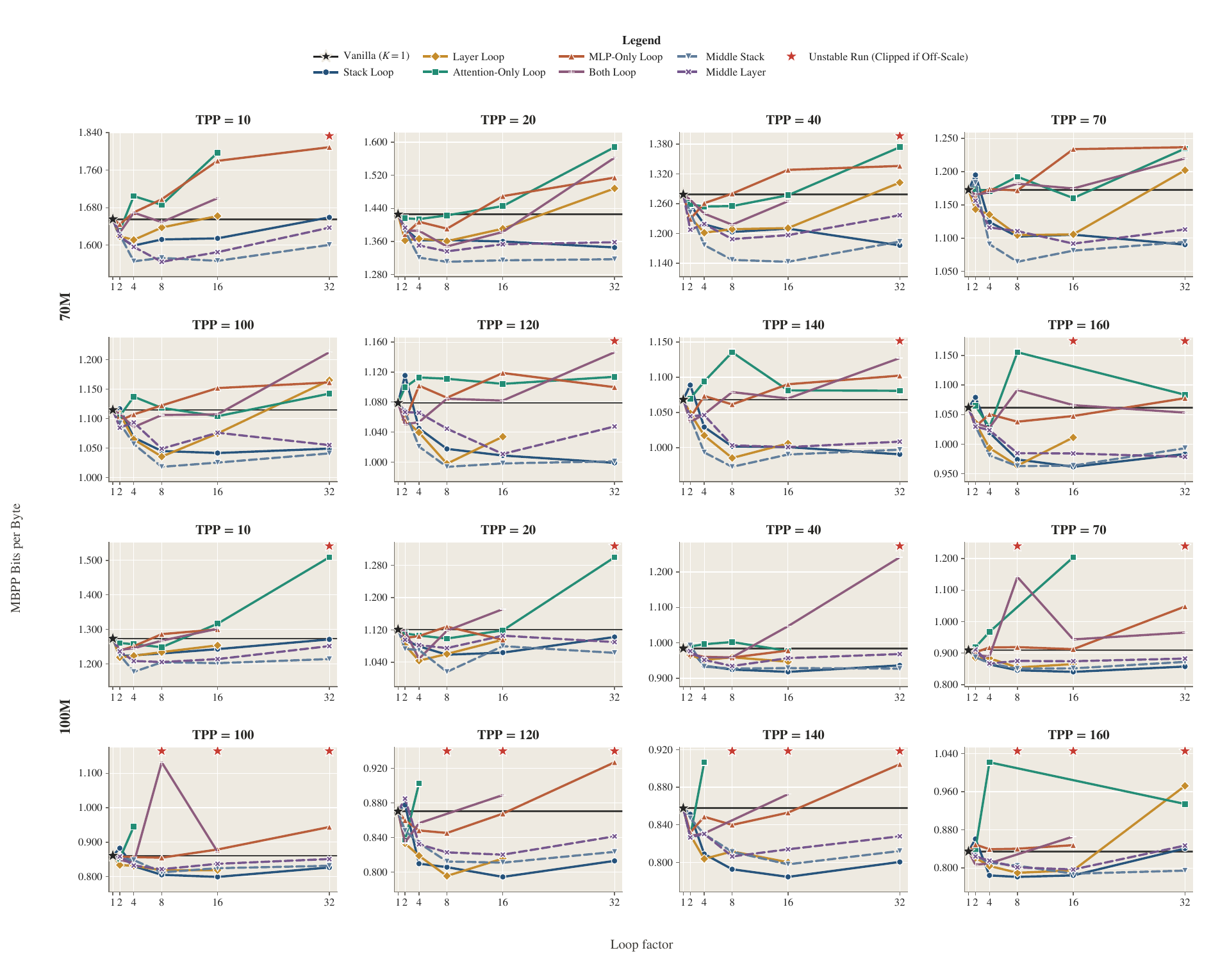}
  \caption{\textbf{MBPP BPB versus loop factor \(L\): 70M and 100M models.}
  Results span TPP 10--160 and loop factors
  \(L\in\{1,2,4,8,16,32\}\). Several looping designs remain competitive,
  while the improvement from increasing \(L\) generally diminishes at larger
  loop factors. Each panel fixes model scale and TPP. BPB measures
  teacher-forced likelihood rather than execution accuracy;
  lower BPB is better. Red stars denote unstable runs as described in
  \autoref{sec:looping-appendix-coverage}.}
  \label{fig:loop-public-mbpp-small}
\end{figure}

\begin{figure}[htbp]
  \centering
  \includegraphics[width=0.70\textwidth,height=0.70\textheight,keepaspectratio]{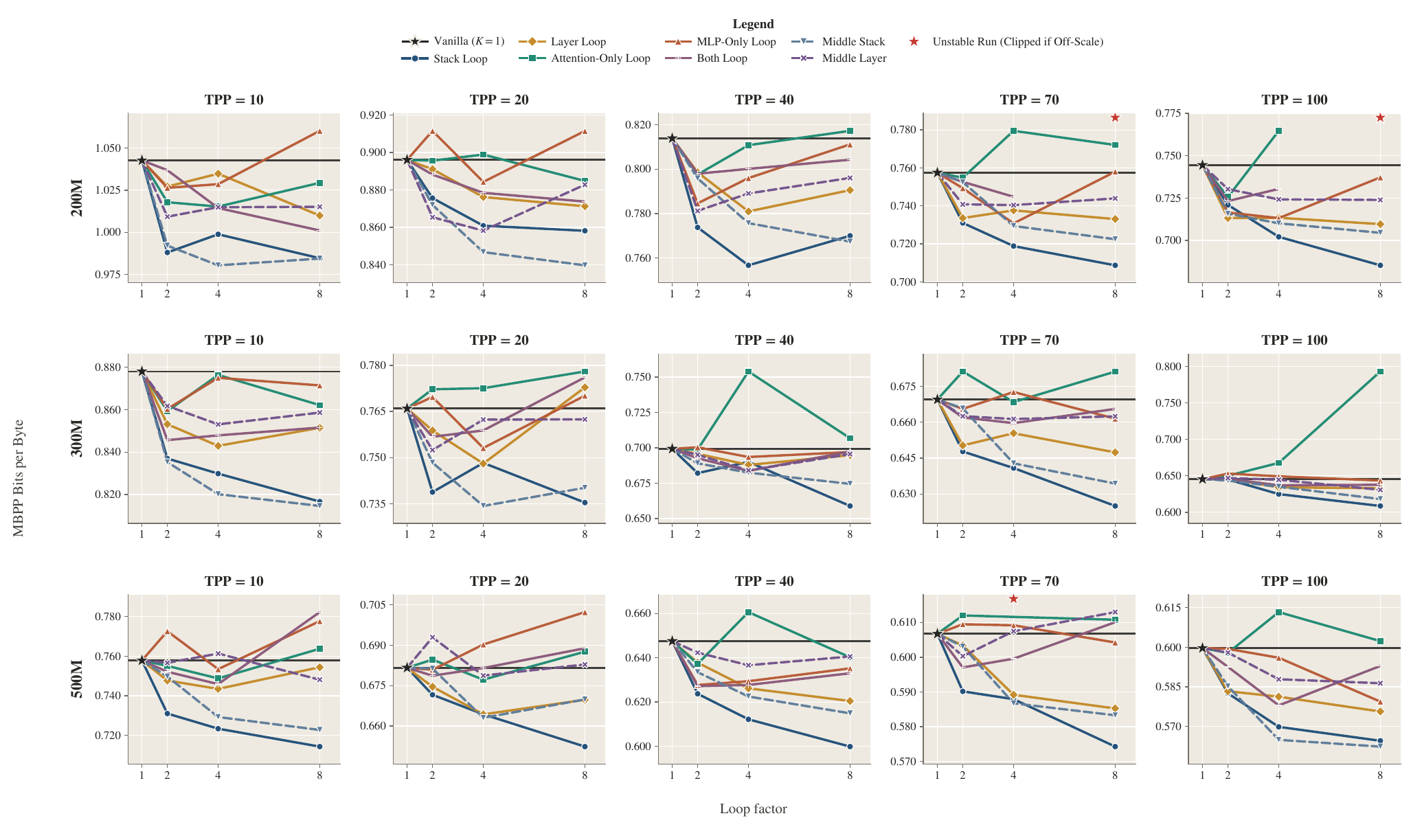}
  \caption{\textbf{MBPP BPB versus loop factor \(L\): 200M, 300M, and 500M models.}
  Results span TPP 10--100 and loop factors
  \(L\in\{1,2,4,8\}\). Full-Stack Looping becomes more favorable as model
  scale and TPP increase, although the preferred design remains more
  task-dependent than on the general pretraining corpora. Each panel fixes
  model scale and TPP. Lower BPB is better.}
  \label{fig:loop-public-mbpp-large}
\end{figure}

The MBPP results in
\autoref{fig:loop-public-mbpp-small} and
\autoref{fig:loop-public-mbpp-large}
show substantial variation across looping designs, particularly at smaller
scales and lower TPP. Full-Stack Looping becomes more favorable as model
scale and TPP increase, but the preference is less consistent than on the
general pretraining corpora. The small-model sweeps also exhibit diminishing
returns as the loop factor \(L\) increases: additional looping produces
progressively smaller reductions in BPB and can eventually cease to improve
quality. These results measure teacher-forced likelihood on MBPP code rather
than execution accuracy or pass@\(k\).

\clearpage

\subsection{Looping at 1B Scale}
\label{sec:looping-public-1b}

We test whether the quality--compute trade-off observed at smaller scales
persists at 1B parameters across DCLM (\autoref{sec:looping-public-1b-dclm}), C4 (\autoref{sec:looping-public-1b-c4}), Dolma v1.5 (\autoref{sec:looping-public-1b-dolma-v15}), Dolma programming
languages (\autoref{sec:looping-public-1b-dolma-programming}), RedPajama (\autoref{sec:looping-public-1b-redpajama}), and WikiText-103 (\autoref{sec:looping-public-1b-wikitext103}). We focus on the designs that were
most competitive in the smaller-scale sweeps: full-stack and middle-stack
looping, together with Per-Layer Looping at shallow loop counts. At TPP 10
and 40, we compare these designs at \(L=2\) and \(L=4\) against the
non-looped \(L=1\) model. The corpus-specific loop-count and training-FLOP
comparisons appear in
\autoref{fig:loop-public-1b-dclm-factor}--\autoref{fig:loop-public-1b-wikitext103-flops}.
Across these corpora, the same broad conclusion holds: looping generally
lowers BPB; Full-Stack Looping tends to achieve the lowest BPB at a matched
loop count, while Middle-Stack Looping obtains gains with fewer training
FLOPs. Thus, even at 1B, the preferred schedule depends on whether the
priority is quality at a fixed stored parameter count or the cost of
additional computation.

\subsubsection{DCLM}
\label{sec:looping-public-1b-dclm}

\begin{figure}[htbp]
  \centering
  \includegraphics[width=0.60\textwidth,height=0.7\textheight,keepaspectratio]{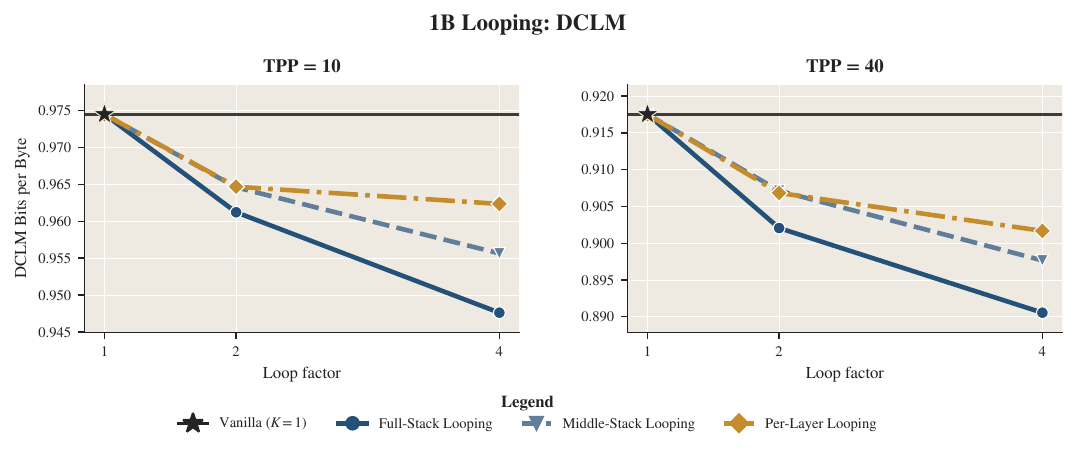}
  \caption{\textbf{DCLM BPB versus loop factor \(L\) at 1B.}
  Full-stack, middle-stack, and Per-Layer Looping are evaluated at
  TPP 10 and 40 over \(L\in\{1,2,4\}\). All three improve over the
  non-looped \(L=1\) model, with Full-Stack Looping attaining the
  lowest BPB at matched loop counts. Lower is better.}
  \label{fig:loop-public-1b-dclm-factor}
\end{figure}

\begin{figure}[htbp]
  \centering
  \includegraphics[width=0.60\textwidth,height=0.7\textheight,keepaspectratio]{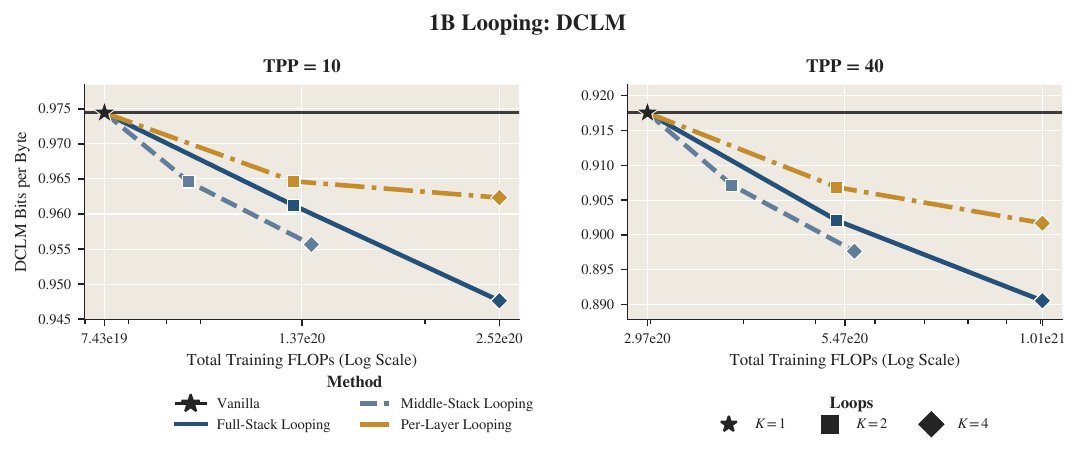}
  \caption{\textbf{DCLM BPB versus training FLOPs at 1B.}
  The same models are compared at TPP 10 and 40. Middle-Stack Looping
  improves BPB with a smaller increase in FLOPs, while full-stack
  looping reaches lower BPB with more compute. The horizontal axis is
  logarithmic; marker shape denotes \(L\). Lower is better.}
  \label{fig:loop-public-1b-dclm-flops}
\end{figure}

On DCLM, all three looping schedules improve over the non-looped model,
and Full-Stack Looping attains the lowest BPB at matched loop counts
(\autoref{fig:loop-public-1b-dclm-factor}). Middle-Stack Looping uses
fewer additional training FLOPs, while Full-Stack Looping can spend more
compute to reach lower BPB
(\autoref{fig:loop-public-1b-dclm-flops}).

\subsubsection{C4}
\label{sec:looping-public-1b-c4}

\begin{figure}[htbp]
  \centering
  \includegraphics[width=0.60\textwidth,height=0.7\textheight,keepaspectratio]{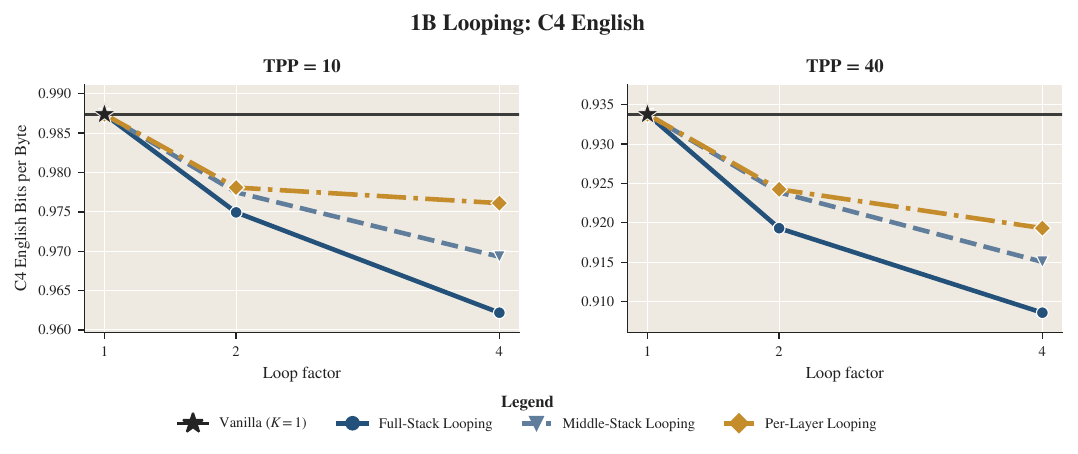}
  \caption{\textbf{C4 BPB versus loop factor \(L\) at 1B.}
  Non-looped, full-stack, middle-stack, and per-layer models are
  compared at TPP 10 and 40 over \(L\in\{1,2,4\}\). Each panel fixes
  TPP; lower BPB is better.}
  \label{fig:loop-public-1b-c4-factor}
\end{figure}

\begin{figure}[htbp]
  \centering
  \includegraphics[width=0.60\textwidth,height=0.7\textheight,keepaspectratio]{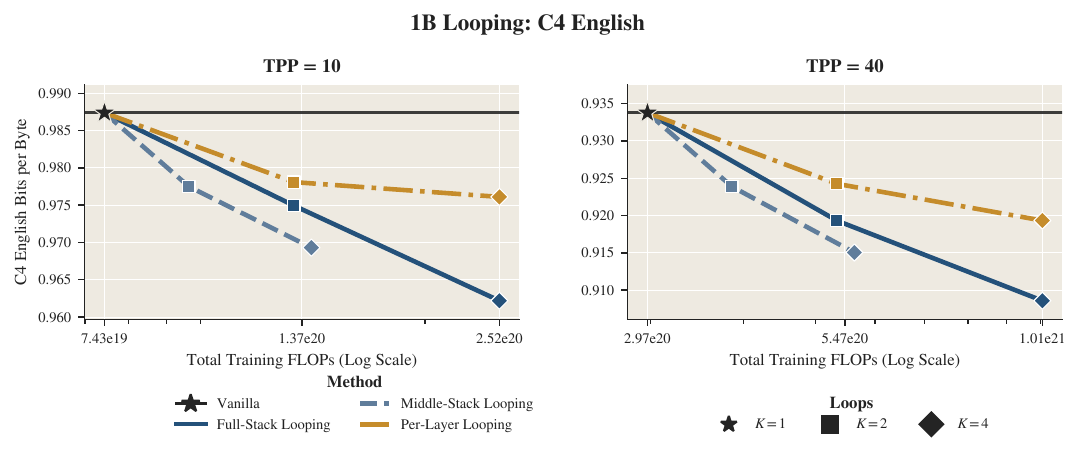}
  \caption{\textbf{C4 BPB versus training FLOPs at 1B.}
  The same checkpoints are plotted against total training FLOPs at
  TPP 10 and 40. The horizontal axis is logarithmic; marker shape
  denotes \(L\). Lower BPB is better.}
  \label{fig:loop-public-1b-c4-flops}
\end{figure}

The C4 panels repeat the matched loop-count and training-compute
comparisons on a second public text corpus
(\autoref{fig:loop-public-1b-c4-factor} and
\autoref{fig:loop-public-1b-c4-flops}).

\subsubsection{Dolma v1.5}
\label{sec:looping-public-1b-dolma-v15}

\begin{figure}[htbp]
  \centering
  \includegraphics[width=0.60\textwidth,height=0.7\textheight,keepaspectratio]{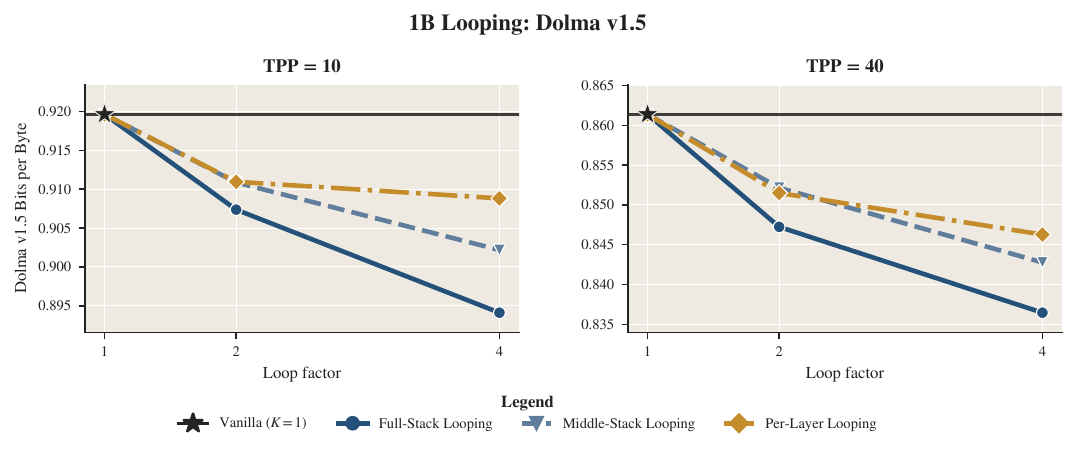}
  \caption{\textbf{Dolma v1.5 BPB versus loop factor \(L\) at 1B.}
  Non-looped, full-stack, middle-stack, and per-layer models are
  compared at TPP 10 and 40 over \(L\in\{1,2,4\}\). Each panel fixes
  TPP; lower BPB is better.}
  \label{fig:loop-public-1b-dolma-v15-factor}
\end{figure}

\begin{figure}[htbp]
  \centering
  \includegraphics[width=0.60\textwidth,height=0.7\textheight,keepaspectratio]{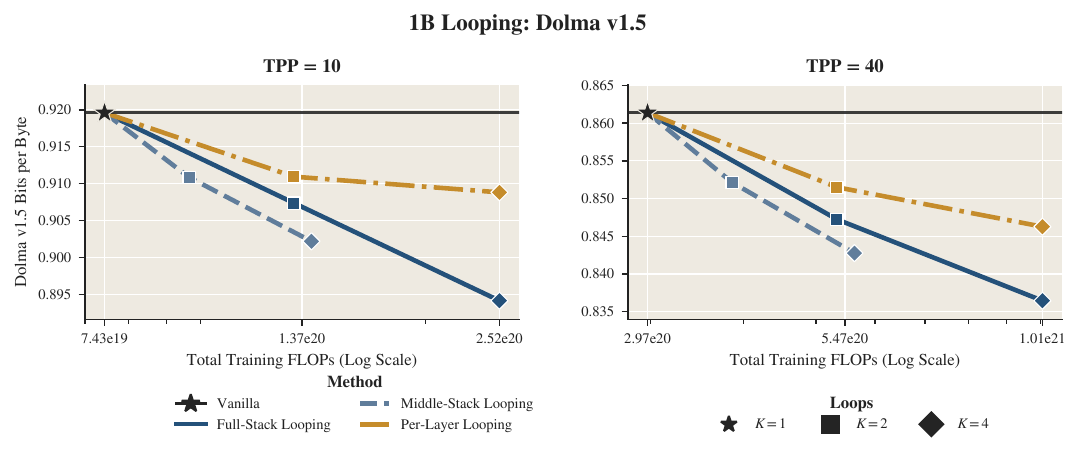}
  \caption{\textbf{Dolma v1.5 BPB versus training FLOPs at 1B.}
  The same checkpoints are plotted against total training FLOPs at
  TPP 10 and 40. The horizontal axis is logarithmic; marker shape
  denotes \(L\). Lower BPB is better.}
  \label{fig:loop-public-1b-dolma-v15-flops}
\end{figure}

We evaluate the same 1B checkpoints on Dolma v1.5 to compare their
quality and compute trade-offs on another pretraining corpus
(\autoref{fig:loop-public-1b-dolma-v15-factor} and
\autoref{fig:loop-public-1b-dolma-v15-flops}).

\subsubsection{Dolma Programming Languages}
\label{sec:looping-public-1b-dolma-programming}

\begin{figure}[htbp]
  \centering
  \includegraphics[width=0.70\textwidth,height=0.70\textheight,keepaspectratio]{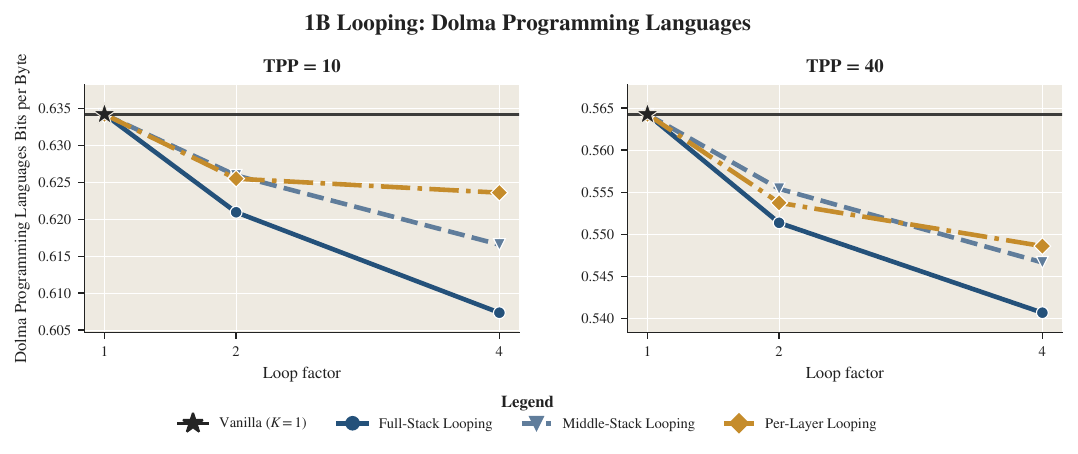}
  \caption{\textbf{Dolma programming-languages BPB versus loop factor \(L\) at 1B.}
  Non-looped, full-stack, middle-stack, and per-layer models are
  compared at TPP 10 and 40 over \(L\in\{1,2,4\}\). Each panel fixes
  TPP; lower BPB is better.}
  \label{fig:loop-public-1b-dolma-programming-factor}
\end{figure}

\begin{figure}[htbp]
  \centering
  \includegraphics[width=0.70\textwidth,height=0.70\textheight,keepaspectratio]{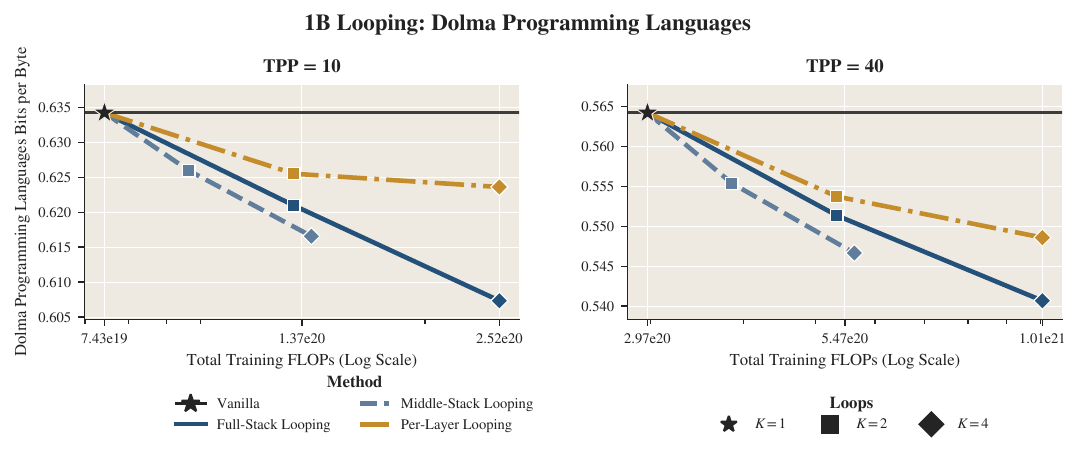}
  \caption{\textbf{Dolma programming-languages BPB versus training FLOPs at 1B.}
  The same checkpoints are plotted against total training FLOPs at
  TPP 10 and 40. The horizontal axis is logarithmic; marker shape
  denotes \(L\). Lower BPB is better.}
  \label{fig:loop-public-1b-dolma-programming-flops}
\end{figure}

The Dolma programming-languages split tests whether the preferred
looping schedule changes when quality is measured on code
(\autoref{fig:loop-public-1b-dolma-programming-factor} and
\autoref{fig:loop-public-1b-dolma-programming-flops}).

\subsubsection{RedPajama}
\label{sec:looping-public-1b-redpajama}

\begin{figure}[htbp]
  \centering
  \includegraphics[width=0.70\textwidth,height=0.70\textheight,keepaspectratio]{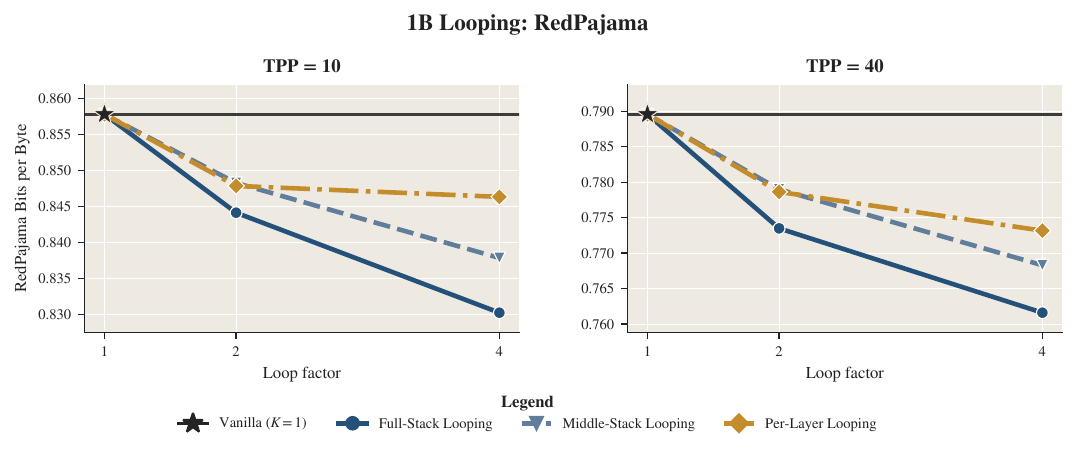}
  \caption{\textbf{RedPajama BPB versus loop factor \(L\) at 1B.}
  Non-looped, full-stack, middle-stack, and per-layer models are
  compared at TPP 10 and 40 over \(L\in\{1,2,4\}\). Each panel fixes
  TPP; lower BPB is better.}
  \label{fig:loop-public-1b-redpajama-factor}
\end{figure}

\begin{figure}[htbp]
  \centering
  \includegraphics[width=0.70\textwidth,height=0.70\textheight,keepaspectratio]{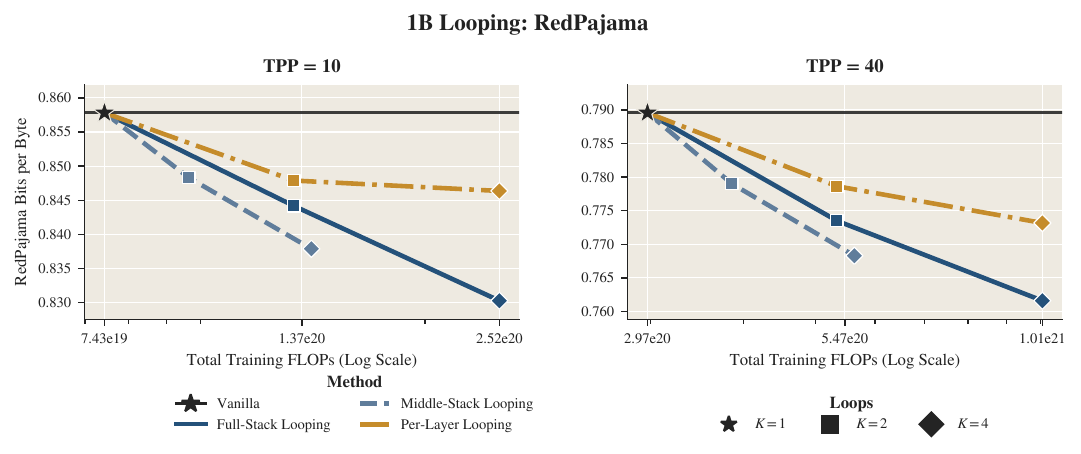}
  \caption{\textbf{RedPajama BPB versus training FLOPs at 1B.}
  The same checkpoints are plotted against total training FLOPs at
  TPP 10 and 40. The horizontal axis is logarithmic; marker shape
  denotes \(L\). Lower BPB is better.}
  \label{fig:loop-public-1b-redpajama-flops}
\end{figure}

The RedPajama panels provide another test of whether the 1B schedule
comparison transfers beyond DCLM
(\autoref{fig:loop-public-1b-redpajama-factor} and
\autoref{fig:loop-public-1b-redpajama-flops}).

\subsubsection{WikiText-103}
\label{sec:looping-public-1b-wikitext103}

\begin{figure}[htbp]
  \centering
  \includegraphics[width=0.70\textwidth,height=0.70\textheight,keepaspectratio]{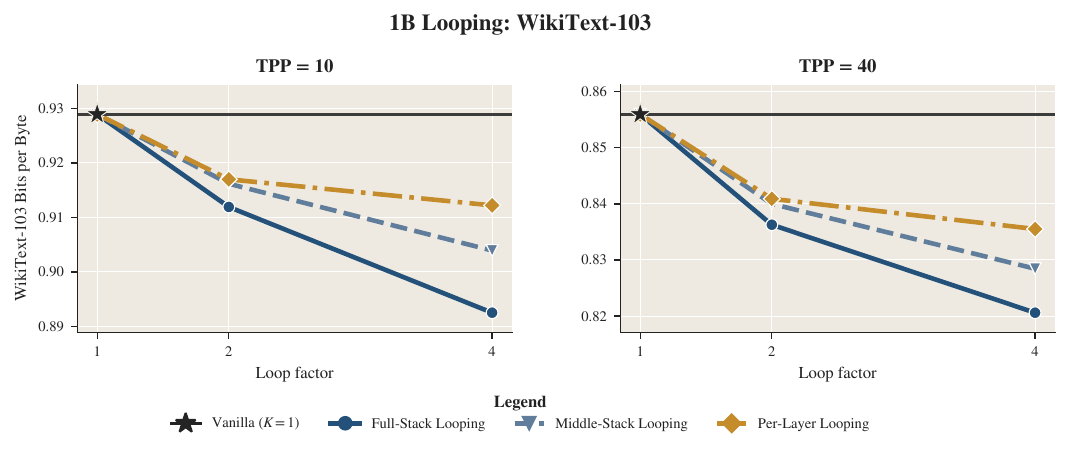}
  \caption{\textbf{WikiText-103 BPB versus loop factor \(L\) at 1B.}
  Non-looped, full-stack, middle-stack, and per-layer models are
  compared at TPP 10 and 40 over \(L\in\{1,2,4\}\). Each panel fixes
  TPP; lower BPB is better.}
  \label{fig:loop-public-1b-wikitext103-factor}
\end{figure}

\begin{figure}[htbp]
  \centering
  \includegraphics[width=0.70\textwidth,height=0.70\textheight,keepaspectratio]{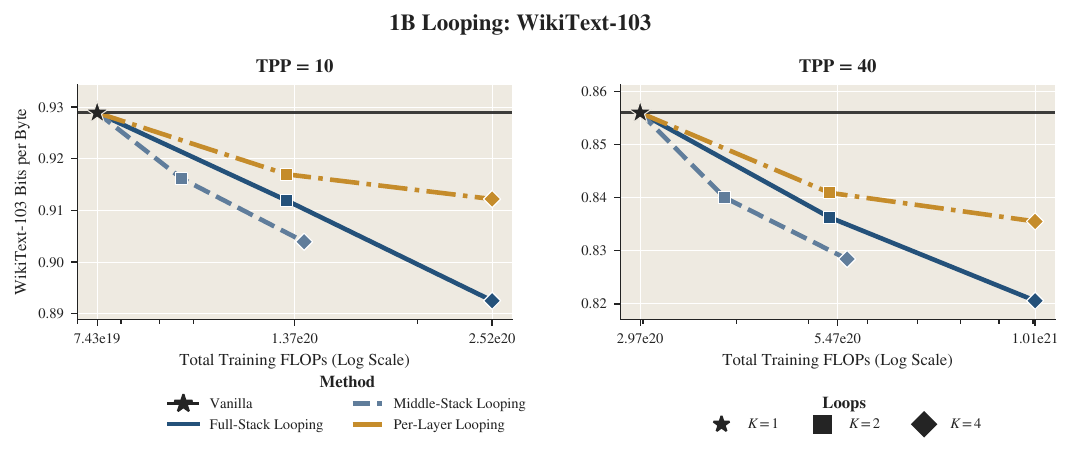}
  \caption{\textbf{WikiText-103 BPB versus training FLOPs at 1B.}
  The same checkpoints are plotted against total training FLOPs at
  TPP 10 and 40. The horizontal axis is logarithmic; marker shape
  denotes \(L\). Lower BPB is better.}
  \label{fig:loop-public-1b-wikitext103-flops}
\end{figure}

Finally, the WikiText-103 panels show the matched 1B comparison on the
remaining public corpus
(\autoref{fig:loop-public-1b-wikitext103-factor} and
\autoref{fig:loop-public-1b-wikitext103-flops}).

\clearpage

\section{Additional Results for Loop-Specific Specialization}
\label{app:loop-lora}

This section provides the additional results supporting
\autoref{sec:loop-lora}. The main text uses 300M full-stack models on DCLM to
show that loop-specific corrections become more useful as the number of loops
increases. Here, we test whether this conclusion holds across model sizes,
training budgets, looping schedules, and evaluation datasets.
\autoref{app:loop-lora-accounting} describes the experimental setup and
parameter and compute accounting. \autoref{app:loop-lora-core-results}
reports the core sweep over 70M--300M models at TPP 10, 40, and 100 on DCLM,
C4 English, WikiText-103, Dolma v1.5, RedPajama, and Dolma Programming
Languages. For each corpus, we examine how the benefit of specialization
changes with loop count, how many parameters it adds, how it compares with
adding independently parameterized layers, and how much training compute it
requires. \autoref{app:loop-lora-math-code} reports teacher-forced likelihood
on GSM8K, MATH, HumanEval, and MBPP;
\autoref{app:loop-lora-across-designs} compares full-stack, per-layer,
middle-stack, and Middle-Layer Looping to determine where specialization is
useful; and \autoref{app:loop-lora-scale-extension} tests whether the
full-stack result persists at 500M and 1B. All results use bits per byte
(BPB), for which lower is better.

\subsection{Experimental Setup}
\label{app:loop-lora-accounting}

\paragraph{Models and coverage.}
The main sweep covers 70M, 100M, 200M, and 300M models at TPP 10, 40, and
100, with loop counts $L\in\{2,4,8\}$. We evaluate full-stack, per-layer,
middle-stack, and Middle-Layer Looping. For every setting, the baseline uses
shared weights in every loop, and we compare it with Loop LoRA, Loop
Matrix, and Loop Softmax. We classify a run as unstable and exclude it when
its globally averaged training loss exceeds 5. This criterion excludes only
the three specialization variants of the 300M per-layer model at $L=8$ and
TPP 100. Their points are omitted, and curves are not drawn across the gaps.
The targeted 500M and 1B experiments use Full-Stack Looping.

\paragraph{Loop-specific specialization.}
The baseline reuses the same weight matrix in every loop. Giving every loop a
different full weight matrix would allow specialization but would largely
remove the parameter savings from weight reuse. We instead retain a shared
weight matrix and add a small loop-specific low-rank correction.

\begin{definitionbox}[label=def:loop-lora-mixing]
{\textbf{Loop-specific specialization.}}
Let $\bm W\in\mathbb{R}^{d_{\mathrm{out}}\times d_{\mathrm{in}}}$ be shared
across $L$ loops. Loop $\ell$ uses
\[
\bm W^{(\ell)}\coloneqq\bm W+\Delta\bm W^{(\ell)}.
\]
Let $\{(\bm A_j,\bm B_j)\}_{j=1}^{L}$ denote $L$ low-rank components, with
$\bm A_j\in\mathbb{R}^{q\times d_{\mathrm{in}}}$ and
$\bm B_j\in\mathbb{R}^{d_{\mathrm{out}}\times q}$.

\textbf{Loop LoRA} assigns one component directly to each loop:
\[
\Delta\bm W^{(\ell)}
\coloneqq
\frac{\alpha}{q}\bm B_\ell\bm A_\ell.
\]

\textbf{Loop Matrix} learns an arbitrary linear combination:
\[
\Delta\bm W^{(\ell)}
\coloneqq
\frac{\alpha}{q}\sum_{j=1}^{L}M_{\ell j}\bm B_j\bm A_j,
\qquad
\bm M\in\mathbb{R}^{L\times L}.
\]
We initialize $\bm M$ to the identity. \\

\textbf{Loop Softmax} uses the same mixture with nonnegative coefficients
that sum to one:
\[
\Delta\bm W^{(\ell)}
\coloneqq
\frac{\alpha}{q}\sum_{j=1}^{L}
\operatorname{softmax}(\bm S_{\ell,:})_j\bm B_j\bm A_j,
\qquad
\bm S\in\mathbb{R}^{L\times L}.
\]
The logits initially favor the component corresponding to each loop.
\end{definitionbox}

Loop LoRA uses a single rank-$q$ component, so
$\operatorname{rank}(\Delta\bm W^{(\ell)})\leq q$. Loop Matrix and Loop
Softmax can instead combine several such components. By subadditivity of
matrix rank,
\[
\operatorname{rank}\!\left(\Delta\bm W^{(\ell)}\right)
\leq
\sum_{j:M_{\ell j}\neq 0}
\operatorname{rank}\!\left(\bm B_j\bm A_j\right)
\leq Lq.
\]
The mixed variants therefore increase the maximum rank that a loop-specific
correction can represent, although a learned correction need not attain this
upper bound. We use $q=8$ and unit scaling, $\alpha=q$, and apply the
corrections to all attention and SwiGLU MLP projections in every looped
Transformer block. The shared weights and corrections are trained jointly
from initialization

\paragraph{Parameter and compute accounting.}
For one adapted projection, the corrections add
$Lq(d_{\mathrm{in}}+d_{\mathrm{out}})$ parameters, plus $L^2$ mixing
coefficients for Loop Matrix or Loop Softmax. At fixed nominal TPP, these
parameters also slightly increase the token budget because we recompute
$D=\mathrm{TPP}\times P_{\mathrm{non\text{-}embed}}$ for each model. Reported
training FLOPs include the realized token count, repeated Transformer
execution, attention-score computation, and the low-rank corrections.

\subsection{Results from 70M to 300M}
\label{app:loop-lora-core-results}
\label{app:loop-lora-public-pretraining}
\label{app:loop-lora-public-depth}

We evaluate the models on DCLM, C4 English, WikiText-103, Dolma v1.5,
RedPajama, Dolma Programming Languages, GSM8K, MATH, HumanEval, and MBPP.

\clearpage

\subsubsection{DCLM}
\label{app:loop-lora-dclm}

As in the main text, we use DCLM as our primary evaluation corpus because it
is the largest corpus in our evaluation suite. Here, we use DCLM to study four
questions: how the benefit of specialization changes with the number of
loops, how many additional parameters it requires, how it compares with
adding independently parameterized layers, and how the comparison changes
when training FLOPs are taken into account. We evaluate 70M--300M full-stack
models at TPP 10, 40, and 100, with the available physical-depth comparison
also including 500M models.

\begin{figure*}[ht]
  \centering
  \begin{minipage}[t]{0.485\textwidth}
    \centering
    \includegraphics[width=\linewidth]{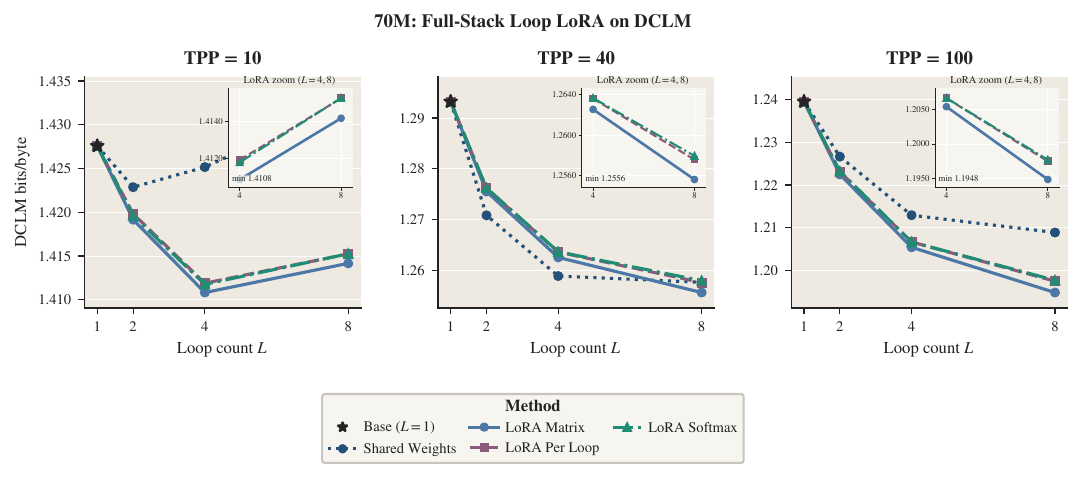}
    \par\small\textbf{(a) 70M}
  \end{minipage}
  \hfill
  \begin{minipage}[t]{0.485\textwidth}
    \centering
    \includegraphics[width=\linewidth]{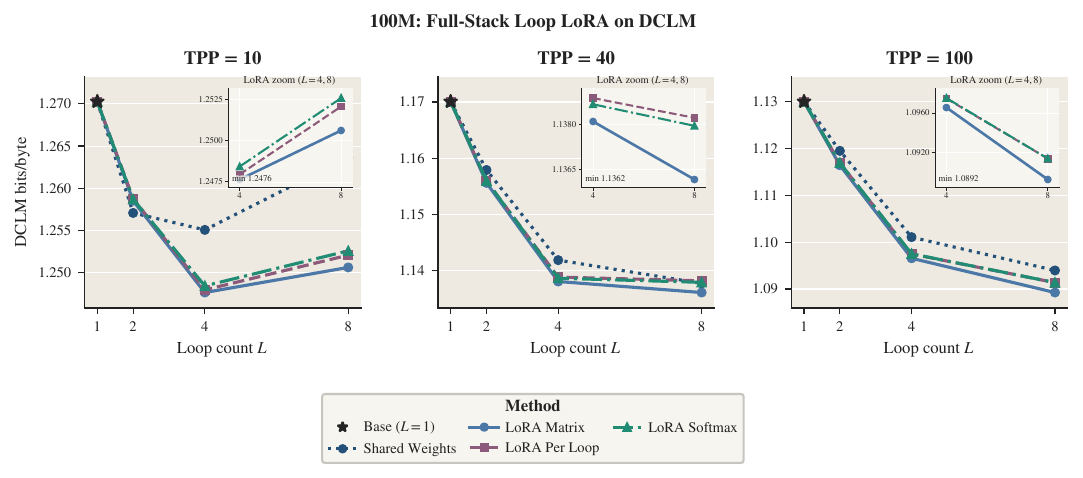}
    \par\small\textbf{(b) 100M}
  \end{minipage}

  \vspace{0.75em}

  \begin{minipage}[t]{0.485\textwidth}
    \centering
    \includegraphics[width=\linewidth]{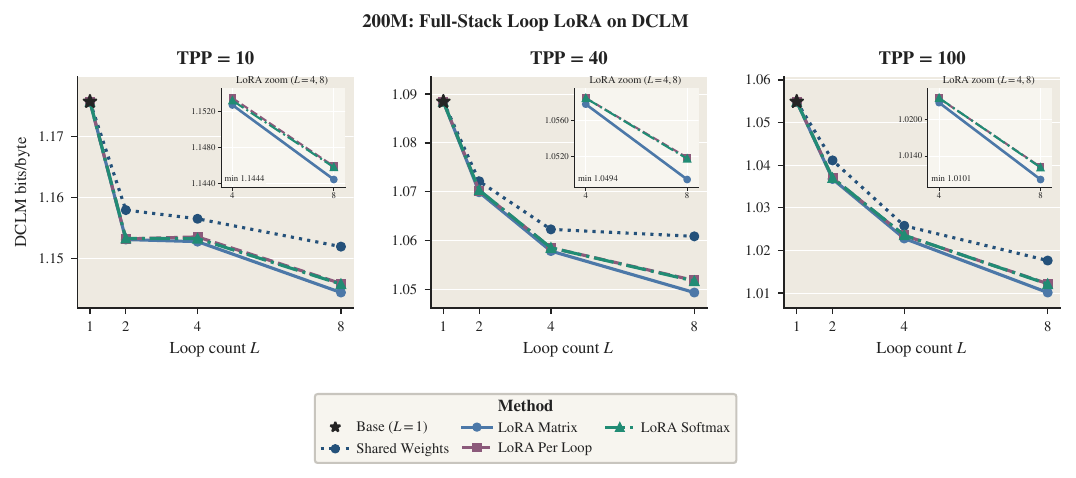}
    \par\small\textbf{(c) 200M}
  \end{minipage}
  \hfill
  \begin{minipage}[t]{0.485\textwidth}
    \centering
    \includegraphics[width=\linewidth]{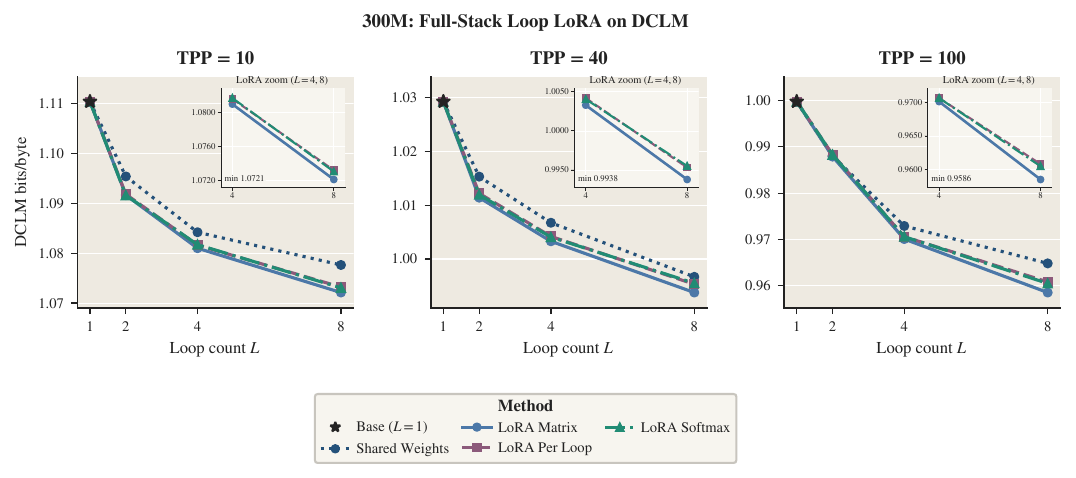}
    \par\small\textbf{(d) 300M}
  \end{minipage}

  \caption{\textbf{Specialization helps more as the number of loops
  increases, consistently across TPP.} DCLM BPB versus loop count for
  70M--300M full-stack models at TPP 10, 40, and 100. Specialization provides
  a small improvement at $L=2$ and clearer gains at $L=4$ and $L=8$.
  Increasing TPP also generally lowers BPB. Lower is better.}
  \label{fig:loop-lora-dclm-loops}
\end{figure*}

\paragraph{Loop count.}
The same loop-count pattern appears at TPP 10, 40, and 100. Specialization
usually provides only a small improvement at $L=2$, while its advantage over
using identical weights becomes clearer at $L=4$ and $L=8$. The three
specialization methods remain closer to one another than to the
identical-weight baseline, although Loop Matrix is most often strongest.
\autoref{fig:loop-lora-dclm-loops} reports the complete loop-count
comparison.

\begin{figure*}[ht]
  \centering
  \begin{minipage}[t]{0.485\textwidth}
    \centering
    \includegraphics[width=\linewidth]{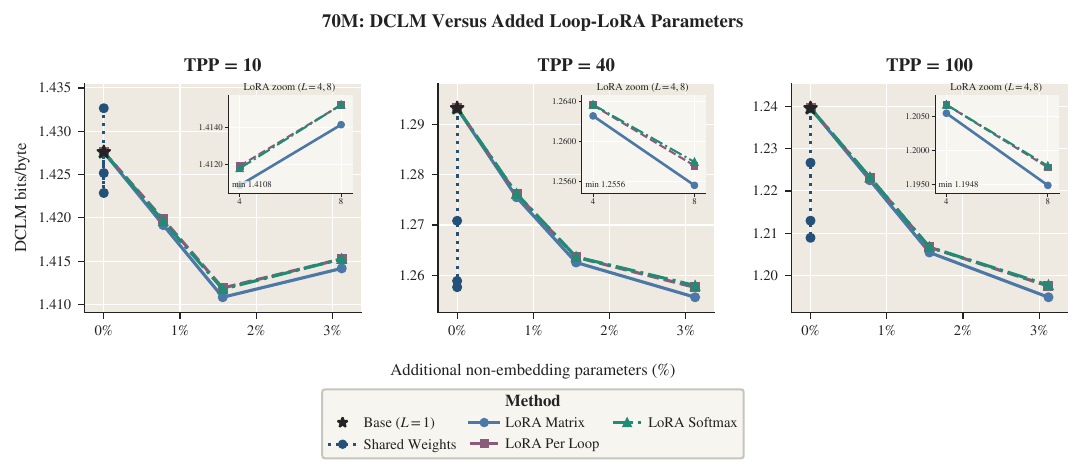}
    \par\small\textbf{(a) 70M}
  \end{minipage}
  \hfill
  \begin{minipage}[t]{0.485\textwidth}
    \centering
    \includegraphics[width=\linewidth]{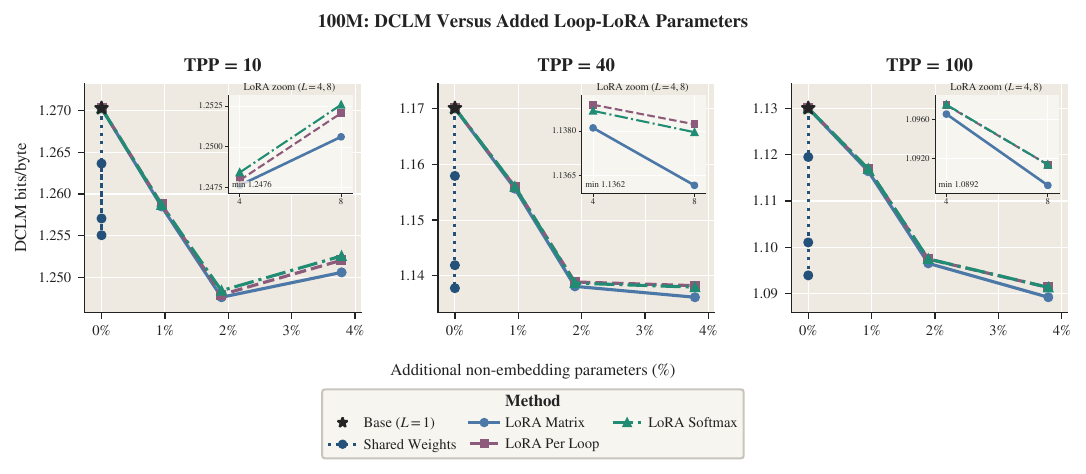}
    \par\small\textbf{(b) 100M}
  \end{minipage}

  \vspace{0.75em}

  \begin{minipage}[t]{0.485\textwidth}
    \centering
    \includegraphics[width=\linewidth]{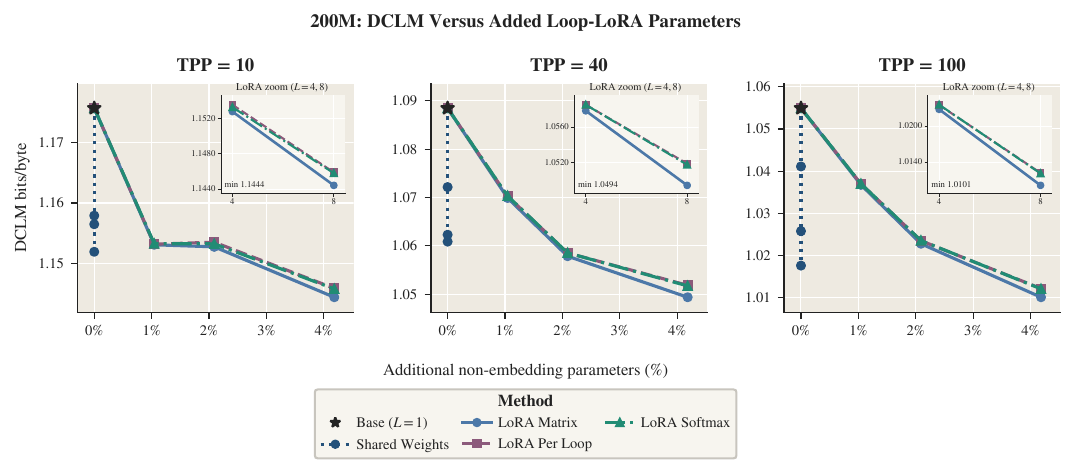}
    \par\small\textbf{(c) 200M}
  \end{minipage}
  \hfill
  \begin{minipage}[t]{0.485\textwidth}
    \centering
    \includegraphics[width=\linewidth]{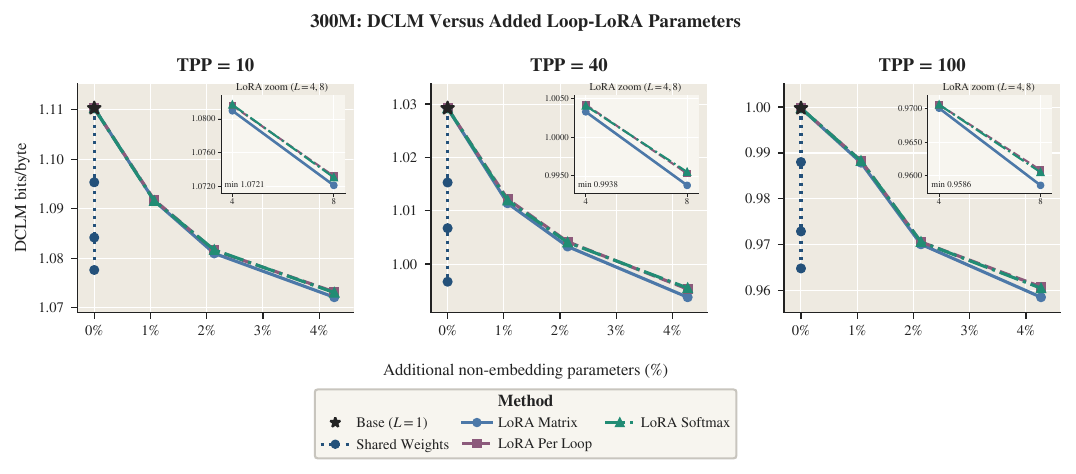}
    \par\small\textbf{(d) 300M}
  \end{minipage}

  \caption{\textbf{Specialization improves DCLM BPB with relatively few
  additional parameters across TPP.} Results cover 70M--300M full-stack
  models at TPP 10, 40, and 100. The horizontal axis reports additional
  non-embedding parameters relative to the identical-weight model. Higher
  TPP generally lowers BPB. Lower is better.}
  \label{fig:loop-lora-dclm-params}
\end{figure*}

\paragraph{Stored parameters.}
Across all three TPP values, loop-specific corrections improve BPB while
adding relatively few stored parameters. Specialization is therefore most
attractive when the number of parameters that can be stored is constrained.
The corresponding parameter comparison appears in
\autoref{fig:loop-lora-dclm-params}.

\begin{figure*}[ht]
  \centering
  \includegraphics[width=0.7\textwidth]{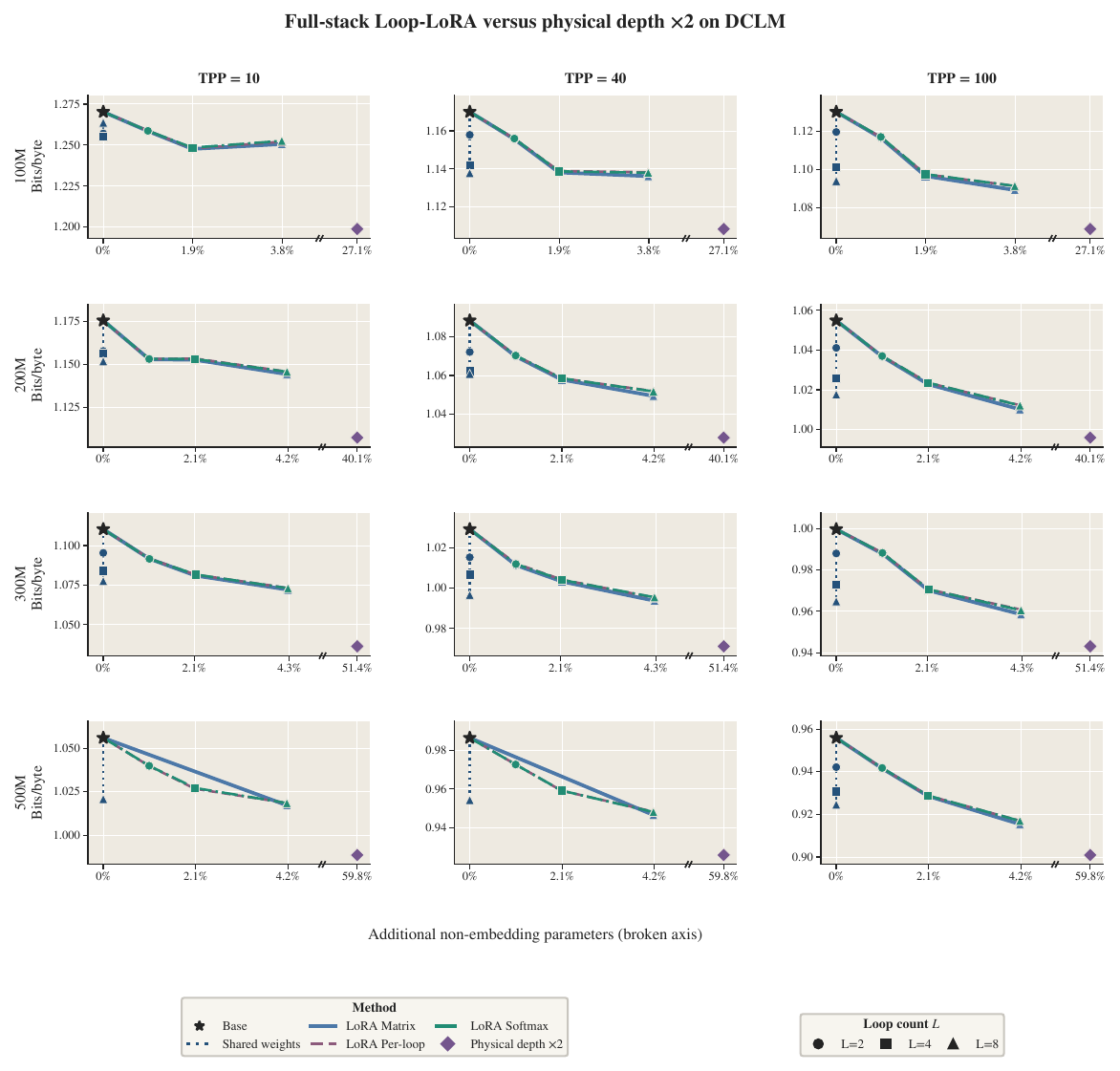}
  \caption{\textbf{Adding independently parameterized layers can achieve
  lower BPB but requires substantially more storage.}
  Rows show 100M, 200M, 300M, and 500M models; columns show TPP 10, 40, and
  100. The purple diamond is the matched doubled-depth model. The final tick
  after the broken axis gives its true increase in non-embedding parameters.
  Lower BPB is better.}
  \label{fig:loop-lora-dclm-depth}
\end{figure*}

\paragraph{Loop-specific specialization or physical depth.}
To place the parameter savings from specialization in context, we compare
against a doubled-depth Transformer. This reference is a conventional
Transformer with twice as many blocks as the corresponding baseline, with
each block using its own independently learned parameters. It therefore adds
depth by storing new layers rather than by repeatedly applying the same
layers. Across the evaluated TPP values, the doubled-depth model can achieve
lower BPB, but requires substantially more stored parameters.
\autoref{fig:loop-lora-dclm-depth} compares these choices directly.
Physical-depth references are available for 100M--500M rather than for the
complete 70M--300M grid used in the other views.

\begin{figure*}[ht]
  \centering
  \begin{minipage}[t]{0.485\textwidth}
    \centering
    \includegraphics[width=\linewidth]{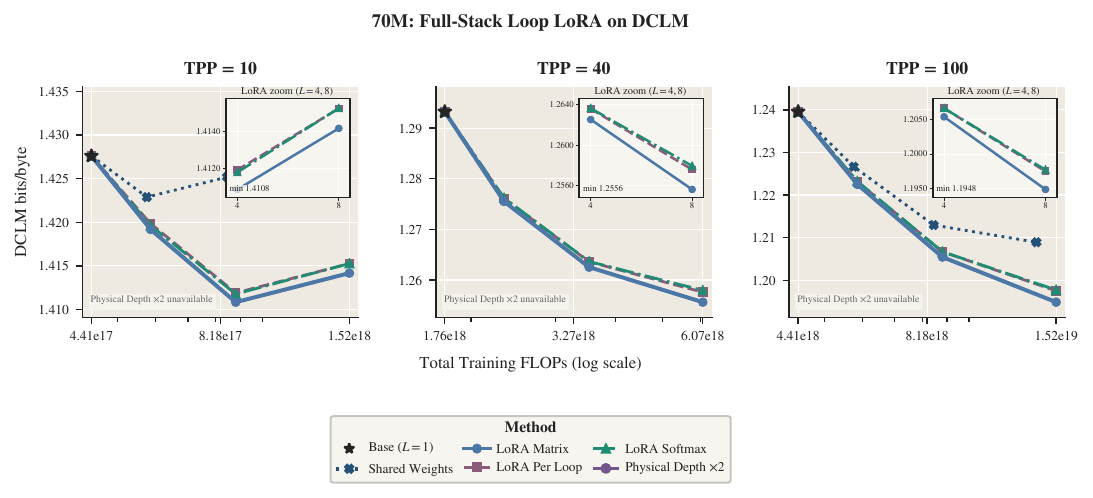}
    \par\small\textbf{(a) 70M}
  \end{minipage}
  \hfill
  \begin{minipage}[t]{0.485\textwidth}
    \centering
    \includegraphics[width=\linewidth]{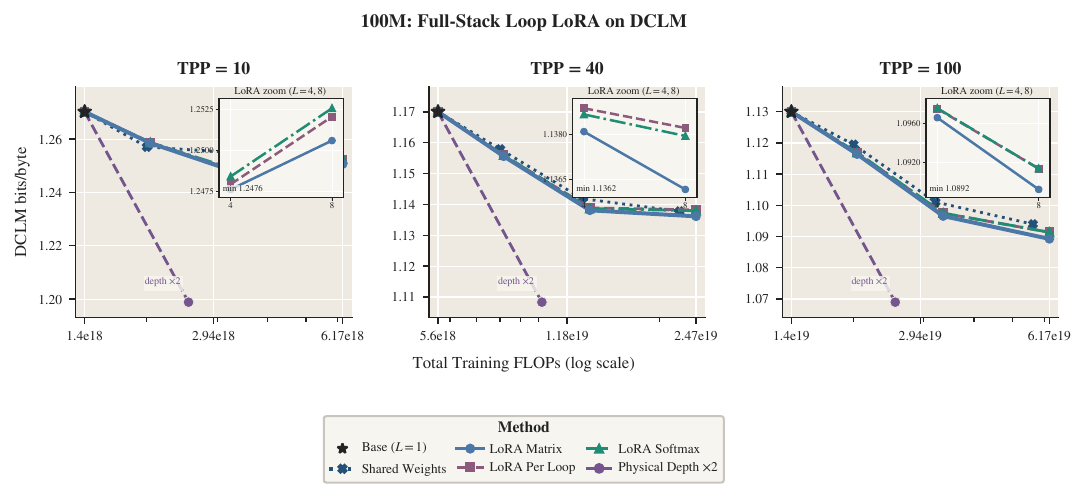}
    \par\small\textbf{(b) 100M}
  \end{minipage}

  \vspace{0.75em}

  \begin{minipage}[t]{0.485\textwidth}
    \centering
    \includegraphics[width=\linewidth]{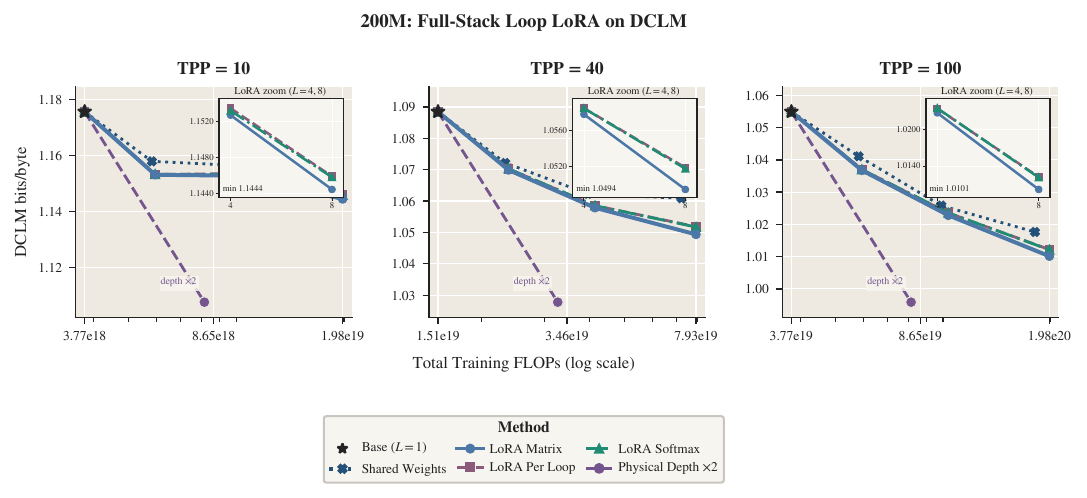}
    \par\small\textbf{(c) 200M}
  \end{minipage}
  \hfill
  \begin{minipage}[t]{0.485\textwidth}
    \centering
    \includegraphics[width=\linewidth]{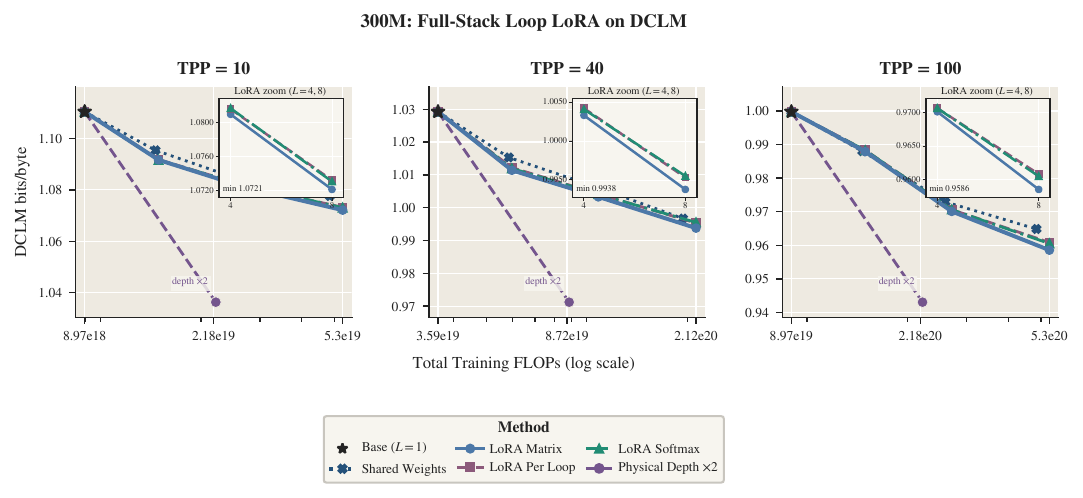}
    \par\small\textbf{(d) 300M}
  \end{minipage}

  \caption{\textbf{When parameters can be stored, additional physical depth
  can be a better use of training FLOPs.} DCLM BPB versus total training
  FLOPs for 70M--300M models at TPP 10, 40, and 100. Training FLOPs include
  repeated loop execution, loop-specific corrections, attention-score
  computation, and the realized token budget. Lower BPB is better.}
  \label{fig:loop-lora-dclm-flops}
\end{figure*}

\paragraph{Training FLOPs.}
This comparison includes the training cost of running the additional loops
and their loop-specific corrections. At a similar number of training FLOPs,
a Transformer with twice as many independently parameterized layers can
achieve lower BPB across the evaluated TPP values. Therefore, if training
compute is limited but storing more parameters is acceptable, training a
deeper Transformer may be a better use of compute than adding and
specializing more loops. The complete comparison is shown in
\autoref{fig:loop-lora-dclm-flops}.

\paragraph{DCLM summary.}
The same qualitative conclusions hold at TPP 10, 40, and 100:
specialization provides a small benefit at $L=2$ and a clearer benefit at
$L=4$ and $L=8$; the precise specialization method matters less than allowing
the loops to specialize; and Loop Matrix is most often the strongest choice.
Increasing TPP also generally lowers BPB for both shared and specialized
models. Thus, training on more tokens improves overall quality without
changing the main conclusion that specialization becomes more useful as the
number of loops increases. In terms of resources, specialization is most
useful when stored parameters are constrained. If training FLOPs are instead
the main constraint and more parameters can be stored, training a deeper
Transformer can provide better quality for the same amount of training
compute.

\clearpage

\subsubsection{C4 English}
\label{app:loop-lora-c4}

C4 English provides a second large web-text evaluation. It therefore tests whether the conclusions drawn from DCLM depend on our primary evaluation corpus. We repeat the four
comparisons used for DCLM: loop count, added parameter storage, additional
independently parameterized layers, and total training FLOPs. We evaluate
70M--300M full-stack models at TPP 10, 40, and 100. The doubled-depth
comparison also includes 500M models.

\begin{figure*}[ht]
  \centering
  \begin{minipage}[t]{0.485\textwidth}
    \centering
    \includegraphics[width=\linewidth]{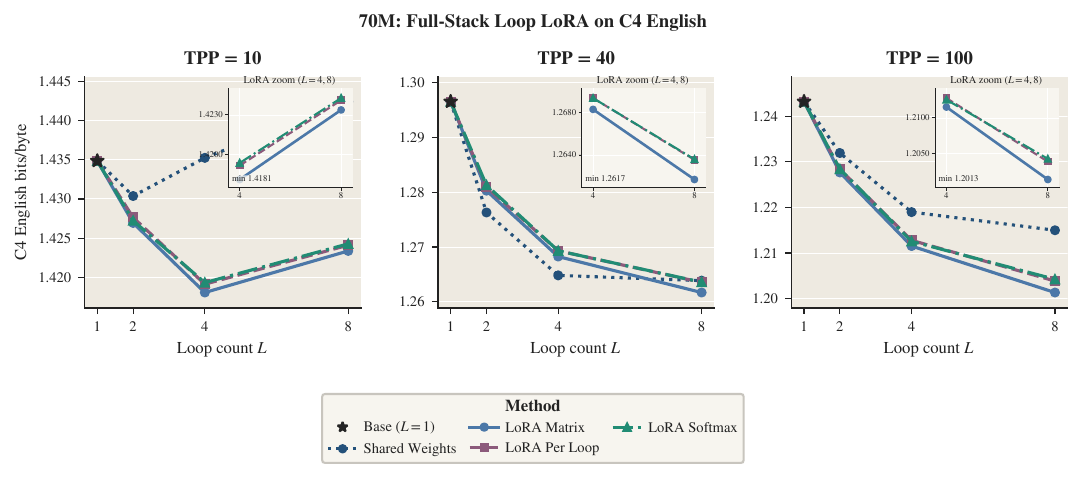}
    \par\small\textbf{(a) 70M}
  \end{minipage}
  \hfill
  \begin{minipage}[t]{0.485\textwidth}
    \centering
    \includegraphics[width=\linewidth]{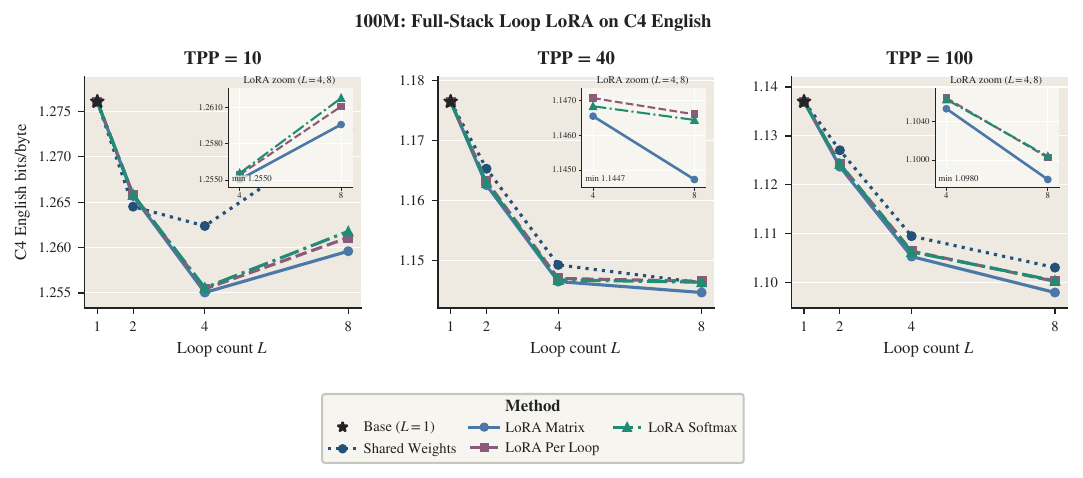}
    \par\small\textbf{(b) 100M}
  \end{minipage}

  \vspace{0.75em}

  \begin{minipage}[t]{0.485\textwidth}
    \centering
    \includegraphics[width=\linewidth]{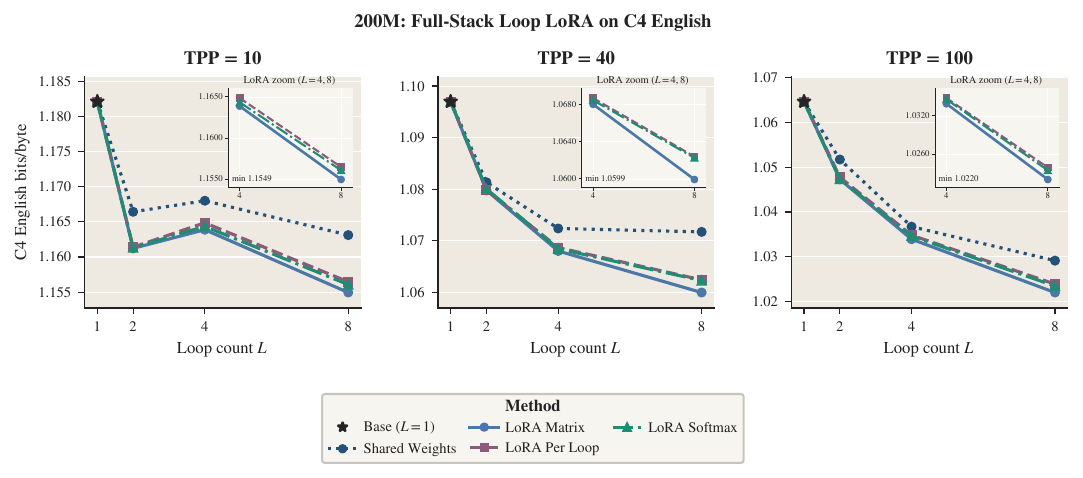}
    \par\small\textbf{(c) 200M}
  \end{minipage}
  \hfill
  \begin{minipage}[t]{0.485\textwidth}
    \centering
    \includegraphics[width=\linewidth]{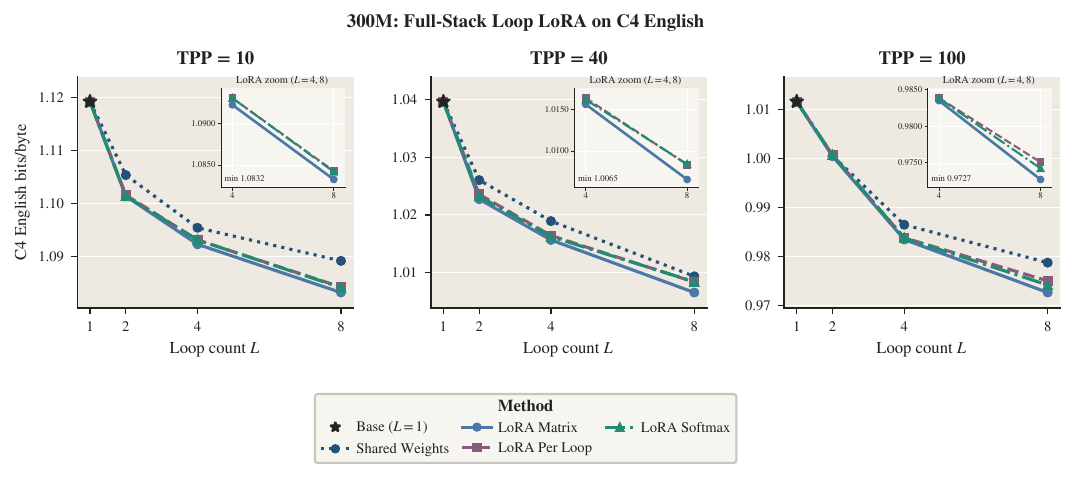}
    \par\small\textbf{(d) 300M}
  \end{minipage}
  \caption{\textbf{Specialization generally helps more as the number of loops increases.} C4 English BPB versus loop count for 70M--300M
  full-stack models at TPP 10, 40, and 100. Specialization is generally more
  useful at $L=4$ and $L=8$ than at $L=2$. Increasing TPP also generally
  lowers BPB. Lower is better.}
  \label{fig:loop-lora-c4-loops}
\end{figure*}

\paragraph{Loop count.}
The same broad pattern appears at TPP 10, 40, and 100. Specialization
usually provides only a small improvement at $L=2$, while its advantage over
using identical weights becomes clearer at $L=4$ and $L=8$. The three
specialization methods remain closer to one another than to the
identical-weight baseline, although Loop Matrix is most often strongest.
Increasing TPP also generally lowers BPB.
\autoref{fig:loop-lora-c4-loops} reports the complete loop-count
comparison.

\begin{figure*}[ht]
  \centering
  \begin{minipage}[t]{0.485\textwidth}
    \centering
    \includegraphics[width=\linewidth]{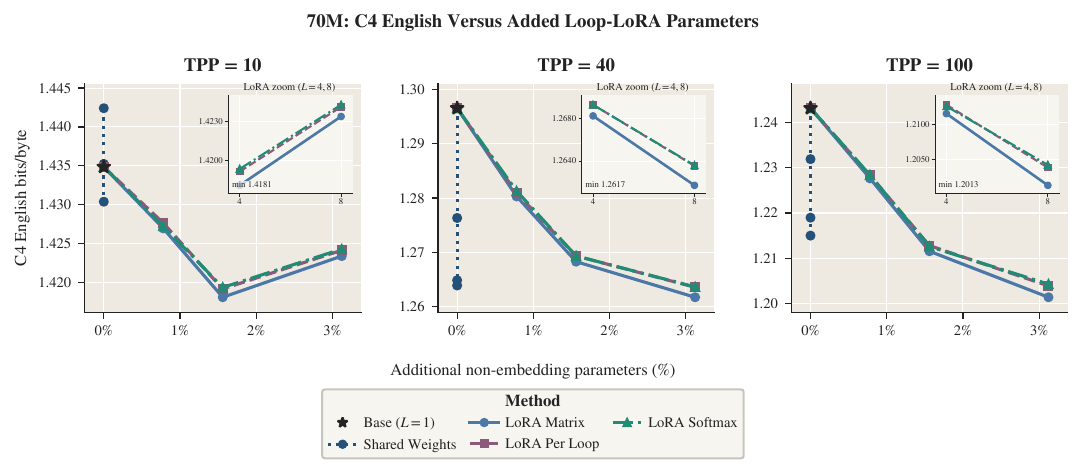}
    \par\small\textbf{(a) 70M}
  \end{minipage}
  \hfill
  \begin{minipage}[t]{0.485\textwidth}
    \centering
    \includegraphics[width=\linewidth]{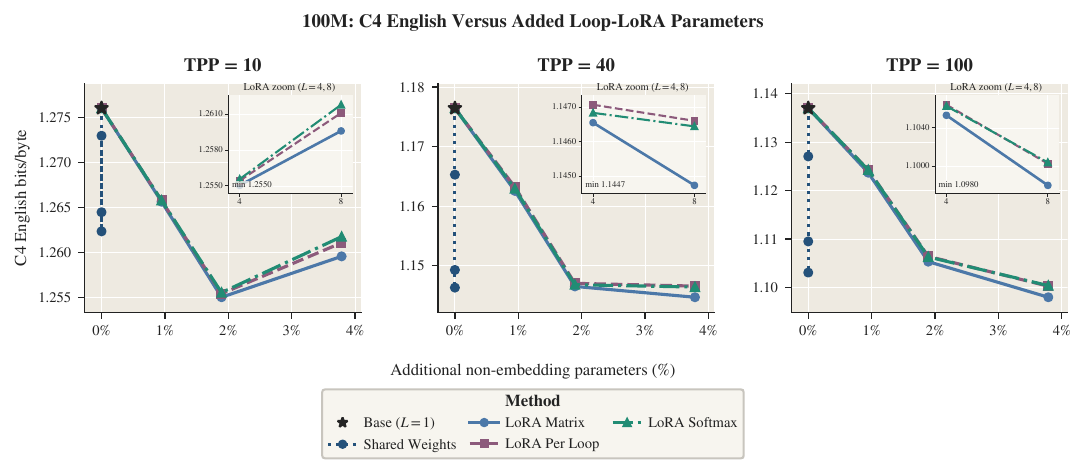}
    \par\small\textbf{(b) 100M}
  \end{minipage}

  \vspace{0.75em}

  \begin{minipage}[t]{0.485\textwidth}
    \centering
    \includegraphics[width=\linewidth]{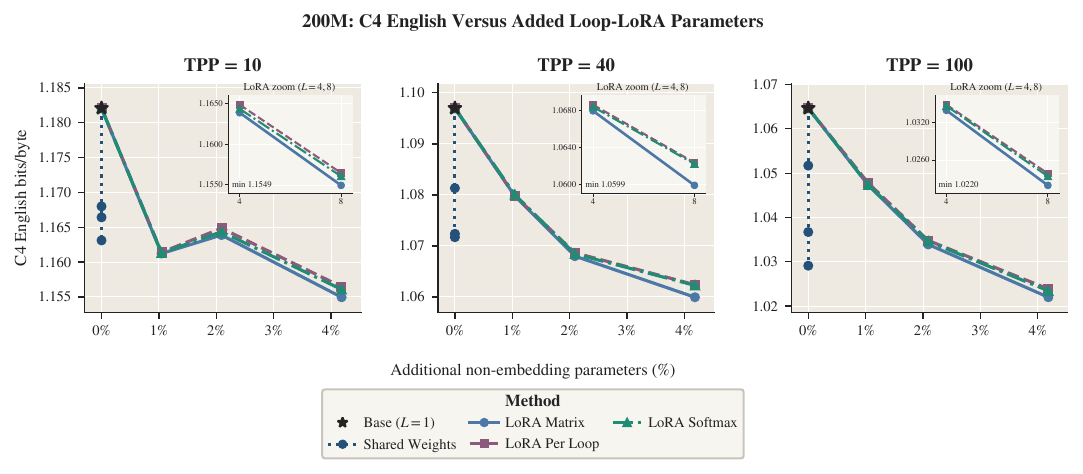}
    \par\small\textbf{(c) 200M}
  \end{minipage}
  \hfill
  \begin{minipage}[t]{0.485\textwidth}
    \centering
    \includegraphics[width=\linewidth]{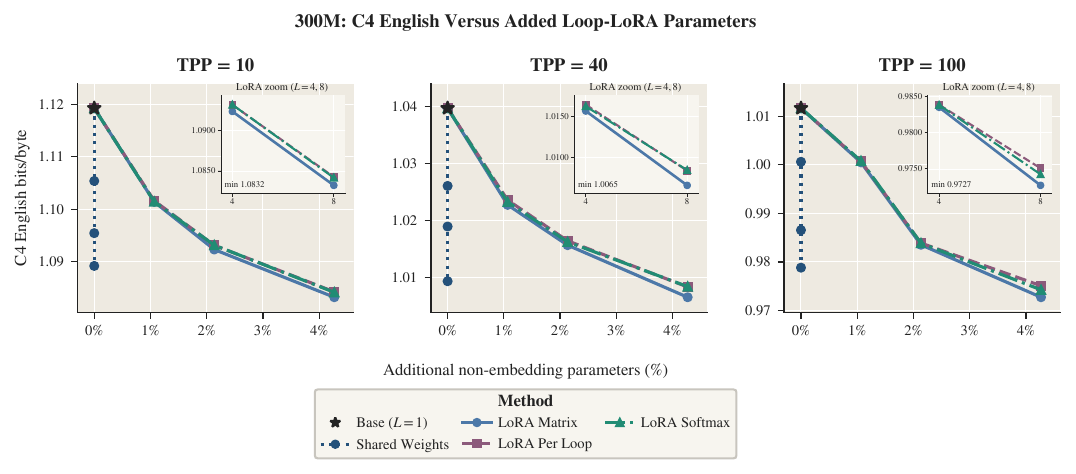}
    \par\small\textbf{(d) 300M}
  \end{minipage}
  \caption{\textbf{Specialization improves C4 English BPB with relatively few additional parameters.} Results cover 70M--300M full-stack models at TPP
  10, 40, and 100. The horizontal axis reports additional non-embedding
  parameters relative to the identical-weight model. Higher TPP generally
  lowers BPB. Lower is better.}
  \label{fig:loop-lora-c4-params}
\end{figure*}

\paragraph{Stored parameters.}
Across all three TPP values, loop-specific corrections improve BPB while adding relatively few stored parameters.
Specialization therefore provides a useful middle ground when parameter
storage is constrained: most weights remain shared, while each loop receives
a small correction. The corresponding comparison appears in
\autoref{fig:loop-lora-c4-params}.

\begin{figure*}[ht]
  \centering
  \includegraphics[width=0.7\textwidth]{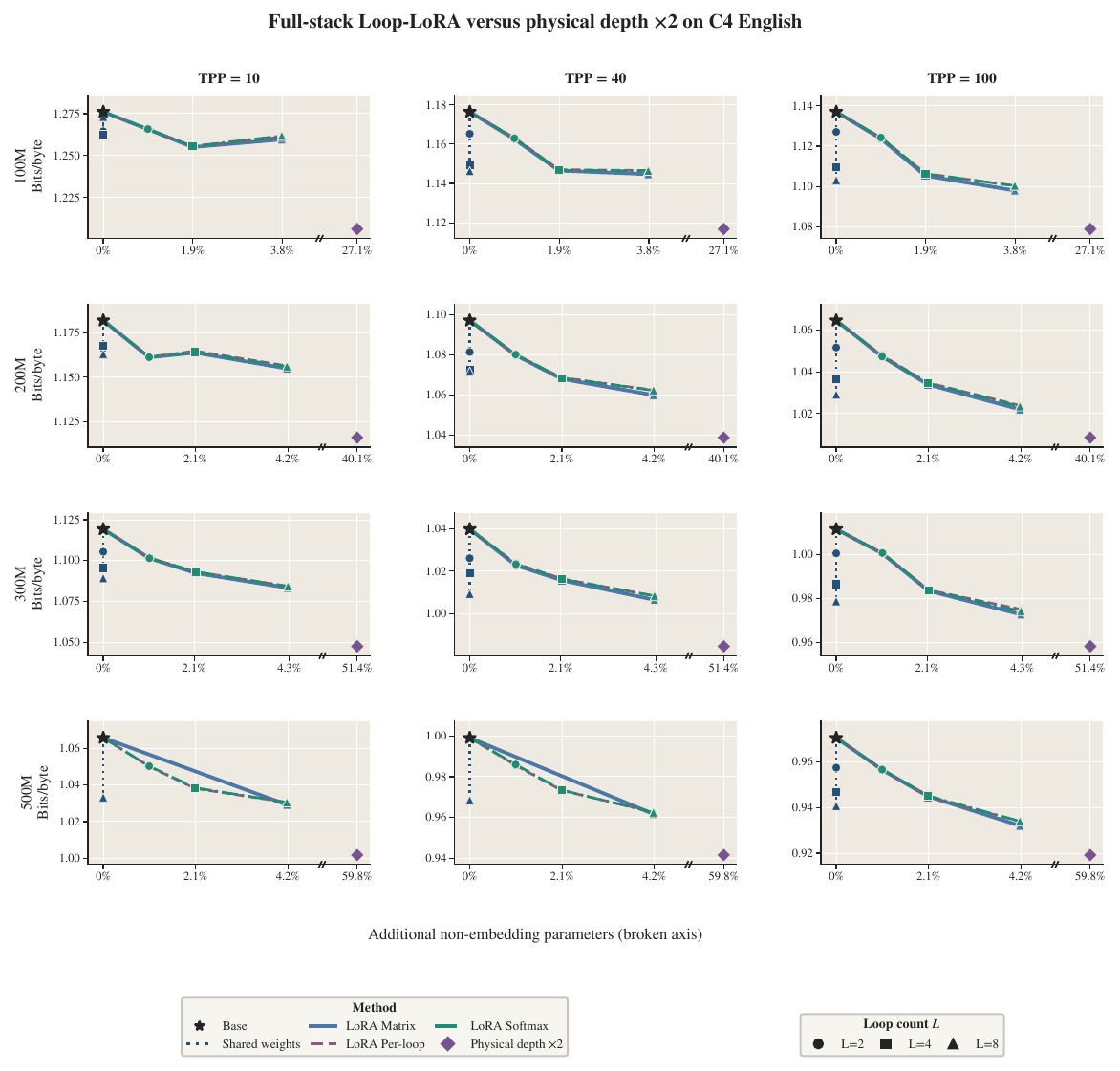}
  \caption{\textbf{Adding independently parameterized layers can achieve
  lower C4 English BPB but requires substantially more storage.}
  Rows show 100M, 200M, 300M, and 500M models; columns show TPP 10, 40, and
  100. The purple diamond is the matched doubled-depth Transformer. The final
  tick after the broken axis gives its true increase in non-embedding
  parameters. Lower BPB is better.}
  \label{fig:loop-lora-c4-depth}
\end{figure*}

\paragraph{Loop-specific specialization or additional Transformer layers.}
To put this parameter saving in context, we use the same doubled-depth
reference as in the DCLM comparison. This reference is a conventional
Transformer with twice as many blocks as the corresponding baseline, and
every block has its own independently learned parameters. It therefore adds
depth by storing new layers rather than by applying the same layers multiple
times. Across the evaluated TPP values, the doubled-depth Transformer can achieve lower BPB, but requires substantially more stored parameters.
\autoref{fig:loop-lora-c4-depth} compares the two choices. The
doubled-depth references cover 100M--500M models rather than the complete
70M--300M grid used in the other comparisons.

\begin{figure*}[ht]
  \centering
  \begin{minipage}[t]{0.485\textwidth}
    \centering
    \includegraphics[width=\linewidth]{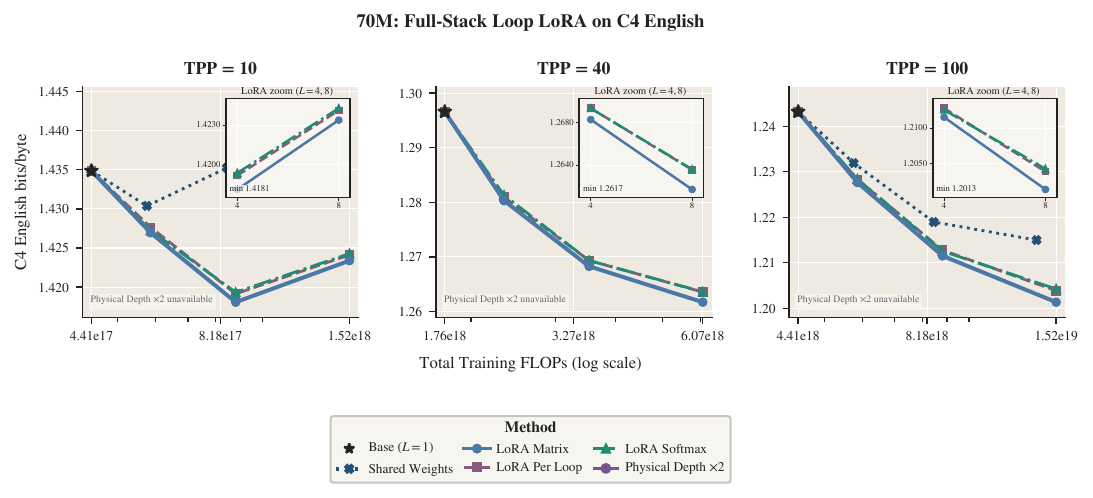}
    \par\small\textbf{(a) 70M}
  \end{minipage}
  \hfill
  \begin{minipage}[t]{0.485\textwidth}
    \centering
    \includegraphics[width=\linewidth]{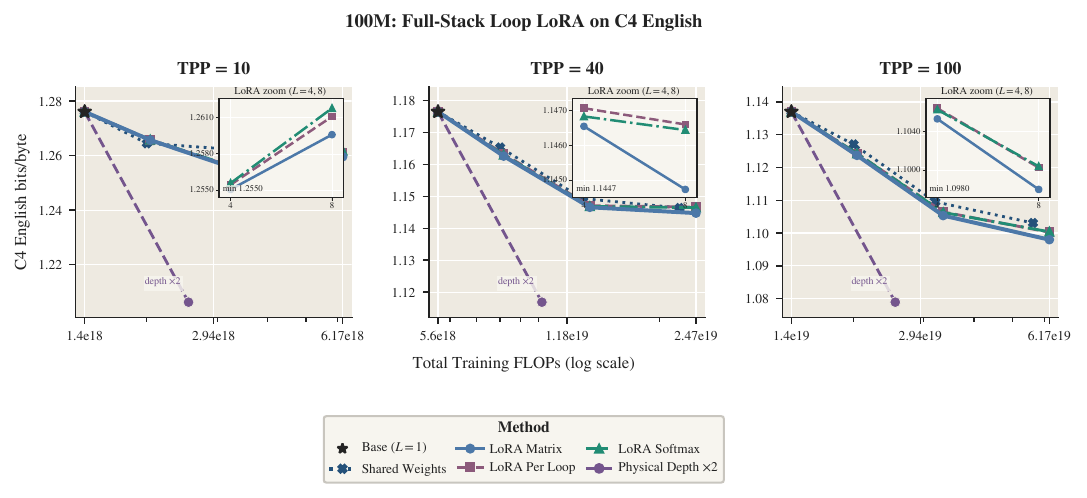}
    \par\small\textbf{(b) 100M}
  \end{minipage}

  \vspace{0.75em}

  \begin{minipage}[t]{0.485\textwidth}
    \centering
    \includegraphics[width=\linewidth]{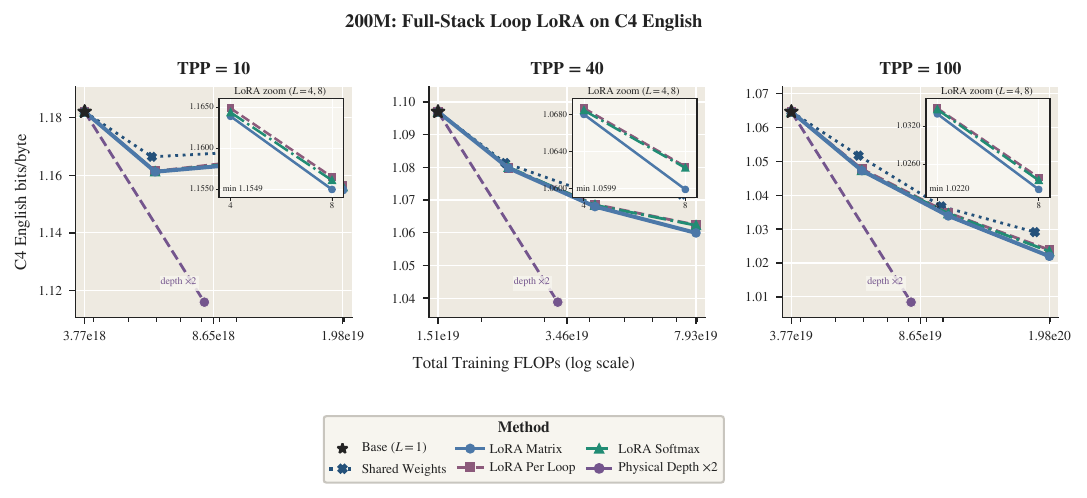}
    \par\small\textbf{(c) 200M}
  \end{minipage}
  \hfill
  \begin{minipage}[t]{0.485\textwidth}
    \centering
    \includegraphics[width=\linewidth]{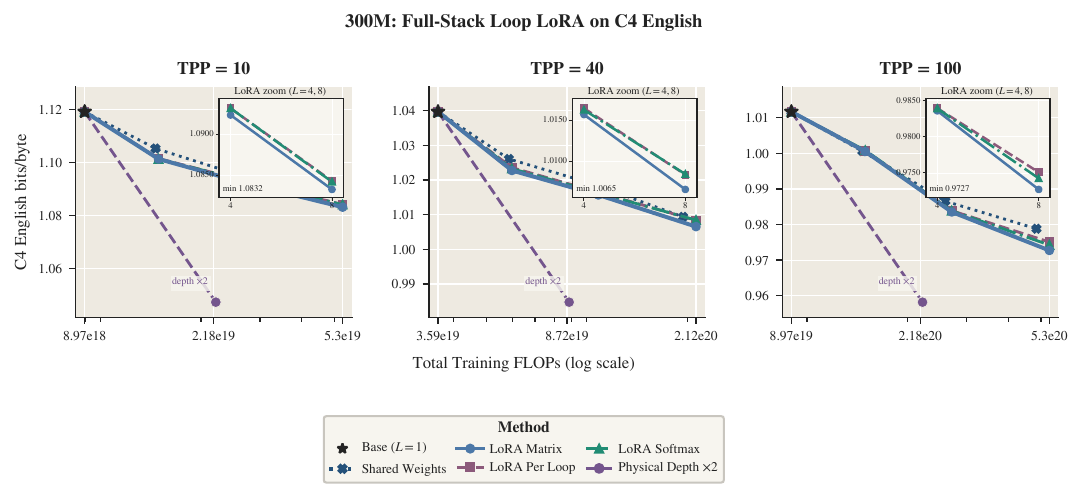}
    \par\small\textbf{(d) 300M}
  \end{minipage}
  \caption{\textbf{Low parameter overhead does not necessarily mean better training-compute efficiency.} C4 English BPB versus total training FLOPs for
  70M--300M models at TPP 10, 40, and 100. Training FLOPs include repeated
  loop execution, loop-specific corrections, attention-score computation,
  and the realized token budget. Lower is better.}
  \label{fig:loop-lora-c4-flops}
\end{figure*}

\paragraph{Training FLOPs.}
The FLOP comparison includes the cost of executing the additional loops and
their loop-specific corrections. At a similar number of training FLOPs, the doubled-depth Transformer can often achieve lower BPB. Therefore, when training
compute is the main constraint and more parameters can be stored, training a
deeper Transformer may be a better use of compute than adding and
specializing more loops. The complete comparison appears in
\autoref{fig:loop-lora-c4-flops}.

\paragraph{C4 English summary.}
The C4 English results reproduce the main DCLM pattern. Specialization provides a small benefit at $L=2$ and a clearer benefit at
$L=4$ and $L=8$, while the precise specialization method matters less than
allowing the loops to specialize. Loop Matrix is most often the strongest
choice.
Increasing TPP generally lowers BPB for both identical-weight and specialized
models, so training on more tokens improves overall quality without changing
the main loop-count pattern. In terms of resources, specialization is most
useful when stored parameters are constrained. If training FLOPs are instead
the main constraint and more parameters can be stored, a deeper Transformer
can provide better quality for the same amount of training compute.

\clearpage

\subsubsection{WikiText-103}
\label{app:loop-lora-wikitext}

WikiText-103 is a smaller, more curated language-modeling corpus than DCLM. It tests whether the specialization result transfers beyond large web-text evaluations. We repeat the four
comparisons used for DCLM: loop count, added parameter storage, additional
independently parameterized layers, and total training FLOPs. We evaluate
70M--300M full-stack models at TPP 10, 40, and 100. The doubled-depth
comparison also includes 500M models.

\begin{figure*}[ht]
  \centering
  \begin{minipage}[t]{0.485\textwidth}
    \centering
    \includegraphics[width=\linewidth]{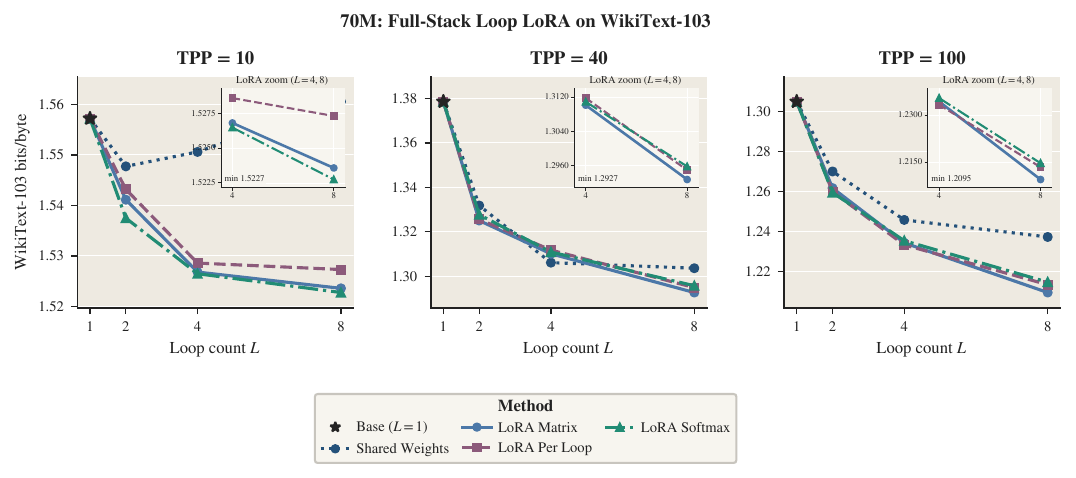}
    \par\small\textbf{(a) 70M}
  \end{minipage}
  \hfill
  \begin{minipage}[t]{0.485\textwidth}
    \centering
    \includegraphics[width=\linewidth]{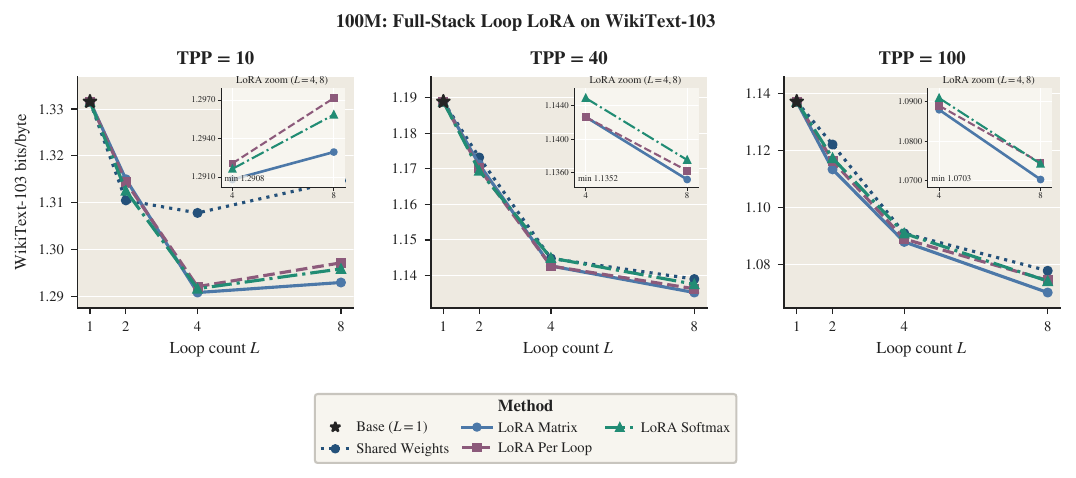}
    \par\small\textbf{(b) 100M}
  \end{minipage}

  \vspace{0.75em}

  \begin{minipage}[t]{0.485\textwidth}
    \centering
    \includegraphics[width=\linewidth]{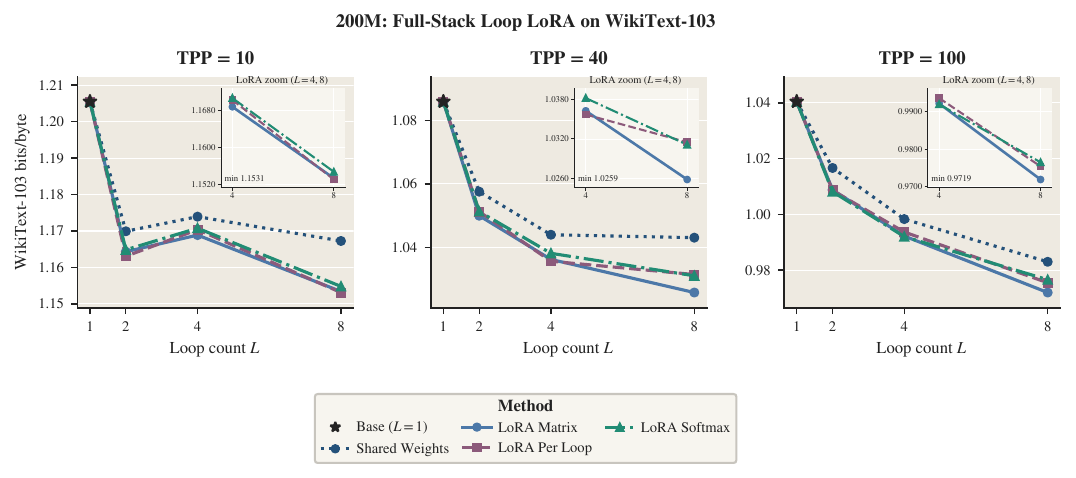}
    \par\small\textbf{(c) 200M}
  \end{minipage}
  \hfill
  \begin{minipage}[t]{0.485\textwidth}
    \centering
    \includegraphics[width=\linewidth]{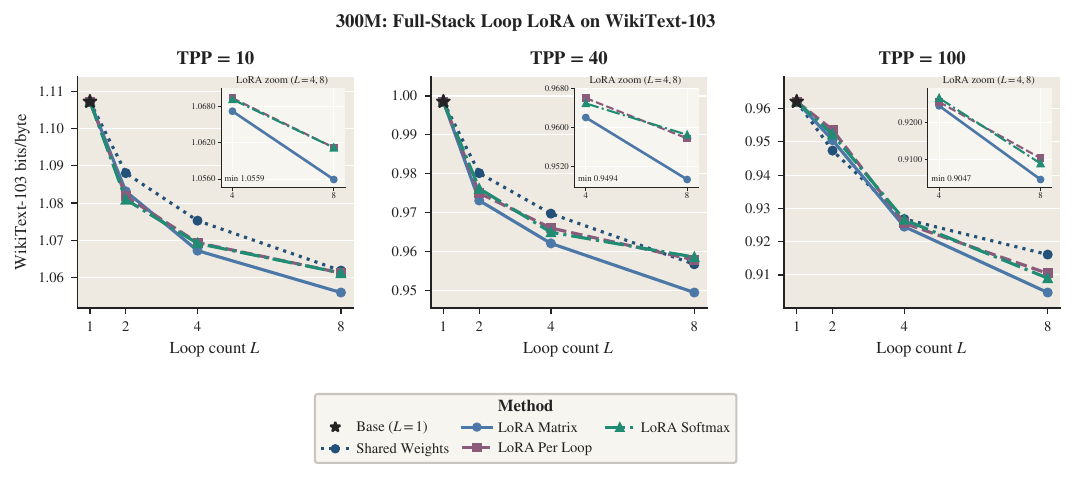}
    \par\small\textbf{(d) 300M}
  \end{minipage}
  \caption{\textbf{Specialization generally helps more as the number of loops increases.} WikiText-103 BPB versus loop count for 70M--300M
  full-stack models at TPP 10, 40, and 100. Specialization is generally more
  useful at $L=4$ and $L=8$ than at $L=2$. Increasing TPP also generally
  lowers BPB. Lower is better.}
  \label{fig:loop-lora-wikitext-loops}
\end{figure*}

\paragraph{Loop count.}
The same qualitative loop-count pattern remains visible at TPP 10, 40,
and 100, although the differences are somewhat more variable than on the
larger web-text corpora. Specialization is usually modest at $L=2$ and becomes
clearer at $L=4$ and $L=8$. Differences among the three specialization
methods are generally smaller than their differences from the
identical-weight baseline, with Loop Matrix often strongest. Increasing TPP
also generally lowers BPB.
\autoref{fig:loop-lora-wikitext-loops} reports the complete loop-count
comparison.

\begin{figure*}[ht]
  \centering
  \begin{minipage}[t]{0.485\textwidth}
    \centering
    \includegraphics[width=\linewidth]{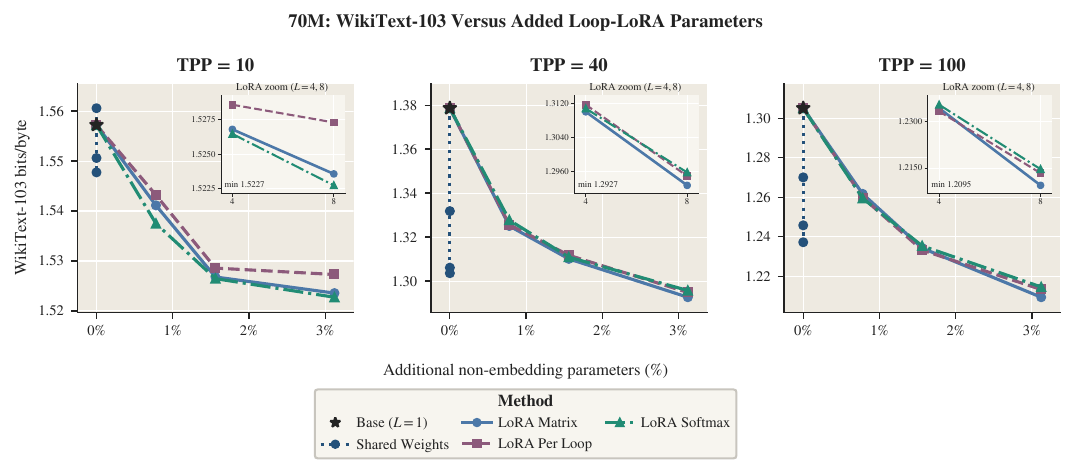}
    \par\small\textbf{(a) 70M}
  \end{minipage}
  \hfill
  \begin{minipage}[t]{0.485\textwidth}
    \centering
    \includegraphics[width=\linewidth]{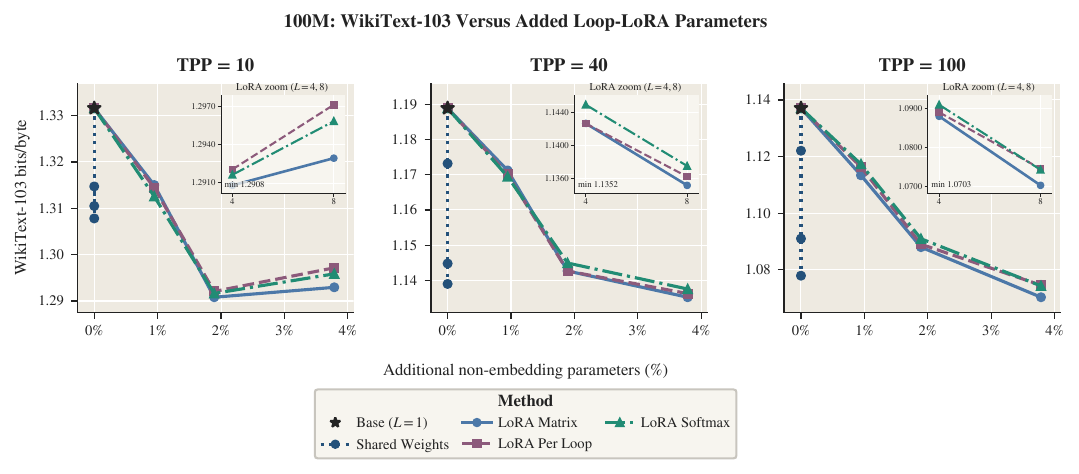}
    \par\small\textbf{(b) 100M}
  \end{minipage}

  \vspace{0.75em}

  \begin{minipage}[t]{0.485\textwidth}
    \centering
    \includegraphics[width=\linewidth]{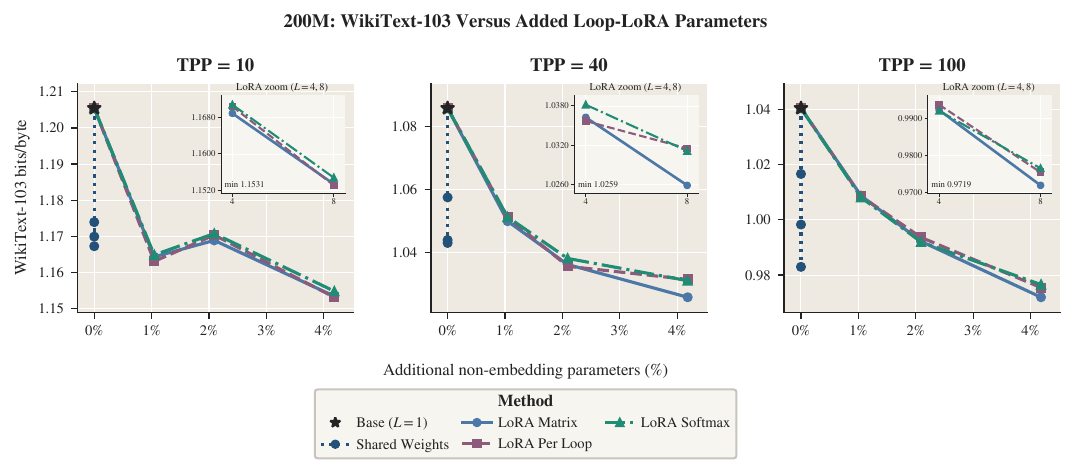}
    \par\small\textbf{(c) 200M}
  \end{minipage}
  \hfill
  \begin{minipage}[t]{0.485\textwidth}
    \centering
    \includegraphics[width=\linewidth]{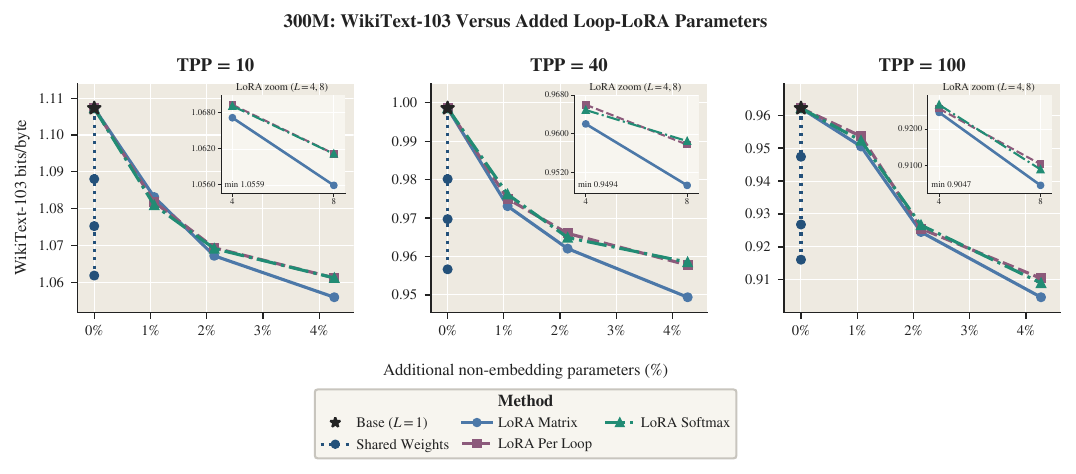}
    \par\small\textbf{(d) 300M}
  \end{minipage}
  \caption{\textbf{Specialization improves WikiText-103 BPB with relatively few additional parameters.} Results cover 70M--300M full-stack models at TPP
  10, 40, and 100. The horizontal axis reports additional non-embedding
  parameters relative to the identical-weight model. Higher TPP generally
  lowers BPB. Lower is better.}
  \label{fig:loop-lora-wikitext-params}
\end{figure*}

\paragraph{Stored parameters.}
Across all three TPP values, loop-specific corrections improve BPB while adding relatively few stored parameters.
Specialization therefore provides a useful middle ground when parameter
storage is constrained: most weights remain shared, while each loop receives
a small correction. The corresponding comparison appears in
\autoref{fig:loop-lora-wikitext-params}.

\begin{figure*}[ht]
  \centering
  \includegraphics[width=0.7\textwidth]{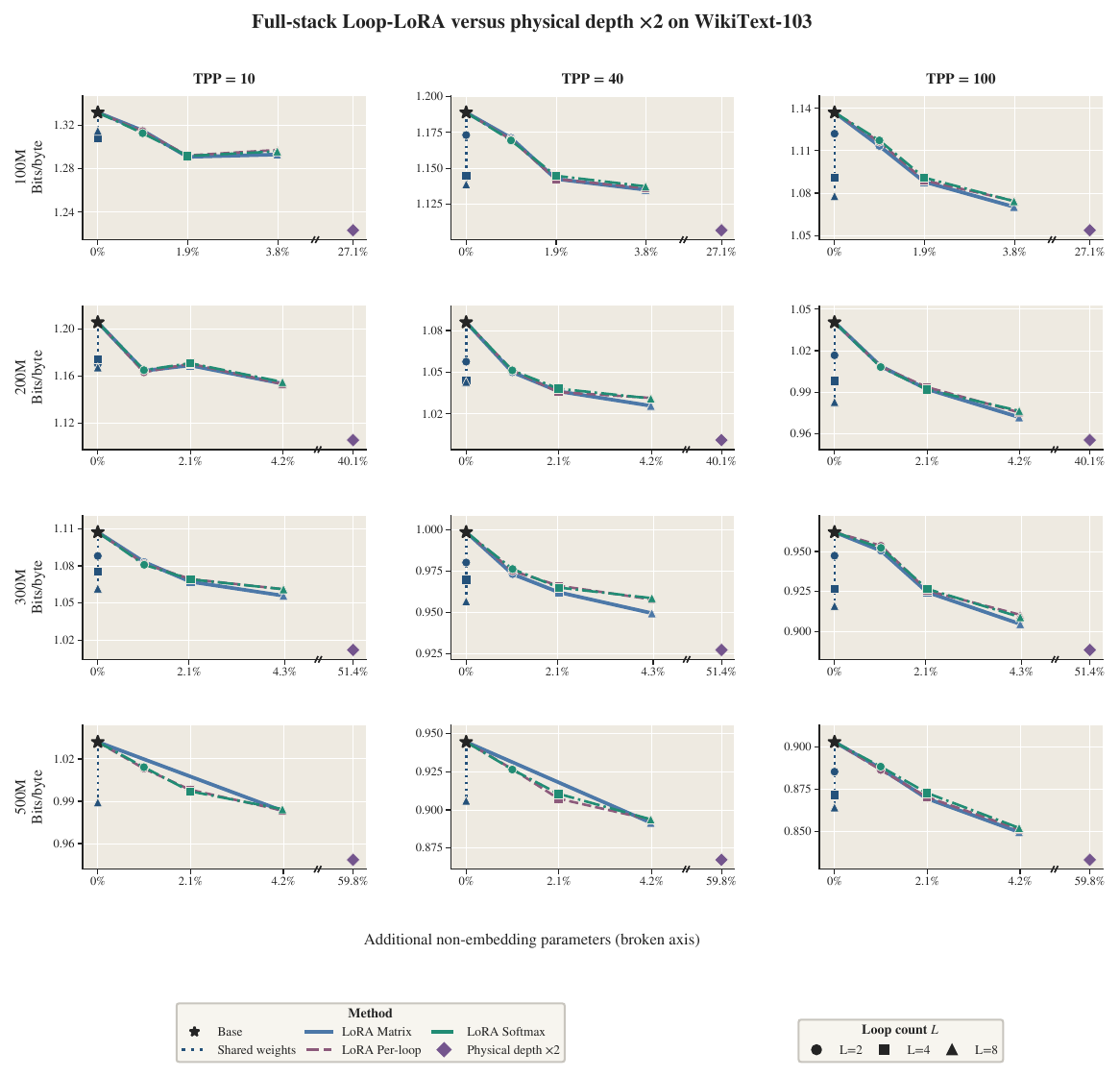}
  \caption{\textbf{Adding independently parameterized layers can achieve
  lower WikiText-103 BPB but requires substantially more storage.}
  Rows show 100M, 200M, 300M, and 500M models; columns show TPP 10, 40, and
  100. The purple diamond is the matched doubled-depth Transformer. The final
  tick after the broken axis gives its true increase in non-embedding
  parameters. Lower BPB is better.}
  \label{fig:loop-lora-wikitext-depth}
\end{figure*}

\paragraph{Loop-specific specialization or additional Transformer layers.}
To put this parameter saving in context, we use the same doubled-depth
reference as in the DCLM comparison. This reference is a conventional
Transformer with twice as many blocks as the corresponding baseline, and
every block has its own independently learned parameters. It therefore adds
depth by storing new layers rather than by applying the same layers multiple
times. Across the evaluated TPP values, the doubled-depth Transformer can achieve lower BPB, but requires substantially more stored parameters.
\autoref{fig:loop-lora-wikitext-depth} compares the two choices. The
doubled-depth references cover 100M--500M models rather than the complete
70M--300M grid used in the other comparisons.

\begin{figure*}[ht]
  \centering
  \begin{minipage}[t]{0.485\textwidth}
    \centering
    \includegraphics[width=\linewidth]{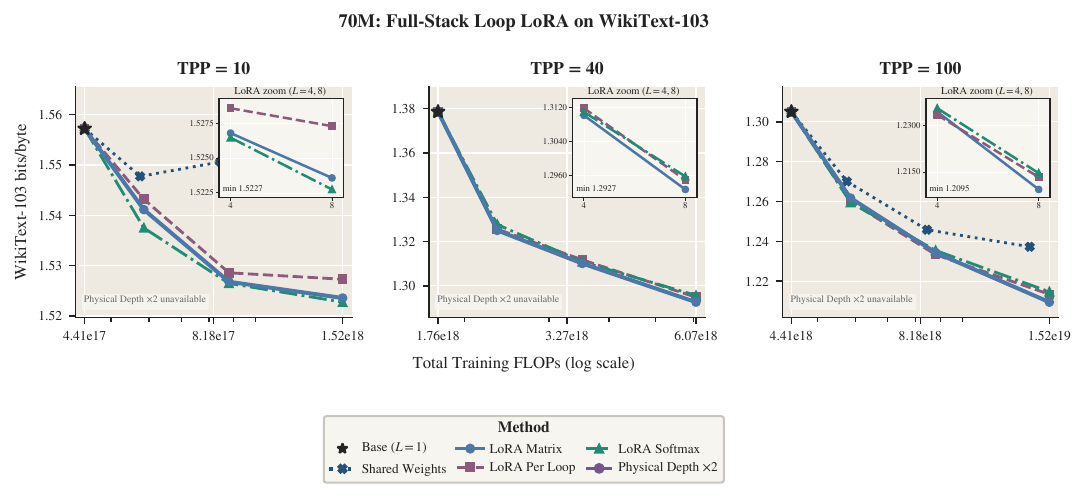}
    \par\small\textbf{(a) 70M}
  \end{minipage}
  \hfill
  \begin{minipage}[t]{0.485\textwidth}
    \centering
    \includegraphics[width=\linewidth]{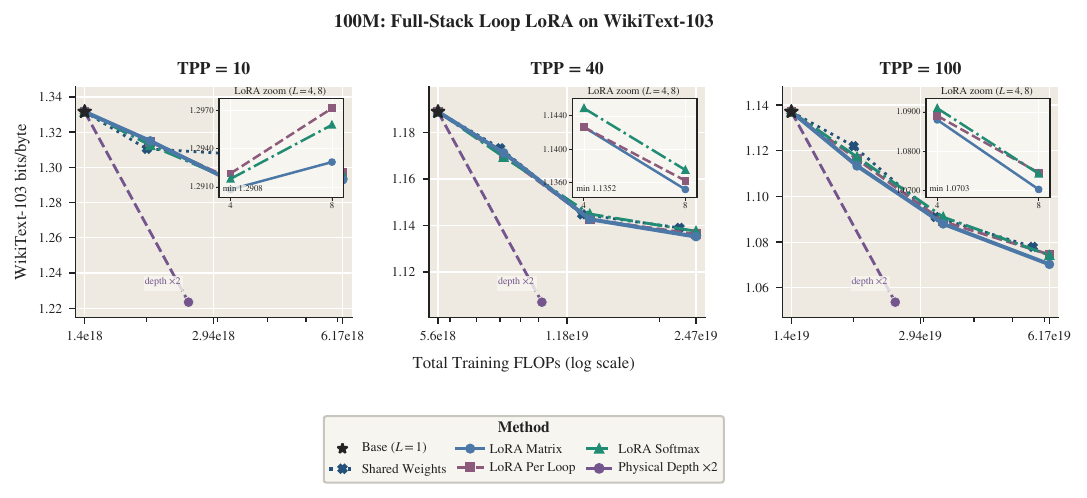}
    \par\small\textbf{(b) 100M}
  \end{minipage}

  \vspace{0.75em}

  \begin{minipage}[t]{0.485\textwidth}
    \centering
    \includegraphics[width=\linewidth]{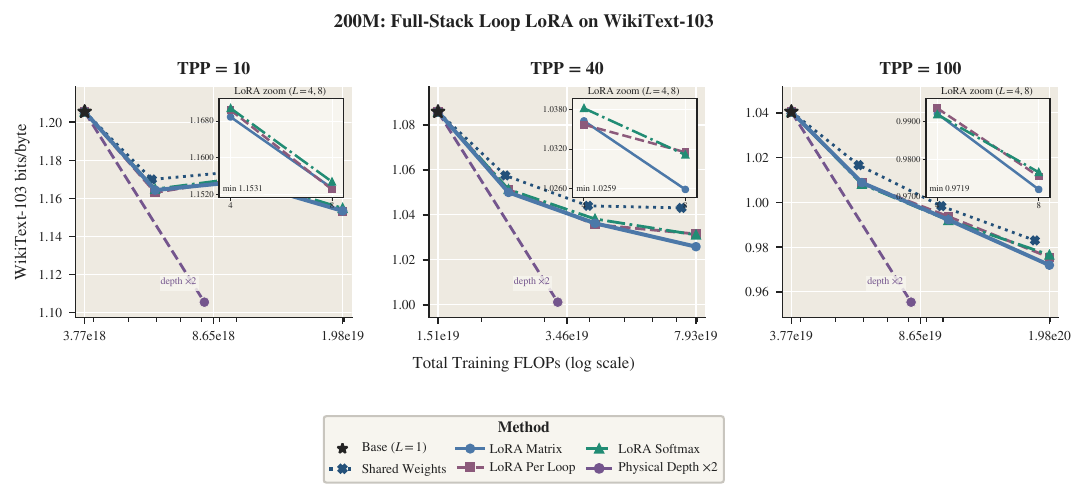}
    \par\small\textbf{(c) 200M}
  \end{minipage}
  \hfill
  \begin{minipage}[t]{0.485\textwidth}
    \centering
    \includegraphics[width=\linewidth]{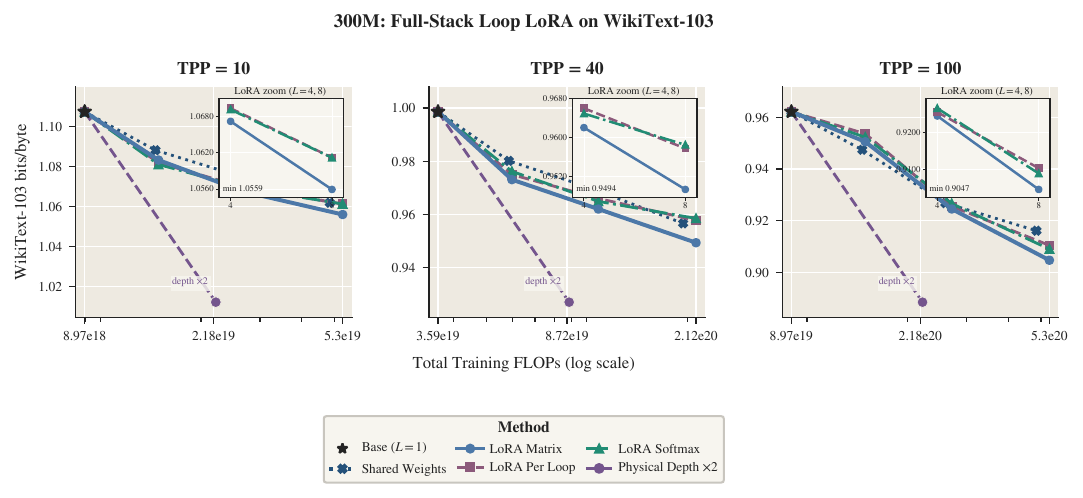}
    \par\small\textbf{(d) 300M}
  \end{minipage}
  \caption{\textbf{Low parameter overhead does not necessarily mean better training-compute efficiency.} WikiText-103 BPB versus total training FLOPs for
  70M--300M models at TPP 10, 40, and 100. Training FLOPs include repeated
  loop execution, loop-specific corrections, attention-score computation,
  and the realized token budget. Lower is better.}
  \label{fig:loop-lora-wikitext-flops}
\end{figure*}

\paragraph{Training FLOPs.}
The FLOP comparison includes the cost of executing the additional loops and
their loop-specific corrections. At a similar number of training FLOPs, the doubled-depth Transformer can often achieve lower BPB. Therefore, when training
compute is the main constraint and more parameters can be stored, training a
deeper Transformer may be a better use of compute than adding and
specializing more loops. The complete comparison appears in
\autoref{fig:loop-lora-wikitext-flops}.

\paragraph{WikiText-103 summary.}
WikiText-103 broadly supports the DCLM conclusion. Specialization provides a small benefit at $L=2$ and a clearer benefit at
$L=4$ and $L=8$, while the precise specialization method matters less than
allowing the loops to specialize. Loop Matrix is most often the strongest
choice.
Increasing TPP generally lowers BPB for both identical-weight and specialized
models, so training on more tokens improves overall quality without changing
the main loop-count pattern. In terms of resources, specialization is most
useful when stored parameters are constrained. If training FLOPs are instead
the main constraint and more parameters can be stored, a deeper Transformer
can provide better quality for the same amount of training compute.

\clearpage

\subsubsection{Dolma v1.5}
\label{app:loop-lora-dolma}

Dolma v1.5 provides a broad language-modeling evaluation corpus with a
different text distribution from DCLM. It tests whether the observed benefits
depend on the composition of the evaluation text. We repeat the four
comparisons used for DCLM: loop count, added parameter storage, additional
independently parameterized layers, and total training FLOPs. We evaluate
70M--300M full-stack models at TPP 10, 40, and 100. The doubled-depth
comparison also includes 500M models.

\begin{figure*}[ht]
  \centering
  \begin{minipage}[t]{0.485\textwidth}
    \centering
    \includegraphics[width=\linewidth]{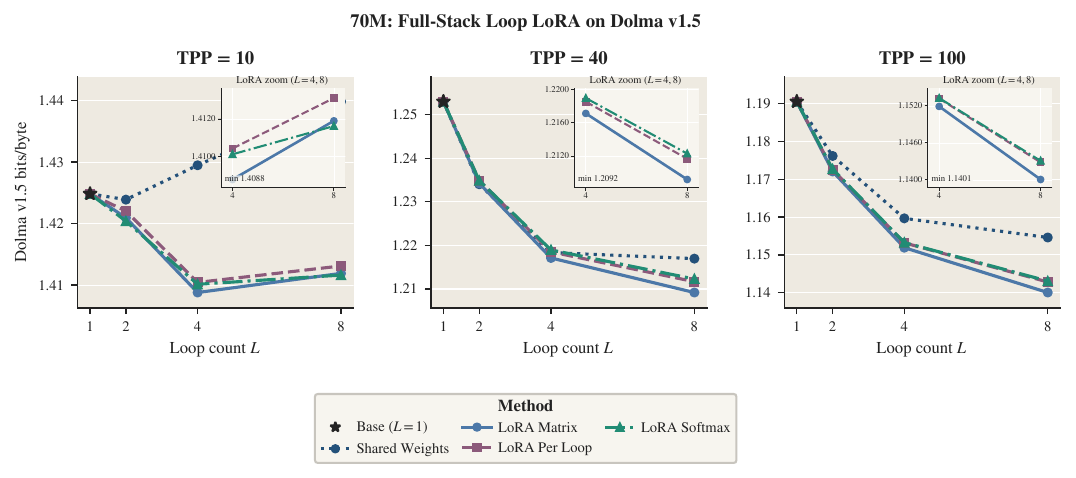}
    \par\small\textbf{(a) 70M}
  \end{minipage}
  \hfill
  \begin{minipage}[t]{0.485\textwidth}
    \centering
    \includegraphics[width=\linewidth]{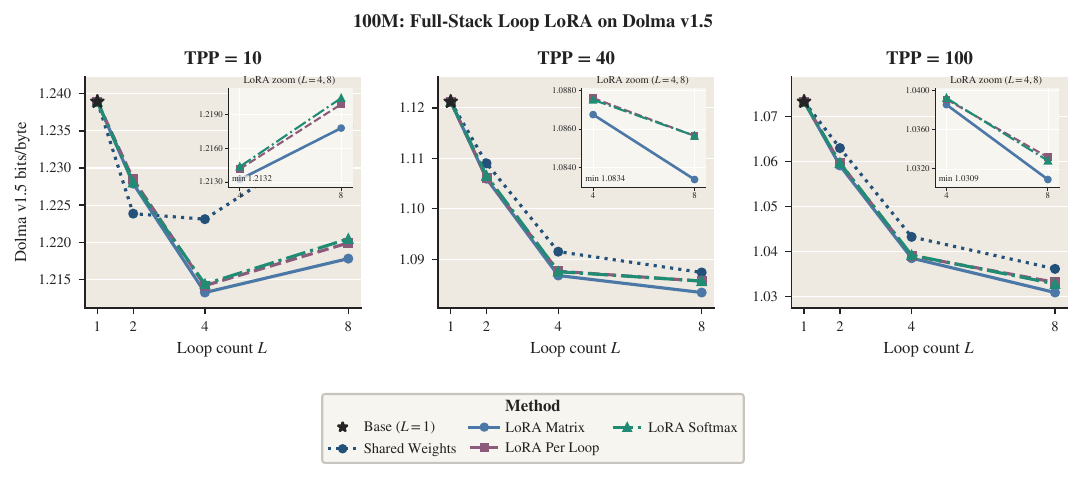}
    \par\small\textbf{(b) 100M}
  \end{minipage}

  \vspace{0.75em}

  \begin{minipage}[t]{0.485\textwidth}
    \centering
    \includegraphics[width=\linewidth]{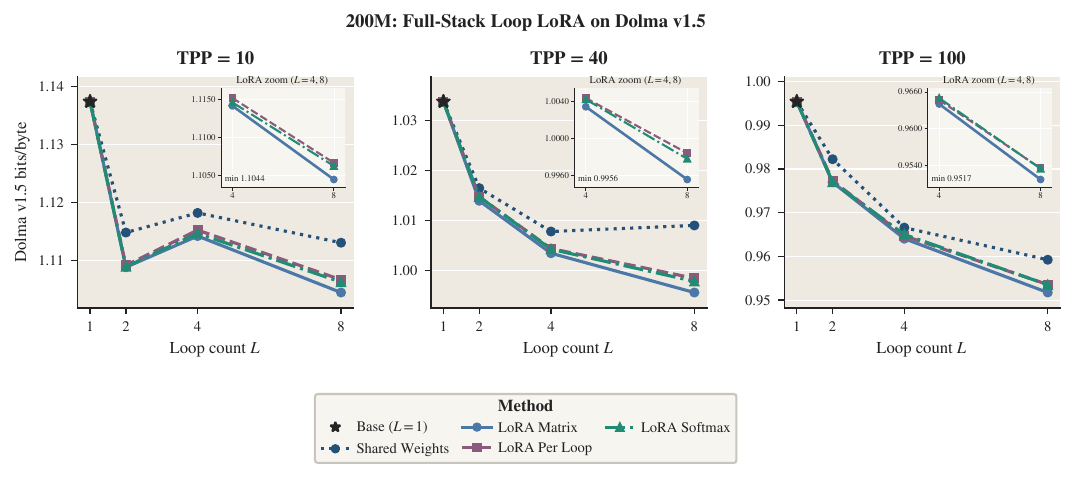}
    \par\small\textbf{(c) 200M}
  \end{minipage}
  \hfill
  \begin{minipage}[t]{0.485\textwidth}
    \centering
    \includegraphics[width=\linewidth]{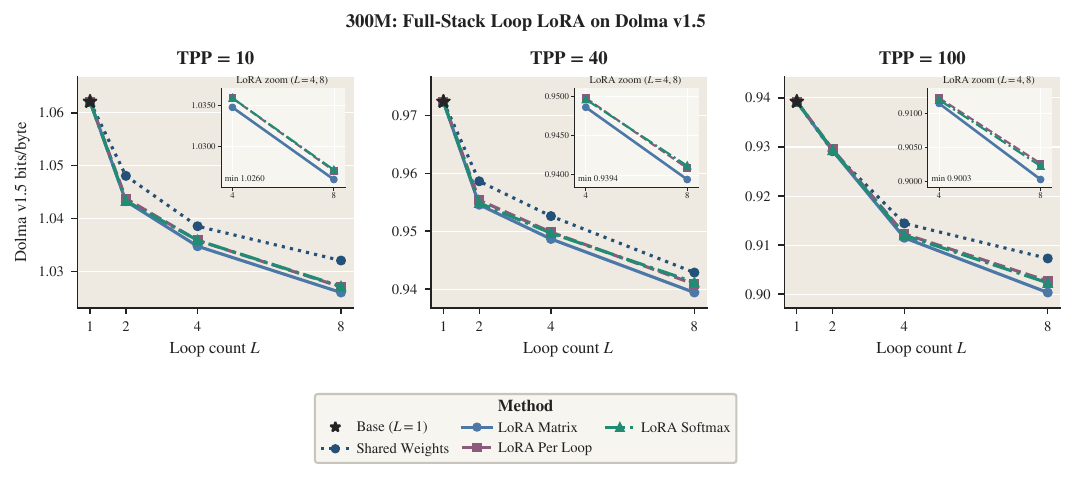}
    \par\small\textbf{(d) 300M}
  \end{minipage}
  \caption{\textbf{Specialization generally helps more as the number of loops increases.} Dolma v1.5 BPB versus loop count for 70M--300M
  full-stack models at TPP 10, 40, and 100. Specialization is generally more
  useful at $L=4$ and $L=8$ than at $L=2$. Increasing TPP also generally
  lowers BPB. Lower is better.}
  \label{fig:loop-lora-dolma-loops}
\end{figure*}

\paragraph{Loop count.}
The same loop-count pattern appears across TPP 10, 40, and 100.
Specialization usually produces only a small improvement at $L=2$, while the
separation from the identical-weight baseline becomes clearer at $L=4$ and
$L=8$. The three specialization methods are generally closer to one another
than to the baseline, and Loop Matrix is most often strongest. Increasing TPP
also generally lowers BPB.
\autoref{fig:loop-lora-dolma-loops} reports the complete loop-count
comparison.

\begin{figure*}[ht]
  \centering
  \begin{minipage}[t]{0.485\textwidth}
    \centering
    \includegraphics[width=\linewidth]{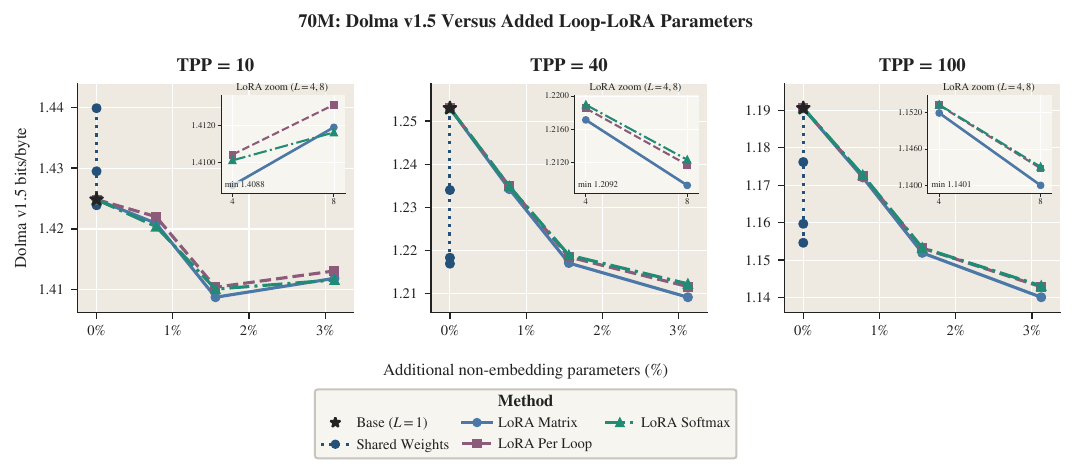}
    \par\small\textbf{(a) 70M}
  \end{minipage}
  \hfill
  \begin{minipage}[t]{0.485\textwidth}
    \centering
    \includegraphics[width=\linewidth]{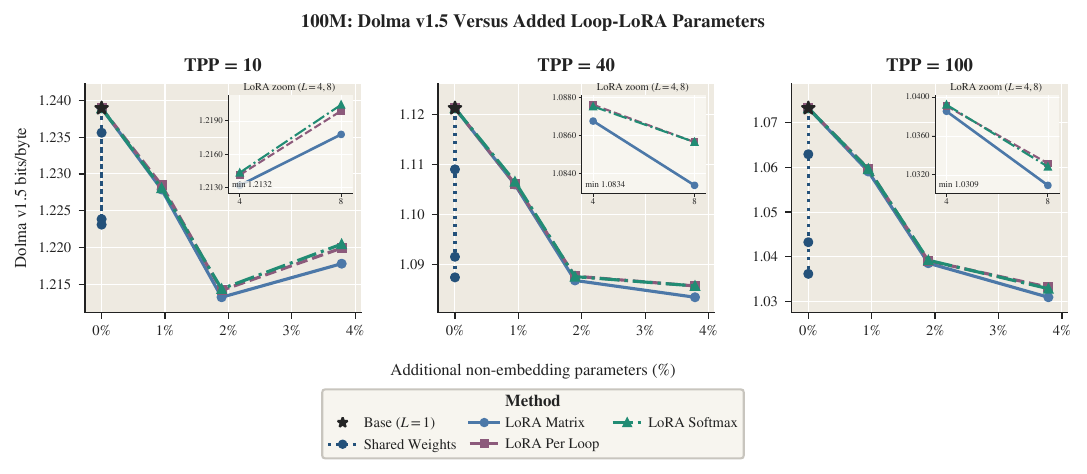}
    \par\small\textbf{(b) 100M}
  \end{minipage}

  \vspace{0.75em}

  \begin{minipage}[t]{0.485\textwidth}
    \centering
    \includegraphics[width=\linewidth]{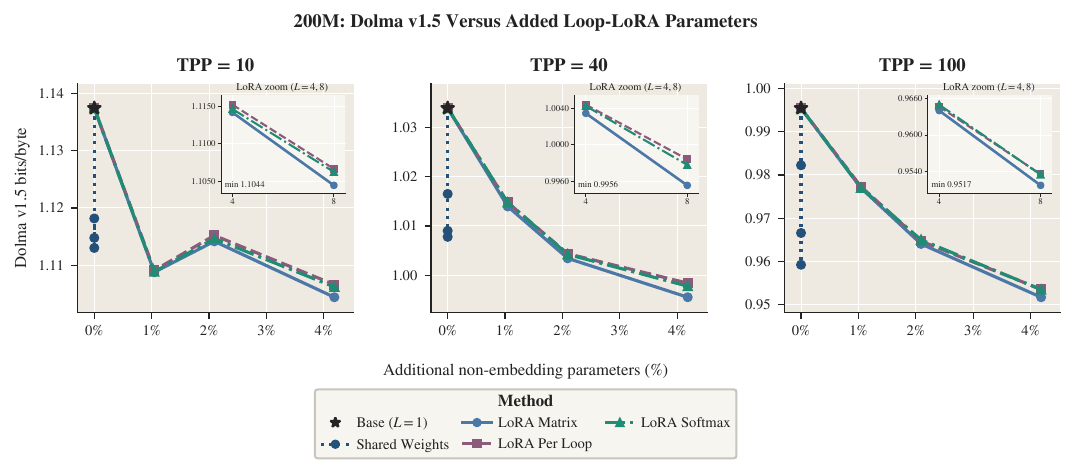}
    \par\small\textbf{(c) 200M}
  \end{minipage}
  \hfill
  \begin{minipage}[t]{0.485\textwidth}
    \centering
    \includegraphics[width=\linewidth]{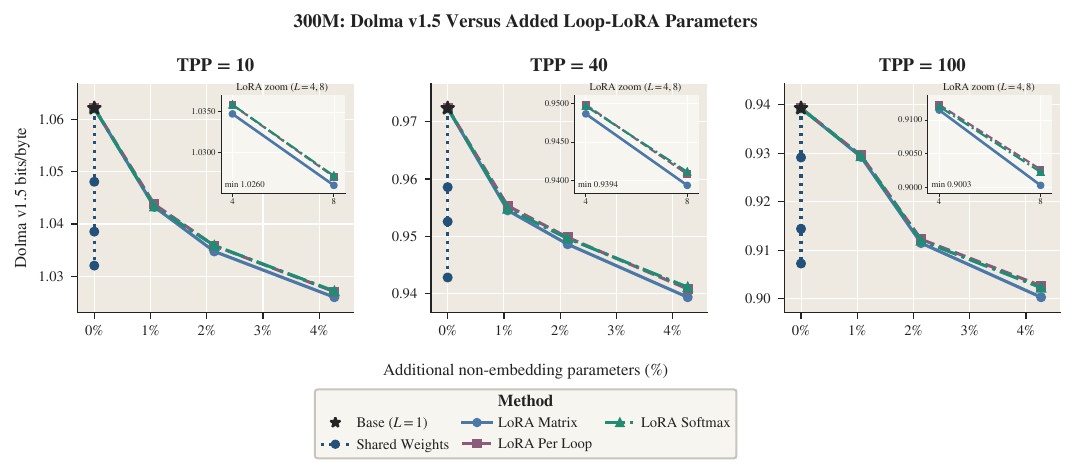}
    \par\small\textbf{(d) 300M}
  \end{minipage}
  \caption{\textbf{Specialization improves Dolma v1.5 BPB with relatively few additional parameters.} Results cover 70M--300M full-stack models at TPP
  10, 40, and 100. The horizontal axis reports additional non-embedding
  parameters relative to the identical-weight model. Higher TPP generally
  lowers BPB. Lower is better.}
  \label{fig:loop-lora-dolma-params}
\end{figure*}

\paragraph{Stored parameters.}
Across all three TPP values, loop-specific corrections improve BPB while adding relatively few stored parameters.
Specialization therefore provides a useful middle ground when parameter
storage is constrained: most weights remain shared, while each loop receives
a small correction. The corresponding comparison appears in
\autoref{fig:loop-lora-dolma-params}.

\begin{figure*}[ht]
  \centering
  \includegraphics[width=0.7\textwidth]{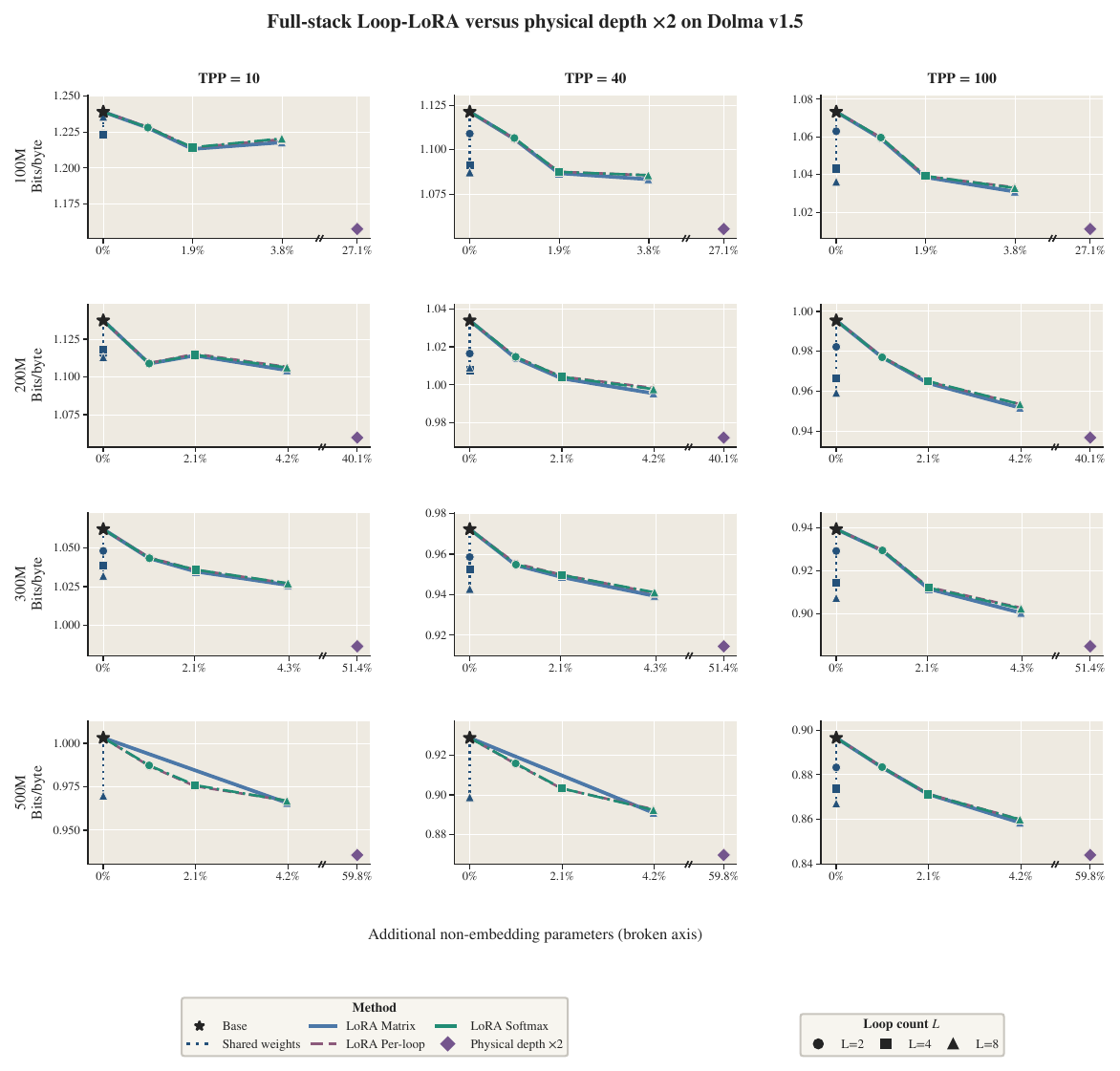}
  \caption{\textbf{Adding independently parameterized layers can achieve
  lower Dolma v1.5 BPB but requires substantially more storage.}
  Rows show 100M, 200M, 300M, and 500M models; columns show TPP 10, 40, and
  100. The purple diamond is the matched doubled-depth Transformer. The final
  tick after the broken axis gives its true increase in non-embedding
  parameters. Lower BPB is better.}
  \label{fig:loop-lora-dolma-depth}
\end{figure*}

\paragraph{Loop-specific specialization or additional Transformer layers.}
To put this parameter saving in context, we use the same doubled-depth
reference as in the DCLM comparison. This reference is a conventional
Transformer with twice as many blocks as the corresponding baseline, and
every block has its own independently learned parameters. It therefore adds
depth by storing new layers rather than by applying the same layers multiple
times. Across the evaluated TPP values, the doubled-depth Transformer can achieve lower BPB, but requires substantially more stored parameters.
\autoref{fig:loop-lora-dolma-depth} compares the two choices. The
doubled-depth references cover 100M--500M models rather than the complete
70M--300M grid used in the other comparisons.

\begin{figure*}[ht]
  \centering
  \begin{minipage}[t]{0.485\textwidth}
    \centering
    \includegraphics[width=\linewidth]{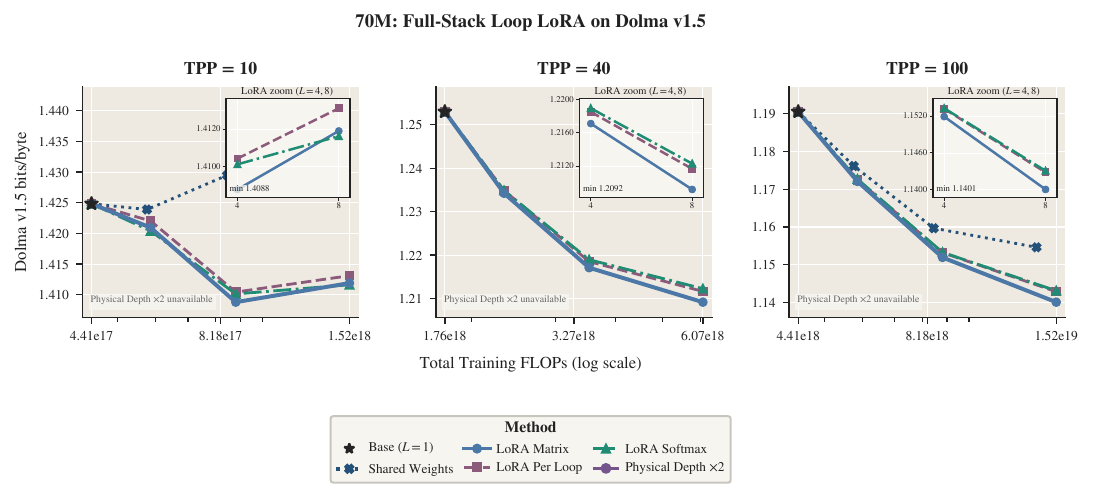}
    \par\small\textbf{(a) 70M}
  \end{minipage}
  \hfill
  \begin{minipage}[t]{0.485\textwidth}
    \centering
    \includegraphics[width=\linewidth]{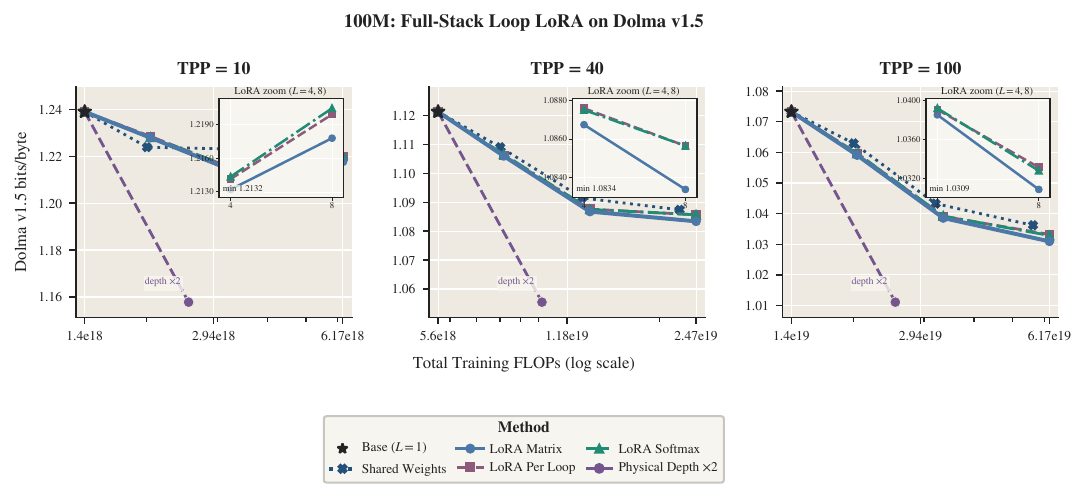}
    \par\small\textbf{(b) 100M}
  \end{minipage}

  \vspace{0.75em}

  \begin{minipage}[t]{0.485\textwidth}
    \centering
    \includegraphics[width=\linewidth]{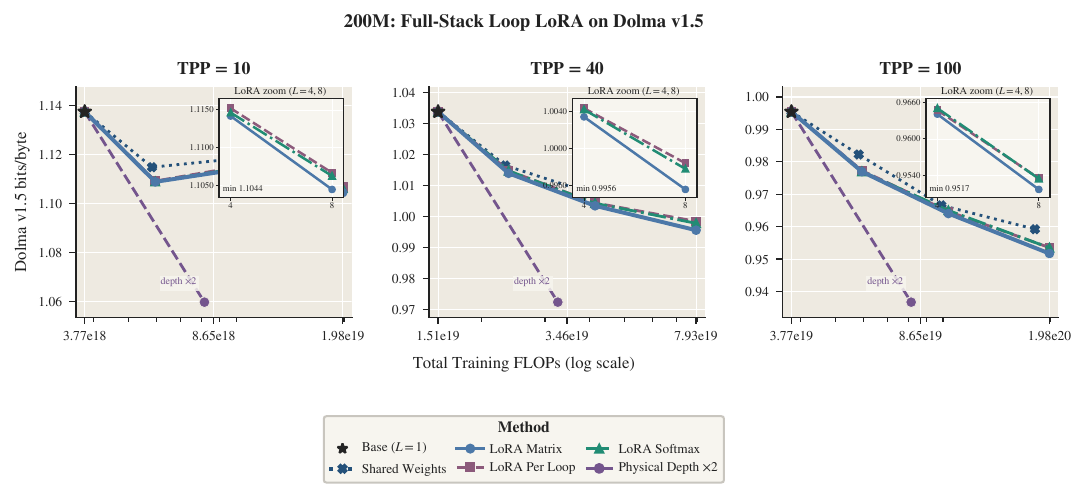}
    \par\small\textbf{(c) 200M}
  \end{minipage}
  \hfill
  \begin{minipage}[t]{0.485\textwidth}
    \centering
    \includegraphics[width=\linewidth]{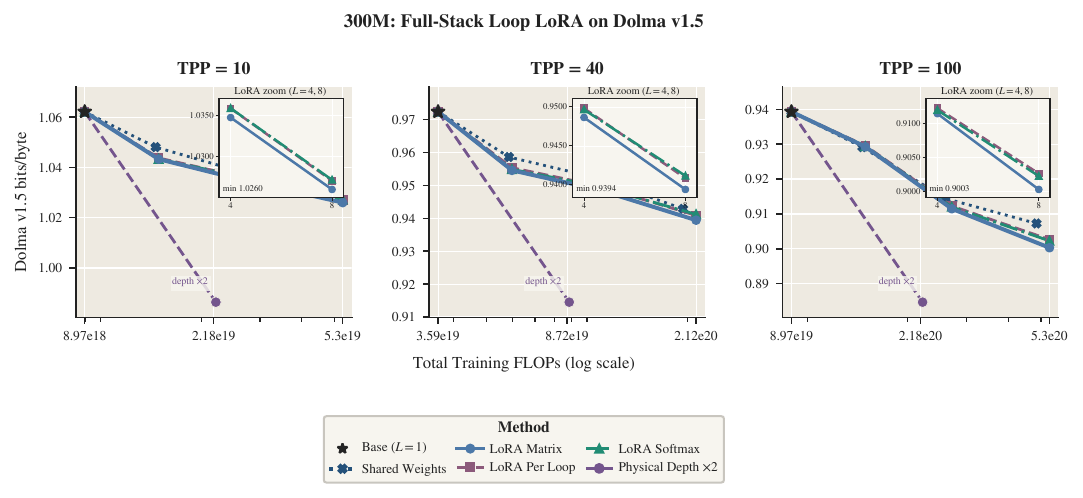}
    \par\small\textbf{(d) 300M}
  \end{minipage}
  \caption{\textbf{Low parameter overhead does not necessarily mean better training-compute efficiency.} Dolma v1.5 BPB versus total training FLOPs for
  70M--300M models at TPP 10, 40, and 100. Training FLOPs include repeated
  loop execution, loop-specific corrections, attention-score computation,
  and the realized token budget. Lower is better.}
  \label{fig:loop-lora-dolma-flops}
\end{figure*}

\paragraph{Training FLOPs.}
The FLOP comparison includes the cost of executing the additional loops and
their loop-specific corrections. At a similar number of training FLOPs, the doubled-depth Transformer can often achieve lower BPB. Therefore, when training
compute is the main constraint and more parameters can be stored, training a
deeper Transformer may be a better use of compute than adding and
specializing more loops. The complete comparison appears in
\autoref{fig:loop-lora-dolma-flops}.

\paragraph{Dolma v1.5 summary.}
Dolma v1.5 yields the same main conclusion. Specialization provides a small benefit at $L=2$ and a clearer benefit at
$L=4$ and $L=8$, while the precise specialization method matters less than
allowing the loops to specialize. Loop Matrix is most often the strongest
choice.
Increasing TPP generally lowers BPB for both identical-weight and specialized
models, so training on more tokens improves overall quality without changing
the main loop-count pattern. In terms of resources, specialization is most
useful when stored parameters are constrained. If training FLOPs are instead
the main constraint and more parameters can be stored, a deeper Transformer
can provide better quality for the same amount of training compute.

\clearpage

\subsubsection{RedPajama}
\label{app:loop-lora-redpajama}

RedPajama provides another broad language-modeling evaluation corpus with a
text distribution distinct from both DCLM and Dolma. It gives an additional
test of whether the result transfers across general-language corpora. We
repeat the four
comparisons used for DCLM: loop count, added parameter storage, additional
independently parameterized layers, and total training FLOPs. We evaluate
70M--300M full-stack models at TPP 10, 40, and 100. The doubled-depth
comparison also includes 500M models.

\begin{figure*}[ht]
  \centering
  \begin{minipage}[t]{0.485\textwidth}
    \centering
    \includegraphics[width=\linewidth]{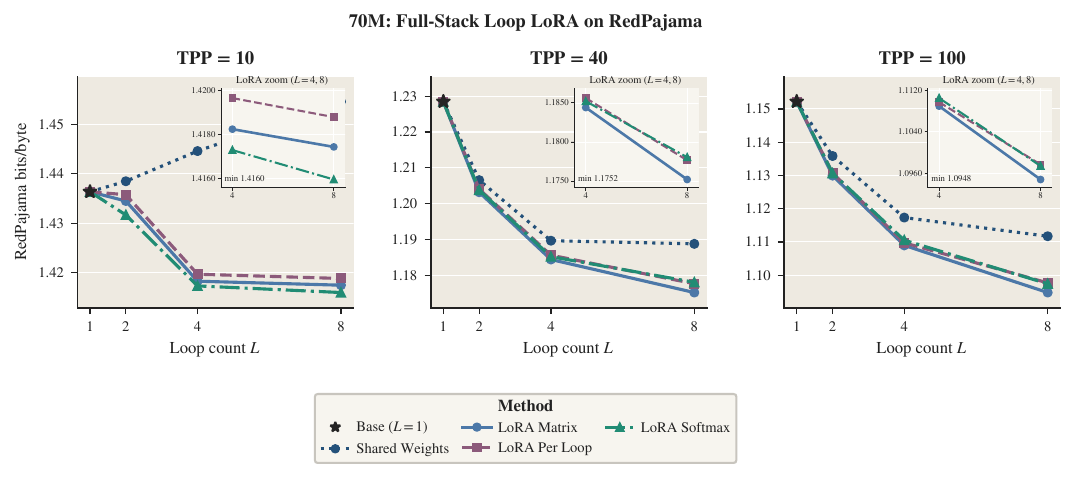}
    \par\small\textbf{(a) 70M}
  \end{minipage}
  \hfill
  \begin{minipage}[t]{0.485\textwidth}
    \centering
    \includegraphics[width=\linewidth]{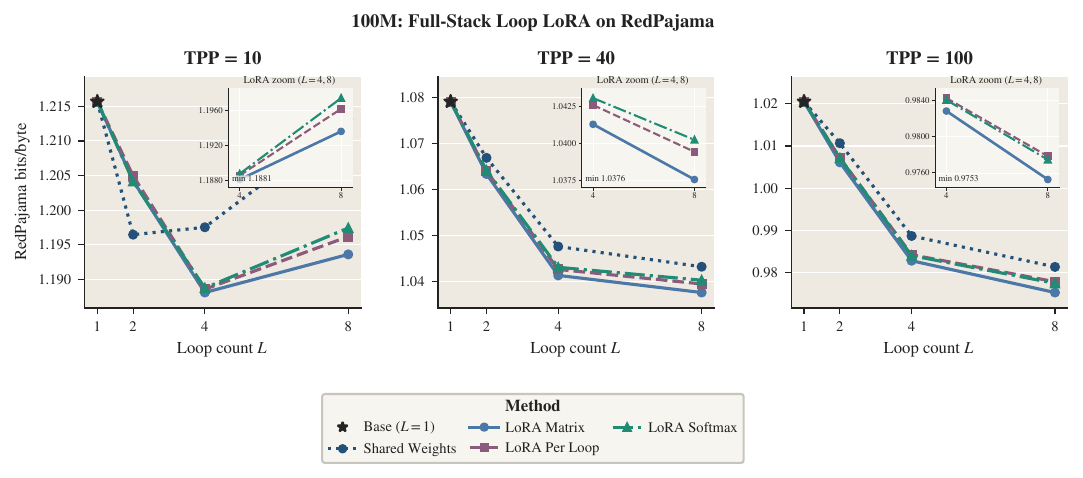}
    \par\small\textbf{(b) 100M}
  \end{minipage}

  \vspace{0.75em}

  \begin{minipage}[t]{0.485\textwidth}
    \centering
    \includegraphics[width=\linewidth]{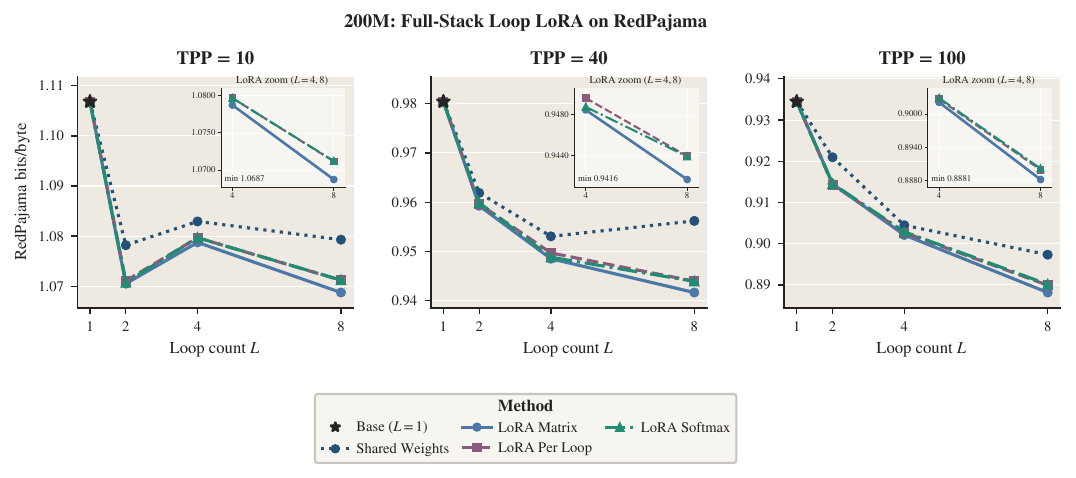}
    \par\small\textbf{(c) 200M}
  \end{minipage}
  \hfill
  \begin{minipage}[t]{0.485\textwidth}
    \centering
    \includegraphics[width=\linewidth]{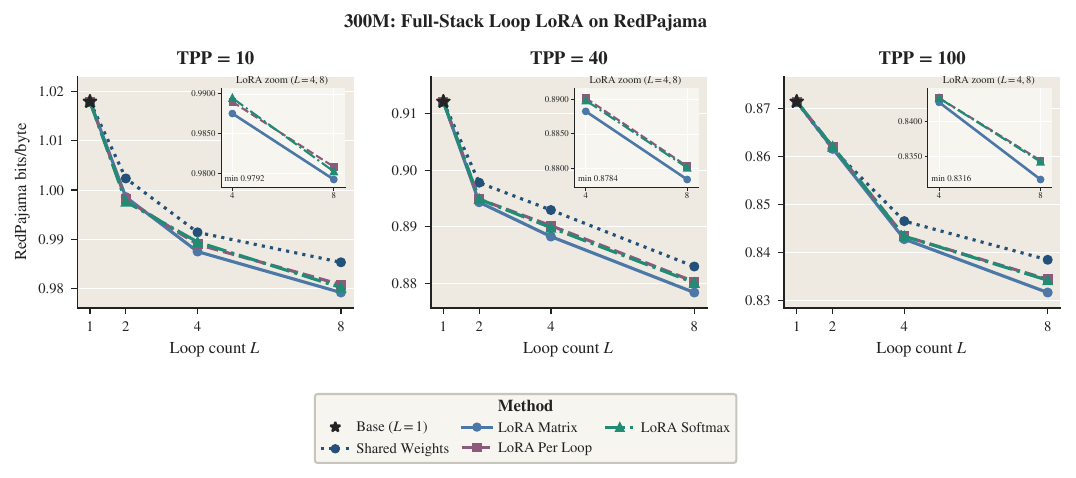}
    \par\small\textbf{(d) 300M}
  \end{minipage}
  \caption{\textbf{Specialization generally helps more as the number of loops increases.} RedPajama BPB versus loop count for 70M--300M
  full-stack models at TPP 10, 40, and 100. Specialization is generally more
  useful at $L=4$ and $L=8$ than at $L=2$. Increasing TPP also generally
  lowers BPB. Lower is better.}
  \label{fig:loop-lora-redpajama-loops}
\end{figure*}

\paragraph{Loop count.}
The results again show only a small benefit from specialization at
$L=2$ and a clearer benefit at $L=4$ and $L=8$, across TPP 10, 40, and 100.
The precise method matters less than allowing the loops to differ, although
Loop Matrix most frequently reaches the lowest BPB. Increasing TPP also
generally improves quality.
\autoref{fig:loop-lora-redpajama-loops} reports the complete loop-count
comparison.

\begin{figure*}[ht]
  \centering
  \begin{minipage}[t]{0.485\textwidth}
    \centering
    \includegraphics[width=\linewidth]{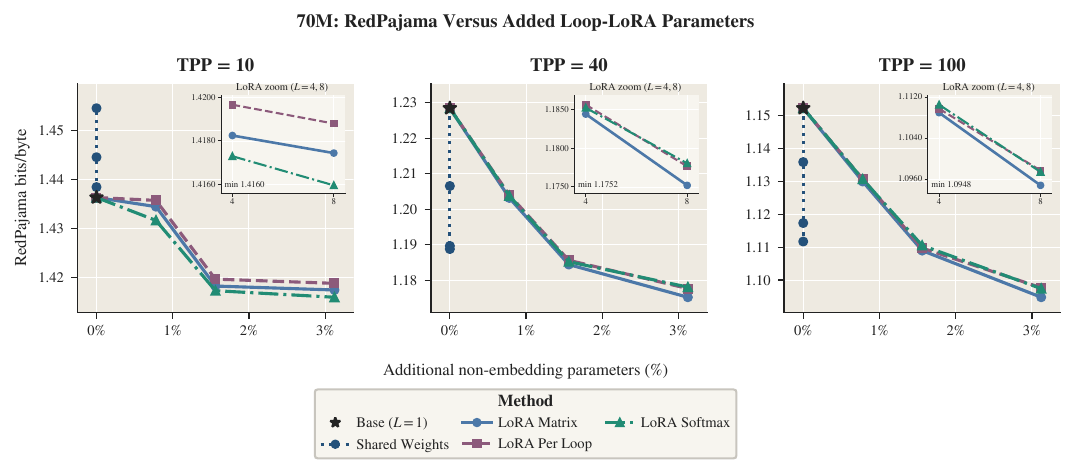}
    \par\small\textbf{(a) 70M}
  \end{minipage}
  \hfill
  \begin{minipage}[t]{0.485\textwidth}
    \centering
    \includegraphics[width=\linewidth]{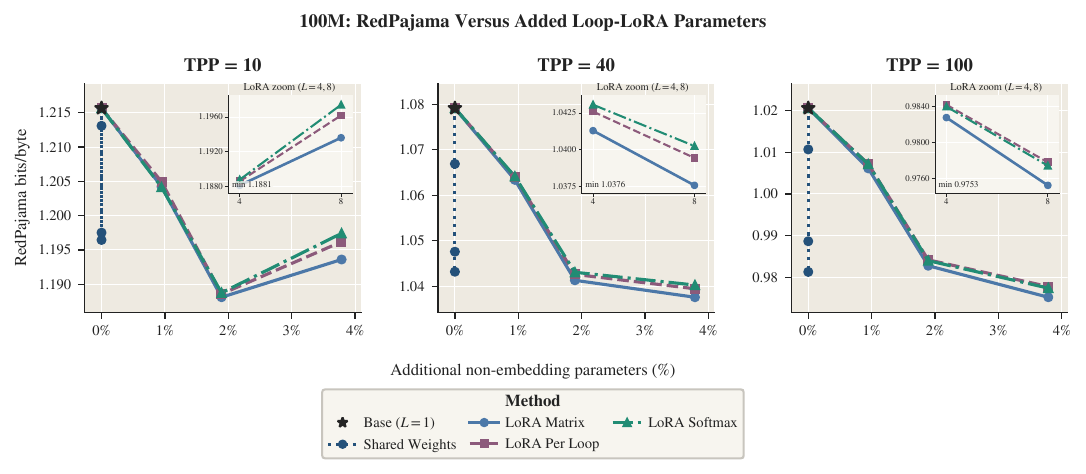}
    \par\small\textbf{(b) 100M}
  \end{minipage}

  \vspace{0.75em}

  \begin{minipage}[t]{0.485\textwidth}
    \centering
    \includegraphics[width=\linewidth]{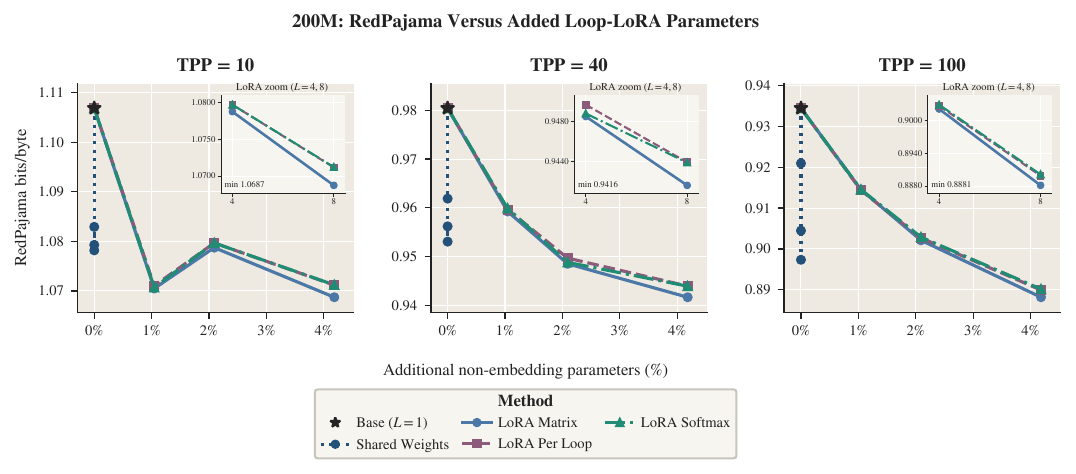}
    \par\small\textbf{(c) 200M}
  \end{minipage}
  \hfill
  \begin{minipage}[t]{0.485\textwidth}
    \centering
    \includegraphics[width=\linewidth]{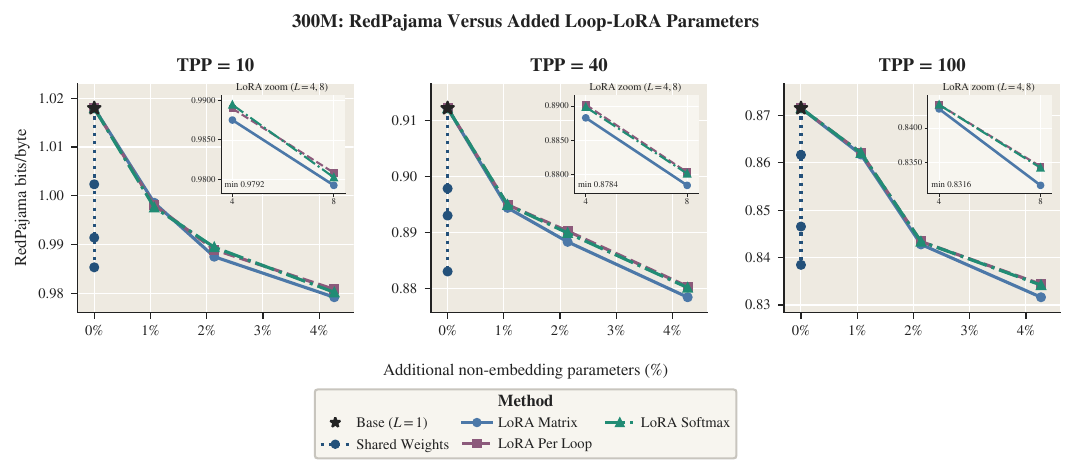}
    \par\small\textbf{(d) 300M}
  \end{minipage}
  \caption{\textbf{Specialization improves RedPajama BPB with relatively few additional parameters.} Results cover 70M--300M full-stack models at TPP
  10, 40, and 100. The horizontal axis reports additional non-embedding
  parameters relative to the identical-weight model. Higher TPP generally
  lowers BPB. Lower is better.}
  \label{fig:loop-lora-redpajama-params}
\end{figure*}

\paragraph{Stored parameters.}
Across all three TPP values, loop-specific corrections improve BPB while adding relatively few stored parameters.
Specialization therefore provides a useful middle ground when parameter
storage is constrained: most weights remain shared, while each loop receives
a small correction. The corresponding comparison appears in
\autoref{fig:loop-lora-redpajama-params}.

\begin{figure*}[ht]
  \centering
  \includegraphics[width=0.7\textwidth]{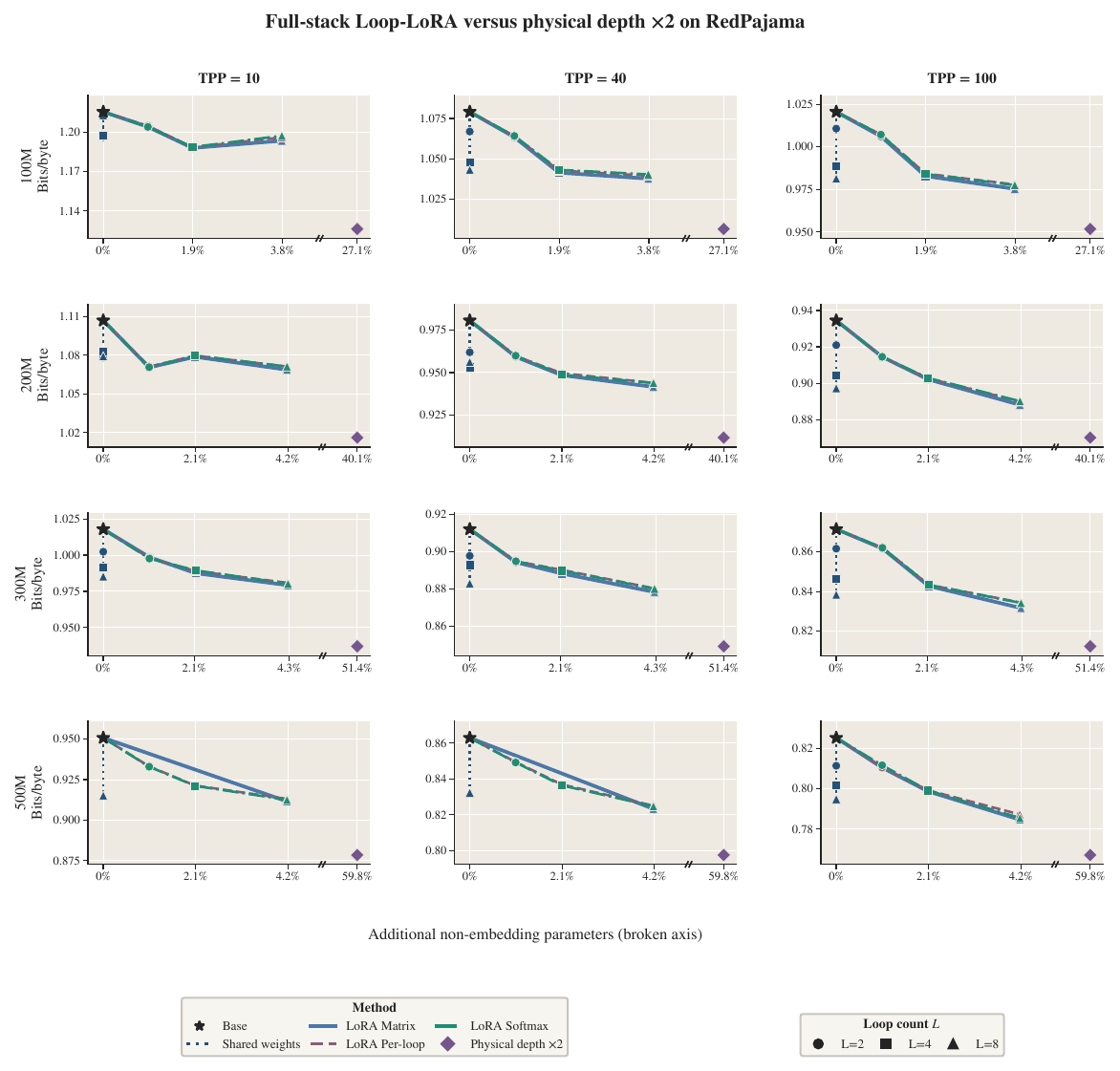}
  \caption{\textbf{Adding independently parameterized layers can achieve
  lower RedPajama BPB but requires substantially more storage.}
  Rows show 100M, 200M, 300M, and 500M models; columns show TPP 10, 40, and
  100. The purple diamond is the matched doubled-depth Transformer. The final
  tick after the broken axis gives its true increase in non-embedding
  parameters. Lower BPB is better.}
  \label{fig:loop-lora-redpajama-depth}
\end{figure*}

\paragraph{Loop-specific specialization or additional Transformer layers.}
To put this parameter saving in context, we use the same doubled-depth
reference as in the DCLM comparison. This reference is a conventional
Transformer with twice as many blocks as the corresponding baseline, and
every block has its own independently learned parameters. It therefore adds
depth by storing new layers rather than by applying the same layers multiple
times. Across the evaluated TPP values, the doubled-depth Transformer can achieve lower BPB, but requires substantially more stored parameters.
\autoref{fig:loop-lora-redpajama-depth} compares the two choices. The
doubled-depth references cover 100M--500M models rather than the complete
70M--300M grid used in the other comparisons.

\begin{figure*}[ht]
  \centering
  \begin{minipage}[t]{0.485\textwidth}
    \centering
    \includegraphics[width=\linewidth]{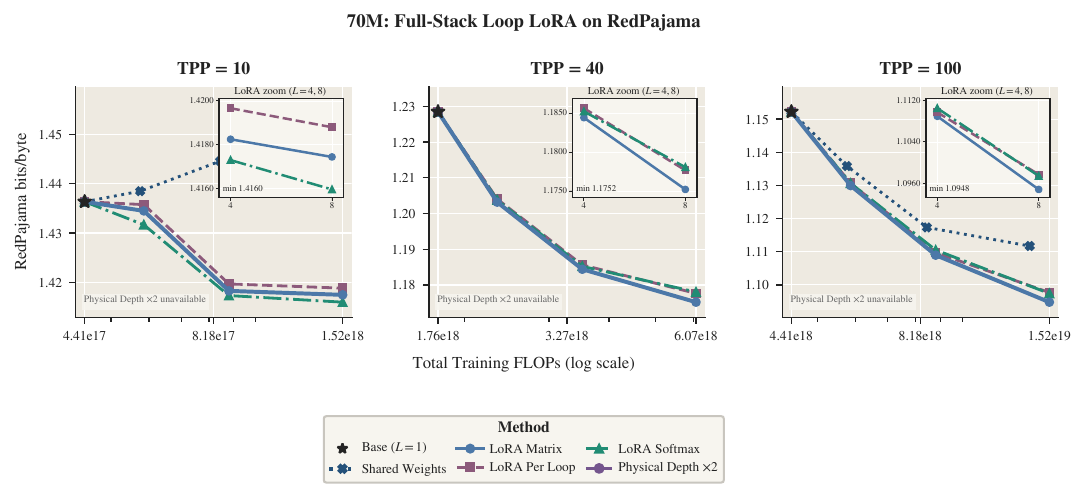}
    \par\small\textbf{(a) 70M}
  \end{minipage}
  \hfill
  \begin{minipage}[t]{0.485\textwidth}
    \centering
    \includegraphics[width=\linewidth]{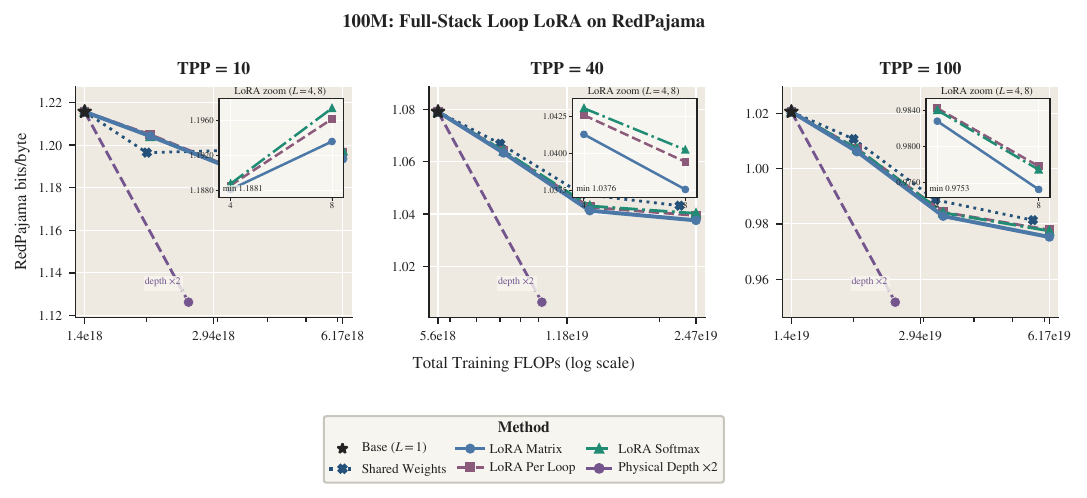}
    \par\small\textbf{(b) 100M}
  \end{minipage}

  \vspace{0.75em}

  \begin{minipage}[t]{0.485\textwidth}
    \centering
    \includegraphics[width=\linewidth]{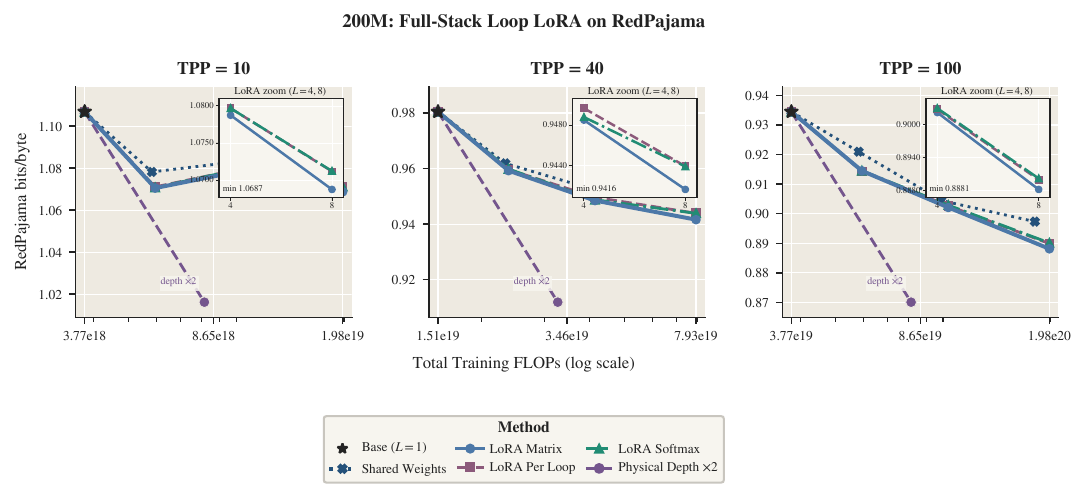}
    \par\small\textbf{(c) 200M}
  \end{minipage}
  \hfill
  \begin{minipage}[t]{0.485\textwidth}
    \centering
    \includegraphics[width=\linewidth]{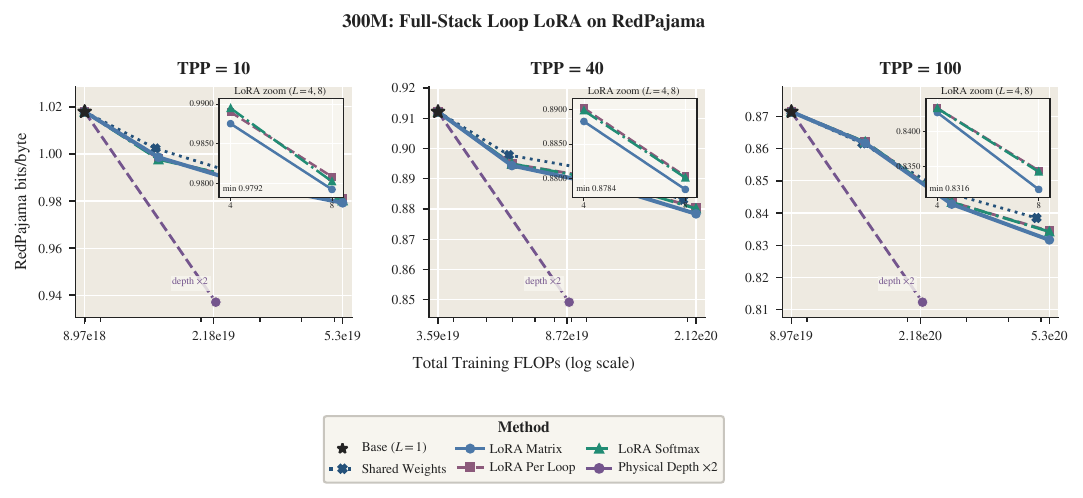}
    \par\small\textbf{(d) 300M}
  \end{minipage}
  \caption{\textbf{Low parameter overhead does not necessarily mean better training-compute efficiency.} RedPajama BPB versus total training FLOPs for
  70M--300M models at TPP 10, 40, and 100. Training FLOPs include repeated
  loop execution, loop-specific corrections, attention-score computation,
  and the realized token budget. Lower is better.}
  \label{fig:loop-lora-redpajama-flops}
\end{figure*}

\paragraph{Training FLOPs.}
The FLOP comparison includes the cost of executing the additional loops and
their loop-specific corrections. At a similar number of training FLOPs, the doubled-depth Transformer can often achieve lower BPB. Therefore, when training
compute is the main constraint and more parameters can be stored, training a
deeper Transformer may be a better use of compute than adding and
specializing more loops. The complete comparison appears in
\autoref{fig:loop-lora-redpajama-flops}.

\paragraph{RedPajama summary.}
RedPajama again supports the same resource-dependent conclusion. Specialization provides a small benefit at $L=2$ and a clearer benefit at
$L=4$ and $L=8$, while the precise specialization method matters less than
allowing the loops to specialize. Loop Matrix is most often the strongest
choice.
Increasing TPP generally lowers BPB for both identical-weight and specialized
models, so training on more tokens improves overall quality without changing
the main loop-count pattern. In terms of resources, specialization is most
useful when stored parameters are constrained. If training FLOPs are instead
the main constraint and more parameters can be stored, a deeper Transformer
can provide better quality for the same amount of training compute.

\clearpage

\subsubsection{Dolma Programming Languages}
\label{app:loop-lora-programming}

Dolma Programming Languages is a code-heavy evaluation corpus, so it tests
whether the specialization result extends beyond general-language corpora.
We repeat the four
comparisons used for DCLM: loop count, added parameter storage, additional
independently parameterized layers, and total training FLOPs. We evaluate
70M--300M full-stack models at TPP 10, 40, and 100. The doubled-depth
comparison also includes 500M models.

\begin{figure*}[ht]
  \centering
  \begin{minipage}[t]{0.485\textwidth}
    \centering
    \includegraphics[width=\linewidth]{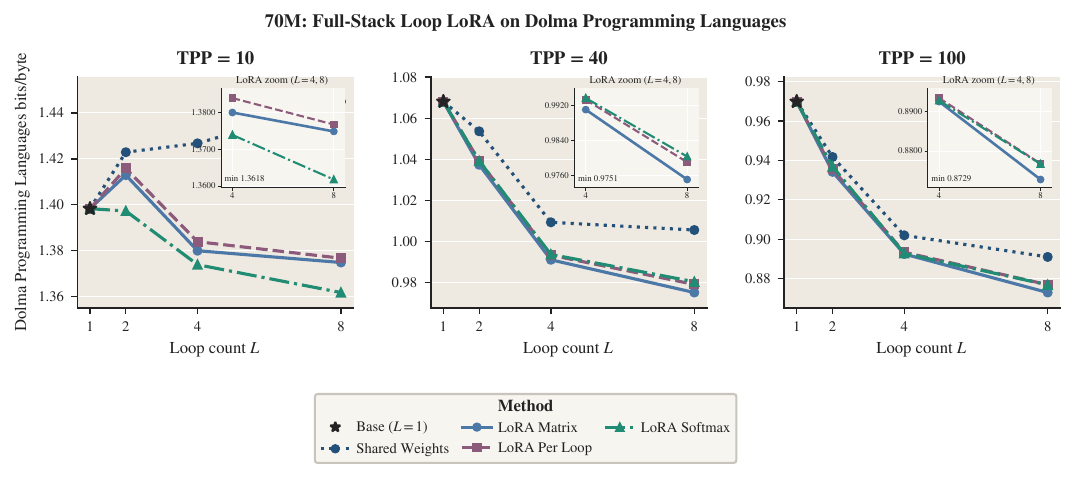}
    \par\small\textbf{(a) 70M}
  \end{minipage}
  \hfill
  \begin{minipage}[t]{0.485\textwidth}
    \centering
    \includegraphics[width=\linewidth]{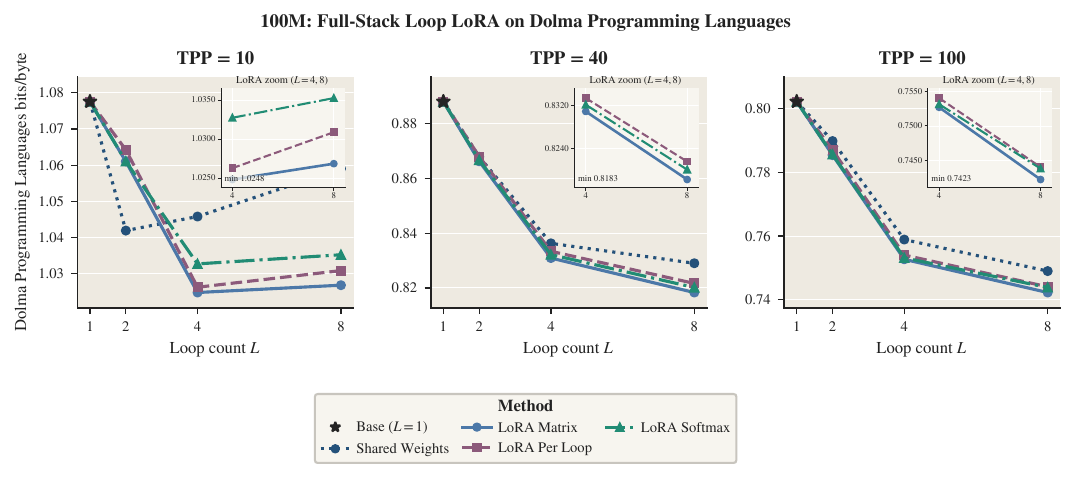}
    \par\small\textbf{(b) 100M}
  \end{minipage}

  \vspace{0.75em}

  \begin{minipage}[t]{0.485\textwidth}
    \centering
    \includegraphics[width=\linewidth]{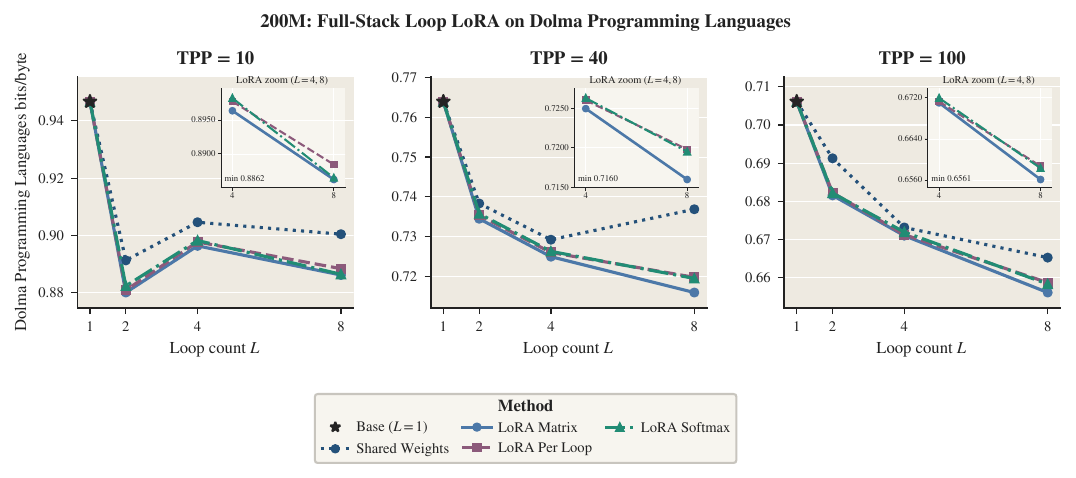}
    \par\small\textbf{(c) 200M}
  \end{minipage}
  \hfill
  \begin{minipage}[t]{0.485\textwidth}
    \centering
    \includegraphics[width=\linewidth]{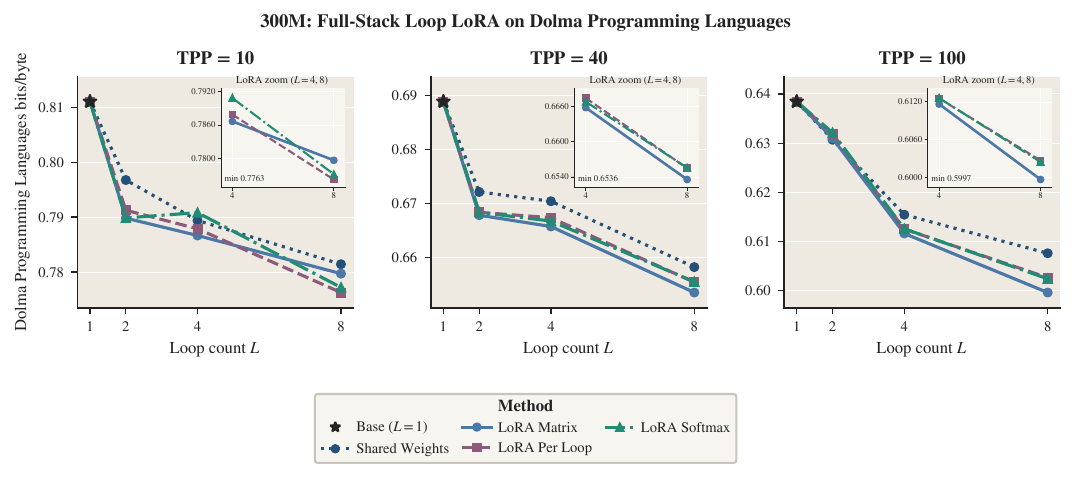}
    \par\small\textbf{(d) 300M}
  \end{minipage}
  \caption{\textbf{Specialization generally helps more as the number of loops increases.} Dolma Programming Languages BPB versus loop count for 70M--300M
  full-stack models at TPP 10, 40, and 100. Specialization is generally more
  useful at $L=4$ and $L=8$ than at $L=2$. Increasing TPP also generally
  lowers BPB. Lower is better.}
  \label{fig:loop-lora-programming-loops}
\end{figure*}

\paragraph{Loop count.}
The code results show the same broad tendency, but with more variation
than on the general-language corpora. Improvements at $L=2$ are small or
inconsistent, while specialization is more often useful at $L=4$ and $L=8$.
The three methods usually remain close, and Loop Matrix is often strong, but
no method is uniformly best in every setting. Increasing TPP generally lowers
BPB.
\autoref{fig:loop-lora-programming-loops} reports the complete loop-count
comparison.

\begin{figure*}[ht]
  \centering
  \begin{minipage}[t]{0.485\textwidth}
    \centering
    \includegraphics[width=\linewidth]{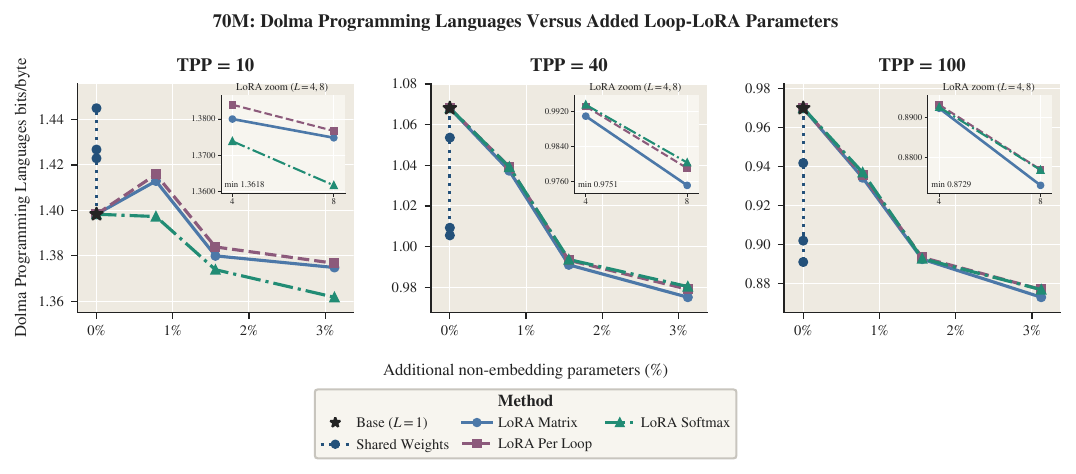}
    \par\small\textbf{(a) 70M}
  \end{minipage}
  \hfill
  \begin{minipage}[t]{0.485\textwidth}
    \centering
    \includegraphics[width=\linewidth]{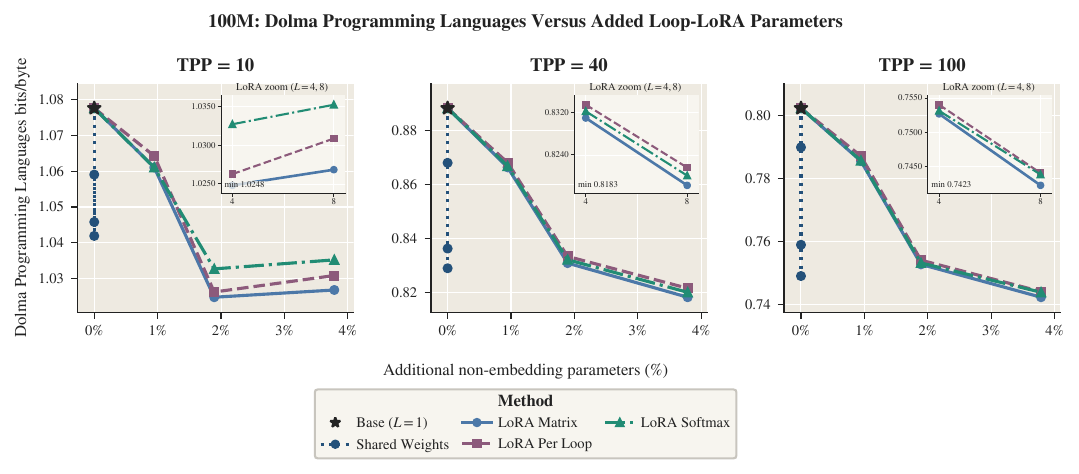}
    \par\small\textbf{(b) 100M}
  \end{minipage}

  \vspace{0.75em}

  \begin{minipage}[t]{0.485\textwidth}
    \centering
    \includegraphics[width=\linewidth]{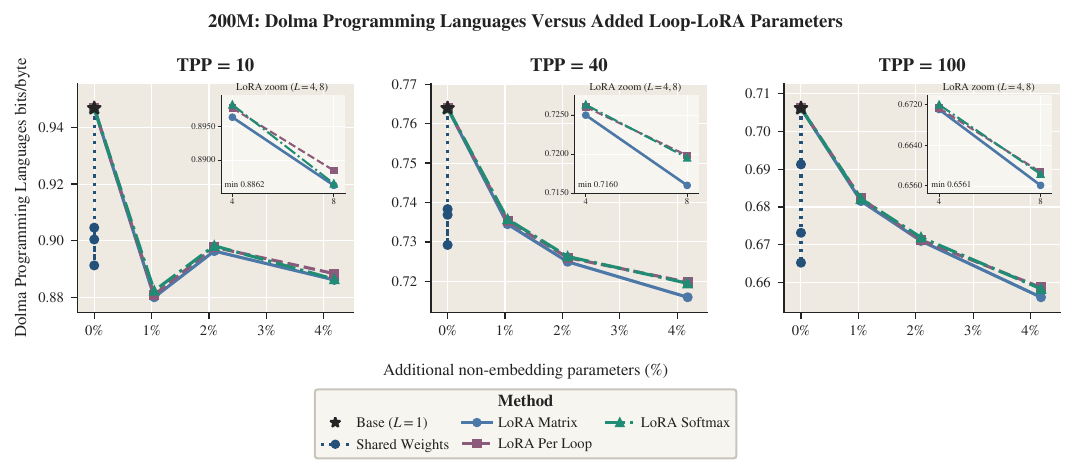}
    \par\small\textbf{(c) 200M}
  \end{minipage}
  \hfill
  \begin{minipage}[t]{0.485\textwidth}
    \centering
    \includegraphics[width=\linewidth]{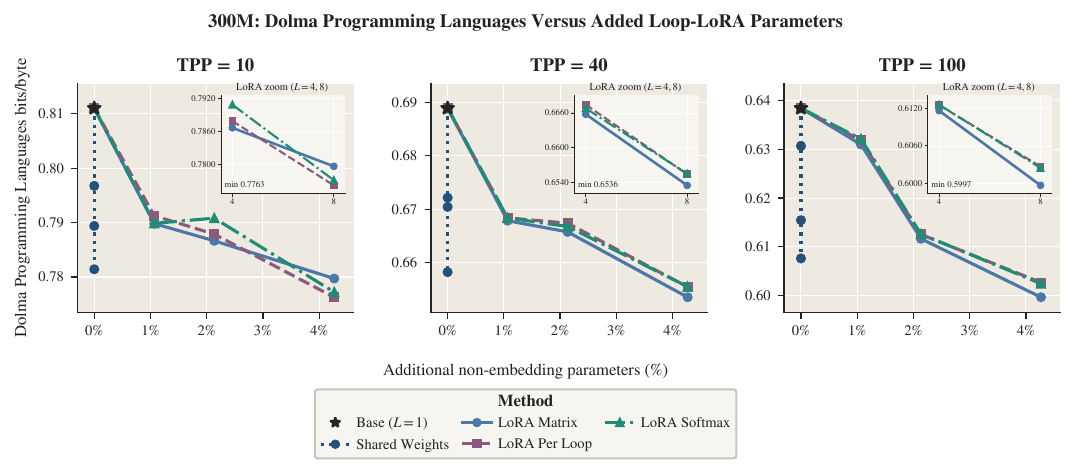}
    \par\small\textbf{(d) 300M}
  \end{minipage}
  \caption{\textbf{Specialization can improve Dolma Programming Languages
  BPB with relatively few additional parameters.} Results cover 70M--300M full-stack models at TPP
  10, 40, and 100. The horizontal axis reports additional non-embedding
  parameters relative to the identical-weight model. Higher TPP generally
  lowers BPB. Lower is better.}
  \label{fig:loop-lora-programming-params}
\end{figure*}

\paragraph{Stored parameters.}
When specialization improves code BPB, it does so with only a small increase in stored parameters.
Specialization therefore provides a useful middle ground when parameter
storage is constrained: most weights remain shared, while each loop receives
a small correction. The corresponding comparison appears in
\autoref{fig:loop-lora-programming-params}.

\begin{figure*}[ht]
  \centering
  \includegraphics[width=0.7\textwidth]{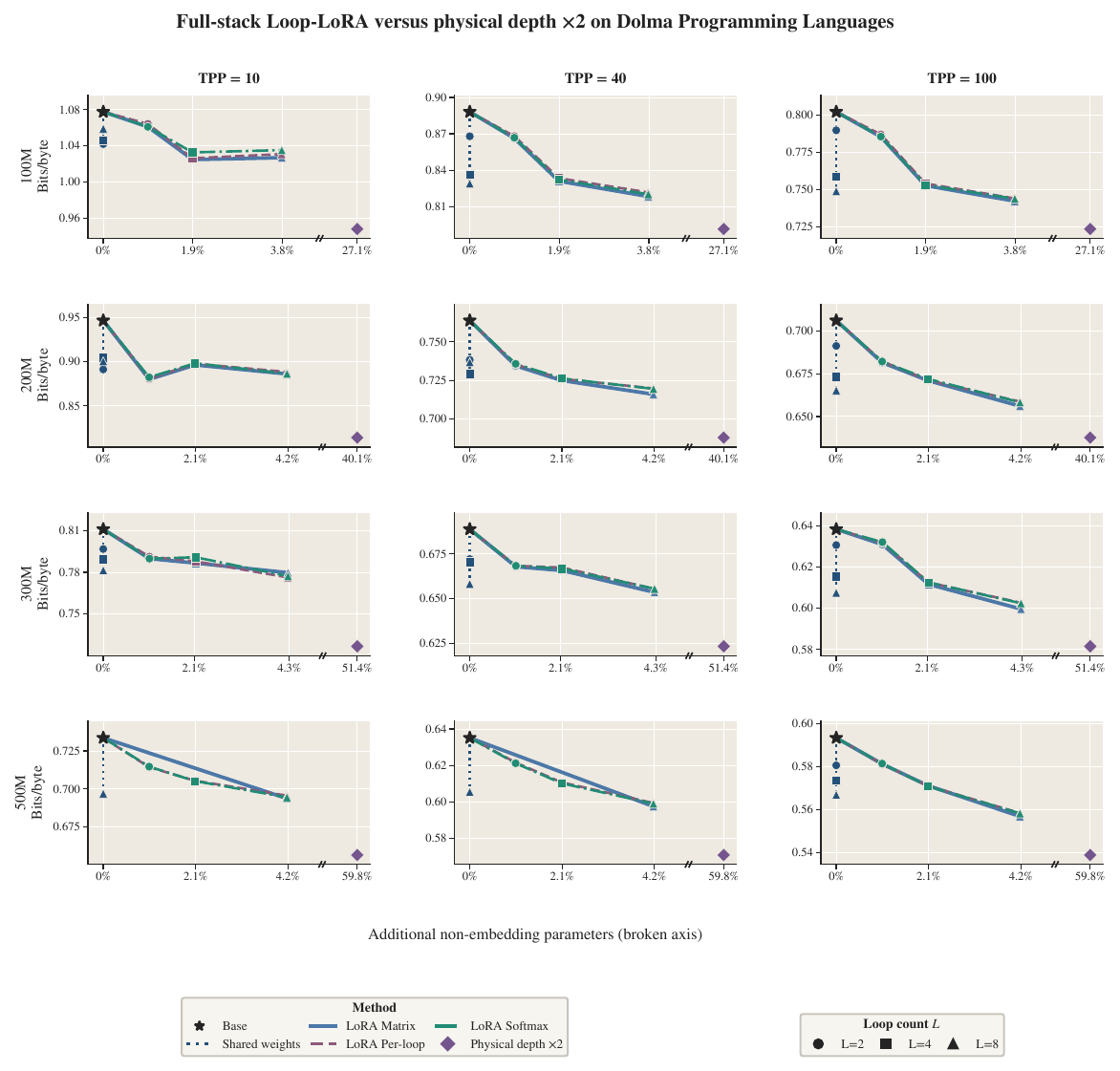}
  \caption{\textbf{Adding independently parameterized layers can achieve
  lower Dolma Programming Languages BPB but requires substantially more storage.}
  Rows show 100M, 200M, 300M, and 500M models; columns show TPP 10, 40, and
  100. The purple diamond is the matched doubled-depth Transformer. The final
  tick after the broken axis gives its true increase in non-embedding
  parameters. Lower BPB is better.}
  \label{fig:loop-lora-programming-depth}
\end{figure*}

\paragraph{Loop-specific specialization or additional Transformer layers.}
To put this parameter saving in context, we use the same doubled-depth
reference as in the DCLM comparison. This reference is a conventional
Transformer with twice as many blocks as the corresponding baseline, and
every block has its own independently learned parameters. It therefore adds
depth by storing new layers rather than by applying the same layers multiple
times. The doubled-depth Transformer can achieve lower BPB in several settings, but requires substantially more stored parameters.
\autoref{fig:loop-lora-programming-depth} compares the two choices. The
doubled-depth references cover 100M--500M models rather than the complete
70M--300M grid used in the other comparisons.

\begin{figure*}[ht]
  \centering
  \begin{minipage}[t]{0.485\textwidth}
    \centering
    \includegraphics[width=\linewidth]{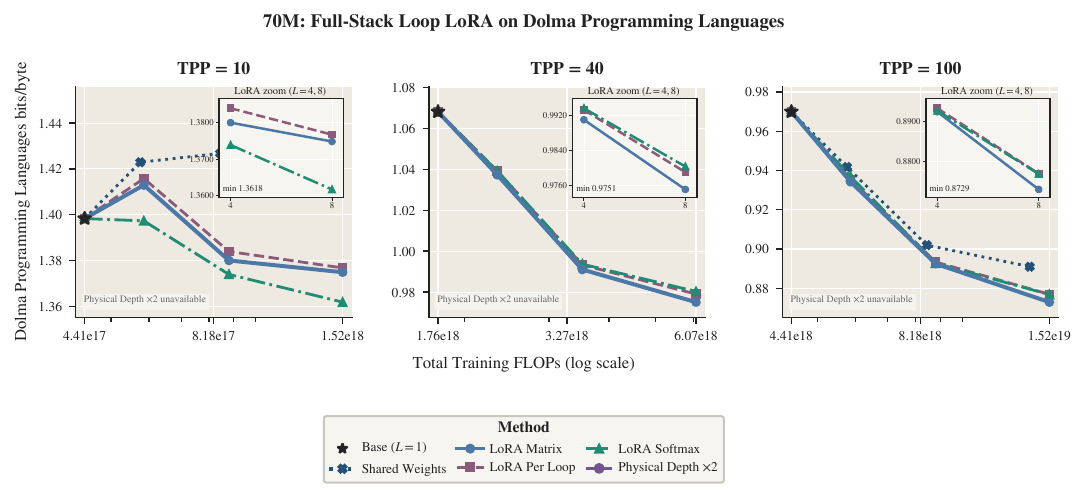}
    \par\small\textbf{(a) 70M}
  \end{minipage}
  \hfill
  \begin{minipage}[t]{0.485\textwidth}
    \centering
    \includegraphics[width=\linewidth]{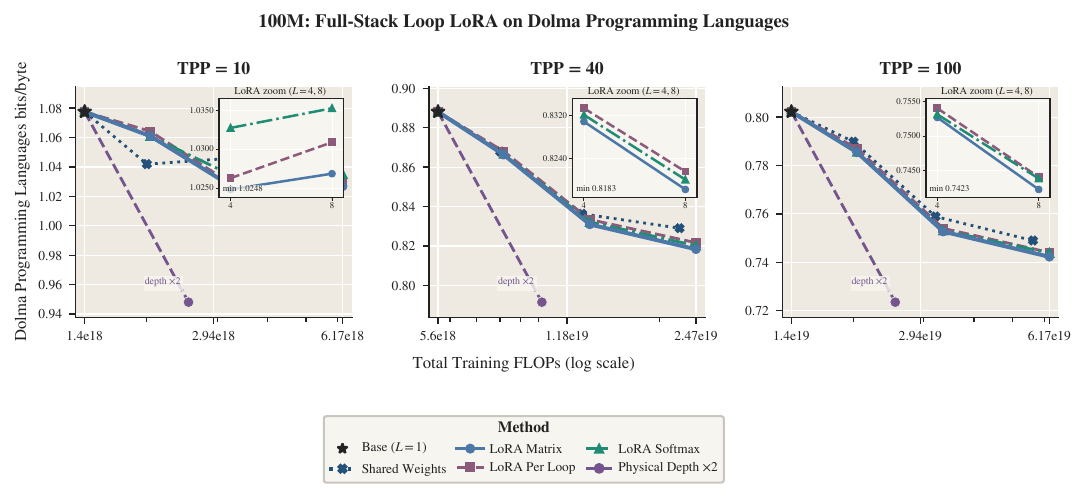}
    \par\small\textbf{(b) 100M}
  \end{minipage}

  \vspace{0.75em}

  \begin{minipage}[t]{0.485\textwidth}
    \centering
    \includegraphics[width=\linewidth]{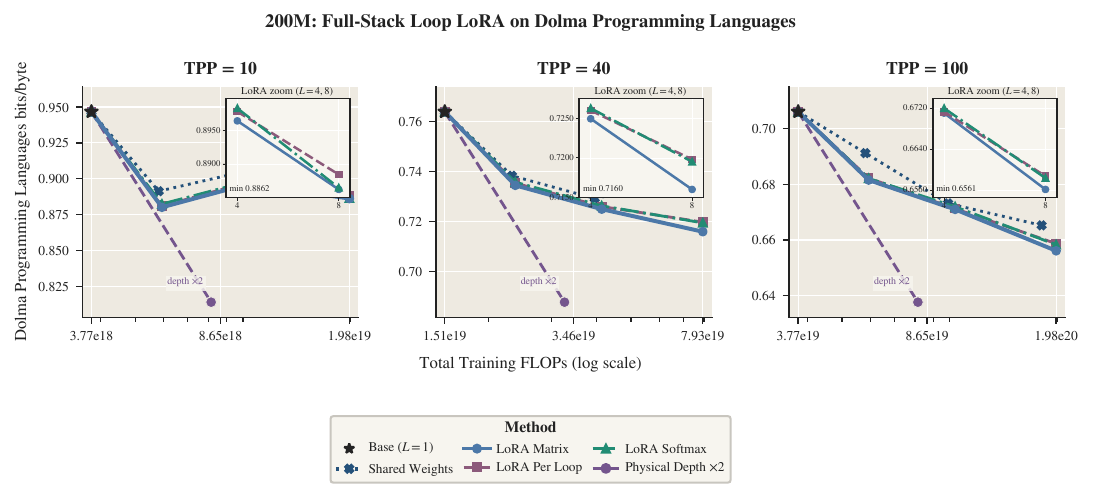}
    \par\small\textbf{(c) 200M}
  \end{minipage}
  \hfill
  \begin{minipage}[t]{0.485\textwidth}
    \centering
    \includegraphics[width=\linewidth]{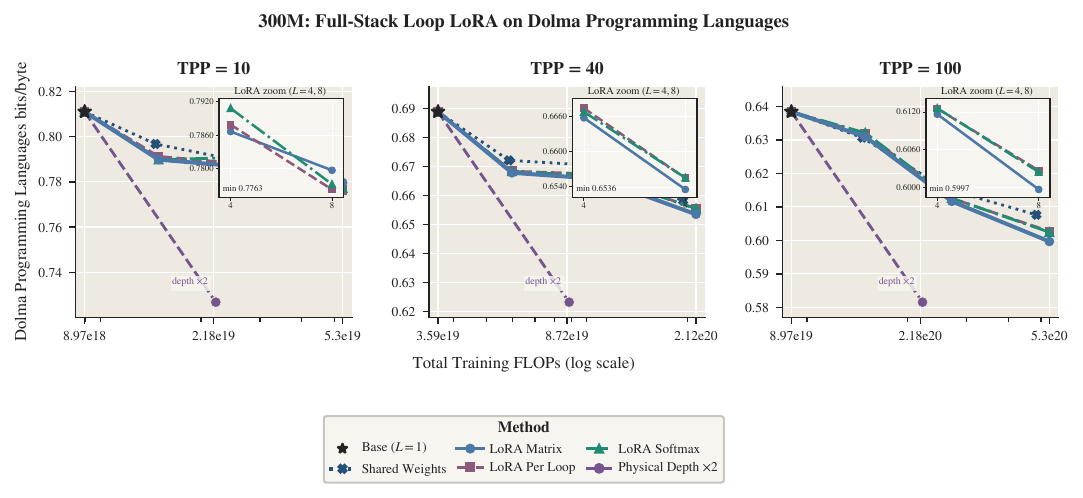}
    \par\small\textbf{(d) 300M}
  \end{minipage}
  \caption{\textbf{Low parameter overhead does not necessarily mean better training-compute efficiency.} Dolma Programming Languages BPB versus total training FLOPs for
  70M--300M models at TPP 10, 40, and 100. Training FLOPs include repeated
  loop execution, loop-specific corrections, attention-score computation,
  and the realized token budget. Lower is better.}
  \label{fig:loop-lora-programming-flops}
\end{figure*}

\paragraph{Training FLOPs.}
The FLOP comparison includes the cost of executing the additional loops and
their loop-specific corrections. At comparable training FLOPs, the doubled-depth Transformer can be stronger in several settings. Therefore, when training
compute is the main constraint and more parameters can be stored, training a
deeper Transformer may be a better use of compute than adding and
specializing more loops. The complete comparison appears in
\autoref{fig:loop-lora-programming-flops}.

\paragraph{Dolma Programming Languages summary.}
The code-heavy evaluation broadly supports the same conclusion, although its gains are less uniform. Specialization is more often useful at $L=4$ and $L=8$ than at $L=2$,
but the improvements vary more across model size and TPP than on the
general-language corpora.
Increasing TPP generally lowers BPB for both identical-weight and specialized
models, so training on more tokens improves overall quality without changing
the main loop-count pattern. In terms of resources, specialization is most
useful when stored parameters are constrained. If training FLOPs are instead
the main constraint and more parameters can be stored, a deeper Transformer
can provide better quality for the same amount of training compute.

\clearpage

\subsection{Mathematical and Code Likelihood Evaluations from 70M to 300M}
\label{app:loop-lora-math-code}

We next test whether the specialization findings extend from broad
language-modeling corpora to mathematical reasoning and code. We evaluate the
same 70M--300M full-stack models at TPP 10, 40, and 100, with loop counts
$L\in\{2,4,8\}$. For GSM8K and MATH, BPB is computed by teacher-forcing the
reference solution. For HumanEval and MBPP, BPB is computed by
teacher-forcing the reference code. As these datasets are smaller and more specialized than the
language-modeling corpora, their comparisons vary more across model size,
TPP, and loop count.

\subsubsection{GSM8K}
\label{app:loop-lora-gsm8k}

GSM8K tests whether the conclusions from the language-modeling corpora
extend to mathematical solution text. We evaluate 70M--300M full-stack models
at TPP 10, 40, and 100. These results measure teacher-forced likelihood of the
reference solution, not final-answer accuracy.

\begin{figure*}[ht]
  \centering
  \includegraphics[width=0.94\textwidth]{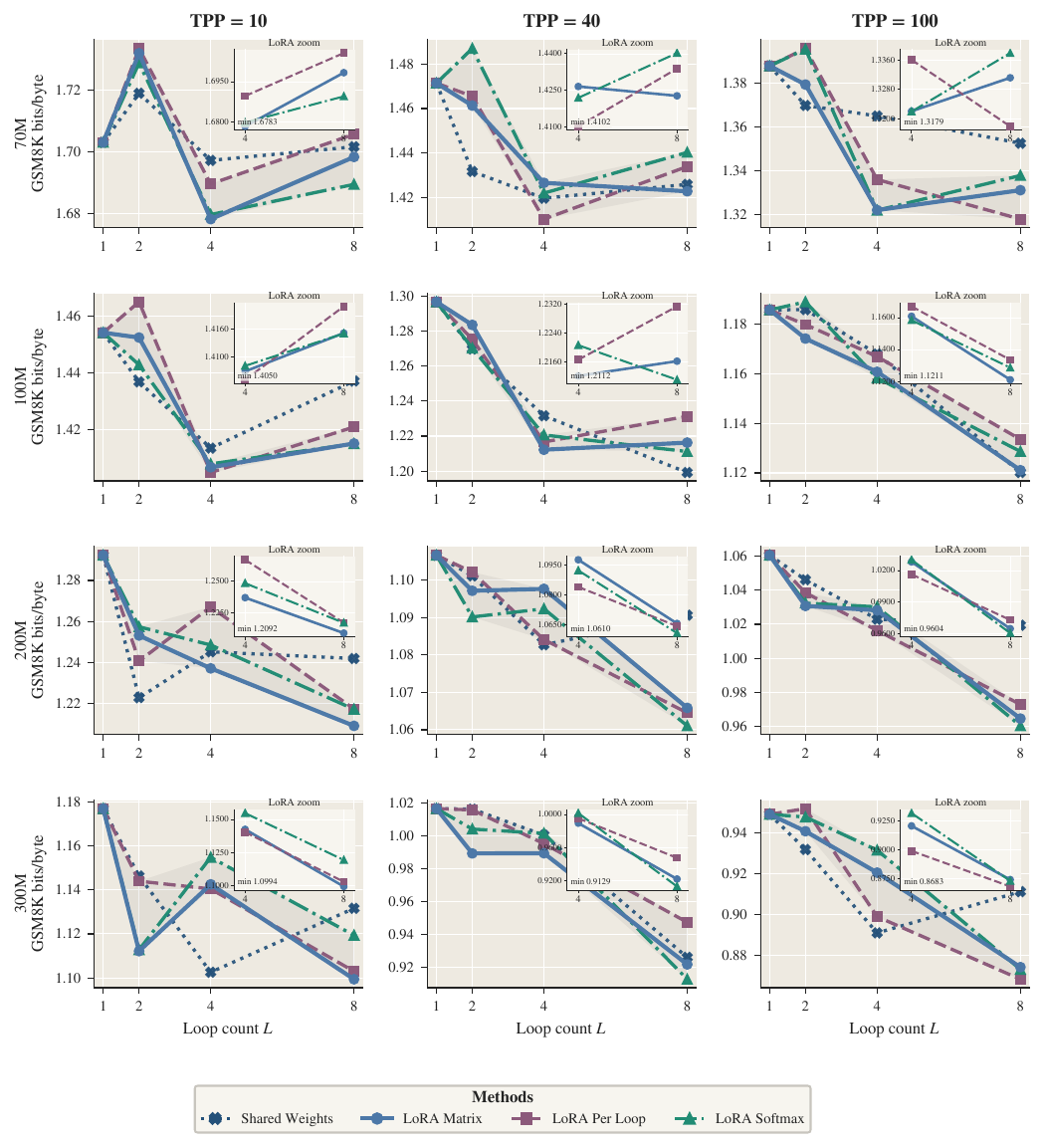}
  \caption{\textbf{Model scale and TPP reliably improve GSM8K likelihood, whereas
  the benefit of specialization depends on the configuration.}
  Teacher-forced GSM8K BPB versus loop count for 70M--300M full-stack models.
  Rows vary model size and columns show TPP 10, 40, and 100. Specialization
  often helps at $L=4$ or $L=8$, but the gain is not uniform. Lower BPB is
  better; these values do not measure final-answer accuracy.}
  \label{fig:loop-lora-gsm8k-loops}
\end{figure*}

\paragraph{Loop count.}
Increasing model size and TPP reliably lowers absolute GSM8K BPB. The
effect of loop-specific specialization is less regular. It often helps at
$L=4$ or $L=8$, whereas at $L=2$ the specialized and identical-weight models
frequently overlap, and specialization can sometimes be worse. No
specialization method dominates throughout the grid. GSM8K therefore supports
a benefit from specialization in many configurations, but not a uniform
improvement at every model size, TPP value, or loop count.
The complete comparison appears in
\autoref{fig:loop-lora-gsm8k-loops}.

\begin{figure*}[ht]
  \centering
  \includegraphics[width=0.94\textwidth]{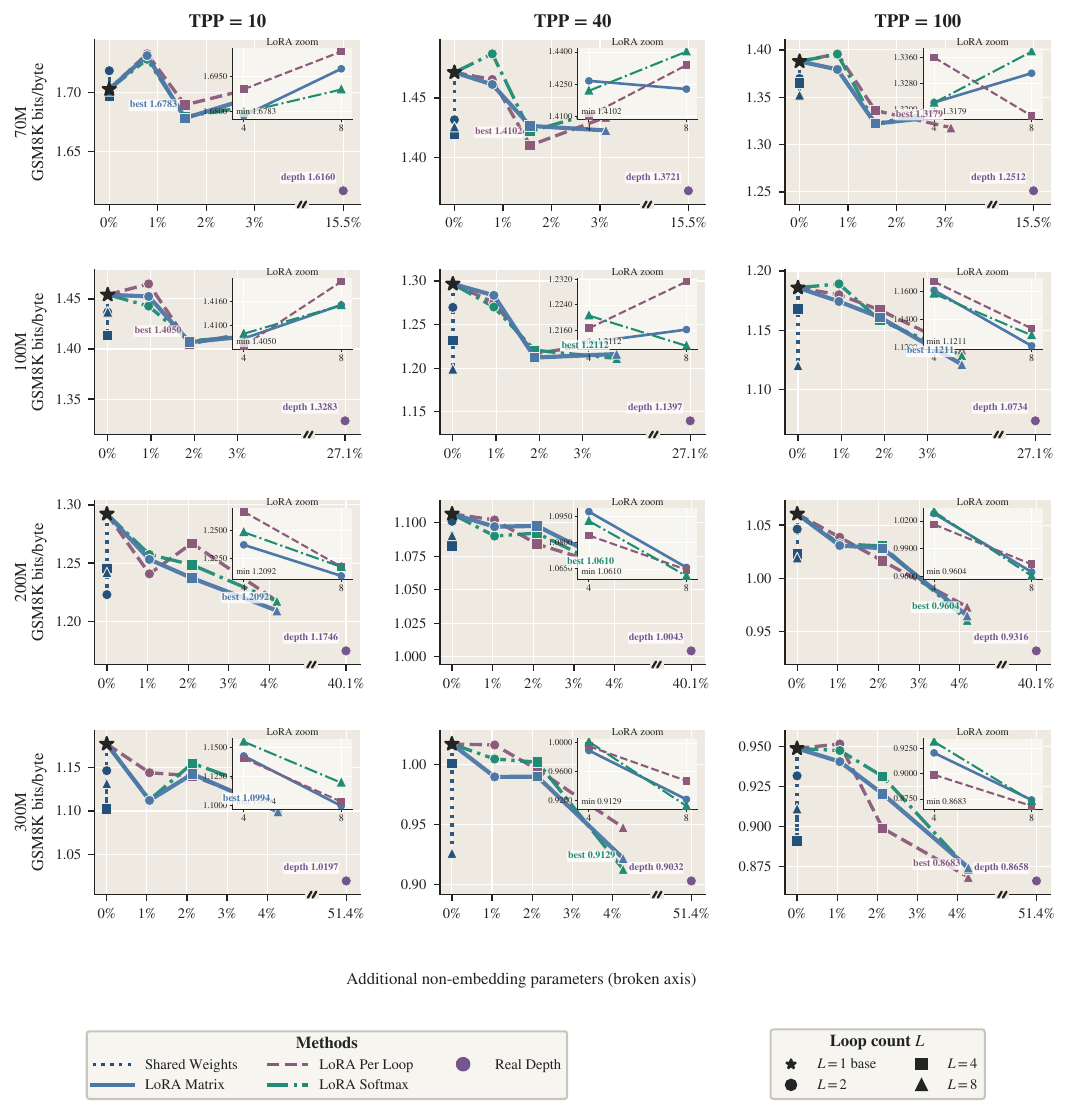}
  \caption{\textbf{The doubled-depth Transformer gives the lowest GSM8K BPB, while
  specialization uses far fewer additional parameters.} Rows show 70M, 100M,
  200M, and 300M models; columns show TPP 10, 40, and 100. The horizontal axis
  reports additional non-embedding parameters relative to the
  identical-weight model. Insets enlarge the specialization range, and the
  separated reference point denotes the doubled-depth Transformer. Lower BPB
  is better.}
  \label{fig:loop-lora-gsm8k-params-depth}
\end{figure*}

\paragraph{Stored parameters and additional Transformer layers.}
Loop-specific corrections occupy only the low-overhead portion of the
parameter axis. Within this range, they sometimes improve BPB substantially,
but the gain remains configuration-dependent. The doubled-depth Transformer,
which has twice as many blocks with independently learned weights, achieves
lower BPB than the looped models in every displayed model-size and TPP panel.
It also requires substantially more stored parameters. Specialization
therefore provides a storage-efficient middle ground, whereas additional
independently parameterized layers are stronger when their storage cost is
acceptable.
The complete comparison appears in
\autoref{fig:loop-lora-gsm8k-params-depth}.

\begin{figure*}[ht]
  \centering
  \includegraphics[width=0.94\textwidth]{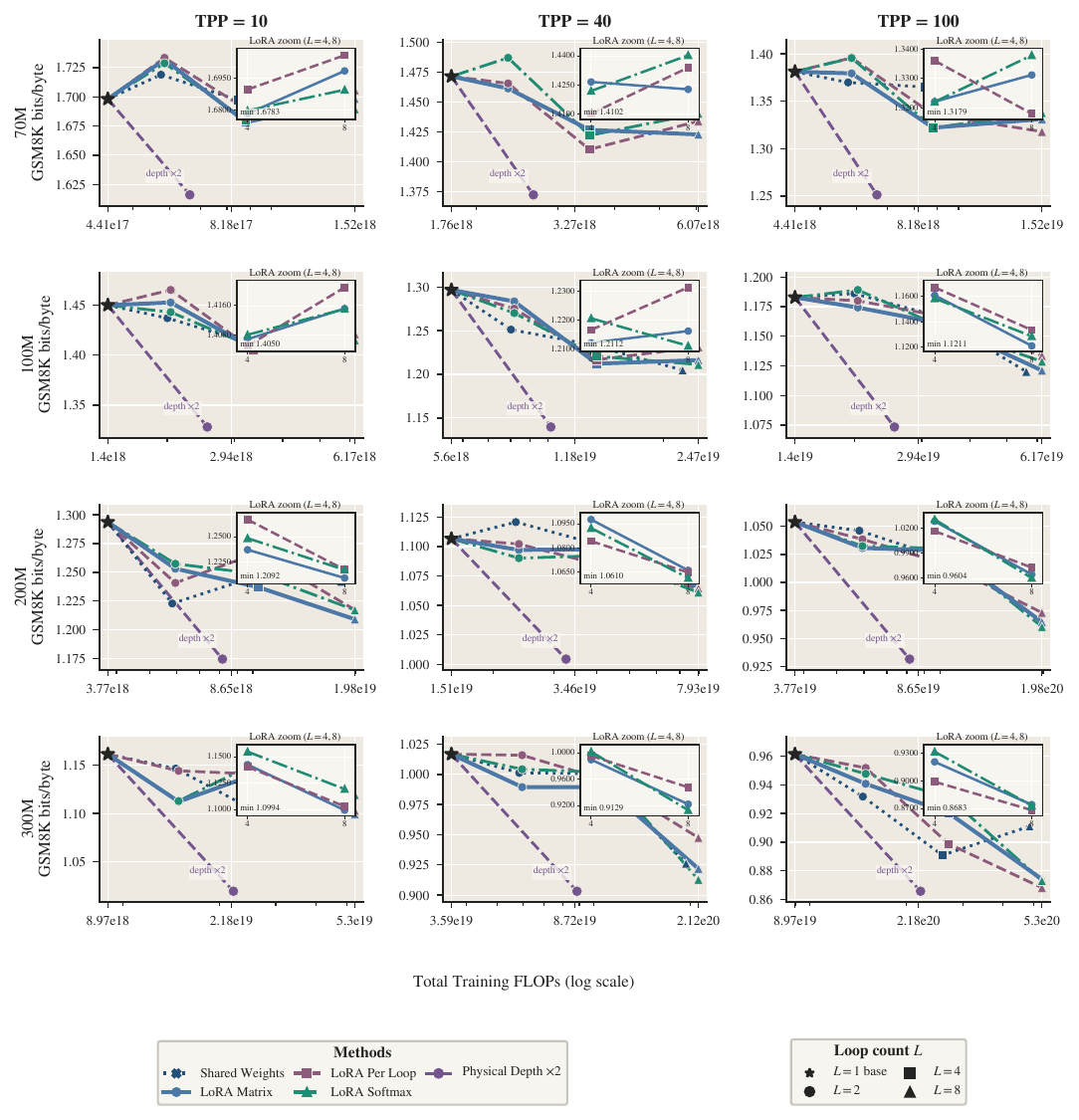}
  \caption{\textbf{At comparable training FLOPs, doubling depth is stronger
  than adding and specializing more loops on GSM8K.} Teacher-forced BPB
  versus total training FLOPs for 70M--300M
  models at TPP 10, 40, and 100. Training FLOPs include repeated loop
  execution, loop-specific corrections, attention-score computation, and the
  realized token budget. The doubled-depth Transformer is included as a
  reference. Lower BPB is better.}
  \label{fig:loop-lora-gsm8k-flops}
\end{figure*}

\paragraph{Training FLOPs.}
The FLOP view reinforces the variability of the GSM8K result. Higher TPP
and larger models lower absolute BPB, but spending more compute on additional
loops does not produce a monotonic improvement in every panel. At a fixed
loop count, specialization can outperform identical weights, although the
best method and the size of the gain vary across model size and TPP. The
doubled-depth Transformer achieves lower BPB than the looped alternatives at
comparable training FLOPs. Thus, when training FLOPs are constrained and the
additional parameters can be stored, doubling the number of independently
parameterized layers is a better use of compute than adding and specializing
more loops.
The complete comparison appears in
\autoref{fig:loop-lora-gsm8k-flops}.

\paragraph{GSM8K summary.}
Model scale and TPP improve GSM8K likelihood much more consistently than
loop-specific specialization. Specialization helps in many settings,
especially among the higher-loop configurations, but the preferred method
and the size of the gain vary. Its clearest advantage is low parameter
overhead: the doubled-depth Transformer achieves lower BPB throughout the
parameter comparison, but requires substantially more storage. Consequently,
specialization is useful when stored parameters are constrained, whereas
doubling depth is preferable when training FLOPs are constrained and the
additional parameters can be stored.

\clearpage

\subsubsection{MATH}
\label{app:loop-lora-math}

MATH provides a second evaluation of mathematical solution text and tests
whether the GSM8K variability persists on a different reasoning corpus. We
again evaluate 70M--300M full-stack models at TPP 10, 40, and 100. These
teacher-forced BPB measurements evaluate the reference-solution likelihood
rather than final-answer accuracy.

\begin{figure*}[ht]
  \centering
  \includegraphics[width=0.94\textwidth]{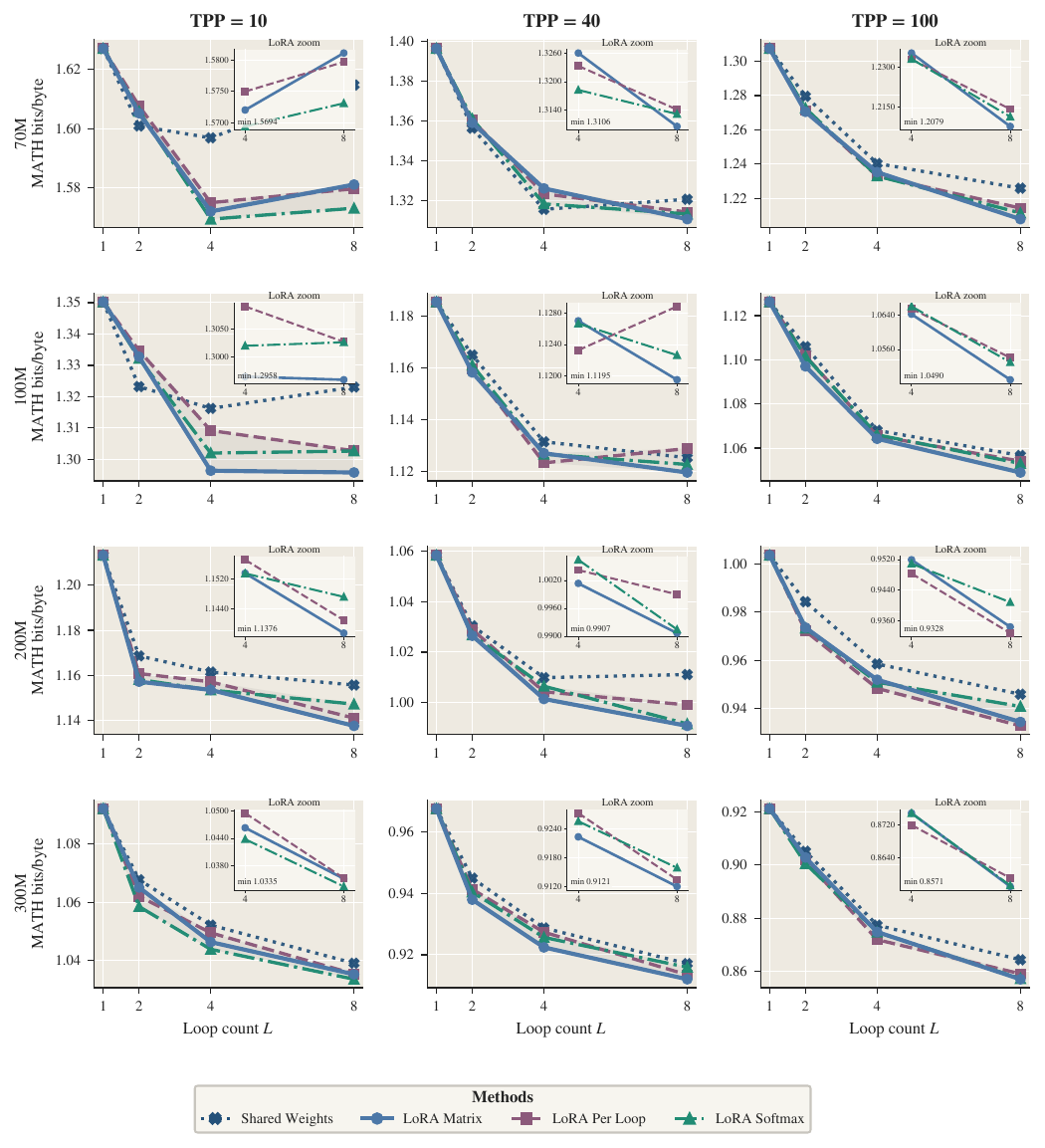}
  \caption{\textbf{MATH likelihood improves with model scale, TPP, and, more
  modestly, loop-specific specialization.} Teacher-forced MATH BPB versus
  loop count for 70M--300M full-stack models. Rows vary model size and columns
  show TPP 10, 40, and 100. Specialization is most consistently useful at
  $L=4$ and $L=8$ beyond the smallest scale. Lower BPB is better; these values
  do not measure final-answer accuracy.}
  \label{fig:loop-lora-math-loops}
\end{figure*}

\paragraph{Loop count.}
Increasing model size and TPP consistently lowers MATH BPB. Additional
loops also generally improve likelihood, and loop-specific specialization
usually provides a further reduction relative to identical weights. The
benefit is small or mixed in some 70M settings, but becomes more consistent at
100M--300M and is most visible at $L=4$ and $L=8$. Loop Matrix is frequently
strongest, although differences among the three methods are usually smaller
than their common improvement over identical weights.
The complete comparison appears in
\autoref{fig:loop-lora-math-loops}.

\begin{figure*}[ht]
  \centering
  \includegraphics[width=0.94\textwidth]{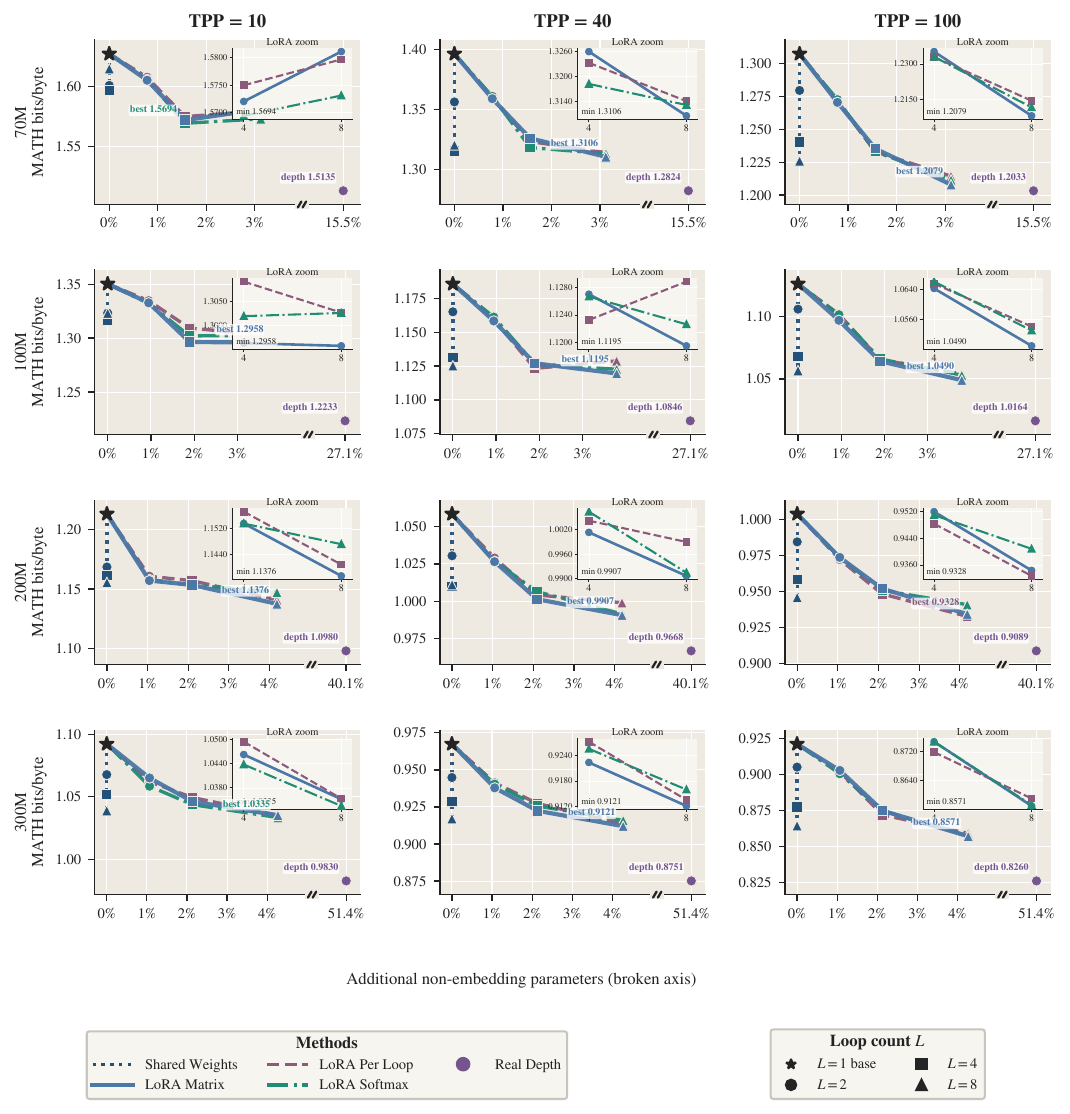}
  \caption{\textbf{Specialization improves MATH BPB with little added storage,
  while the doubled-depth Transformer remains strongest.} Rows show 70M,
  100M, 200M, and 300M models; columns show TPP 10, 40, and 100. The
  horizontal axis reports additional non-embedding parameters relative to the
  identical-weight model. Insets enlarge the specialization range, and the
  separated reference point denotes the doubled-depth Transformer. Lower BPB
  is better.}
  \label{fig:loop-lora-math-params-depth}
\end{figure*}

\paragraph{Stored parameters and additional Transformer layers.}
Across model sizes and TPP values, loop-specific corrections usually
improve MATH BPB while adding only a small number of parameters. The
doubled-depth Transformer has twice as many blocks, each with independently
learned weights. It achieves the lowest BPB in every displayed panel, but
lies far beyond the specialization methods on the parameter axis.
Specialization is therefore attractive when stored parameters are limited;
additional independently parameterized layers remain stronger when their
storage cost is acceptable.
The complete comparison appears in
\autoref{fig:loop-lora-math-params-depth}.

\begin{figure*}[ht]
  \centering
  \includegraphics[width=0.94\textwidth]{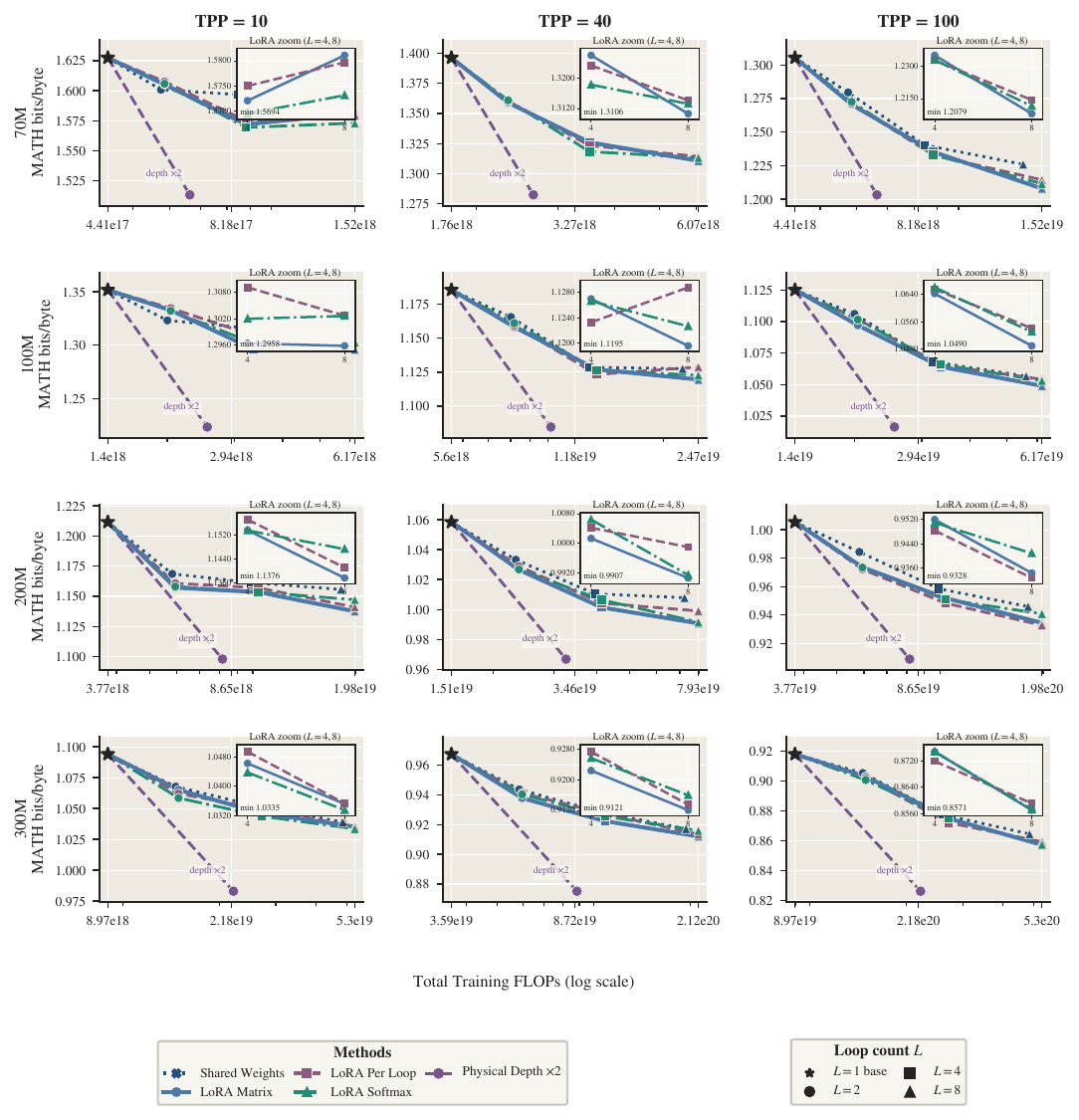}
  \caption{\textbf{Doubling depth provides lower MATH BPB than adding and
  specializing loops at comparable training FLOPs.} Teacher-forced BPB versus total training
  FLOPs for 70M--300M models at TPP 10, 40, and 100. Training FLOPs include
  repeated loop execution, loop-specific corrections, attention-score
  computation, and the realized token budget. The doubled-depth Transformer
  is included as a reference. Lower BPB is better.}
  \label{fig:loop-lora-math-flops}
\end{figure*}

\paragraph{Training FLOPs.}
Unlike GSM8K, MATH BPB generally decreases as training FLOPs increase
through additional loops. At the same loop count, specialization then
provides a smaller but comparatively consistent improvement over identical
weights, particularly at $L=4$ and $L=8$. The dominant absolute improvements
still come from model scale, TPP, and the additional loop computation; the
loop-specific corrections refine those gains. Even so, the doubled-depth
Transformer achieves lower BPB at comparable training FLOPs. If training
FLOPs are the primary constraint and more parameters can be stored, doubling
depth is therefore preferable to adding and specializing more loops.
The complete comparison appears in
\autoref{fig:loop-lora-math-flops}.

\paragraph{MATH summary.}
MATH provides more consistent evidence for loop-specific specialization
than GSM8K. The benefit is clearest at $L=4$ and $L=8$ for 100M--300M models
and persists across TPP values, while Loop Matrix is frequently strongest.
Higher TPP and larger model size remain the most reliable sources of absolute
improvement. Specialization offers a favorable storage trade-off, whereas the
doubled-depth Transformer achieves lower BPB with substantially more
parameters. Thus, specialization is preferable under a storage constraint,
while doubling depth is preferable when training FLOPs are constrained and
the additional parameters can be stored.

\clearpage

\subsubsection{HumanEval}
\label{app:loop-lora-humaneval}

HumanEval tests whether the specialization behavior extends to code. We
evaluate 70M--300M full-stack models at TPP 10, 40, and 100 by
teacher-forcing the reference implementation. BPB therefore measures
reference-code likelihood, not whether sampled programs pass the unit tests,
and should not be interpreted as execution accuracy or pass@$k$.

\begin{figure*}[ht]
  \centering
  \includegraphics[width=0.94\textwidth]{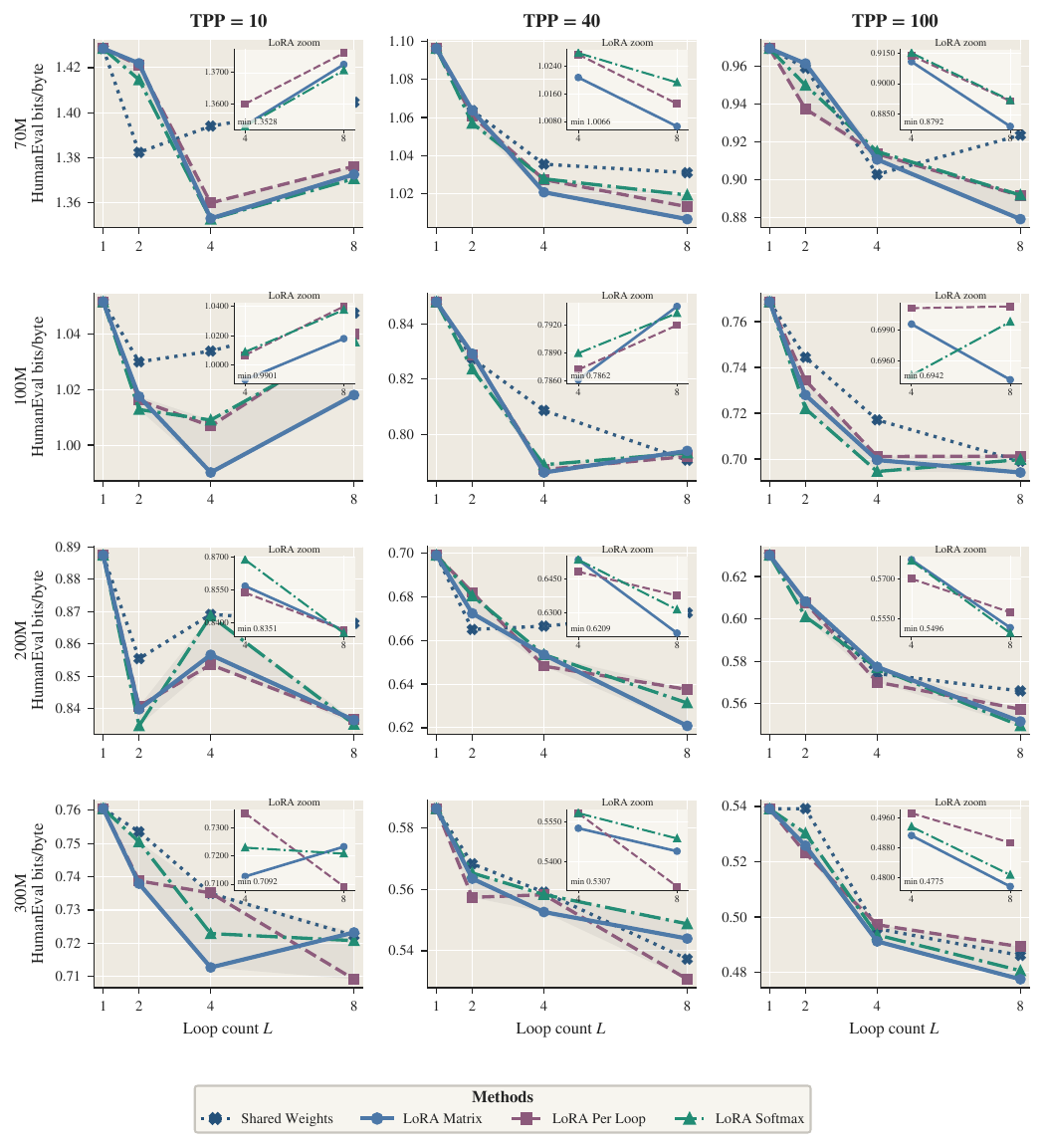}
  \caption{\textbf{Specialization usually improves HumanEval likelihood, but the
  best loop count and method depend on model size and TPP.} Absolute
  teacher-forced BPB versus loop count for 70M--300M full-stack models. Rows
  vary model size and columns show TPP 10, 40, and 100. Lower BPB is better;
  these values do not measure execution accuracy or pass@$k$.}
  \label{fig:loop-lora-humaneval-loops}
\end{figure*}

\paragraph{Loop count.}
Across model sizes and TPP values, at least one specialization method
usually improves on the corresponding identical-weight model. The separation
is generally clearest at $L=4$ or $L=8$, but neither the best loop count nor
the best method is uniform across the grid. In particular, the low-TPP results
are sometimes non-monotonic in loop count. Increasing TPP substantially
lowers absolute BPB without removing this configuration dependence.
The complete comparison appears in
\autoref{fig:loop-lora-humaneval-loops}.

\begin{figure*}[ht]
  \centering
  \includegraphics[width=0.94\textwidth]{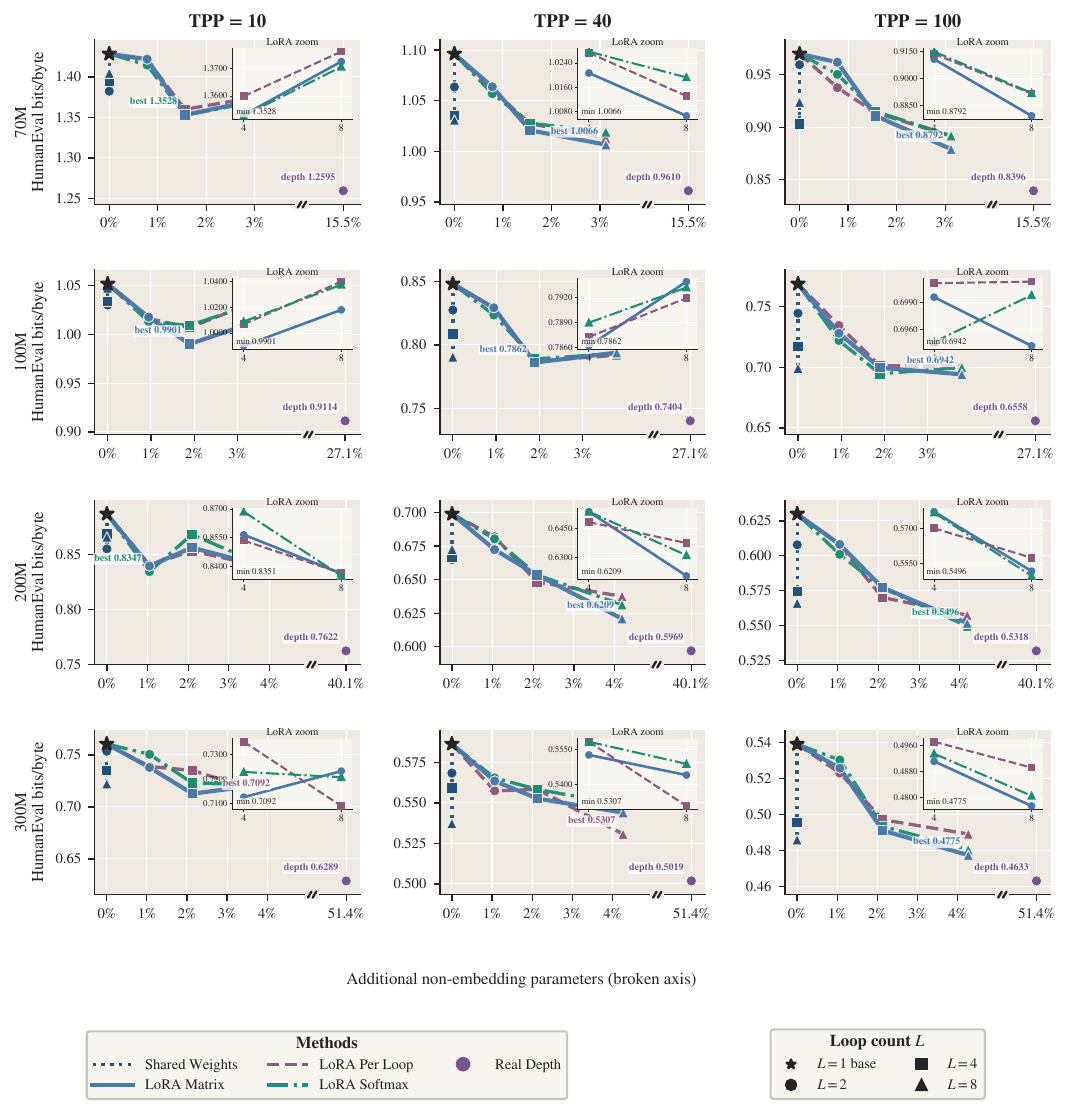}
  \caption{\textbf{The doubled-depth Transformer achieves lower HumanEval BPB,
  while specialization improves identical weights with far less storage.}
  Results cover 70M--300M full-stack models at TPP 10, 40, and 100. The
  horizontal axis reports additional non-embedding parameters relative to the
  identical-weight model. Insets enlarge the specialization range, and the
  separated reference point denotes the doubled-depth Transformer. Lower BPB
  is better.}
  \label{fig:loop-lora-humaneval-params-depth}
\end{figure*}

\paragraph{Stored parameters and additional Transformer layers.}
The loop-specific corrections add relatively few stored parameters. The
doubled-depth Transformer instead has twice as many blocks, with independently
learned weights in every block. It attains lower HumanEval BPB in every
displayed model-size and TPP panel, but requires a much larger increase in
storage. Specialization is therefore a storage-efficient way to improve the
identical-weight model, rather than a replacement for additional depth when
those parameters can be stored.
The complete comparison appears in
\autoref{fig:loop-lora-humaneval-params-depth}.

\begin{figure*}[ht]
  \centering
  \includegraphics[width=0.94\textwidth]{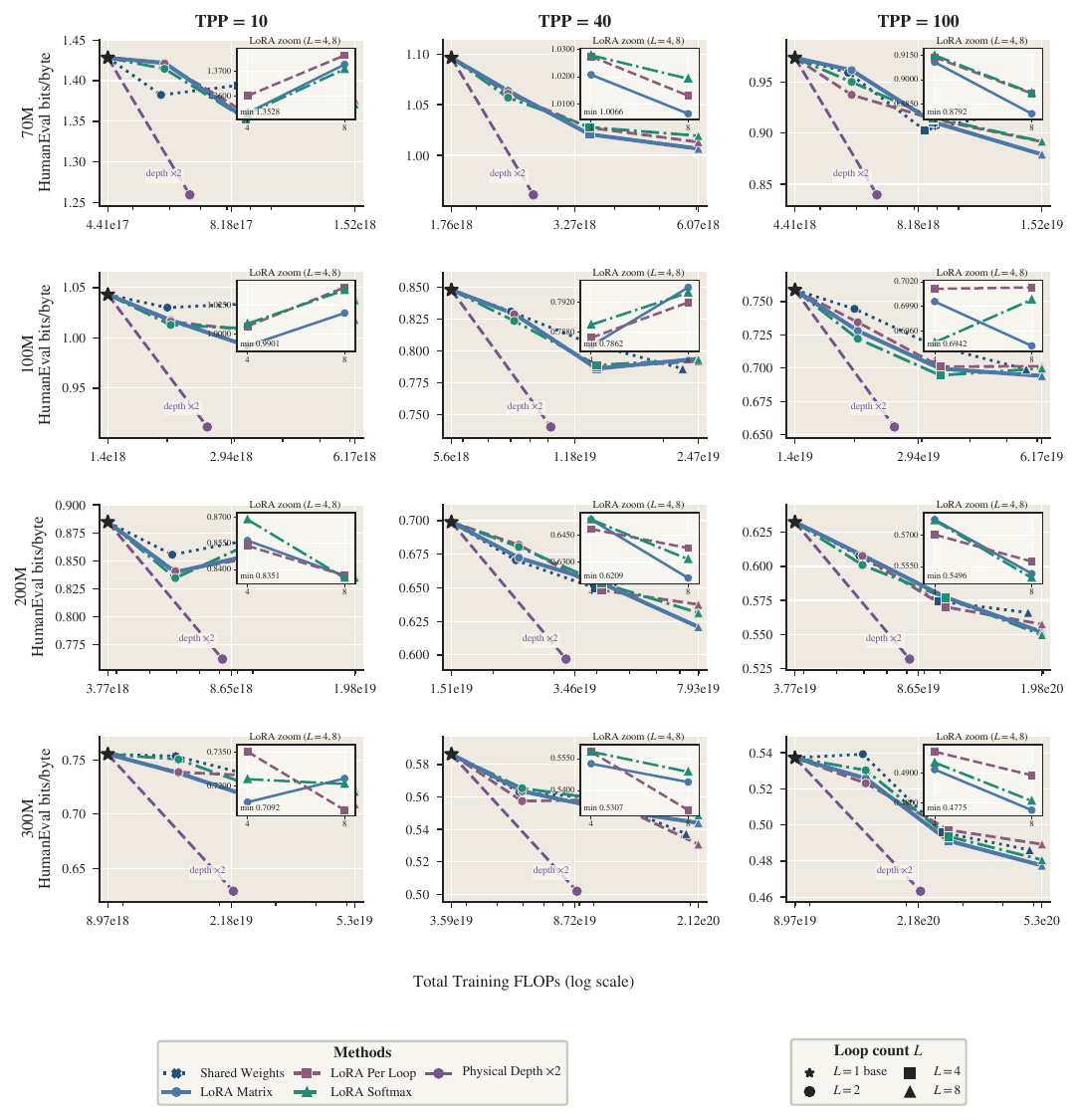}
  \caption{\textbf{At comparable training FLOPs, doubling depth gives lower
  HumanEval BPB than adding and specializing more loops.} HumanEval BPB versus total training FLOPs for
  70M--300M full-stack models at TPP 10, 40, and 100. Training FLOPs include
  repeated loop execution, loop-specific corrections, attention-score
  computation, and the realized token budget. The doubled-depth Transformer
  is included as a reference. Lower BPB is better.}
  \label{fig:loop-lora-humaneval-flops}
\end{figure*}

\paragraph{Training FLOPs.}
Within the looped models, spending more training compute through
additional loops generally lowers HumanEval BPB. Specialization often
improves on identical weights at the same loop count, but the size of the gain
and the strongest method vary across model size and TPP. At comparable
training FLOPs, however, the doubled-depth Transformer achieves lower BPB.
When training FLOPs are constrained and its additional parameters can be
stored, doubling depth is therefore a better use of compute than adding and
specializing more loops.
The complete comparison appears in
\autoref{fig:loop-lora-humaneval-flops}.

\paragraph{HumanEval summary.}
Larger models and higher TPP consistently improve absolute HumanEval
likelihood. Specialization provides an additional benefit in most
configurations, especially at larger loop counts, although the preferred
method varies. The doubled-depth Transformer reaches lower BPB throughout the
parameter comparison, at a substantially larger storage cost. These
conclusions concern teacher-forced code likelihood only. Specialization is
the better choice when stored parameters are constrained; doubling depth is
the better choice when training FLOPs are constrained and more parameters can
be stored.

\clearpage

\subsubsection{MBPP}
\label{app:loop-lora-mbpp}

MBPP provides a second code-likelihood evaluation and tests whether the
HumanEval conclusions persist on another code corpus. We evaluate the same
70M--300M full-stack models at TPP 10, 40, and 100. MBPP BPB is computed by
teacher-forcing the reference code and should not be interpreted as execution
accuracy or pass@$k$.

\begin{figure*}[ht]
  \centering
  \includegraphics[width=0.94\textwidth]{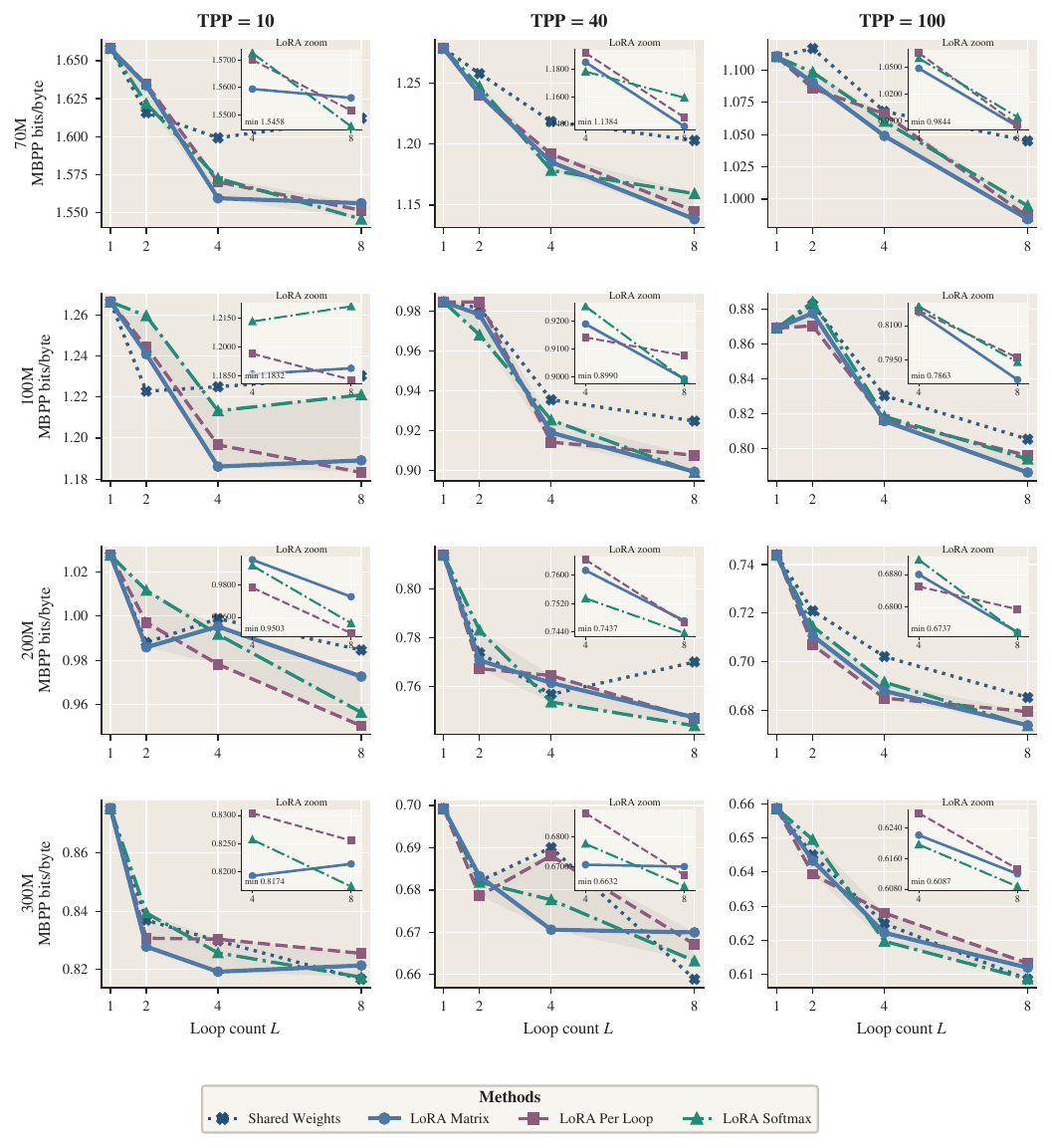}
  \caption{\textbf{Specialization improves MBPP likelihood in most settings, while
  the best loop count and method remain configuration-dependent.} Absolute
  teacher-forced BPB versus loop count for 70M--300M full-stack models. Rows
  vary model size and columns show TPP 10, 40, and 100. Lower BPB is better;
  these values do not measure execution accuracy or pass@$k$.}
  \label{fig:loop-lora-mbpp-loops}
\end{figure*}

\paragraph{Loop count.}
Specialization improves on identical weights in most of the MBPP grid,
with the largest-loop configurations frequently attaining the lowest BPB. The
trend is not universal: the best point can occur at $L=4$, and identical
weights remain competitive in a few panels. The ranking among Loop Matrix,
Loop LoRA, and Loop Softmax also changes with model size and TPP. By contrast,
increasing TPP clearly reduces absolute BPB at every model size.
The complete comparison appears in
\autoref{fig:loop-lora-mbpp-loops}.

\begin{figure*}[ht]
  \centering
  \includegraphics[width=0.94\textwidth]{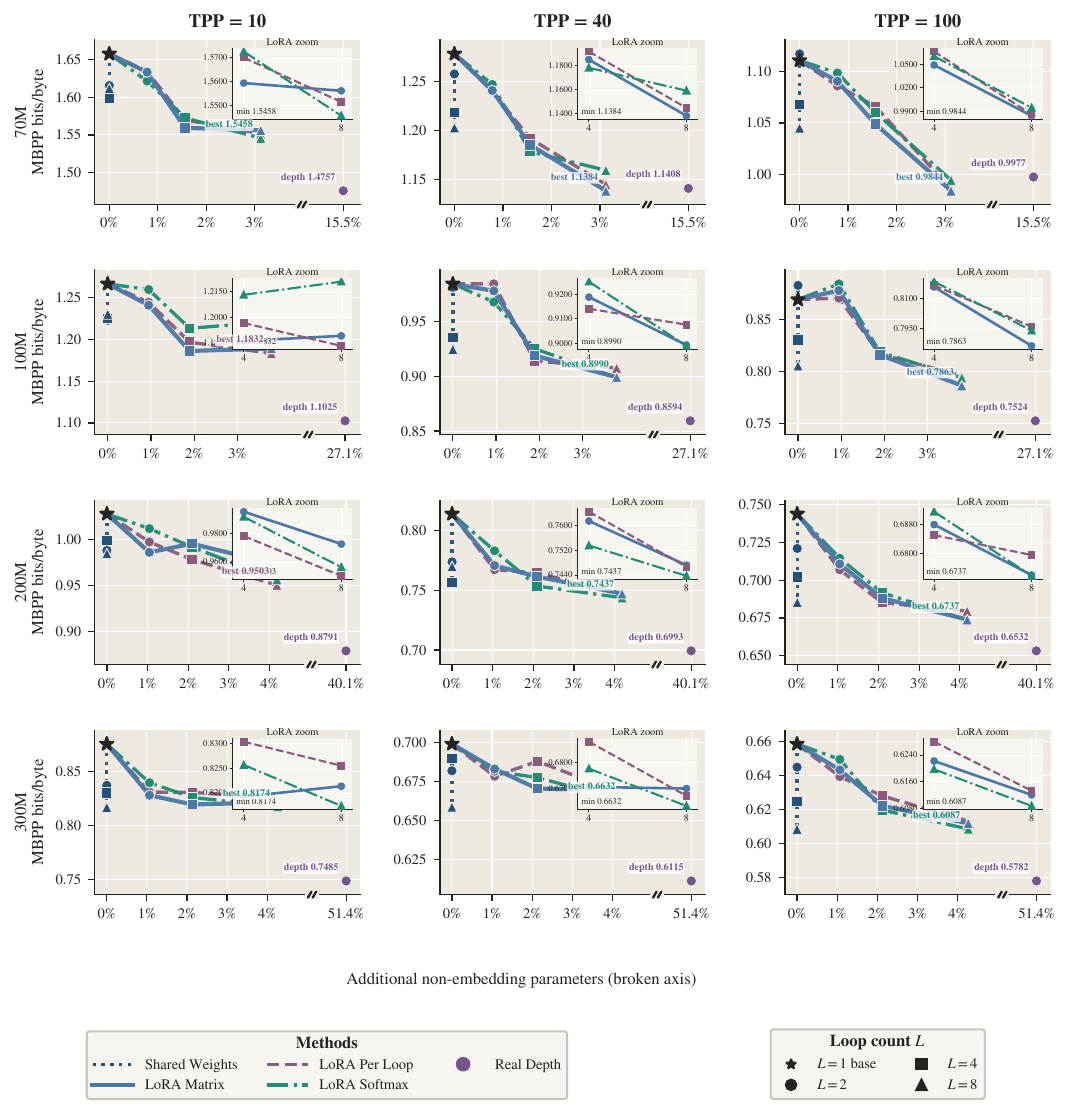}
  \caption{\textbf{The doubled-depth Transformer usually reaches lower MBPP BPB,
  while specialization is competitive at 70M and requires far less storage.}
  Results cover 70M--300M full-stack models at TPP 10, 40, and 100. The
  horizontal axis reports additional non-embedding parameters relative to the
  identical-weight model. Insets enlarge the specialization range, and the
  separated reference point denotes the doubled-depth Transformer. Lower BPB
  is better.}
  \label{fig:loop-lora-mbpp-params-depth}
\end{figure*}

\paragraph{Stored parameters and additional Transformer layers.}
The loop-specific corrections add relatively few stored parameters. From
100M through 300M, the doubled-depth Transformer, which has twice as many
independently parameterized blocks, achieves lower MBPP BPB in every displayed
TPP column. The 70M comparison is more mixed: specialization is slightly
stronger at TPP 40 and 100, while doubled depth is stronger at TPP 10. Thus,
the quality advantage of additional depth is common but not universal on
MBPP, whereas the parameter savings from specialization hold throughout the
grid.
The complete comparison appears in
\autoref{fig:loop-lora-mbpp-params-depth}.

\begin{figure*}[ht]
  \centering
  \includegraphics[width=0.94\textwidth]{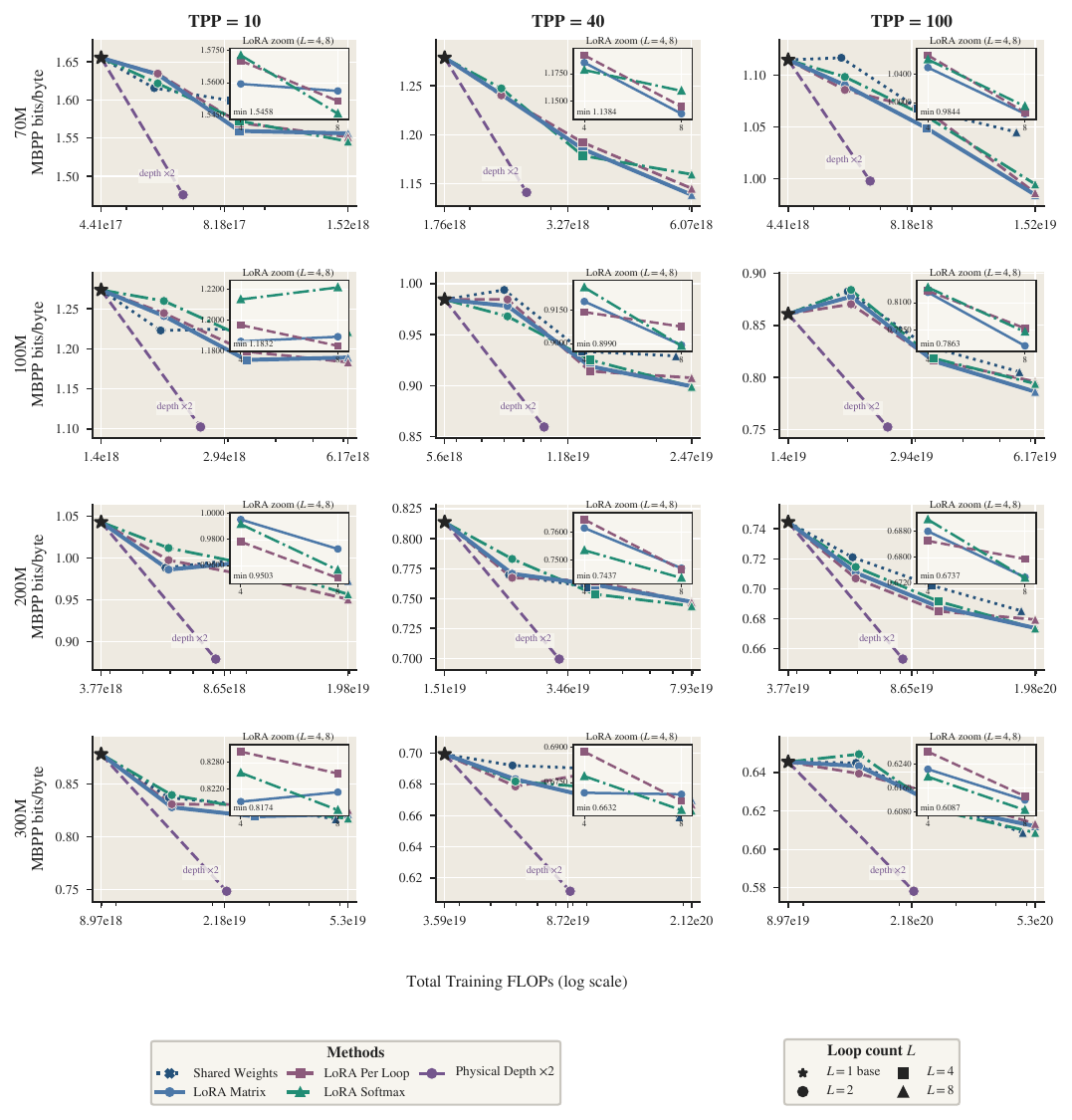}
  \caption{\textbf{Doubling depth gives lower MBPP BPB than adding and
  specializing loops at comparable training FLOPs.} MBPP BPB versus total training FLOPs for 70M--300M full-stack
  models at TPP 10, 40, and 100. Training FLOPs include repeated loop
  execution, loop-specific corrections, attention-score computation, and the
  realized token budget. The doubled-depth Transformer is included as a
  reference. Lower BPB is better.}
  \label{fig:loop-lora-mbpp-flops}
\end{figure*}

\paragraph{Training FLOPs.}
Increasing the number of loops generally lowers MBPP BPB despite the
additional training FLOPs. Loop-specific corrections frequently improve on
identical weights at the same loop count, although the improvement is not
present in every panel and no method dominates the complete grid. At
comparable training FLOPs, the doubled-depth Transformer achieves lower BPB
than the looped alternatives. Thus, when training FLOPs are constrained and
the additional parameters can be stored, doubling depth is the better choice.
The complete comparison appears in
\autoref{fig:loop-lora-mbpp-flops}.

\paragraph{MBPP summary.}
MBPP supports the same qualified conclusion as HumanEval:
loop-specific specialization usually improves identical weights, particularly
at larger loop counts, but the preferred loop count and method depend on model
size and TPP. Additional independently parameterized depth is usually
stronger in absolute likelihood, while specialization requires substantially
less storage and can be competitive at 70M. These results describe
teacher-forced likelihood, not generated-code correctness. Specialization is
preferable when stored parameters are constrained; doubling depth is
preferable when training FLOPs are constrained and more parameters can be
stored.

\clearpage

\subsection{Where Is Loop-Specific Specialization Useful?}
\label{app:loop-lora-across-designs}

\begin{figure*}[ht]
  \centering
  \includegraphics[width=0.6\textwidth]{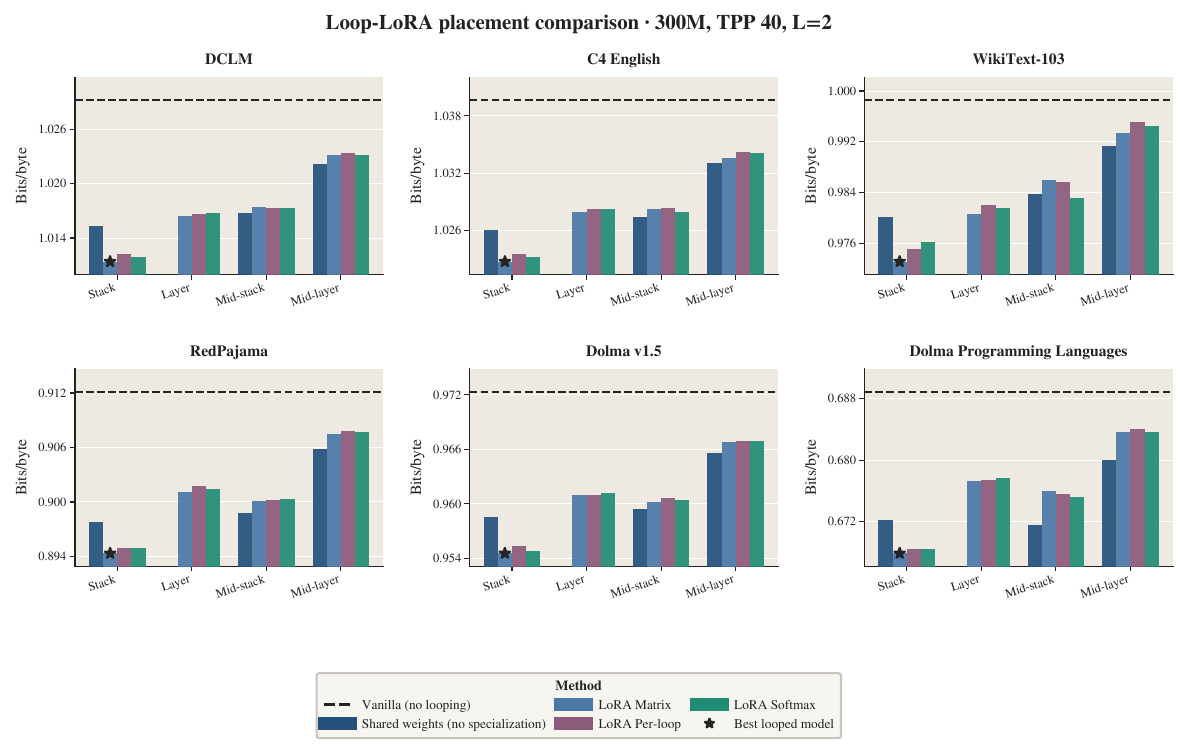}
  \caption{\textbf{At $L=2$, Full-Stack Looping is strongest and receives
  the clearest benefit from specialization.} Results use 300M models at TPP
  40. Each panel shows one evaluation corpus. Within
  each looping schedule, bars compare Loop Matrix, Loop LoRA, and Loop
  Softmax; stars mark the lowest BPB. Per-layer specialization is less
  consistently beneficial. Lower BPB is better.}
  \label{fig:loop-lora-public-designs-k2}
\end{figure*}

\begin{figure*}[ht]
  \centering
  \includegraphics[width=0.6\textwidth]{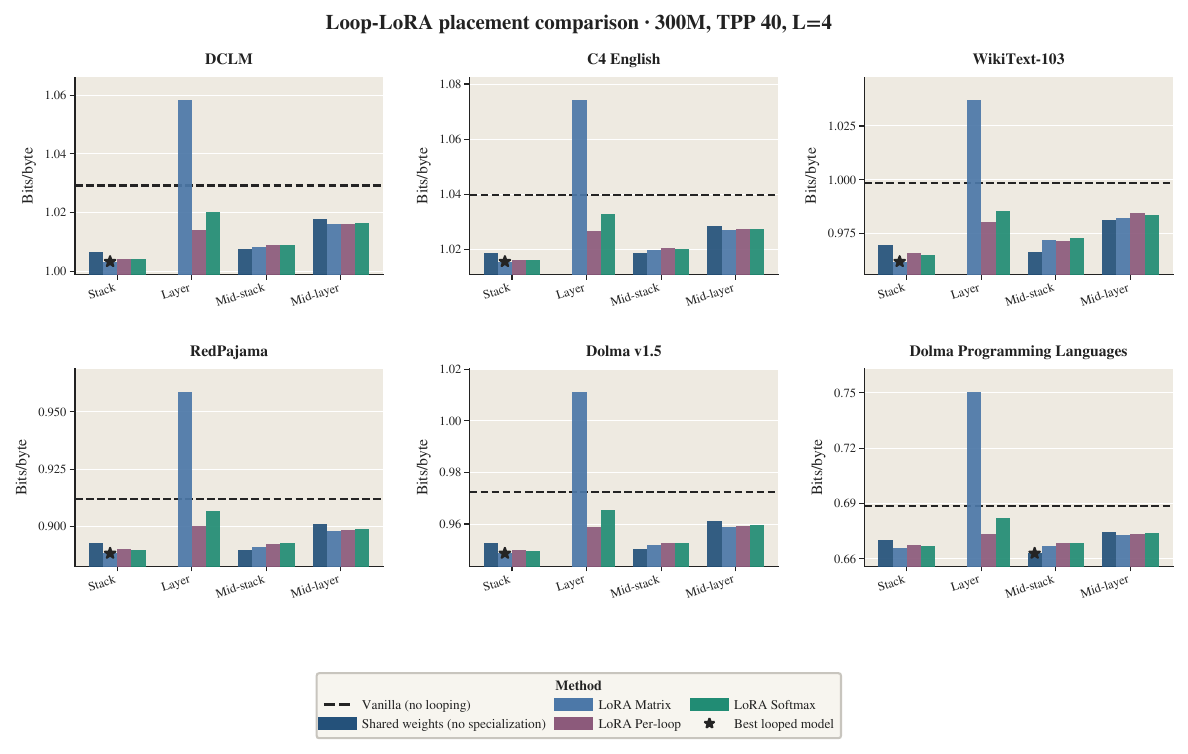}
  \caption{\textbf{At $L=4$, specialization further strengthens full-stack
  looping but remains less useful for Per-Layer Looping.} Results use 300M
  models at TPP 40. Each panel shows one evaluation corpus. Within
  each looping schedule, bars compare Loop Matrix, Loop LoRA, and Loop
  Softmax; stars mark the lowest BPB. Lower BPB is better.}
  \label{fig:loop-lora-public-designs-k4}
\end{figure*}

\begin{figure*}[ht]
  \centering
  \includegraphics[width=0.6\textwidth]{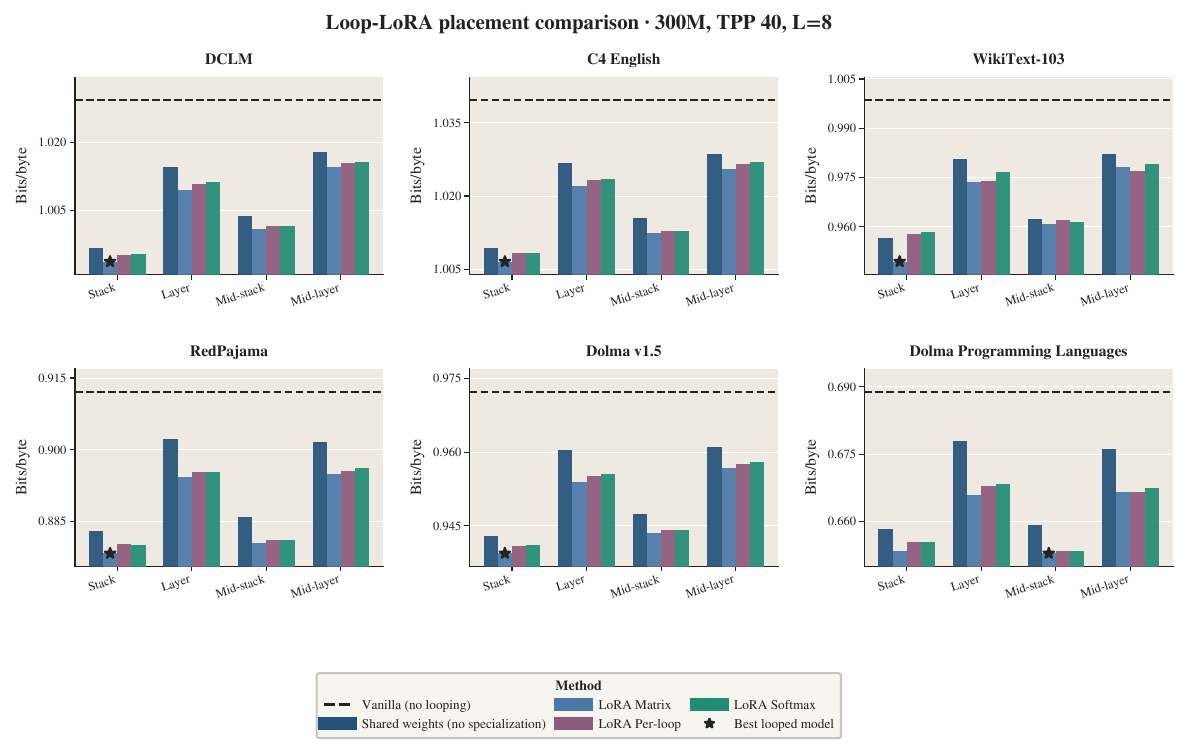}
  \caption{\textbf{At $L=8$, Full-Stack Looping obtains the largest gains
  from specialization, while per-layer gains remain inconsistent.} Results
  use 300M models at TPP 40. Each panel shows one evaluation corpus. Within
  each looping schedule, bars compare Loop Matrix, Loop LoRA, and Loop
  Softmax; stars mark the lowest BPB. Lower BPB is better.}
  \label{fig:loop-lora-public-designs-k8}
\end{figure*}

The preceding sections use Full-Stack Looping to compare loop count, parameter
storage, additional depth, and training FLOPs within a common architecture.
They show that specialization generally improves quality, with its benefits
becoming clearest at higher TPP and larger loop counts. We now ask whether the
same benefit appears when computation is looped elsewhere in the Transformer. We compare full-stack, per-layer, middle-stack, and Middle-Layer Looping in a
representative setting: 300M models at TPP 40, with loop counts
$L\in\{2,4,8\}$, evaluated on all six language-modeling corpora.
\autoref{fig:loop-lora-public-designs-k2},
\autoref{fig:loop-lora-public-designs-k4}, and
\autoref{fig:loop-lora-public-designs-k8} show a consistent hierarchy.
Full-Stack Looping achieves the strongest absolute quality and receives the
largest gains from specialization. Middle-stack and Middle-Layer Looping can
also benefit, but generally by less. Per-Layer Looping is the clearest
exception: loop-specific corrections do not improve it consistently. The
differences among looping schedules are also larger than the differences
among Loop LoRA, Loop Matrix, and Loop Softmax. Thus, choosing where to loop
has a larger effect than choosing the precise specialization method, although
specialization provides an additional benefit for the stronger schedules.

\paragraph{Takeaway.}
Loop placement is the more important design choice. Full-Stack Looping gives
the strongest quality and benefits most from loop-specific specialization,
particularly at larger loop counts and higher TPP. Middle-stack and
Middle-Layer Looping receive smaller gains, while Per-Layer Looping does not
benefit consistently. Once a strong looping schedule has been selected,
specialization provides an additional improvement; the precise choice among
Loop LoRA, Loop Matrix, and Loop Softmax is comparatively less important.

\clearpage

\subsection{Does Specialization Persist at 500M and 1B?}
\label{app:loop-lora-scale-extension}

We finally ask whether the specialization result from the 70M--300M sweep
persists at larger model scales. At 500M, we evaluate Loop LoRA, Loop Matrix,
and Loop Softmax with $L\in\{2,4,8\}$ at TPP 10, 40, and 100. At 1B, all
three methods are available through $L=8$ at TPP 10 and 40 and through $L=4$
at TPP 100; the $L=8$, TPP-100 runs are unavailable. Missing observations are
left blank, and lines are not connected across them.

Across DCLM, C4 English, WikiText-103, Dolma v1.5, RedPajama, and Dolma
Programming Languages, the earlier conclusion continues to hold.
Loop-specific specialization generally improves identical-weight full-stack
looping, and its benefit remains clearest at larger loop counts. Differences
among Loop LoRA, Loop Matrix, and Loop Softmax are usually smaller than their
shared improvement over identical weights, although the absolute BPB gains
are generally smaller than in the 70M--300M sweep. Thus, the value of allowing
loops to specialize does not disappear as model size increases.

This extension focuses on one question: whether loop-specific specialization
continues to help as the model becomes larger. The parameter and FLOP views
quantify the additional storage and training cost of the corrections.
We believe that increasing the parameter count, for example by adding
independently parameterized Transformer layers, would also improve quality.
Here, however, we isolate whether loop-specific specialization remains useful
as model size increases while keeping the Full-Stack Looping design fixed. We
do not have 500M or 1B results on GSM8K, MATH, HumanEval, or MBPP, so the scale
conclusion is limited to the six language-modeling corpora.

\subsubsection{DCLM}

\begin{figure*}[ht]
  \centering
  \includegraphics[width=0.94\textwidth]{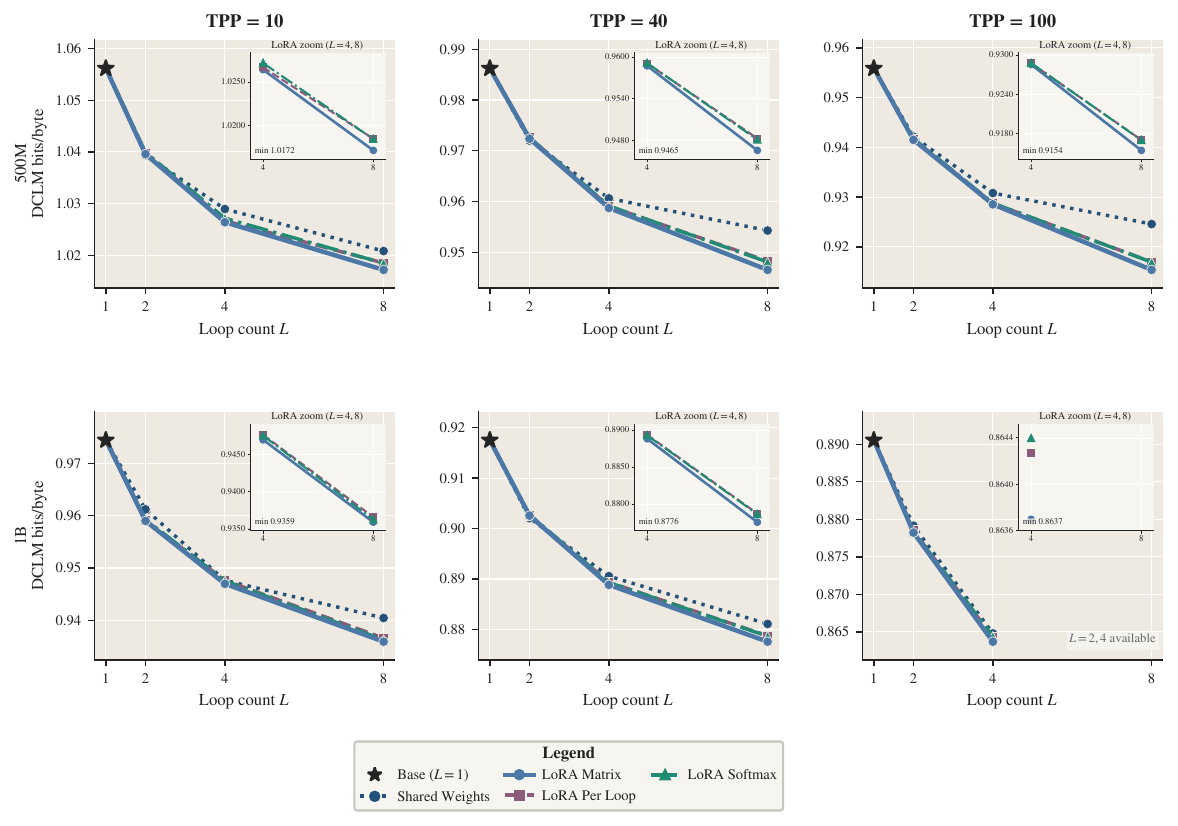}
  \caption{\textbf{Loop-specific specialization remains useful at 500M and
  1B on DCLM.} Full-stack results are shown against loop count for the
  completed TPP settings. The improvement is generally clearest at the larger
  loop counts, although the absolute differences are smaller than in the
  70M--300M sweep. Blank or disconnected cells are unavailable and are not
  imputed. Lower BPB is better.}
  \label{fig:loop-lora-dclm-scale-loops}
\end{figure*}

\begin{figure*}[ht]
  \centering
  \includegraphics[width=0.94\textwidth]{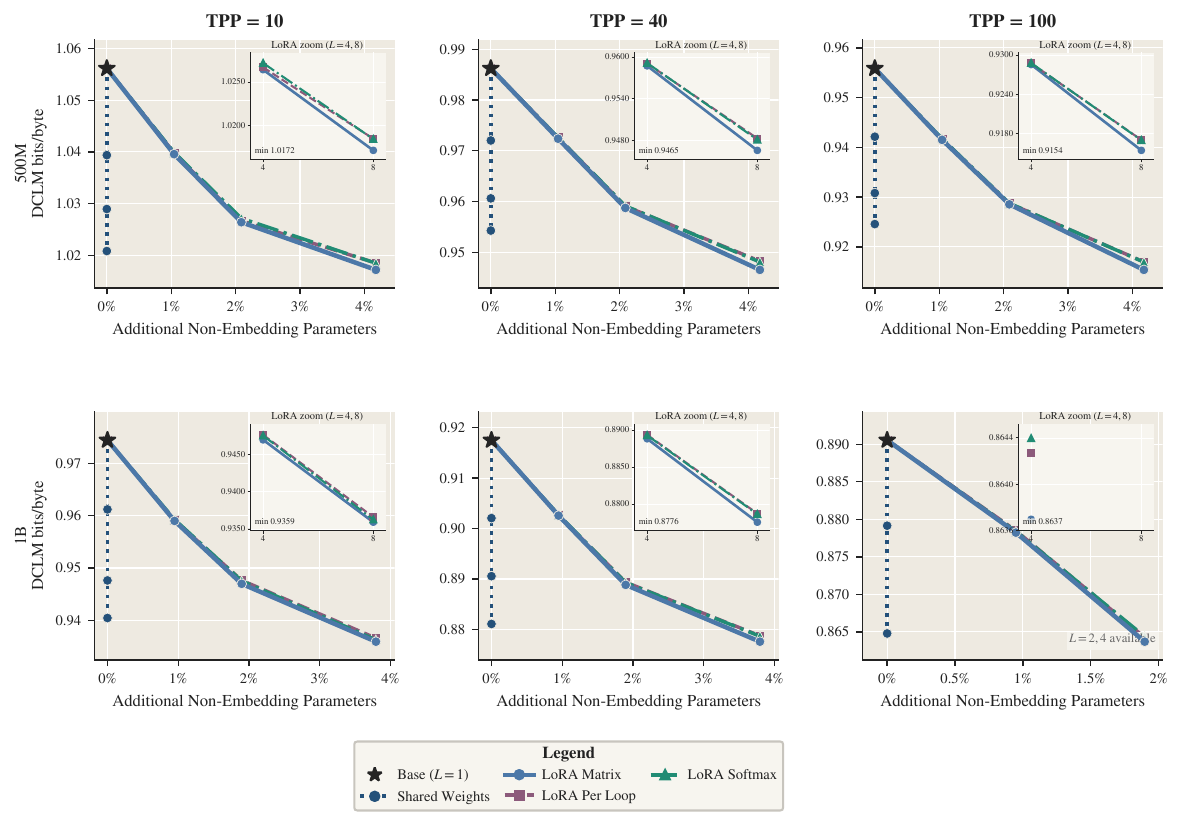}
  \caption{\textbf{The larger-scale specialization gains require relatively
  few additional stored parameters on DCLM.} BPB is plotted against
  additional non-embedding parameters relative to the identical-weight
  full-stack model. Lower BPB is better.}
  \label{fig:loop-lora-dclm-scale-params}
\end{figure*}

\begin{figure*}[ht]
  \centering
  \includegraphics[width=0.94\textwidth]{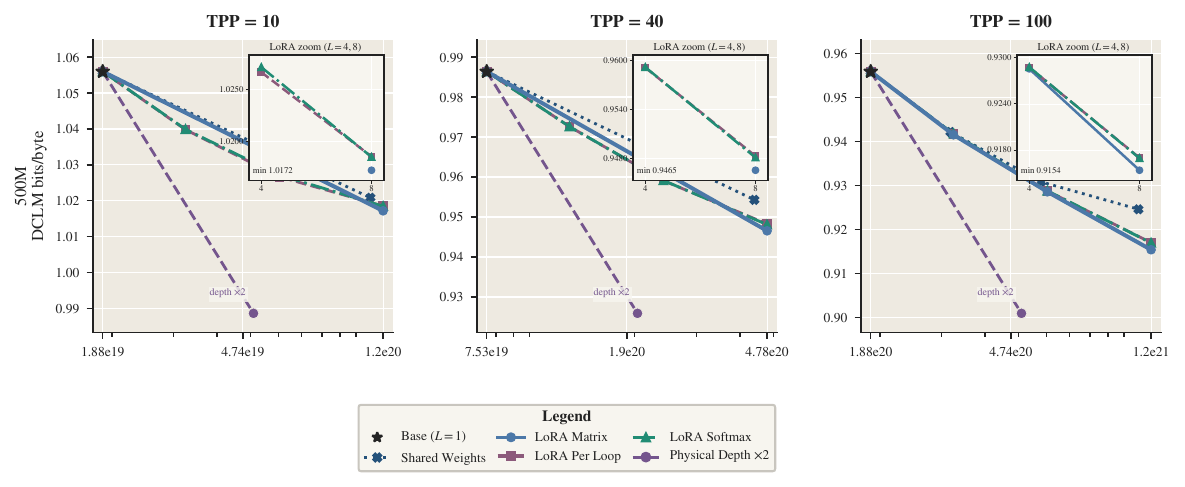}
  \caption{\textbf{Specialization continues to improve Full-Stack Looping
  after accounting for its additional training cost on DCLM.} Training
  FLOPs include repeated execution, loop-specific corrections,
  attention-score computation, and realized training tokens. Lower BPB is
  better.}
  \label{fig:loop-lora-dclm-scale-flops}
\end{figure*}

The DCLM results retain the same broad pattern observed at smaller
scales: specialization generally improves identical-weight full-stack
looping, with clearer gains at the larger loop counts. The three views are
reported in \autoref{fig:loop-lora-dclm-scale-loops},
\autoref{fig:loop-lora-dclm-scale-params}, and
\autoref{fig:loop-lora-dclm-scale-flops}.

\paragraph{Dataset takeaway.}
Even at 500M and 1B, loop-specific specialization generally improves DCLM
BPB, with the clearest gains at larger loop counts.

\clearpage

\subsubsection{C4 English}

\begin{figure*}[ht]
  \centering
  \includegraphics[width=0.8\textwidth]{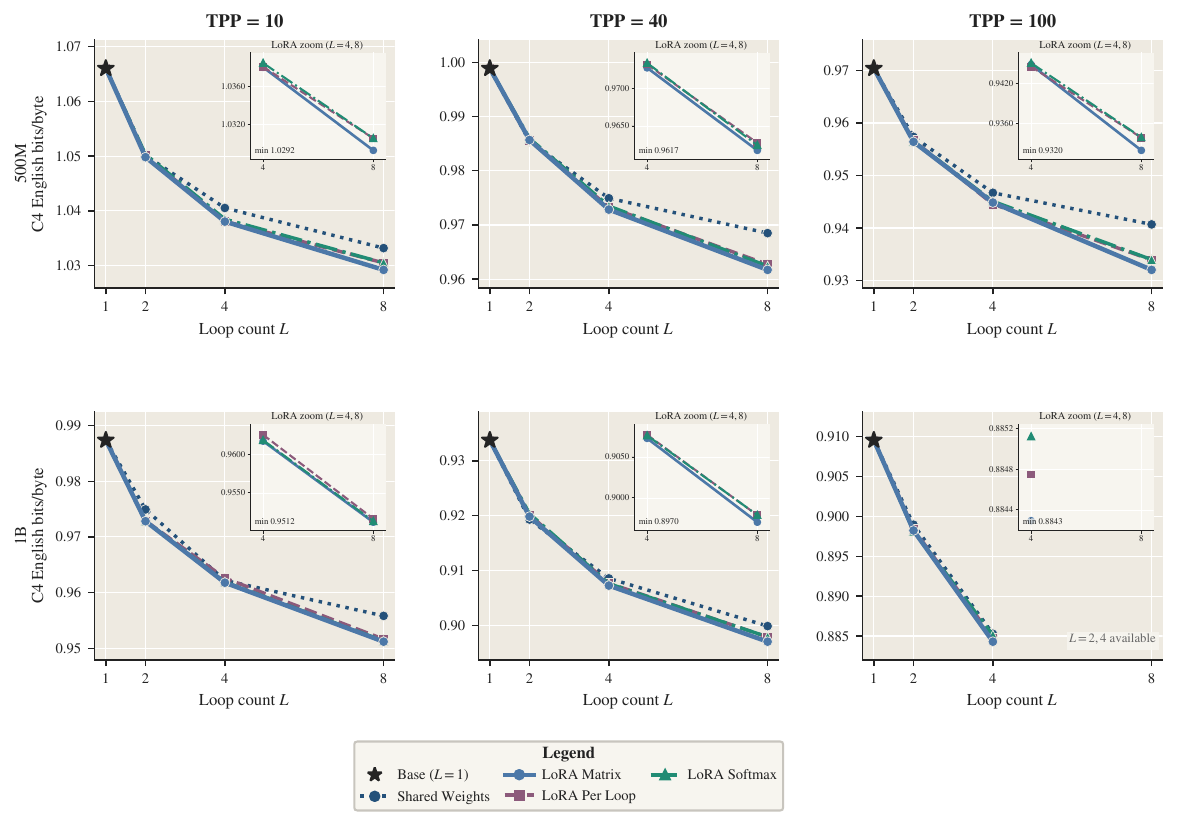}
  \caption{\textbf{Loop-specific specialization remains useful at 500M and
  1B on C4 English.} Full-stack results are shown against loop count for the
  completed TPP settings. The improvement is generally clearest at the larger
  loop counts, although the absolute differences are smaller than in the
  70M--300M sweep. Blank or disconnected cells are unavailable and are not
  imputed. Lower BPB is better.}
  \label{fig:loop-lora-c4-scale-loops}
\end{figure*}

\begin{figure*}[ht]
  \centering
  \includegraphics[width=0.8\textwidth]{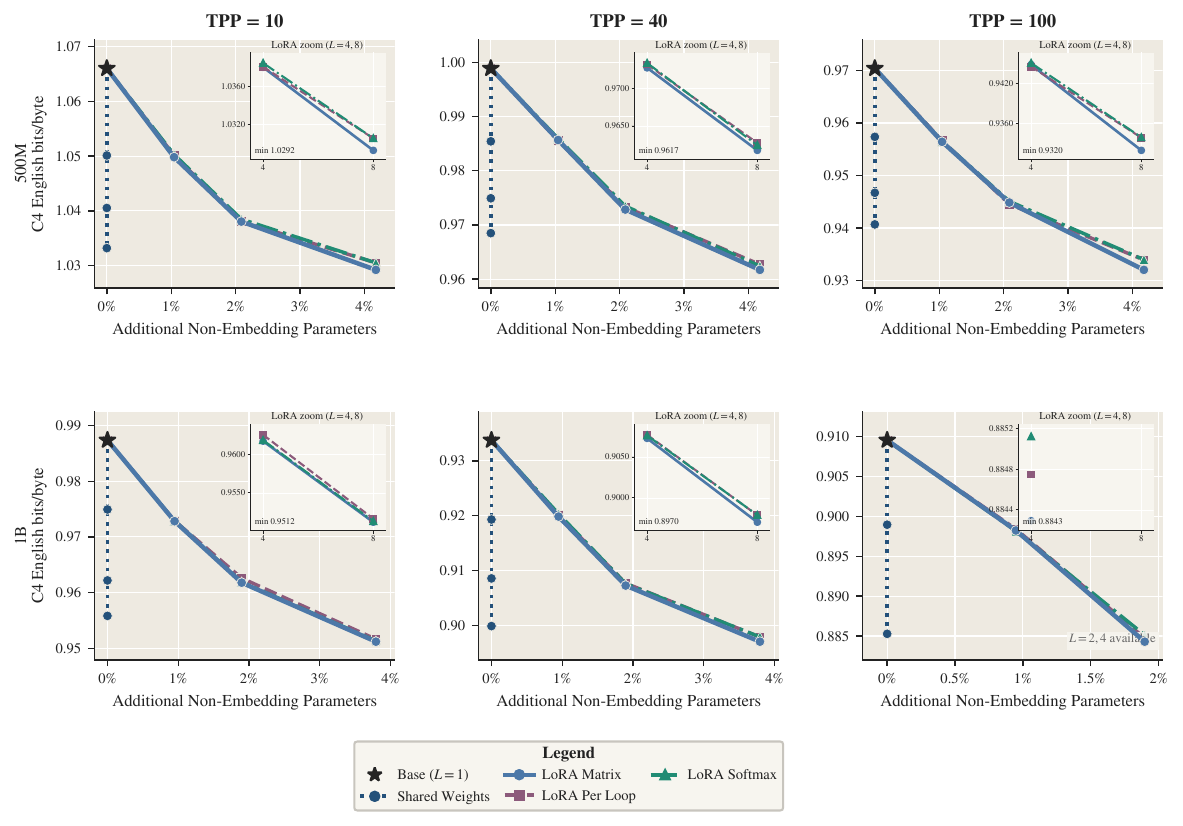}
  \caption{\textbf{The larger-scale specialization gains require relatively
  few additional stored parameters on C4 English.} BPB is plotted against
  additional non-embedding parameters relative to the identical-weight
  full-stack model. Lower BPB is better.}
  \label{fig:loop-lora-c4-scale-params}
\end{figure*}

\begin{figure*}[ht]
  \centering
  \includegraphics[width=0.8\textwidth]{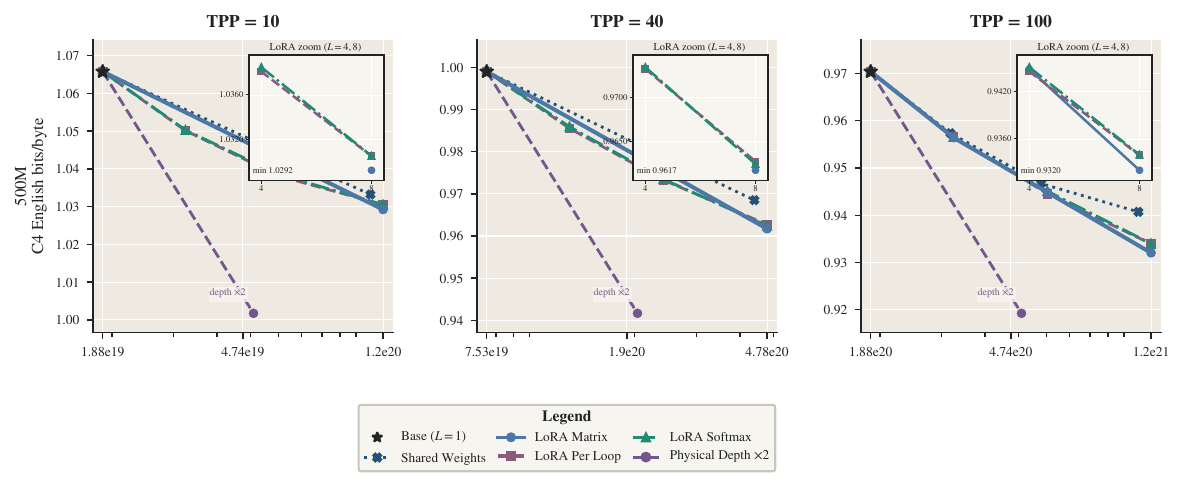}
  \caption{\textbf{Specialization continues to improve Full-Stack Looping
  after accounting for its additional training cost on C4 English.} Training
  FLOPs include repeated execution, loop-specific corrections,
  attention-score computation, and realized training tokens. Lower BPB is
  better.}
  \label{fig:loop-lora-c4-scale-flops}
\end{figure*}

The C4 English results retain the same broad pattern observed at smaller
scales: specialization generally improves identical-weight full-stack
looping, with clearer gains at the larger loop counts. The three views are
reported in \autoref{fig:loop-lora-c4-scale-loops},
\autoref{fig:loop-lora-c4-scale-params}, and
\autoref{fig:loop-lora-c4-scale-flops}.

\paragraph{Dataset takeaway.}
The larger C4 English models show the same result: specialization generally
improves Full-Stack Looping, particularly when the weights are looped more
times.

\clearpage

\subsubsection{WikiText-103}

\begin{figure*}[ht]
  \centering
  \includegraphics[width=0.8\textwidth]{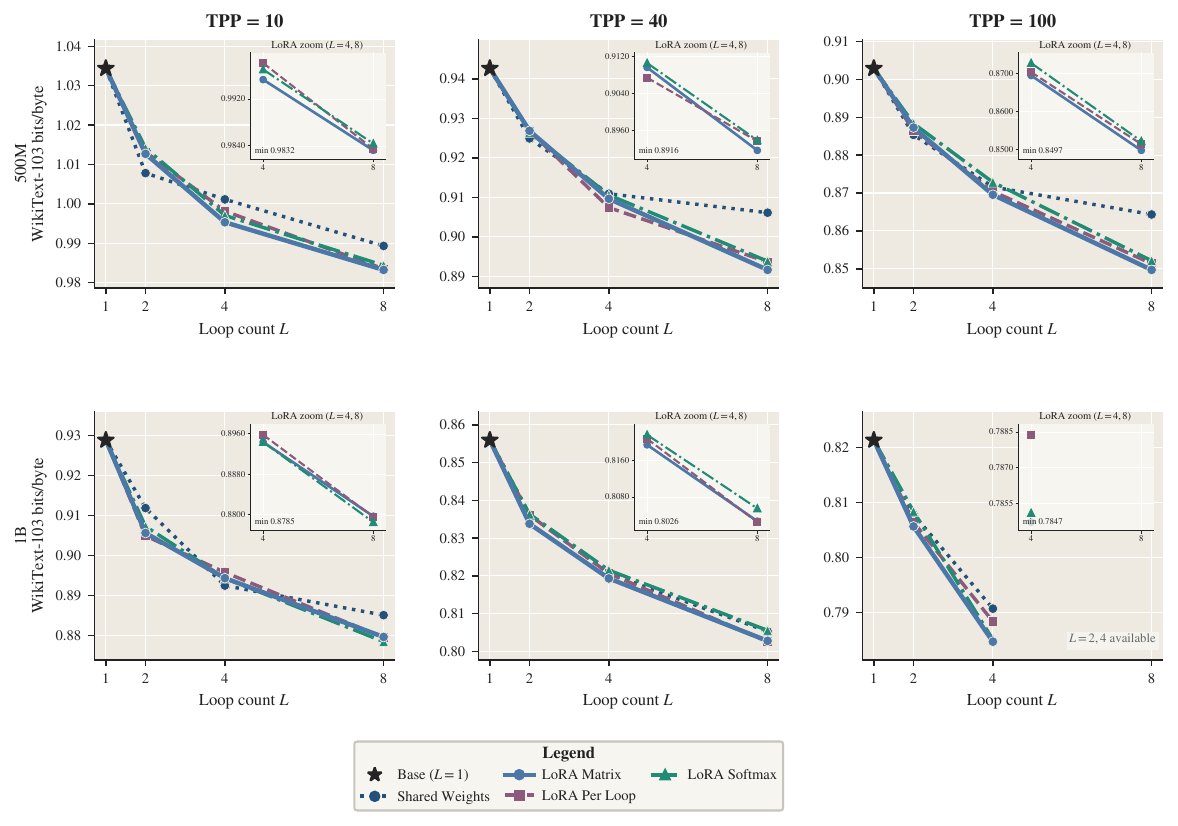}
  \caption{\textbf{Loop-specific specialization remains useful at 500M and
  1B on WikiText-103.} Full-stack results are shown against loop count for the
  completed TPP settings. The improvement is generally clearest at the larger
  loop counts, although the absolute differences are smaller than in the
  70M--300M sweep. Blank or disconnected cells are unavailable and are not
  imputed. Lower BPB is better.}
  \label{fig:loop-lora-wikitext-scale-loops}
\end{figure*}

\begin{figure*}[ht]
  \centering
  \includegraphics[width=0.8\textwidth]{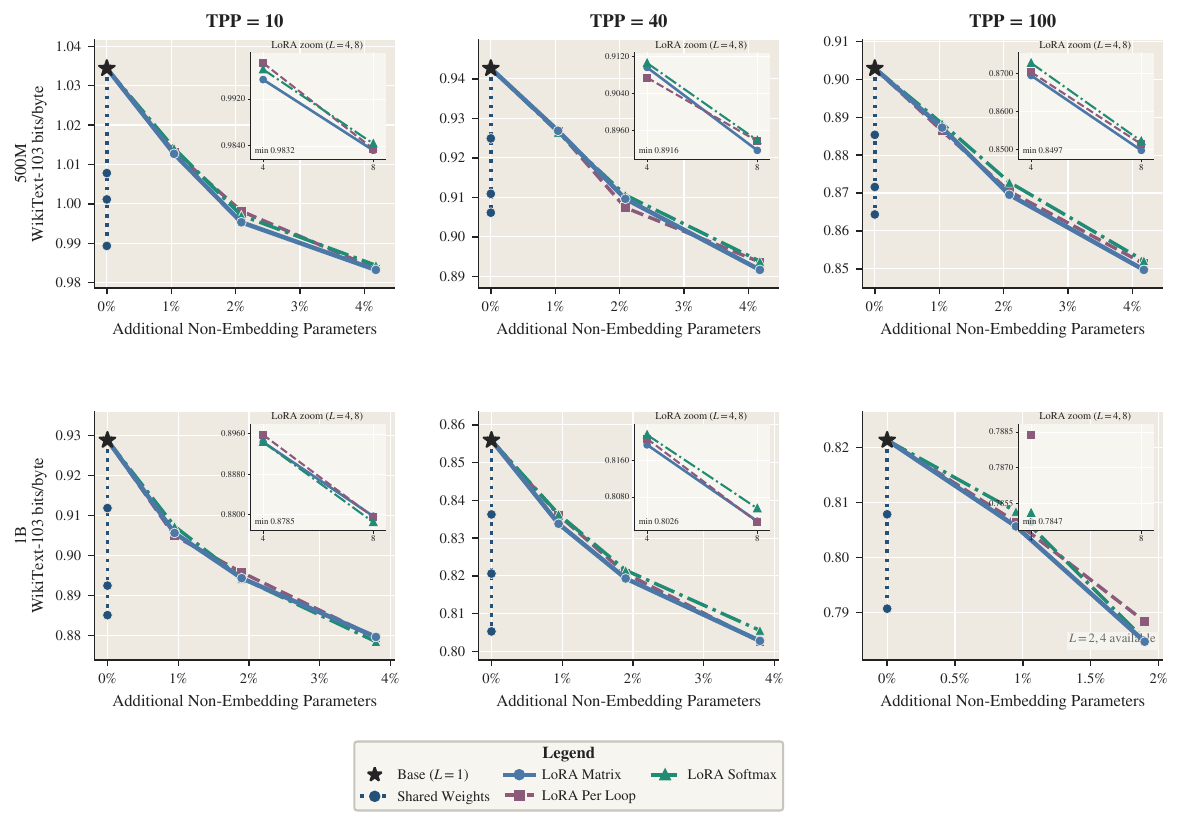}
  \caption{\textbf{The larger-scale specialization gains require relatively
  few additional stored parameters on WikiText-103.} BPB is plotted against
  additional non-embedding parameters relative to the identical-weight
  full-stack model. Lower BPB is better.}
  \label{fig:loop-lora-wikitext-scale-params}
\end{figure*}

\begin{figure*}[ht]
  \centering
  \includegraphics[width=0.8\textwidth]{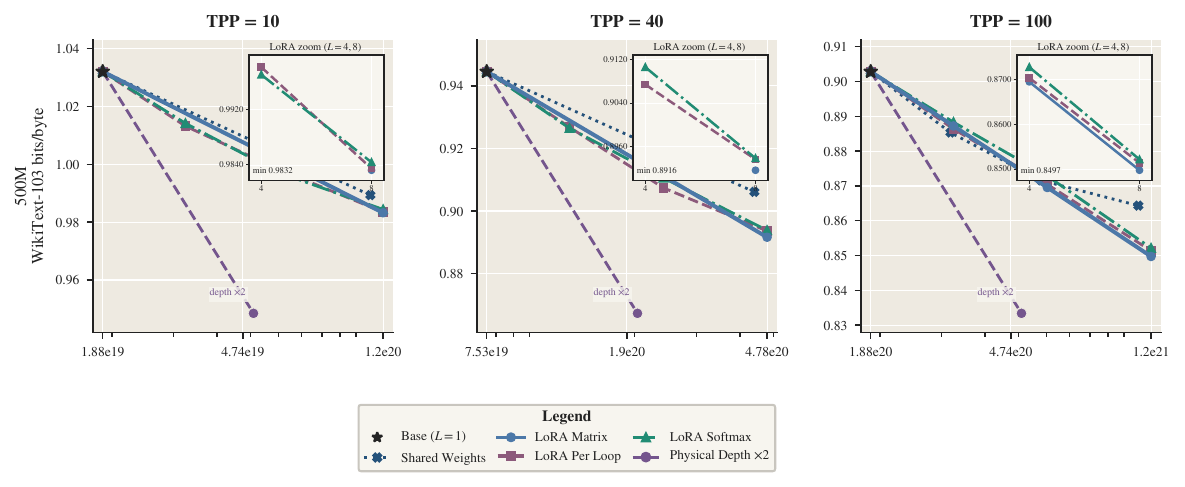}
  \caption{\textbf{Specialization continues to improve Full-Stack Looping
  after accounting for its additional training cost on WikiText-103.} Training
  FLOPs include repeated execution, loop-specific corrections,
  attention-score computation, and realized training tokens. Lower BPB is
  better.}
  \label{fig:loop-lora-wikitext-scale-flops}
\end{figure*}

The WikiText-103 results retain the same broad pattern observed at smaller
scales: specialization generally improves identical-weight full-stack
looping, with clearer gains at the larger loop counts. The three views are
reported in \autoref{fig:loop-lora-wikitext-scale-loops},
\autoref{fig:loop-lora-wikitext-scale-params}, and
\autoref{fig:loop-lora-wikitext-scale-flops}.

\paragraph{Dataset takeaway.}
Specialization also remains beneficial on WikiText-103 at 500M and 1B, with
the clearest improvements at the larger loop counts.

\clearpage

\subsubsection{Dolma v1.5}

\begin{figure*}[ht]
  \centering
  \includegraphics[width=0.8\textwidth]{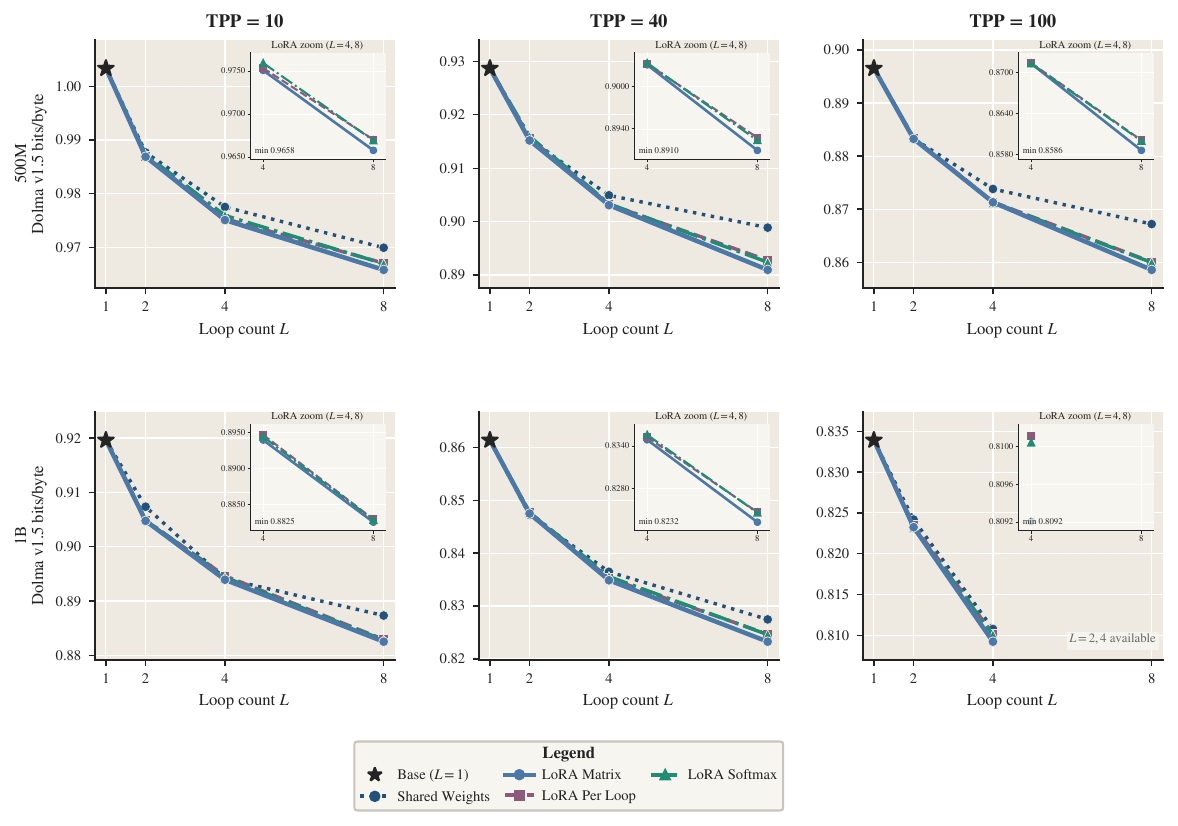}
  \caption{\textbf{Loop-specific specialization remains useful at 500M and
  1B on Dolma v1.5.} Full-stack results are shown against loop count for the
  completed TPP settings. The improvement is generally clearest at the larger
  loop counts, although the absolute differences are smaller than in the
  70M--300M sweep. Blank or disconnected cells are unavailable and are not
  imputed. Lower BPB is better.}
  \label{fig:loop-lora-dolma-scale-loops}
\end{figure*}

\begin{figure*}[ht]
  \centering
  \includegraphics[width=0.8\textwidth]{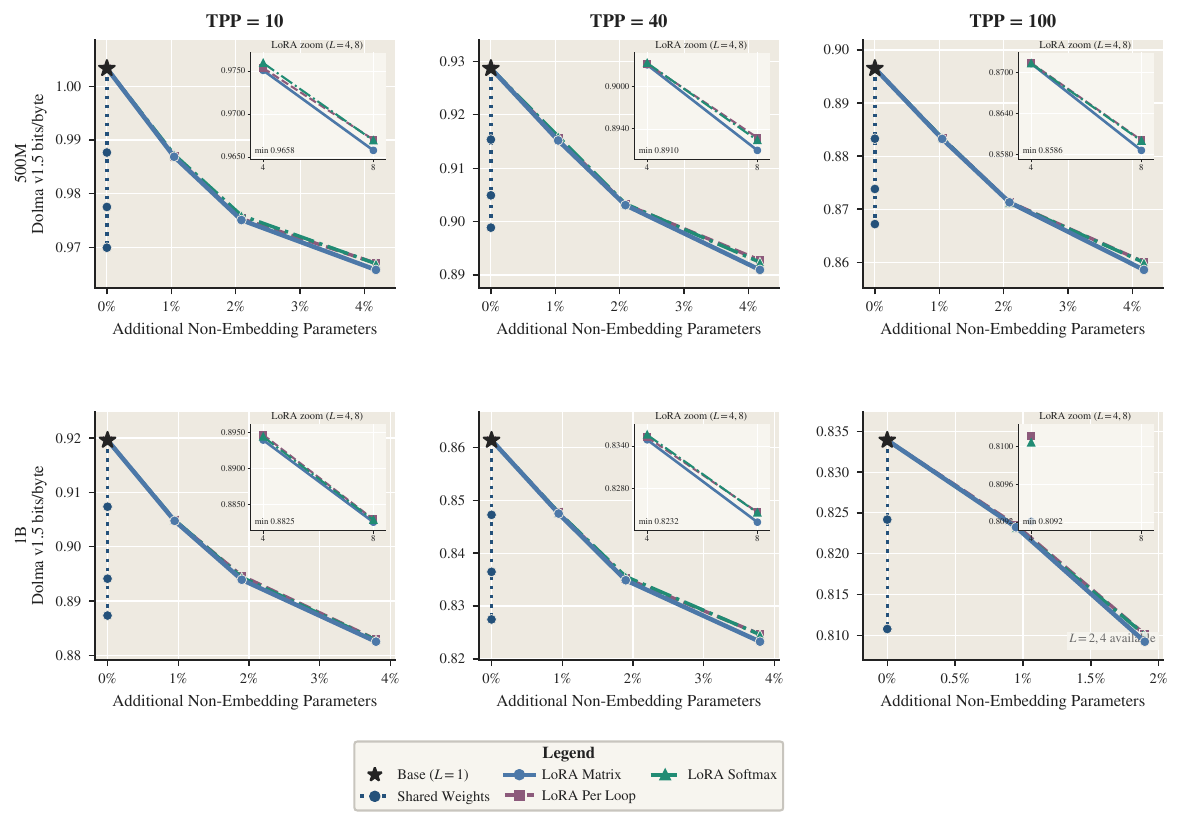}
  \caption{\textbf{The larger-scale specialization gains require relatively
  few additional stored parameters on Dolma v1.5.} BPB is plotted against
  additional non-embedding parameters relative to the identical-weight
  full-stack model. Lower BPB is better.}
  \label{fig:loop-lora-dolma-scale-params}
\end{figure*}

\begin{figure*}[ht]
  \centering
  \includegraphics[width=0.8\textwidth]{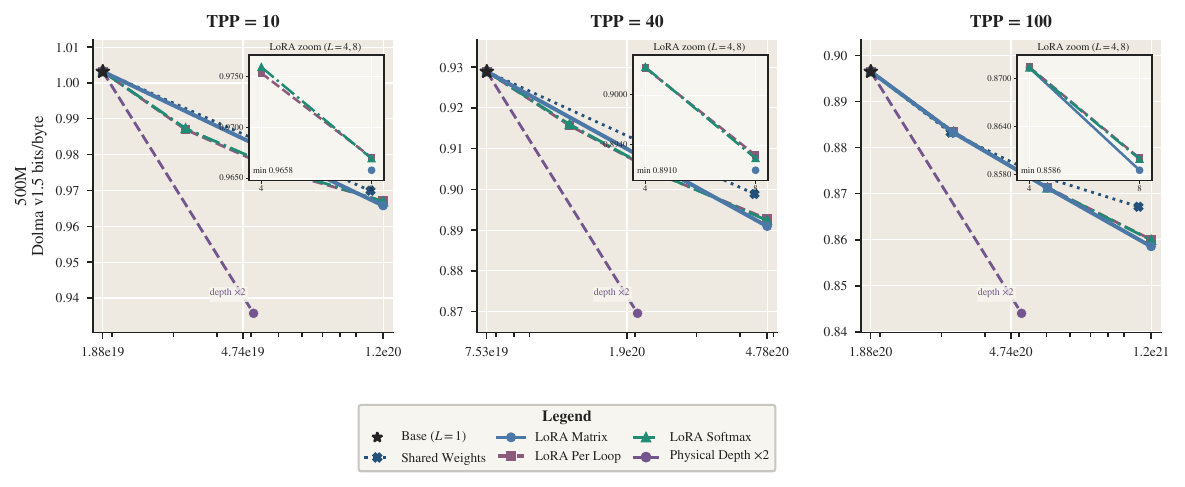}
  \caption{\textbf{Specialization continues to improve Full-Stack Looping
  after accounting for its additional training cost on Dolma v1.5.} Training
  FLOPs include repeated execution, loop-specific corrections,
  attention-score computation, and realized training tokens. Lower BPB is
  better.}
  \label{fig:loop-lora-dolma-scale-flops}
\end{figure*}

The Dolma v1.5 results retain the same broad pattern observed at smaller
scales: specialization generally improves identical-weight full-stack
looping, with clearer gains at the larger loop counts. The three views are
reported in \autoref{fig:loop-lora-dolma-scale-loops},
\autoref{fig:loop-lora-dolma-scale-params}, and
\autoref{fig:loop-lora-dolma-scale-flops}.

\paragraph{Dataset takeaway.}
The Dolma v1.5 results likewise show that loop-specific specialization
continues to improve larger full-stack looped models.

\clearpage

\subsubsection{RedPajama}

\begin{figure*}[ht]
  \centering
  \includegraphics[width=0.8\textwidth]{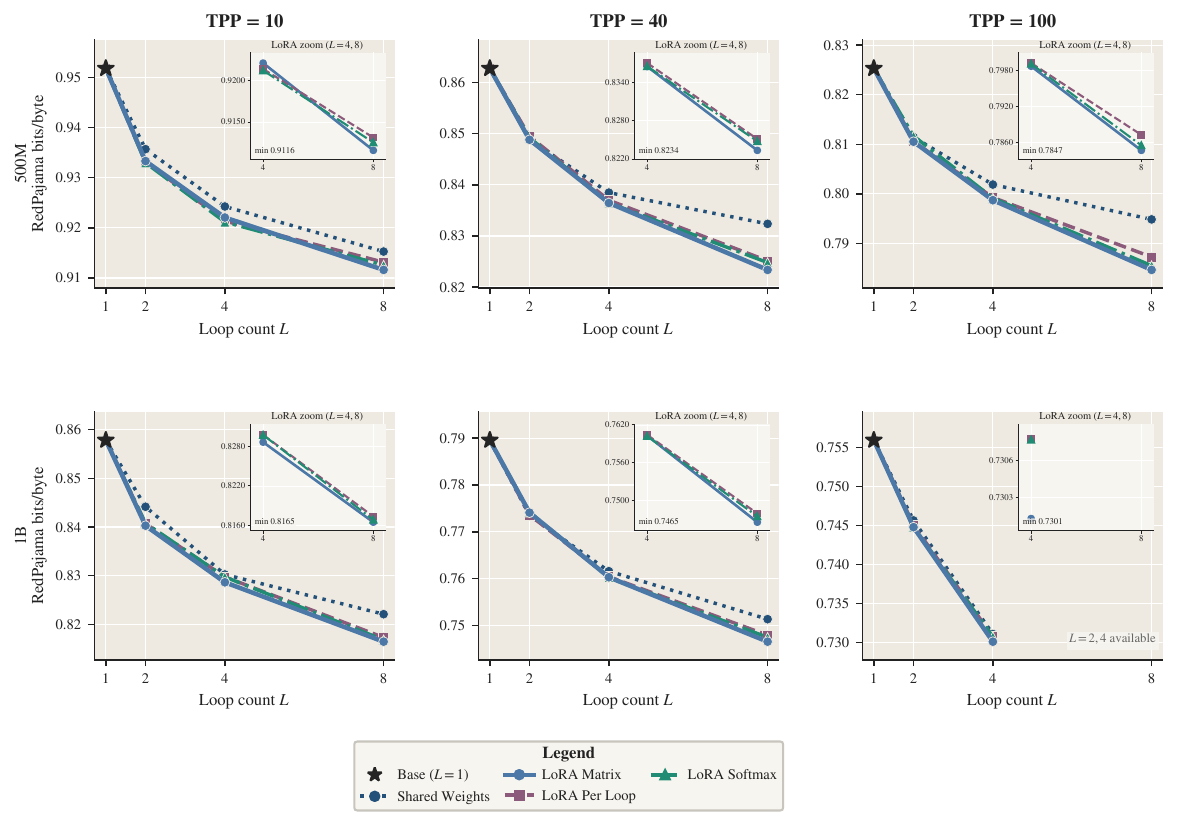}
  \caption{\textbf{Loop-specific specialization remains useful at 500M and
  1B on RedPajama.} Full-stack results are shown against loop count for the
  completed TPP settings. The improvement is generally clearest at the larger
  loop counts, although the absolute differences are smaller than in the
  70M--300M sweep. Blank or disconnected cells are unavailable and are not
  imputed. Lower BPB is better.}
  \label{fig:loop-lora-redpajama-scale-loops}
\end{figure*}

\begin{figure*}[ht]
  \centering
  \includegraphics[width=0.8\textwidth]{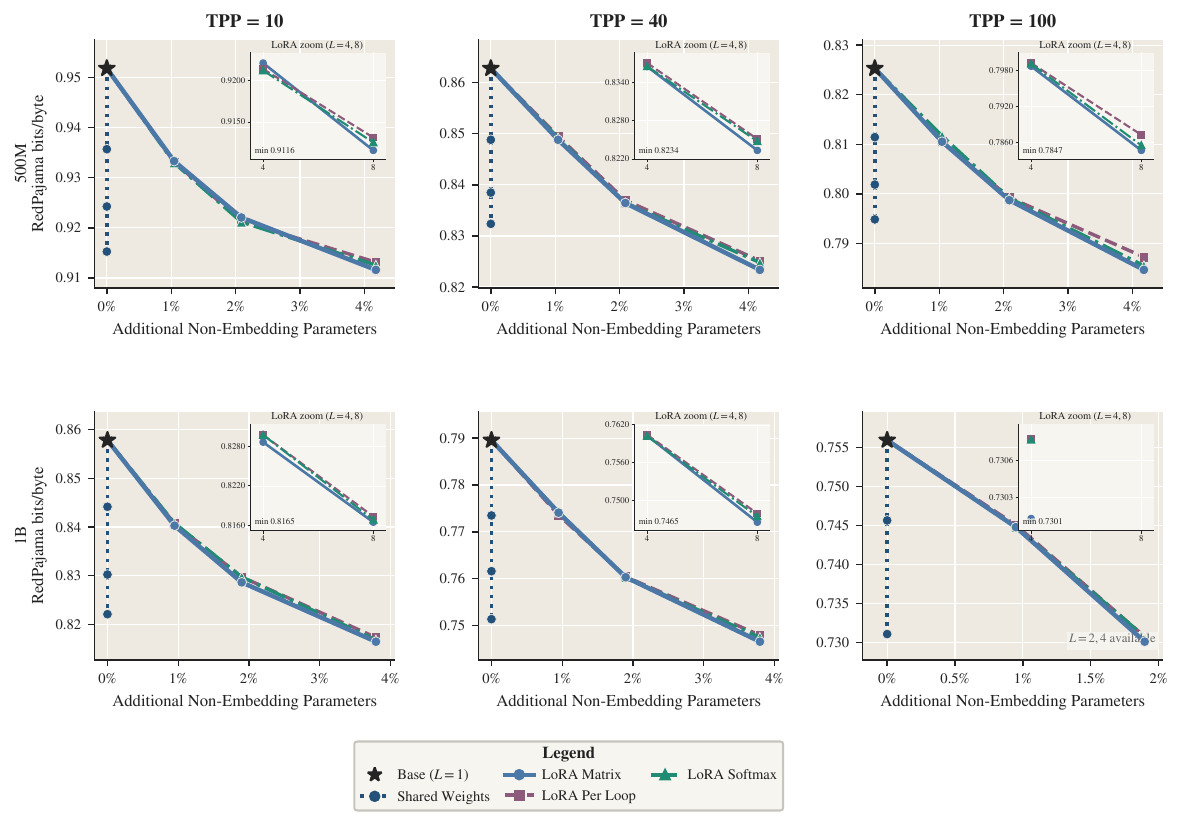}
  \caption{\textbf{The larger-scale specialization gains require relatively
  few additional stored parameters on RedPajama.} BPB is plotted against
  additional non-embedding parameters relative to the identical-weight
  full-stack model. Lower BPB is better.}
  \label{fig:loop-lora-redpajama-scale-params}
\end{figure*}

\begin{figure*}[ht]
  \centering
  \includegraphics[width=0.8\textwidth]{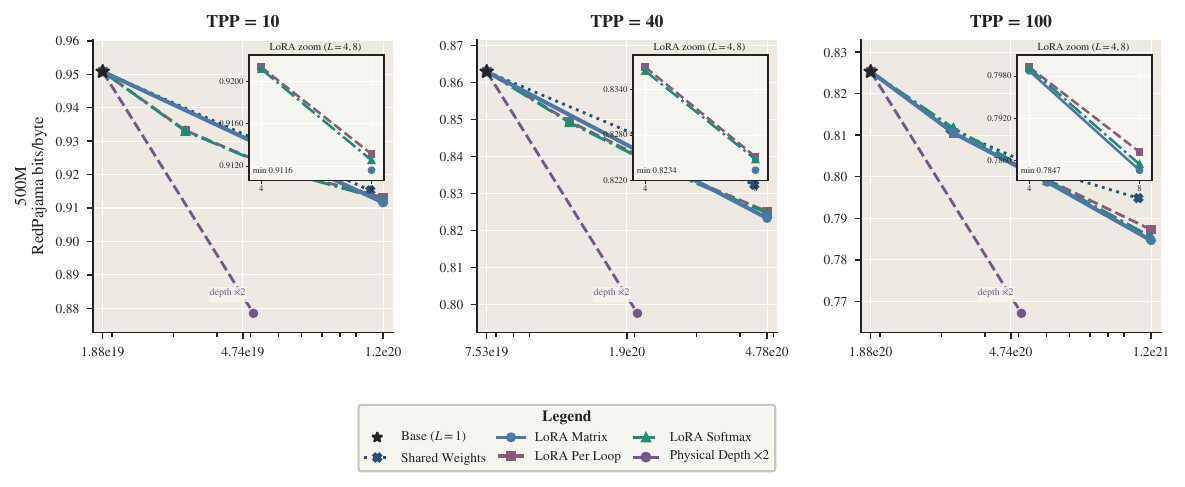}
  \caption{\textbf{Specialization continues to improve Full-Stack Looping
  after accounting for its additional training cost on RedPajama.} Training
  FLOPs include repeated execution, loop-specific corrections,
  attention-score computation, and realized training tokens. Lower BPB is
  better.}
  \label{fig:loop-lora-redpajama-scale-flops}
\end{figure*}

The RedPajama results retain the same broad pattern observed at smaller
scales: specialization generally improves identical-weight full-stack
looping, with clearer gains at the larger loop counts. The three views are
reported in \autoref{fig:loop-lora-redpajama-scale-loops},
\autoref{fig:loop-lora-redpajama-scale-params}, and
\autoref{fig:loop-lora-redpajama-scale-flops}.

\paragraph{Dataset takeaway.}
On RedPajama, specialization again provides an additional benefit over using
identical weights in every loop, including at the larger model scales.

\clearpage

\subsubsection{Dolma Programming Languages}

\begin{figure*}[ht]
  \centering
  \includegraphics[width=0.8\textwidth]{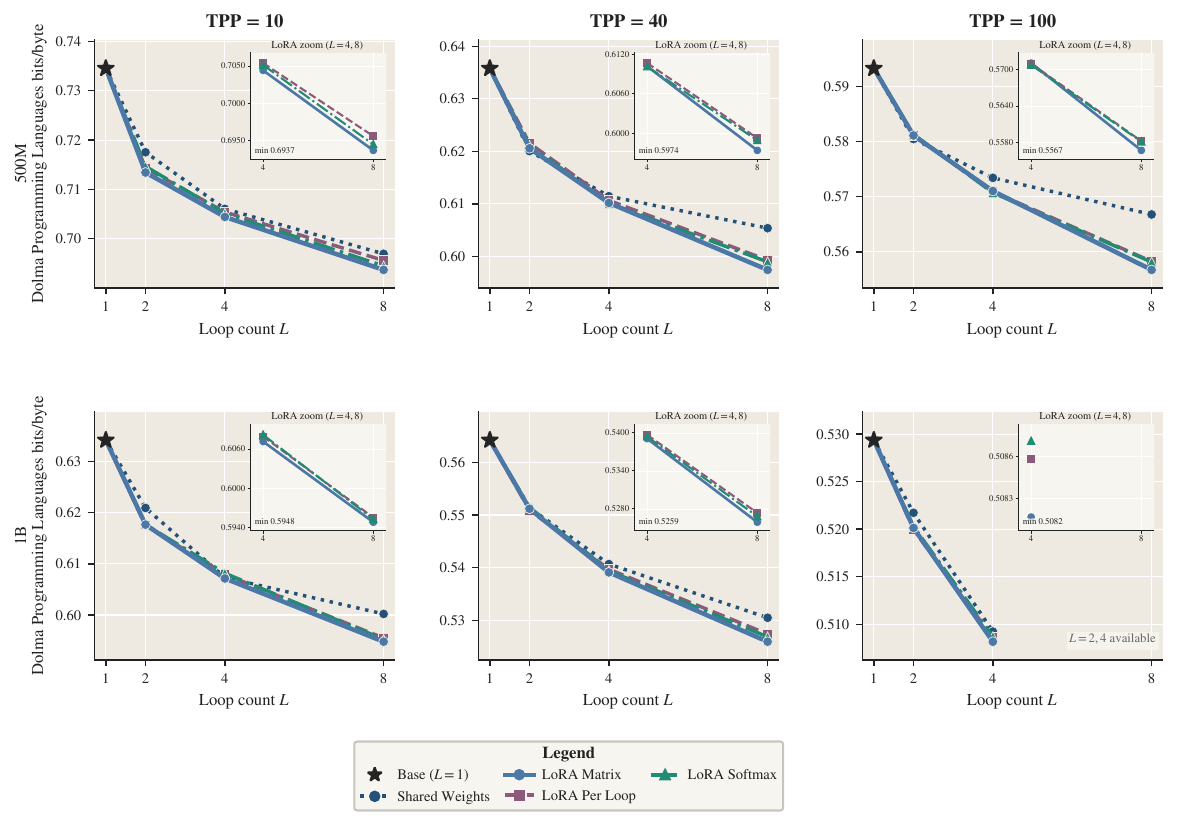}
  \caption{\textbf{Loop-specific specialization remains useful at 500M and
  1B on Dolma Programming Languages.} Full-stack results are shown against loop count for the
  completed TPP settings. The improvement is generally clearest at the larger
  loop counts, although the absolute differences are smaller than in the
  70M--300M sweep. Blank or disconnected cells are unavailable and are not
  imputed. Lower BPB is better.}
  \label{fig:loop-lora-programming-scale-loops}
\end{figure*}

\begin{figure*}[ht]
  \centering
  \includegraphics[width=0.8\textwidth]{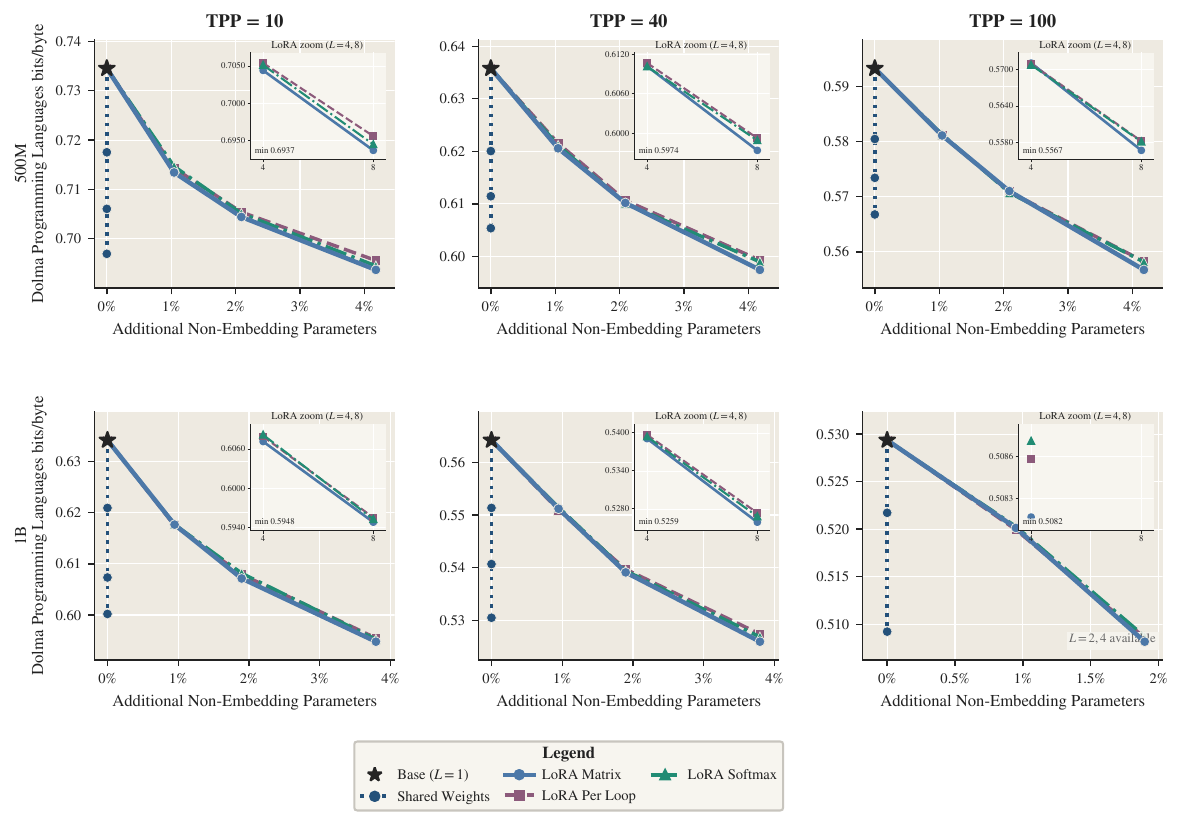}
  \caption{\textbf{The larger-scale specialization gains require relatively
  few additional stored parameters on Dolma Programming Languages.} BPB is plotted against
  additional non-embedding parameters relative to the identical-weight
  full-stack model. Lower BPB is better.}
  \label{fig:loop-lora-programming-scale-params}
\end{figure*}

\begin{figure*}[ht]
  \centering
  \includegraphics[width=0.8\textwidth]{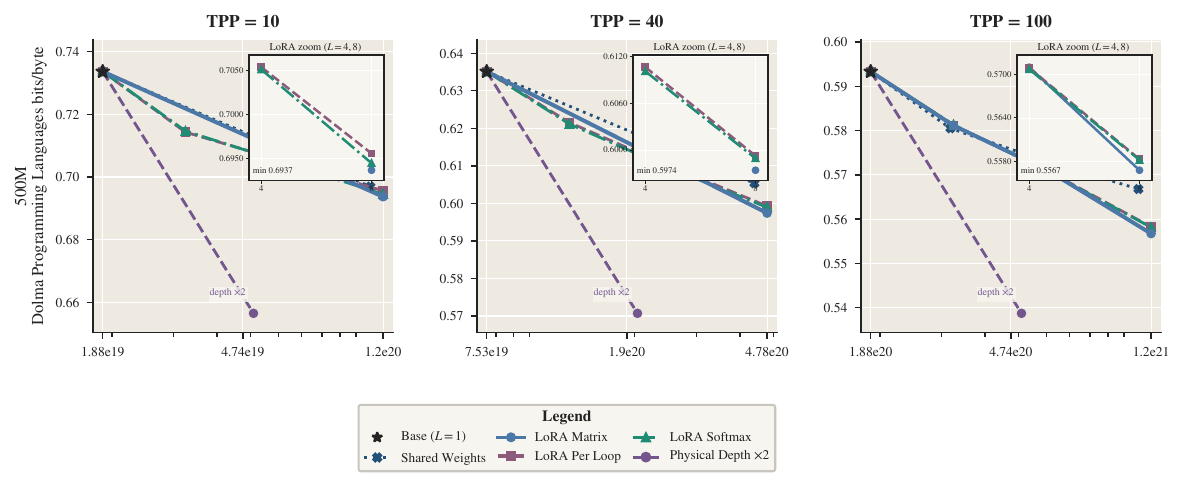}
  \caption{\textbf{Specialization continues to improve Full-Stack Looping
  after accounting for its additional training cost on Dolma Programming Languages.} Training
  FLOPs include repeated execution, loop-specific corrections,
  attention-score computation, and realized training tokens. Lower BPB is
  better.}
  \label{fig:loop-lora-programming-scale-flops}
\end{figure*}

The Dolma Programming Languages results retain the same broad pattern observed at smaller
scales: specialization generally improves identical-weight full-stack
looping, with clearer gains at the larger loop counts. The three views are
reported in \autoref{fig:loop-lora-programming-scale-loops},
\autoref{fig:loop-lora-programming-scale-params}, and
\autoref{fig:loop-lora-programming-scale-flops}.

\paragraph{Dataset takeaway.}
The same conclusion holds on Dolma Programming Languages: specialization
generally improves larger looped models, showing that the benefit extends to
code-heavy language modeling.

\paragraph{Takeaway.}
The main specialization conclusion persists at 500M and 1B: small
loop-specific corrections continue to improve Full-Stack Looping, especially
when the same weights are reused more times. The precise specialization
method remains less important than allowing the loops to differ at all, but
the gains become smaller as model scale increases. Thus, specialization
remains useful at scale. We believe that increasing the parameter count, for
example by adding independently parameterized Transformer layers, would also
improve quality. The purpose of this experiment, however, is to isolate
whether specialization itself continues to help as the model grows.

\clearpage

\clearpage

\clearpage
\section{Additional Results for Parallel Looping}
\label{app:plt}

This appendix provides the additional experiments supporting
\autoref{sec:plt}. As described in \autoref{sec:background_plt}, Parallel
Loop Transformers (PLTs) reorganize looping so that, after a short warm-up,
the computations for different loops can be evaluated together during
autoregressive decoding. Evaluating several loop inputs with the same weights
can increase \emph{arithmetic intensity}, the amount of arithmetic performed
for each byte read from memory, because one set of weights can be reused
across those inputs. PLT does not remove any of the loop computations. During
training, the model still runs all $L$ loops, and the training FLOPs increase
accordingly. We evaluate model quality and account for training FLOPs here;
we do not measure wall-clock decoding latency or throughput.

Our analysis follows a single diagnostic sequence. We first reconstruct the
original PLT design and identify the quality loss introduced by its shifted
transition. We call the rule that combines the current-token representation
with the shifted state the \emph{re-injection rule}. We then measure these two
representations, normalize them before they are combined, and test whether the
combination needs a learned gate or additional sources of context. Finally, we
evaluate the resulting design across model scale, loop count, training budget,
evaluation corpus, stored parameters, and training FLOPs. Unless stated
otherwise, quality is teacher-forced bits per byte (BPB) on the DCLM
validation set, and lower is better. We do not fill in missing observations or
connect curves across missing points.

Throughout this appendix, \emph{Vanilla (No Looping)} denotes the unlooped
baseline, and \emph{Full-Stack Looping} denotes ordinary same-token looping.
For readability, we call our selected parallel design
\emph{shifted PLT (ours)}. It separately normalizes the current-token and
shifted recurrent inputs and combines them with a rank-8 linear gate. Its
same-token counterpart is the \emph{unshifted control}; this control cannot
run its loops concurrently and is included only to measure the modeling cost
of shifting information across token positions. We reserve \emph{Original
PLT} for the earlier design that combines cross-loop parallelism, KV sharing,
and gated sliding-window attention (G-SWA). Loop count is denoted by $L$, and
an individual loop is indexed by $\ell$.

\subsection{Which Part of PLT Causes the Quality Loss?}
\label{app:plt-design-progression}

We use the Cross-Loop Parallelism (CLP) transition defined in
\autoref{def:app-plt-transition} and reconstruct PLT one component at a time. We
first separate the source of the recurrent state from the attention-cache
construction used around it.

\begin{definitionbox}[label=def:app-plt-reinjection-sources]
{\textbf{Shifted, unshifted, and Full-Stack transitions. }}
Let $\bm e_n$ be the embedding of token $n$, let
$\bm h_n^{(\ell-1)}$ be its output from loop $\ell-1$, and let
$\bm b_n^{(\ell)}$ be the input to loop $\ell$. For $\ell>1$, we compare
\begin{align*}
\text{shifted:}\quad
\bm b_n^{(\ell)}
&=\bm e_n+\bm h_{n-1}^{(\ell-1)},\\
\text{unshifted:}\quad
\bm b_n^{(\ell)}
&=\bm e_n+\bm h_n^{(\ell-1)},\\
\text{Full-Stack:}\quad
\bm b_n^{(\ell)}
&=\bm h_n^{(\ell-1)}.
\end{align*}
The shifted transition receives the preceding token's state, removing the
same-token dependency between loops and allowing their computations to be
evaluated together during decoding. The unshifted and Full-Stack transitions
both use the current token's preceding-loop state and therefore execute the
loops sequentially. They differ only in whether $\bm e_n$ is explicitly
re-injected at every loop.
\end{definitionbox}

We now focus on the different cache construction choices; see
\autoref{sec:background_plt} for details.

\begin{definitionbox}[label=def:app-plt-cache-constructions]
{\textbf{PLT cache constructions.} }
All three designs below use the shifted transition in
\autoref{def:app-plt-reinjection-sources}; they differ only in the attention
state retained by later loops.\\

\textbf{Shifted CLP} maintains a separate sequence-wide KV cache for every
loop.\\

\textbf{CLP + KV sharing} retains a sequence-wide cache only for the first
loop and shares it with all later loops.\\

\textbf{Original PLT} supplements the shared first-loop cache with a private
sliding-window cache for each later loop and uses gated sliding-window
attention (G-SWA) to combine the shared global and loop-local context.
\end{definitionbox}

\begin{figure*}[ht]
  \centering
  \includegraphics[width=0.92\textwidth]{
    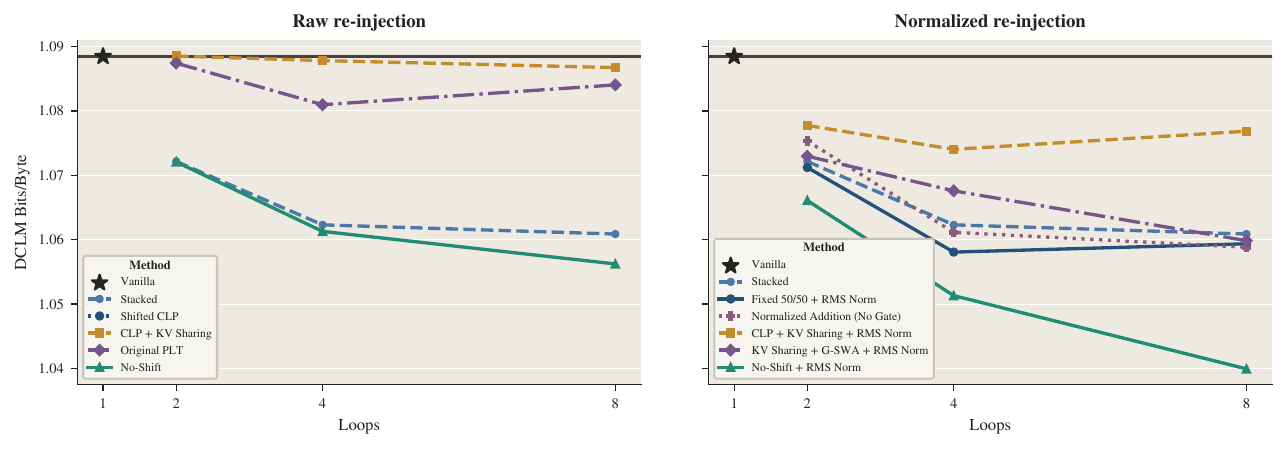
  }\par
  \vspace{0.5em}
  \includegraphics[width=0.92\textwidth]{
    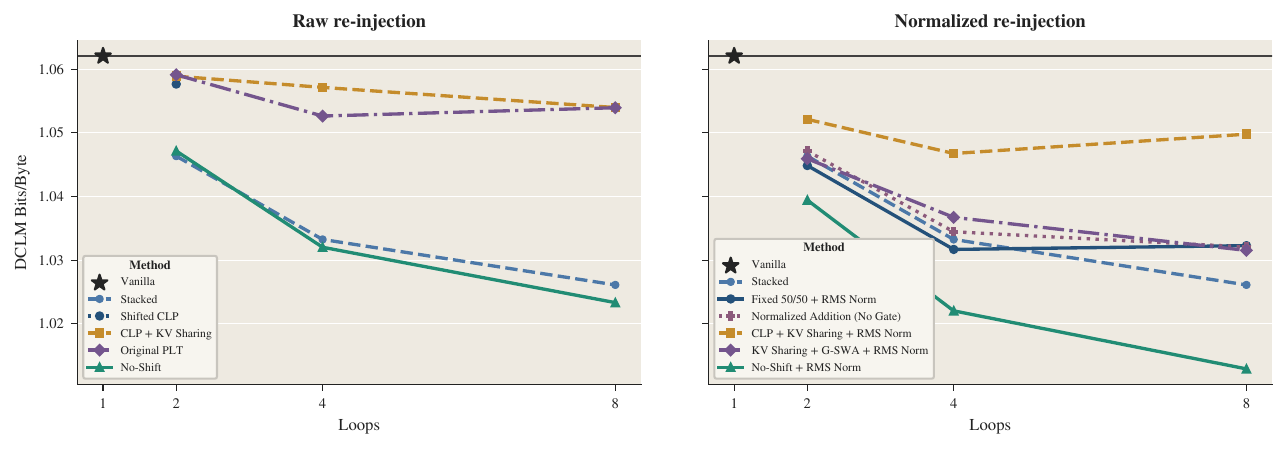
  }
  \caption{\textbf{The shifted re-injection interface is the main source of
  PLT's quality loss at 200M.} DCLM BPB at TPP 40 (top) and TPP 80 (bottom)
  for $L\in\{2,4,8\}$. The left panels isolate the unnormalized PLT designs.
  The unshifted control closely tracks sequential Full-Stack Looping, whereas
  raw Shifted CLP can fail to match even Vanilla. The right panels show that
  separately normalizing and then controlling the mixture of the two inputs
  provides the largest recovery. Lower BPB is better.}
  \label{fig:app-plt-design-200m}
\end{figure*}

\begin{figure*}[ht]
  \centering
  \includegraphics[width=0.92\textwidth]{
    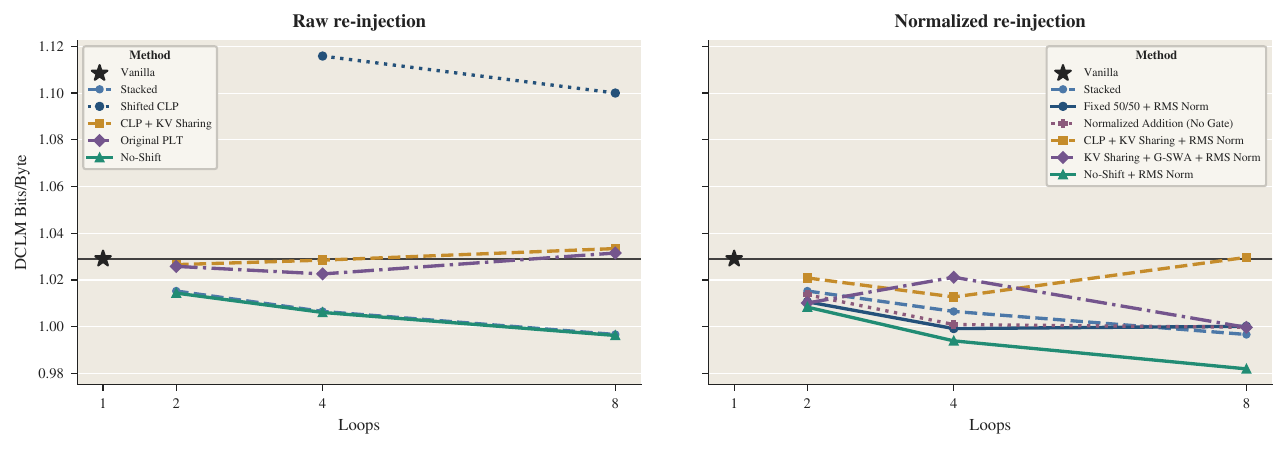
  }\par
  \vspace{0.5em}
  \includegraphics[width=0.92\textwidth]{
    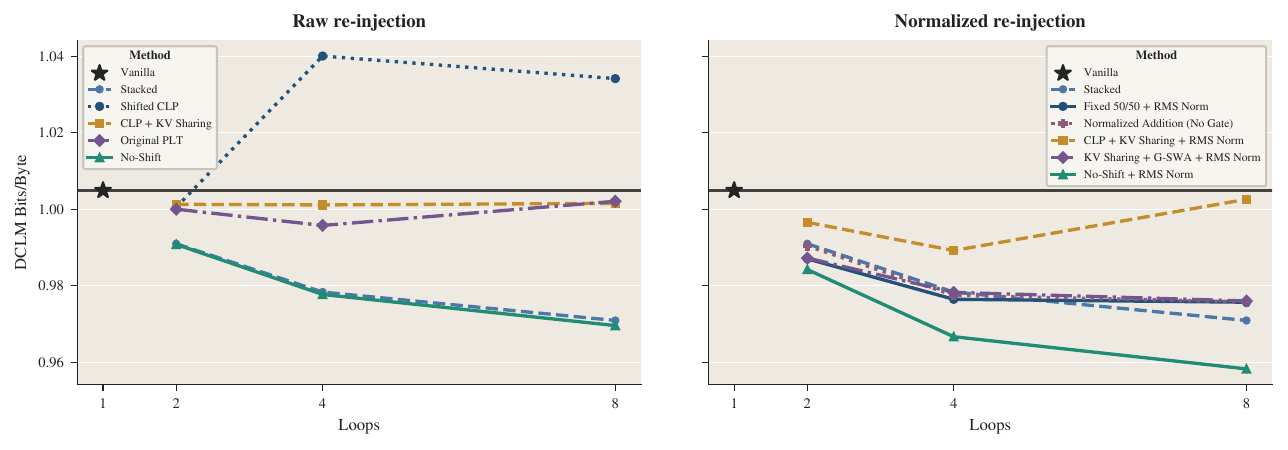
  }
  \caption{\textbf{The same design trade-off persists at 300M.}
  DCLM BPB at TPP 40 (top) and TPP 80 (bottom) for
  $L\in\{2,4,8\}$. The normalized unshifted control is strongest but remains
  sequential. Among the shifted designs, a fixed 50/50 gate with a separate
  KV cache for every loop generally gives the strongest quality. KV sharing
  alone remains weak, while adding G-SWA closes much of the gap and can
  approach Full-Stack quality at larger loop counts. Lower BPB is better.}
  \label{fig:app-plt-design-300m}
\end{figure*}

\paragraph{The cross-token shift is the main source of PLT's quality loss.}
The left panels of \autoref{fig:app-plt-design-200m} and
\autoref{fig:app-plt-design-300m} compare these cache constructions using the
original, unnormalized re-injection rule. The unshifted control closely tracks
ordinary Full-Stack Looping because both preserve the same-token dependency
between successive loops. This confirms that the shared Transformer benefits
from additional loops when the state remains aligned with the current token.
Neither construction, however, provides PLT's inference-efficiency mechanism:
each token must complete one loop before entering the next.

Raw Shifted CLP behaves very differently. Its quality degrades as the number
of loops increases and frequently fails even to match Vanilla (No Looping).
This failure is more severe than the behavior reported in prior work. The
original PLT paper finds gains over a non-looped Transformer
\citep{wu2025parallel}, while LoopCoder-v2 reports a sharp optimum at two total
loops: the second loop improves performance, but models with three or more
loops regress \citep{yang2026loopcoder}. This behavior is initially surprising:
additional loops perform more computation, yet the model can become worse.

This gap matters because the goal is to obtain the quality benefit of ordinary
Full-Stack Looping while allowing the loops to run concurrently during
decoding. The raw shifted transition does not achieve that goal. KV sharing
and G-SWA recover part of the loss, but neither closes the gap to the unshifted
and Full-Stack controls. Because all three PLT variants use the same shifted
state, these comparisons point to the way that state is combined with the
current-token embedding, rather than the cache construction alone, as the
main source of the problem. In the next subsection, we
examine why this normalization helps and show that the failure arises from a
scale mismatch between the two inputs (see \autoref{app:plt-normalization} and \autoref{app:plt-current-token}). For now, we focus on the empirical result:
separately normalizing the current-token embedding and the shifted state (see \autoref{def:app-plt-normalized-mixing}) before
combining them recovers much of the lost quality. 

The right panels in \autoref{fig:app-plt-design-200m} and
\autoref{fig:app-plt-design-300m} retain the corresponding recurrent source and cache
construction but change how the recurrent state is combined with the current
token.

\begin{definitionbox}[label=def:app-plt-normalized-mixing]
{\textbf{Normalized and gated re-injection.} } Using notation from \autoref{def:app-plt-reinjection-sources}, let $\bm r_n^{(\ell)}$ denote the recurrent input:
\begin{align*}
\bm r_n^{(\ell)}
&=
\begin{cases}
\bm h_{n-1}^{(\ell-1)}, & \text{shifted},\\
\bm h_n^{(\ell-1)}, & \text{unshifted}.
\end{cases}
\end{align*}
For any representation $\bm x\in\mathbb{R}^d$, we define RMS normalization as
\begin{align*}
\operatorname{RMSNorm}(\bm x)
&\coloneqq
\frac{\bm x}
{\sqrt{d^{-1}\lVert\bm x\rVert_2^2+\epsilon}},
\end{align*}
where $\epsilon$ is a small constant for numerical stability. RMS normalization
divides a representation by its root-mean-square magnitude. Applying it to the
two inputs separately places them on comparable scales before they interact,
without forcing their directions or information content to be the same. We
therefore define
\begin{align*}
\bar{\bm e}_n
&=\operatorname{RMSNorm}(\bm e_n), &
\bar{\bm r}_n^{(\ell)}
&=\operatorname{RMSNorm}(\bm r_n^{(\ell)}),
\end{align*}
and combine them using
\begin{align*}
\bm b_n^{(\ell)}
&=\bm g_n^{(\ell)}\odot\bar{\bm e}_n
+\bigl(\bm 1-\bm g_n^{(\ell)}\bigr)
  \odot\bar{\bm r}_n^{(\ell)}.
\end{align*}
\textbf{Fixed 50/50 + RMS Norm} uses the parameter-free gate
$\bm g_n^{(\ell)}=\tfrac{1}{2}\bm 1$, and therefore computes
\begin{align*}
\bm b_n^{(\ell)}
&=\tfrac{1}{2}\bar{\bm e}_n
 +\tfrac{1}{2}\bar{\bm r}_n^{(\ell)}.
\end{align*} \\

\textbf{Normalized Addition (No Gate)} instead computes
\begin{align*}
\bm b_n^{(\ell)}
&=\bar{\bm e}_n+\bar{\bm r}_n^{(\ell)}.
\end{align*}
Thus, it applies the same separate RMS normalization but directly adds the two
streams. Unless a curve is labeled \emph{Normalized Addition}, every
normalized design in the right panels uses the fixed 50/50 re-injection gate
and differs only in its recurrent source or KV-cache construction.
\end{definitionbox}

\paragraph{Normalization and controlled mixing repair the shifted interface.}
Normalization improves nearly every shifted construction in the right panels,
but Normalized Addition does not match the consistency of the fixed gated
mixture. The unshifted normalized model performs best overall, confirming that
same-token state is easiest to reuse, but it remains sequential during
decoding. The normalized shifted model with the fixed 50/50 gate is generally
the strongest parallelizable alternative. The main repair is therefore not
RMS normalization alone, but separately normalizing the two streams and
controlling how much each contributes.

\paragraph{Normalization closes the quality gap caused by the cross-token shift.}
The quality loss is the increase in DCLM BPB that occurs when ordinary
Full-Stack Looping is replaced by the raw shifted PLT transition. As shown in
the left panels, raw Shifted CLP performs worse than Full-Stack Looping and can
even perform worse than Vanilla (No Looping). The right panels retain the same
cross-token shift but normalize the current-token embedding and shifted state
before combining them. This substantially lowers BPB across nearly every PLT
variant and brings the shifted models much closer to Full-Stack Looping.
Directly adding the two normalized inputs helps, but averaging them with the
fixed 50/50 rule is more reliable. Thus, the main gain comes from placing the
two inputs on comparable scales and controlling how much each contributes,
rather than from removing the cross-token shift.

\paragraph{Parallel looping can recover the quality of ordinary looping.}
At both 200M and 300M, the strongest shifted results generally come from the
normalized fixed 50/50 model with a separate KV cache for every loop. In our settings, this design matches and sometimes slightly outperforms
Full-Stack Looping while still allowing the loops to run concurrently during
decoding. The normalized unshifted model can perform even better, but it does
not provide this inference benefit because its loops remain sequential. These
comparisons show that using the preceding token's state need not sacrifice the
quality gains of ordinary looping when the current-token embedding and shifted
state are normalized and combined carefully.

The loop-count trend also differs from LoopCoder-v2
\citep{yang2026loopcoder}. That work finds that two total loops improve over a
non-looped model, whereas three or more loops reduce performance. In our
normalized shifted models, quality generally improves or remains stable from
$L=2$ to $L=4$ and $L=8$; we do not observe the same sharp decline after two
loops. This suggests that keeping the two inputs at comparable scales helps
later loops remain useful. The two studies use different model sizes, training
data, and evaluations, so this is not a direct comparison, and we do not claim
that improvements will continue beyond the loop counts tested here.

\paragraph{KV-cache sharing still trades memory for quality.}
Using a separate KV cache for every loop gives the strongest shifted results,
but its memory use grows with $L$. Sharing only the first loop's full cache,
shown as \emph{CLP + KV Sharing + RMS Norm}, performs substantially worse even
after normalization. G-SWA offers a better balance: it combines the shared
first-loop cache with a small private window for each later loop, often
approaches the separate-cache model, and can match Full-Stack Looping at larger
loop counts. In the experiments that follow, we use a separate KV cache for
every loop so that we can study the effect of normalization without also
changing the information available to attention. The G-SWA results show that
the same normalization can also be combined with a more memory-efficient
cache design.

\begin{takeawaybox}
\textbf{Normalization is the key intervention that makes parallel looping
competitive.} Separately normalizing the current-token embedding and the
shifted recurrent state allows the shifted model to match Full-Stack Looping
in our best settings while retaining PLT's inference-time parallelism. When
the shift is removed, the same normalized transition can outperform
Full-Stack Looping, although this unshifted control remains sequential during
decoding. Whereas LoopCoder-v2 reports that performance peaks at two total
loops and regresses with three or more \citep{yang2026loopcoder}, our normalized
shifted models show neither the same saturation at $L=2$ nor systematic
regression over $L\in\{2,4,8\}$. These results suggest that stabilizing the
scales of the two representations allows later loops to remain useful and
that the two-loop limit is not intrinsic to parallel looping.
\end{takeawaybox}

These comparisons identify the re-injection interface as the central modeling
problem and motivate examining the scales of the current-token representation
and shifted state directly.

\subsection{Why Does Normalization Help PLT?}
\label{app:plt-normalization}

The preceding section showed that raw shifted PLT loses quality relative to
Full-Stack Looping, while separately normalizing its two inputs recovers much
of this loss. We now ask why normalization matters. We hypothesize that the
loop representations and embeddings differ substantially in scale. To test
this, we measure the magnitudes of the token embedding and shifted state across
training budgets, loops, model sizes, and PLT cache designs. Raw shifted PLT
adds two representations with different roles. The embedding
$\bm e_n$ corresponds to token $n$, while the shifted state
$\bm h_{n-1}^{(\ell-1)}$ denotes the representation for token $n-1$ at the
preceding loop. Raw addition gives both inputs a coefficient of one, but this
does not make their contributions equal when their magnitudes differ. If one
input is much larger, its scale can strongly influence the result. We
therefore measure whether the two inputs have comparable magnitudes before
they are combined.

\paragraph{Measuring the scale difference.}
For a token representation $\bm x_n\in\mathbb{R}^d$, we measure its
root-mean-square magnitude across the $d$ hidden dimensions:
\begin{equation}
  \operatorname{RMS}(\bm x_n)
  =
  \sqrt{\frac{1}{d}\sum_{j=1}^{d}x_{n,j}^2}
  =
  \frac{\lVert\bm x_n\rVert_2}{\sqrt d}.
  \label{eq:app-plt-rms}
\end{equation}
For each loop, we compute this quantity for every hidden state and token
embedding. We average the hidden-state values across tokens, average the
embedding values across tokens, and then take the ratio of these two averages:
\begin{equation}
  \rho^{(\ell)}
  =
  \frac{
    \frac{1}{N}\sum_{n=1}^{N}
    \operatorname{RMS}\!\left(\bm h_n^{(\ell)}\right)
  }{
    \frac{1}{N}\sum_{n=1}^{N}
    \operatorname{RMS}\!\left(\bm e_n\right)
  },
  \label{eq:app-plt-rms-ratio}
\end{equation}
where $N$ is the total number of evaluated tokens. Thus, we take the ratio
after averaging; we do not compute a separate ratio for every token and then
average those ratios. Every checkpoint is evaluated on the same fixed batch
of four 2,048-token DCLM documents. This experiment measures the internal
representations of the models; it is not an additional likelihood evaluation.

\begin{figure*}[ht]
  \centering
  \includegraphics[width=0.94\textwidth]{
    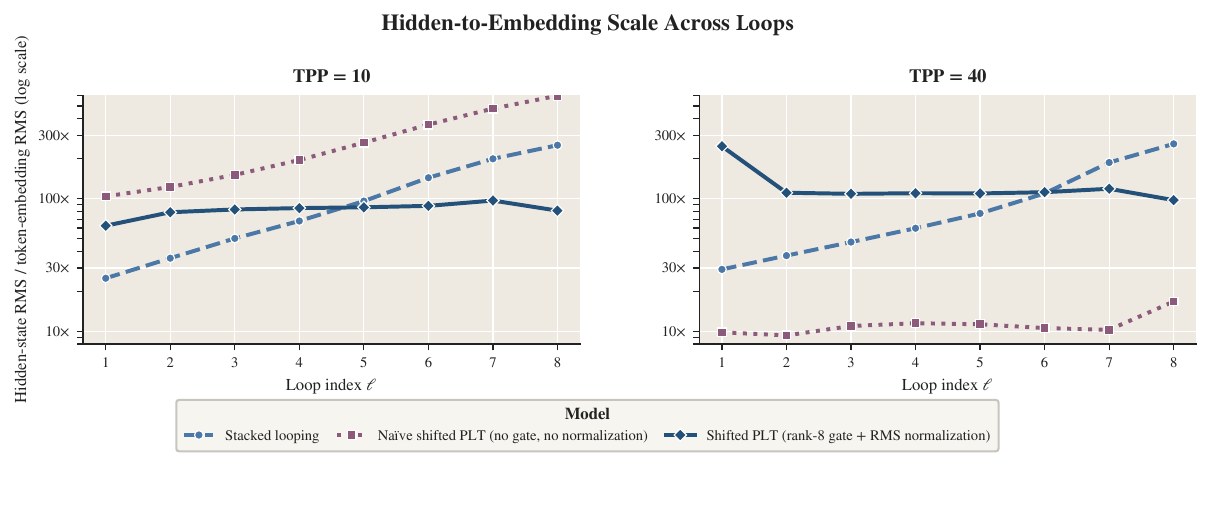
  }
\caption{\textbf{The scale difference changes with training budget and loop
depth.} Hidden-state RMS divided by token-embedding RMS for final 100M,
$L=8$ checkpoints evaluated on the same DCLM sample. Each point is the ratio
of the mean hidden-state RMS across tokens to the mean embedding RMS across
tokens. The figure compares Full-Stack Looping, raw shifted PLT, and
normalized shifted PLT. The raw shifted model uses neither normalization nor
a gate before combining its inputs. The normalized reference uses the
rank-8 linear gate defined in \autoref{app:plt-gates}. The $y$-axis uses
logarithmic spacing, but the tick labels show the underlying RMS ratios, not
their logarithms.}
  \label{fig:app-plt-rms-budget}
\end{figure*}

\paragraph{The scale difference changes with training budget and loop depth.}
We first fix the model size at 100M and compare different training budgets for
models with eight loops. As shown in \autoref{fig:app-plt-rms-budget}, the
hidden states are consistently much larger than the token embeddings. The
magnitude of this difference also changes across training budgets and loops. Raw
addition therefore does not combine the two inputs in a consistent balance:
which input contributes more to the sum can change with both training budget
and loop depth. This finding is consistent with the scale difference
contributing to the lower quality of raw shifted PLT. Full-Stack Looping and
normalized shifted PLT are included as references.

\begin{figure*}[ht]
  \centering
  \includegraphics[width=0.96\textwidth]{
    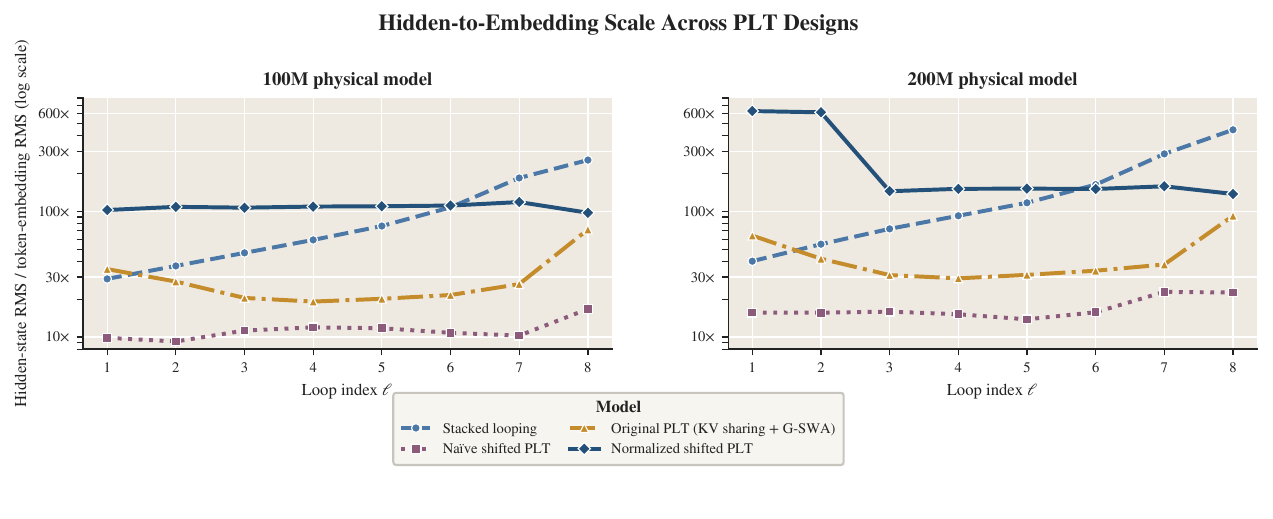
  }
\caption{\textbf{KV sharing and G-SWA do not remove the scale difference.}
Hidden-state RMS divided by token-embedding RMS at TPP 40 and $L=8$ for
final 100M and 200M checkpoints evaluated on the same DCLM token batch.
The $y$-axis uses logarithmic spacing, but the tick labels show the underlying
RMS ratios, not their logarithms. Original PLT is the unnormalized design with
KV sharing and G-SWA. For the normalized model, the plotted quantity is the
hidden state produced by the Transformer. Normalization is applied only when
this state is combined with the embedding at the next loop, so it does not
force the plotted ratio to equal one. The normalized reference uses the rank-8
linear gate defined in \autoref{app:plt-gates}.}
  \label{fig:app-plt-rms-models}
\end{figure*}

\paragraph{The scale difference remains across model sizes and cache designs.}
The first experiment could reflect a particular model size or PLT design. We
therefore fix the training budget and compare 100M and 200M models, including
Original PLT with KV sharing and G-SWA. As shown in
\autoref{fig:app-plt-rms-models}, the hidden states remain substantially larger
than the embeddings at both model sizes and in both raw shifted CLP and
Original PLT. KV sharing and G-SWA change how the difference develops across
loops, but do not make the two inputs similar in scale before they are added.
The problem is therefore not limited to one model size, nor is it removed by
changing the cache and attention design. A normalized model can still produce a hidden state whose RMS is larger than
the embedding RMS. We normalize the two inputs only when
they are about to be combined at the next loop; we do not constrain the size
of the Transformer's output. The relevant question is therefore whether the
inputs are placed on comparable scales before they are combined, not whether
the Transformer must produce an output with an RMS ratio of one.

Together with the quality improvements in
\autoref{fig:app-plt-design-200m} and
\autoref{fig:app-plt-design-300m}, these measurements support differences in
input scale as an important contributor to the failure of raw addition. They
do not show that scale is the only cause, but they motivate normalizing the
embedding and shifted state immediately before combining them.

\begin{takeawaybox}
In raw shifted PLT, the embedding and shifted hidden state can differ
substantially in scale, and the size of this mismatch varies with training
budget and loop depth. The preceding ablations show that separate
normalization and learned gating both improve model quality. Together, they
bring shifted PLT close to the performance of Full-Stack Looping, while the
unshifted PLT variant surpasses Full-Stack Looping. These results show that
PLT can achieve strong performance and that controlling the relative scales
and interaction of the two inputs is particularly important when using the
shifted formulation.
\end{takeawaybox}

\clearpage

\subsection{Does PLT Need a Learned Gate?}
\label{app:plt-gates}

Separate RMS normalization puts the current-token embedding and the shifted hidden state on comparable scales, but it does not specify how the two should be combined. The embedding anchors the computation to the current token, while the shifted state carries information produced by the previous loop for the preceding token. The appropriate balance between them may change across tokens, hidden dimensions, and loops. A fixed 50/50 mixture always gives them equal weight, whereas a learned gate can adjust their contributions based on their contents. We test whether this flexibility is useful by keeping the normalization fixed and varying only the way the two inputs are combined. To determine how much the next loop should use from each source, we introduce
a set of mixing weights between the current-token embedding and the shifted
hidden representation. Recall that
$\bar{\bm e}_n$ is the normalized embedding of the current token, while
$\bar{\bm r}_n^{(\ell)}$ is the normalized hidden representation produced by
the preceding loop for the previous token. Loop $\ell$ receives
\begin{equation}
\bm b_n^{(\ell)}
=
\bm g_n^{(\ell)}\odot\bar{\bm e}_n
+
\left(\bm 1-\bm g_n^{(\ell)}\right)
\odot\bar{\bm r}_n^{(\ell)},
\qquad
\bm g_n^{(\ell)}\in[0,1]^d,
\label{eq:app-plt-gated-mixture}
\end{equation}
where $\odot$ denotes elementwise multiplication. We call
$\bm g_n^{(\ell)}$ the \emph{gate} because it controls how much information
is taken from each input. For a given hidden dimension, a gate value of one
selects the current-token embedding, a value of zero selects the shifted
hidden representation, and a value of one half averages the two. For the learned gates, we predict these mixing weights from both inputs. We
concatenate them as
$\bm z_n^{(\ell)}
=[\bar{\bm e}_n;\bar{\bm r}_n^{(\ell)}]$
and use $\bm z_n^{(\ell)}$ to compute $\bm g_n^{(\ell)}$. We compare these
learned gates with a fixed 50/50 mixture, which sets
$\bm g_n^{(\ell)}=\tfrac{1}{2}\bm 1$ for every token and loop.

\begin{definitionbox}[label=def:app-plt-gates]
{\textbf{Ways of combining the normalized inputs. }}
We compare four choices with increasing flexibility. \\

\textbf{Fixed 50/50:}
\begin{align*}
  \bm g_n^{(\ell)}
  &= \tfrac{1}{2}\bm 1.
\end{align*}
Every token, hidden dimension, and loop receives the same fixed mixture. \\

\textbf{Learned scalar:}
\begin{align*}
  g_n^{(\ell)}
  &= \sigma\!\left(\bm w^\top\bm z_n^{(\ell)}+b\right).
\end{align*}
The resulting scalar is shared across hidden dimensions but can vary with the
inputs at each token and loop. \\

\textbf{Rank-8 linear:}
\begin{align}
  \bm g_n^{(\ell)}
  &= \sigma\!\left(
       \bm W_{\mathrm{up}}\bm W_{\mathrm{down}}
       \bm z_n^{(\ell)}+\bm b
     \right),
  \label{eq:app-plt-gate}
\end{align}
which produces a different coefficient for every hidden dimension through a
rank-8 linear map. \\

\textbf{Rank-8 MLP:}
\begin{align*}
  \bm g_n^{(\ell)}
  &= \sigma\!\left(
       \bm W_{\mathrm{up}}
       \operatorname{ReLU}\!\left(
         \bm W_{\mathrm{down}}\bm z_n^{(\ell)}
       \right)+\bm b
     \right).
\end{align*}
This gate adds a nonlinearity between the same two rank-8 projections. For
both rank-8 gates,
$\bm W_{\mathrm{down}}\in\mathbb{R}^{8\times2d}$ and
$\bm W_{\mathrm{up}}\in\mathbb{R}^{d\times8}$. We initialize
$\bm W_{\mathrm{up}}$ and $\bm b$ to zero, so both gates begin at the fixed
50/50 mixture.
\end{definitionbox}

\begin{figure*}[ht]
  \centering
  \includegraphics[width=0.8\textwidth]{
    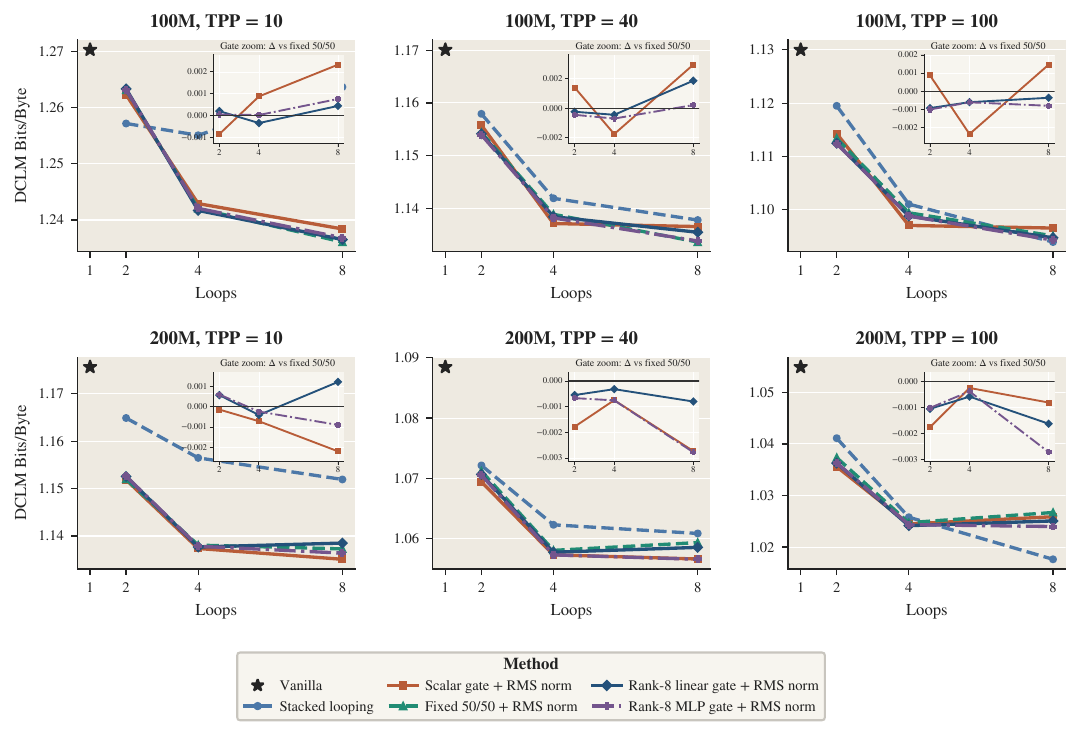
  }
  \caption{\textbf{Gate choice has little effect once the inputs are
  normalized.} DCLM BPB for fixed 50/50, scalar, rank-8 linear, and rank-8 MLP
  mixing at 100M and 200M. Rows show model size and columns show training
  budget; each panel compares loop counts. Insets report matched BPB
  differences from the normalized fixed 50/50 baseline, with negative values
  favoring the learned gate. The rank-8 linear gate improves over fixed 50/50
  in most matched settings, but the differences are small and should be
  treated as directional because most settings use one training seed. Lower
  BPB is better.}
  \label{fig:app-plt-gates-small}
\end{figure*}

\begin{figure*}[ht]
  \centering
  \includegraphics[width=0.8\textwidth]{
    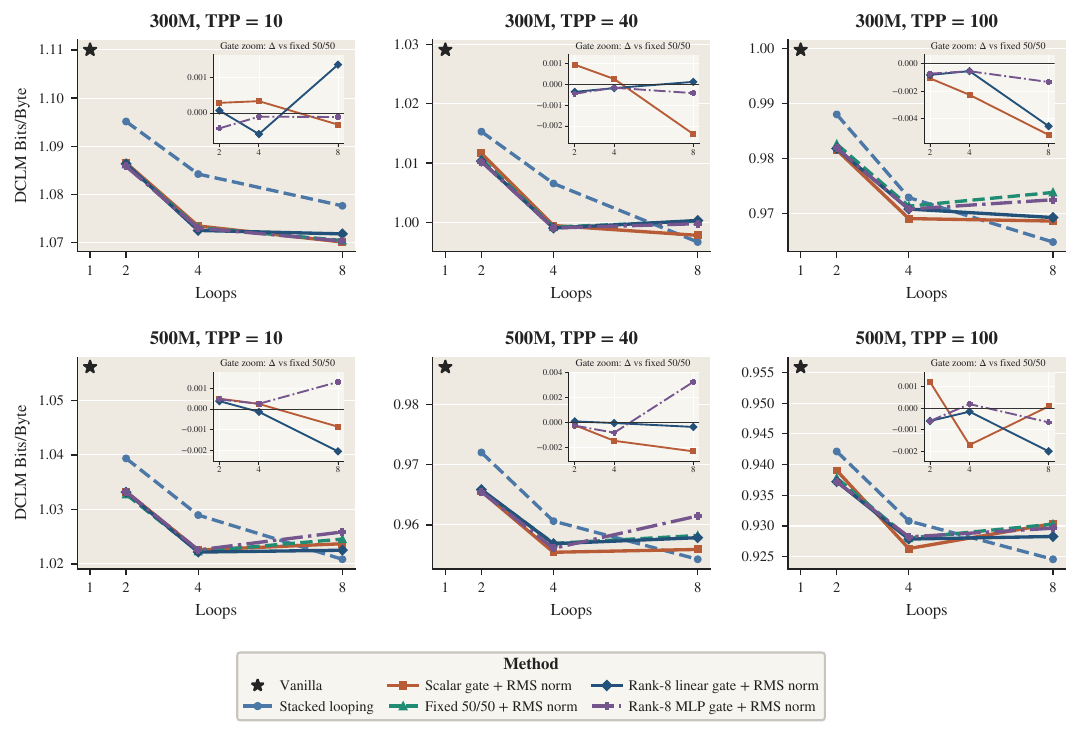
  }
  \caption{\textbf{The gate comparison remains close at larger model sizes.}
  DCLM BPB for the same four mixing rules at 300M and 500M. Rows show model
  size and columns show training budget; insets report matched differences
  from the normalized fixed 50/50 baseline. The rank-8 linear gate is again
  generally favorable, but only by a small margin. Lower BPB and negative
  inset values are better.}
  \label{fig:app-plt-gates-large}
\end{figure*}

\paragraph{Gate choice matters less than normalization.}
We compare the four mixing rules on 100M--500M models at TPP 10, 40, and
100, with $L\in\{2,4,8\}$. Results for the 100M and 200M models appear in
\autoref{fig:app-plt-gates-small}, and results for the 300M and 500M models
appear in \autoref{fig:app-plt-gates-large}. Once the embedding and shifted
hidden representation are normalized separately, all four methods perform
similarly across model sizes, training budgets, and loop counts. The rank-8
linear gate performs best in most matched comparisons, but its improvement
over fixed 50/50 mixing is small. We therefore view the results as favoring
the rank-8 linear gate without demonstrating a substantial advantage. More
importantly, they show that normalization matters much more than the particular
method used to combine the normalized inputs.

We use the rank-8 linear gate in the experiments that follow, including our
1B models. Although the gate comparison used to select it covers models from
100M to 500M, its effectiveness at 1B is evaluated separately in
\autoref{fig:plt-1b-loops}. The small differences among the mixing rules
suggest that fixed 50/50 mixing would lead to the same qualitative
conclusions. The rank-8 gate adds $25d$ parameters: $16d$ for
$\bm W_{\mathrm{down}}$, $8d$ for $\bm W_{\mathrm{up}}$, and $d$ for the
bias. These parameters are shared across loops, so their number does not
increase with $L$. Across the 100M--1B models in which we use the gate, they
account for at most $0.0142\%$ of the non-embedding parameters. At 1B, the
gate adds 51,200 parameters, corresponding to approximately $0.0053\%$ of
the model's non-embedding parameters. It therefore has a negligible effect
on both the stored parameter count and the training-token budget determined
from TPP.

\begin{takeawaybox}
Once the two inputs are normalized, no mixing rule clearly dominates: learned
gates provide only marginal gains and often perform similarly to fixed 50/50
mixing. We use the rank-8 linear gate as a common default because it allows
the mixture to vary across tokens and hidden dimensions, remains consistently
competitive, and adds negligible complexity. We do not interpret this choice
as evidence that the rank-8 gate is essential, and expect fixed 50/50 mixing
to yield the same qualitative conclusions. The gate is shared across loops
and adds less than $0.015\%$ to the non-embedding parameter count at every
scale from 100M to 1B.
\end{takeawaybox}

Having found that the precise gating mechanism has little effect, we keep RMS
normalization and the rank-8 linear gate fixed in the remaining ablations. We
then ask which information must be provided to each loop. Specifically, we
compare using the preceding loop's hidden representation from either the
previous token or the current token, and test each case with and without the
current-token embedding. These comparisons separate the effect of PLT's
cross-token shift from the benefit of explicitly supplying information about
the token currently being processed.

\clearpage

\subsection{Why Does Shifted Looping Need the Current-Token Embedding?}
\label{app:plt-current-token}

Having fixed how the two inputs are combined, we next ask which inputs each
loop needs. Ordinary Full-Stack Looping passes the hidden representation of
the current token from one loop to the next. PLT instead uses the
representation produced by the preceding loop for the previous token and
combines it with the embedding of the current token. This change allows the
loops to run concurrently during decoding, but it also changes the
information available to each loop. We separate these two choices by varying
whether the hidden representation comes from the previous or current token
and whether the current-token embedding is included.

When a hidden representation and an embedding are both present, we normalize
them separately and combine them using the rank-8 linear gate selected in
\autoref{app:plt-gates}. The final condition below is deliberately different:
it omits both normalization and the gate so that it exactly reproduces
ordinary Full-Stack Looping. A bar denotes RMS normalization as defined in
\autoref{def:app-plt-normalized-mixing}.

\begin{definitionbox}[label=def:app-plt-path-controls]
\textbf{Inputs provided to later loops.}
For loops $\ell>1$, we compare four choices. \\

\textbf{Previous-token representation with the current-token embedding:}
\begin{align*}
  \bm b_n^{(\ell)}
  &=
  \bm g_n^{(\ell)}\odot\bar{\bm e}_n
  +
  (\bm 1-\bm g_n^{(\ell)})\odot
  \bar{\bm h}_{n-1}^{(\ell-1)}.
\end{align*}

\textbf{Current-token representation with the current-token embedding:}
\begin{align*}
  \bm b_n^{(\ell)}
  &=
  \bm g_n^{(\ell)}\odot\bar{\bm e}_n
  +
  (\bm 1-\bm g_n^{(\ell)})\odot
  \bar{\bm h}_{n}^{(\ell-1)}.
\end{align*}

\textbf{Previous-token representation without the current-token embedding:}
\begin{align*}
  \bm b_n^{(\ell)}
  &= \bar{\bm h}_{n-1}^{(\ell-1)}.
\end{align*}

\textbf{Current-token representation without the current-token embedding:}
\begin{align*}
  \bm b_n^{(\ell)}
  &= \bm h_n^{(\ell-1)}.
\end{align*}

The final choice is ordinary Full-Stack Looping: each loop receives the
representation produced by the preceding loop for the same token.
\end{definitionbox}

The two choices that use
$\bm h_{n-1}^{(\ell-1)}$ retain PLT's ability to run the loops concurrently
during decoding. The choices that use $\bm h_n^{(\ell-1)}$ require the current
token to finish one loop before entering the next and therefore remain
sequential.

\begin{figure*}[ht]
  \centering
  \includegraphics[width=0.88\textwidth]{
    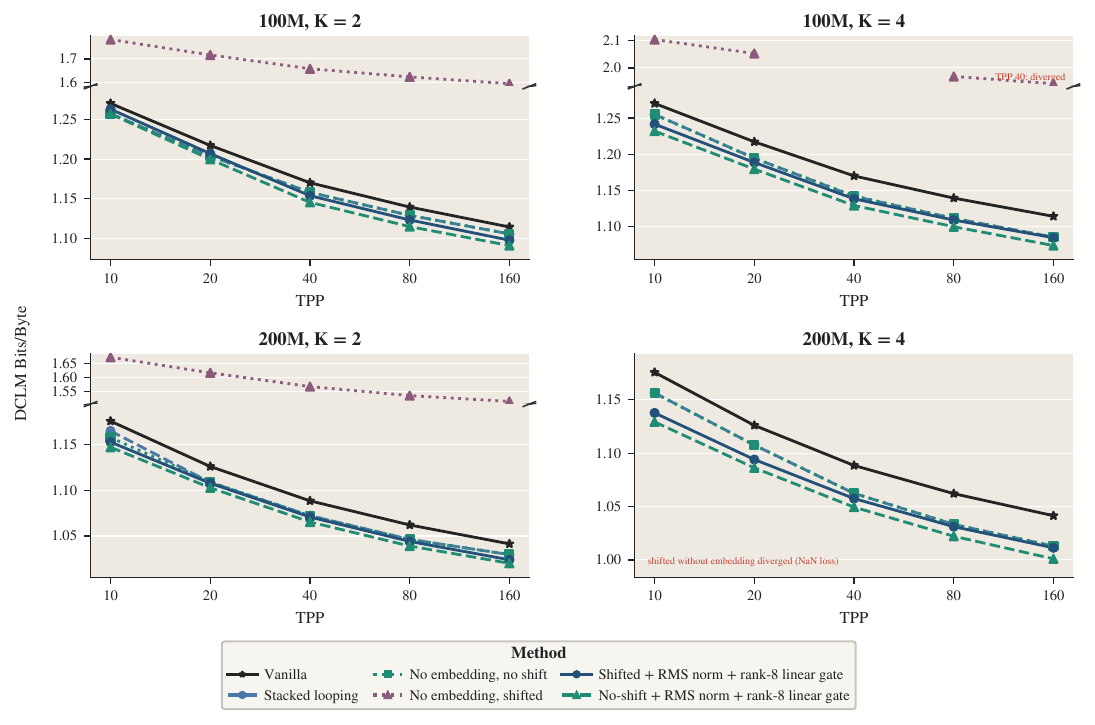
  }
  \caption{\textbf{Shifted looping requires the current-token embedding.}
  DCLM BPB across training budgets for the four inputs in
  \autoref{def:app-plt-path-controls}. The current-token representation
  without the embedding closely matches the independent Full-Stack
  implementation, providing a check of the implementation. Adding the
  embedding improves both the previous-token and current-token variants. When
  the representation comes from the previous token, removing the embedding
  produces results worse than Vanilla (No Looping), and several runs diverge.
  The two conditions with both inputs use separate RMS normalization and the
  rank-8 linear gate. Missing or diverged results are not filled in. Lower BPB
  is better.}
  \label{fig:app-plt-embedding}
\end{figure*}

\paragraph{The Full-Stack control verifies the implementation.}
The current-token condition without the embedding is mathematically
equivalent to ordinary Full-Stack Looping. Its performance closely matches
our independent Full-Stack implementation in
\autoref{fig:app-plt-embedding}. This confirms that the comparison reproduces
ordinary looping when the cross-token shift and additional embedding are both
removed.

\paragraph{The current-token embedding is especially important after the
shift.}
Including the current-token embedding improves both the previous-token and
current-token variants. Its effect is much larger when the hidden
representation comes from the previous token. Without the embedding, the
shifted model performs worse than Vanilla (No Looping), and several runs
diverge. Because the two conditions that combine an embedding and a hidden
representation use the same normalization and gate, this difference isolates
the information provided by the current-token embedding.

\paragraph{Why might the embedding be necessary?}
The representation at position $n-1$ is computed before the model observes
the token at position $n$. It contains the context used to predict the next
token, but it does not specify which token actually occurred. Passing this
representation to the next loop without $\bm e_n$ therefore asks the model to
continue processing information derived from the previous context without
directly telling it the identity of the current token. Repeating this across
several loops may cause the representation to move progressively farther from
the token being processed.

Providing $\bm e_n$ at every loop gives the model the identity of the observed
token and may keep the repeated computation tied to the current position. We
hypothesize that this is why the embedding improves training stability and
quality, particularly as the number of shifted loops increases. The
experiment establishes that the embedding is necessary in our shifted models,
although it does not by itself prove this explanation.

\begin{takeawaybox}
Ordinary Full-Stack Looping is recovered when each loop receives only the
current token's representation from the preceding loop, which verifies our
implementation. When that representation instead comes from the previous
token, the current-token embedding becomes essential: including it improves
quality, while removing it performs worse than no looping and can cause
training to diverge. We hypothesize that the previous-token representation
reflects what the preceding context predicts, whereas the embedding tells
each loop which token actually occurred.
\end{takeawaybox}

For the remaining shifted-PLT experiments, we therefore use separate RMS
normalization, the rank-8 linear gate, and the current-token embedding. We
next ask whether supplying additional information from earlier tokens provides
any further benefit.

\clearpage

\section{Evaluating PLT Across Datasets and Resource Budgets}
\label{app:plt-evaluation}

The previous section identified the PLT design used in the remaining
experiments: the current-token embedding and the hidden representation from
the preceding loop are normalized separately and combined using the rank-8
linear gate. We now keep this design fixed and evaluate it more broadly. We
compare two versions throughout. \emph{Shifted PLT} uses the previous token's
representation and allows the loops to run concurrently during decoding.
\emph{Unshifted PLT} uses the current token's representation from the
preceding loop and must run its loops one after another. It therefore serves
as a quality reference rather than an inference-efficient alternative. We
first compare the two versions across evaluation datasets at 1B. We then vary
model size, loop count, and training budget on DCLM, before comparing quality
against stored parameters and training FLOPs.

\subsection{How Does PLT Perform Across Evaluation Datasets at 1B?}
\label{app:plt-public-evals}

\begin{figure*}[ht]
  \centering
  \includegraphics[width=0.94\textwidth]{
    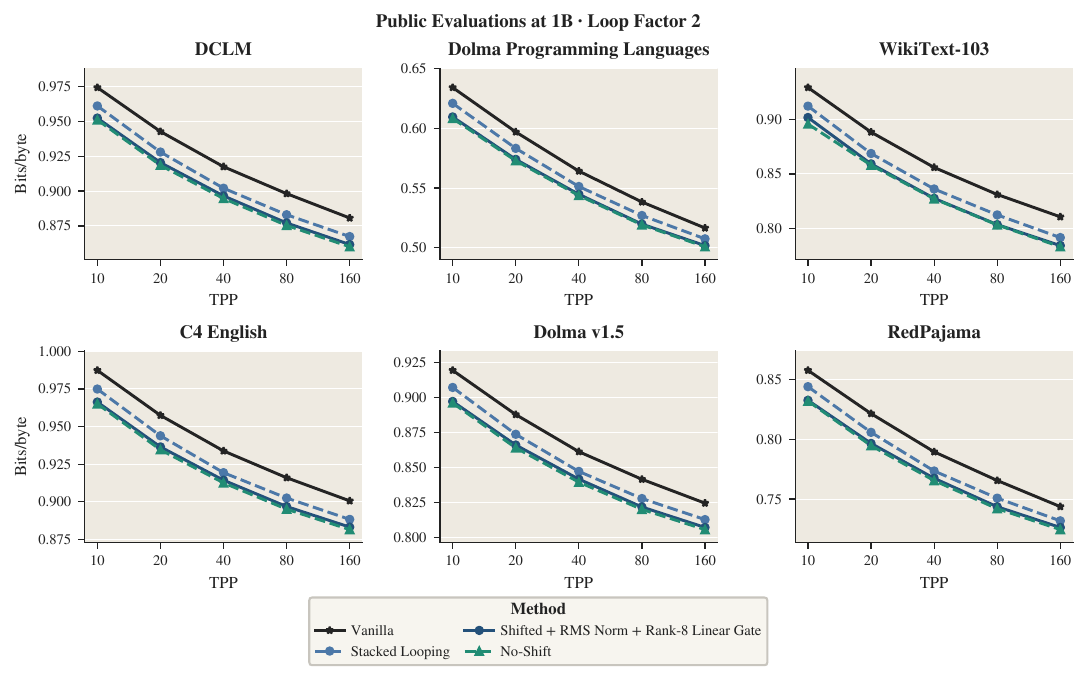
  }
  \caption{\textbf{Normalized shifted PLT preserves the benefit of looping at
  1B across six evaluation datasets.} At $L=2$, shifted PLT closely matches
  unshifted PLT and performs better than ordinary Full-Stack Looping, which in
  turn performs better than Vanilla (No Looping). Shifted PLT retains the
  parallel decoding schedule, while unshifted PLT and Full-Stack Looping
  remain sequential. The panels show BPB across TPP 10--160 on DCLM, C4
  English, WikiText-103, Dolma v1.5, RedPajama, and Dolma programming
  languages. The same trained checkpoints are evaluated on every dataset, and
  each panel uses its own y-axis. Lower BPB is better.}
  \label{fig:app-plt-public-summary}
\end{figure*}

We begin by testing whether the behavior observed on DCLM extends to other
evaluation datasets. \autoref{fig:app-plt-public-summary} fixes the model size
at 1B and the loop count at $L=2$, and compares Vanilla (No Looping), ordinary
Full-Stack Looping, shifted PLT, and unshifted PLT across TPP 10--160. We
evaluate the same trained checkpoints on DCLM, C4 English, WikiText-103,
Dolma v1.5, RedPajama, and Dolma programming languages, so only the evaluation
data change across panels. Lower BPB is better.

Across the six datasets, shifted PLT closely matches unshifted PLT and
performs better than ordinary Full-Stack Looping, which in turn performs
better than Vanilla (No Looping). Unlike unshifted PLT and Full-Stack Looping,
shifted PLT retains the ability to run its loops concurrently during decoding.
The same pattern at 1B and across different datasets shows that the result is
not limited to smaller models or to DCLM. It also reinforces the conclusion of
the preceding section: separately normalizing the two inputs and supplying the
current-token embedding are important for preserving quality when the hidden
representation is shifted across tokens.

\subsection{How Does PLT Change with Model Size and Loop Count?}
\label{app:plt-loop-sweeps}
\label{app:plt-loop-count-details}

\begin{figure*}[ht]
  \centering
  \includegraphics[width=0.88\textwidth]{
    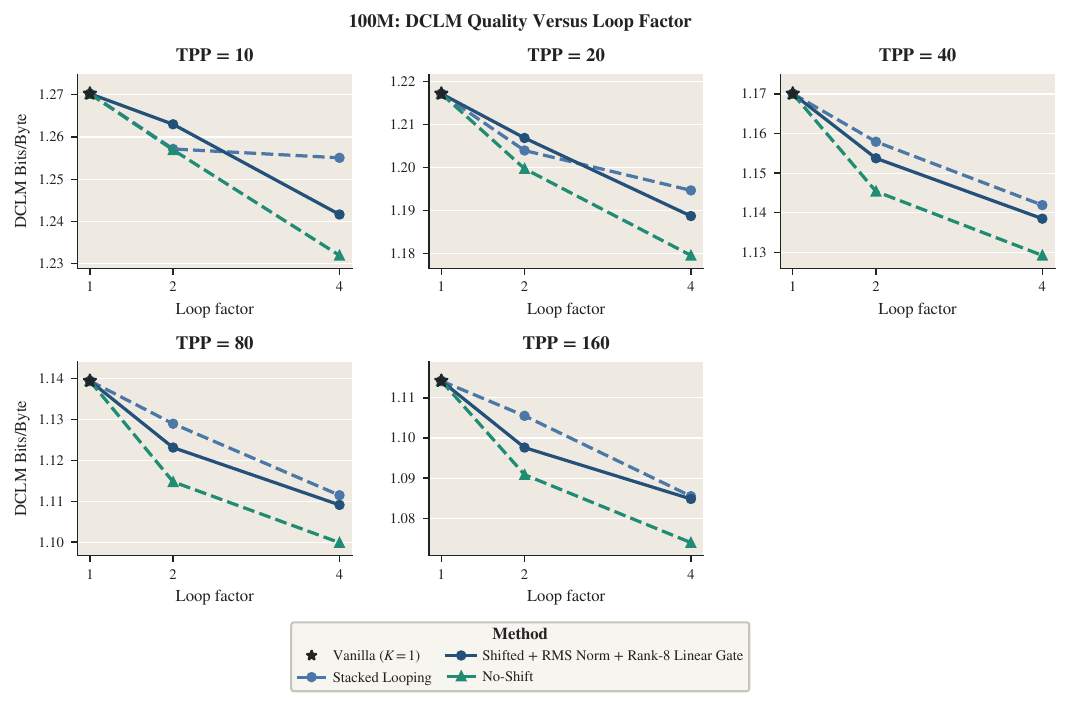
  }\par
  \vspace{0.5em}
  \includegraphics[width=0.88\textwidth]{
    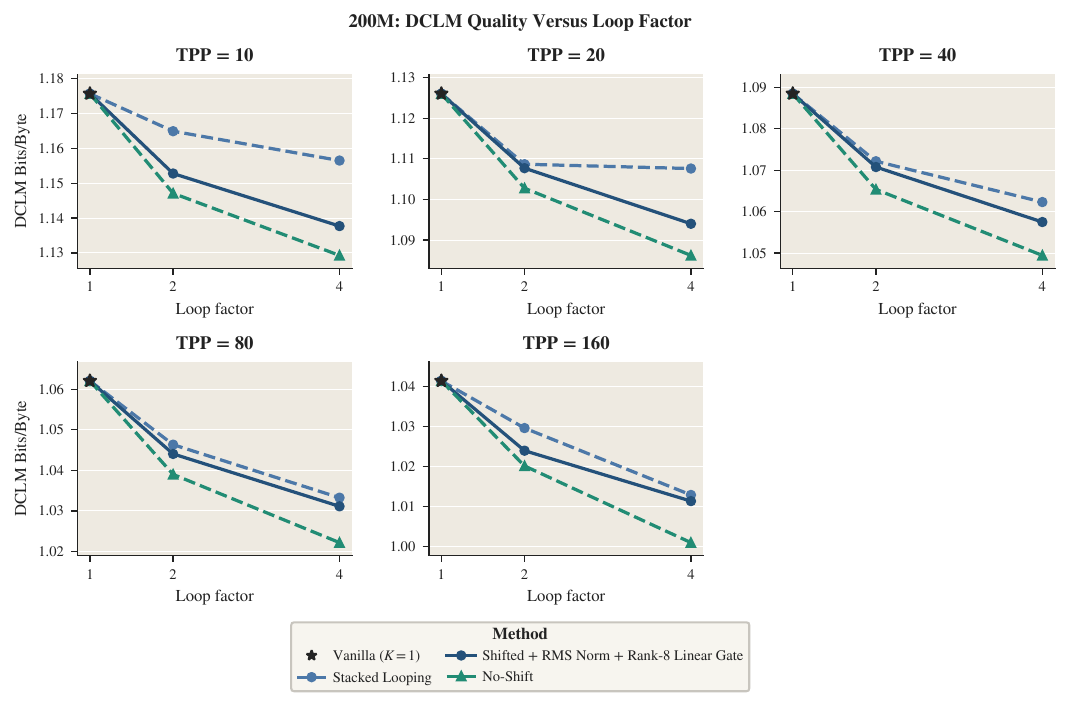
  }
  \caption{\textbf{DCLM quality across loop counts at 100M and 200M.}
  The top row shows 100M and the bottom row shows 200M. Each panel fixes TPP
  and varies $L$. The comparison includes Vanilla (No Looping), ordinary
  Full-Stack Looping, shifted PLT, and unshifted PLT where available. The
  relative ordering varies more at these smaller model sizes. Lower BPB is
  better.}
  \label{fig:app-plt-loops-small}
\end{figure*}

\begin{figure*}[ht]
  \centering
  \includegraphics[width=0.88\textwidth]{
    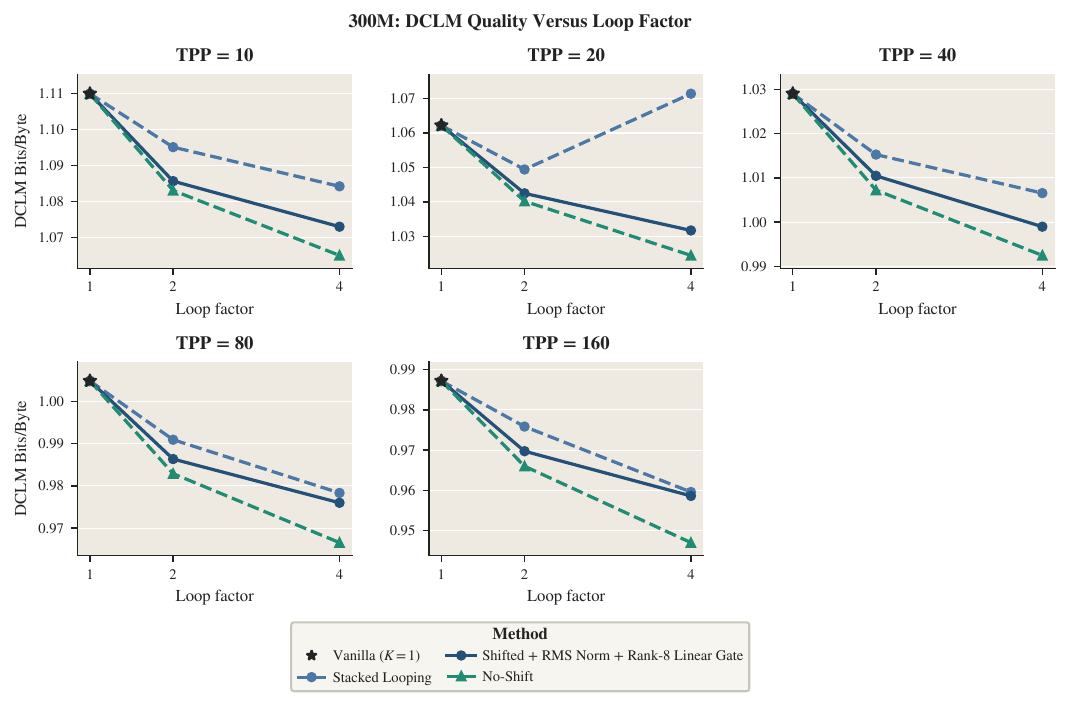
  }\par
  \vspace{0.5em}
  \includegraphics[width=0.88\textwidth]{
    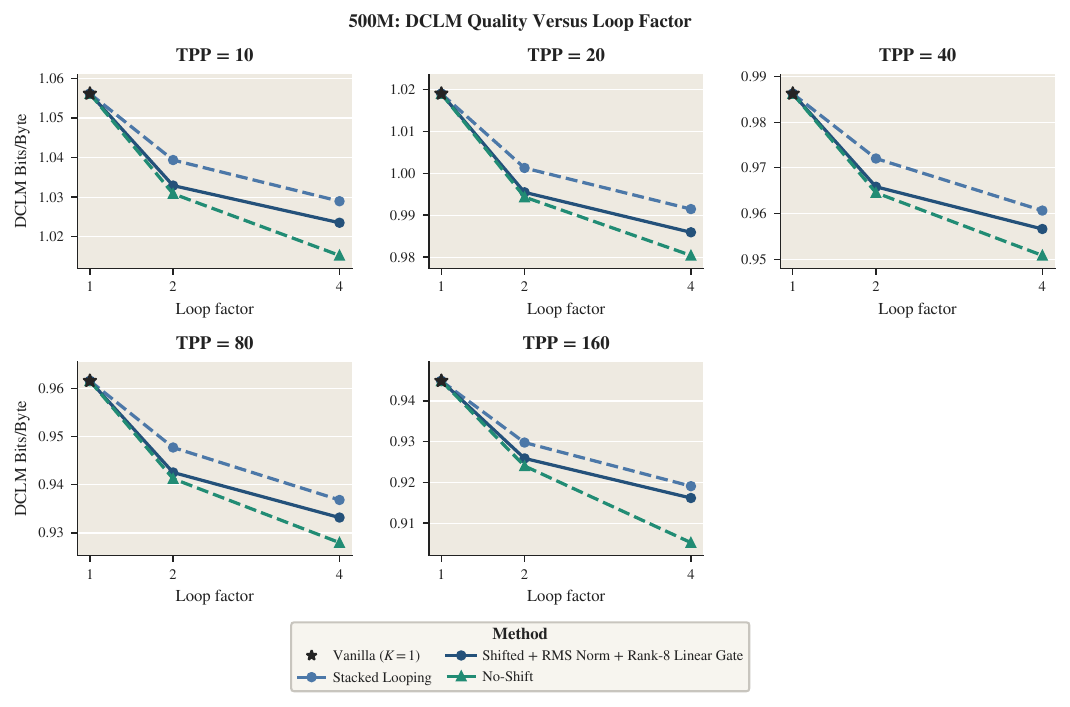
  }
  \caption{\textbf{DCLM quality across loop counts at 300M and 500M.}
  The top row shows 300M and the bottom row shows 500M. Each panel fixes TPP
  and varies $L$. Shifted PLT is more consistently competitive with ordinary
  Full-Stack Looping at these larger model sizes, while unshifted PLT provides
  the stronger sequential reference. Lower BPB is better.}
  \label{fig:app-plt-loops-large}
\end{figure*}

We next return to DCLM and examine how the two PLT variants change with model
size and loop count. Each panel in \autoref{fig:app-plt-loops-small} and
\autoref{fig:app-plt-loops-large} fixes the model size and TPP while varying
$L$. The figures compare Vanilla (No Looping), ordinary Full-Stack Looping,
shifted PLT, and unshifted PLT.

\paragraph{The effect of the shift depends on model size and loop count.}
Unshifted PLT achieves the lowest BPB throughout these comparisons. At the
smaller model sizes, a gap between shifted and unshifted PLT is already visible
at $L=2$, although both generally improve over ordinary Full-Stack Looping and
Vanilla (No Looping). As model size increases, shifted and unshifted PLT become
much closer at $L=2$. The same convergence is not apparent at $L=4$, where
shifted PLT continues to trail the unshifted version.

\paragraph{Why might the gap remain at larger loop counts?}
We hypothesize that this difference arises from what each additional loop is
asked to refine. Unshifted PLT passes the current token's representation from
one loop to the next, so every loop continues refining a representation of the
same token. Shifted PLT instead receives a representation produced for the
previous token. This representation summarizes the preceding context and can
be viewed as a prediction of the representation needed at the current
position. The current-token embedding anchors the computation to the token
that actually occurred, but this context-based prediction may still be a
noisier starting point than the current token's own previous-loop
representation. Larger models appear better able to compensate for this at
$L=2$. At $L=4$, however, information is passed across tokens repeatedly, so
small errors may build up and preserve the gap from unshifted PLT.

\clearpage

\subsection{How Does PLT Change with Model Scale and Training Budget?}
\label{app:plt-tpp-sweeps}

\begin{figure*}[ht]
  \centering
  \includegraphics[width=0.88\textwidth]{
    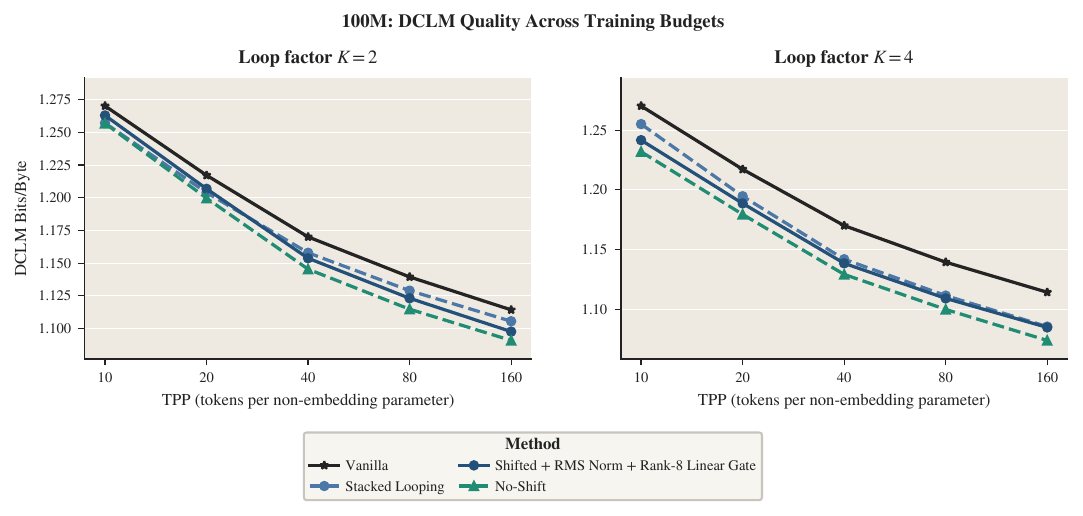
  }\par
  \vspace{0.5em}
  \includegraphics[width=0.88\textwidth]{
    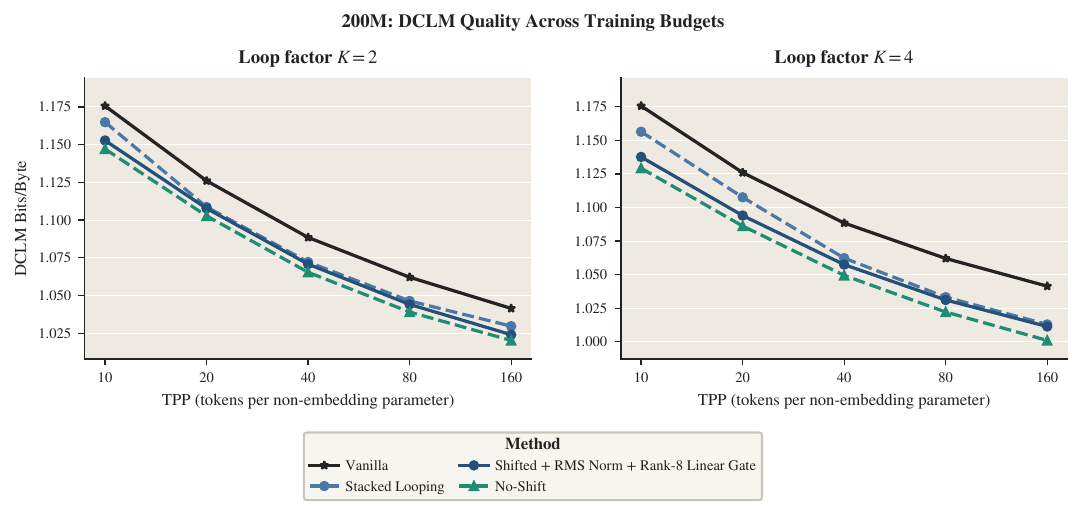
  }
  \caption{\textbf{DCLM quality across training budgets at 100M and 200M.}
  The top row shows 100M and the bottom row shows 200M. Each panel fixes the
  loop count and varies TPP from 10 to 160 for Full-Stack Looping, shifted
  PLT, and unshifted PLT. Unshifted PLT achieves the lowest BPB, while shifted
  PLT usually matches or improves upon Full-Stack Looping. Lower BPB is
  better.}
  \label{fig:app-plt-tpp-small}
\end{figure*}

\begin{figure*}[ht]
  \centering
  \includegraphics[width=0.88\textwidth]{
    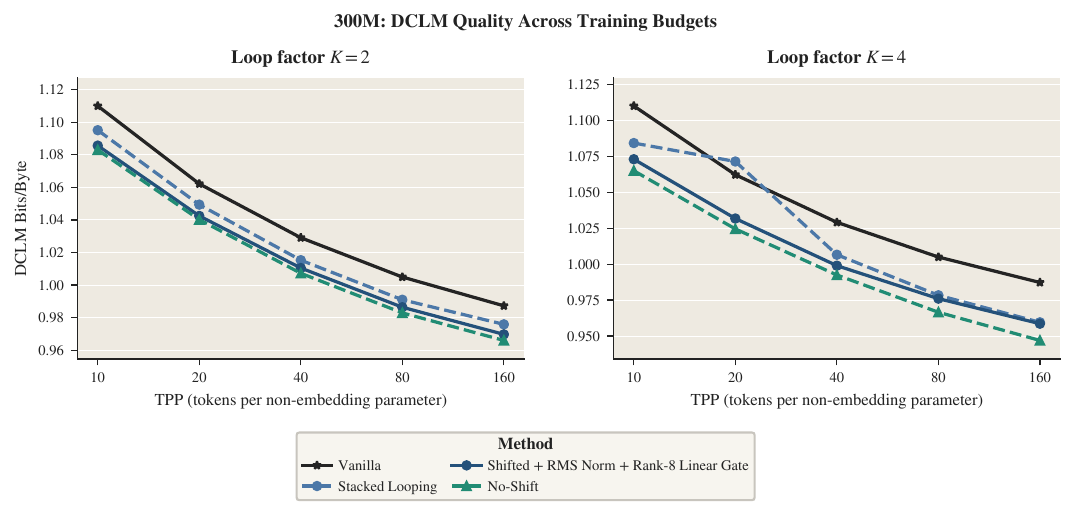
  }\par
  \vspace{0.5em}
  \includegraphics[width=0.88\textwidth]{
    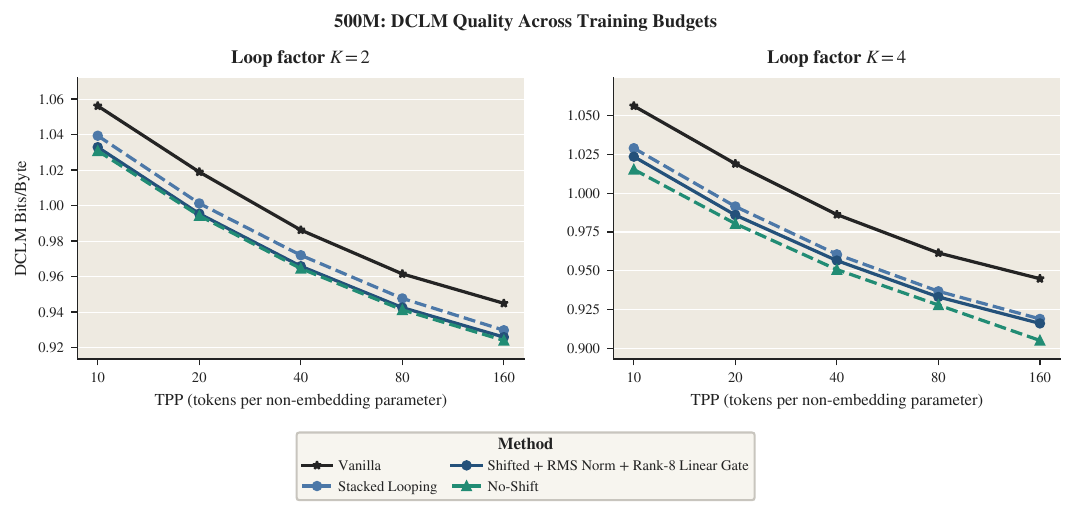
  }
  \caption{\textbf{DCLM quality across training budgets at 300M and 500M.}
  The top row shows 300M and the bottom row shows 500M. Each panel fixes the
  loop count and varies TPP from 10 to 160 for Full-Stack Looping, shifted
  PLT, and unshifted PLT. At $L=2$, shifted PLT closely tracks unshifted PLT
  across the evaluated TPP range; at $L=4$, a clearer gap remains. Shifted PLT
  usually matches or improves upon Full-Stack Looping while retaining its
  parallel decoding schedule. Lower BPB is better.}
  \label{fig:app-plt-tpp-large}
\end{figure*}

We next ask how these conclusions change with model scale and training budget.
To do so, we reorganize the same DCLM runs by TPP. Each panel in
\autoref{fig:app-plt-tpp-small} and \autoref{fig:app-plt-tpp-large} fixes the
loop count and varies TPP from 10 to 160.

The results show the same pattern as the loop-count view above. Unshifted PLT
achieves the lowest BPB throughout the evaluated settings. At $L=2$, the gap
between shifted and unshifted PLT narrows as model size increases; at the
larger scales, shifted PLT closely tracks unshifted PLT across the full TPP
range. At $L=4$, a clearer gap remains across training budgets. Increasing
model scale therefore reduces the cost of shifting the hidden representation
across tokens when only two loops are used, but does not fully remove it when
the shift is repeated over more loops.

Even with this remaining gap, shifted PLT usually matches or improves upon
ordinary Full-Stack Looping across TPP values. This is important because
Full-Stack Looping must run its loops sequentially during decoding, whereas
shifted PLT allows them to run concurrently. These results therefore show
that shifted PLT can retain the quality benefit of ordinary looping while
remaining more efficient during inference. They are also consistent with our
earlier finding that preserving the current-token embedding and separately
normalizing it and the shifted hidden representation are important for making
shifted PLT work.

\clearpage

\subsection{Stored Parameters and Training FLOPs}
\label{app:plt-compute}

The preceding comparisons fix model size and TPP, but adding loops still
increases training FLOPs because the Transformer must run once for every loop.
We therefore compare the models in two ways. First, we hold TPP fixed and ask
how quality changes with stored parameters and training FLOPs. We then allow
TPP to vary and compare the models at similar total training FLOPs.

\begin{figure*}[ht]
  \centering
  \includegraphics[width=0.88\textwidth]{
    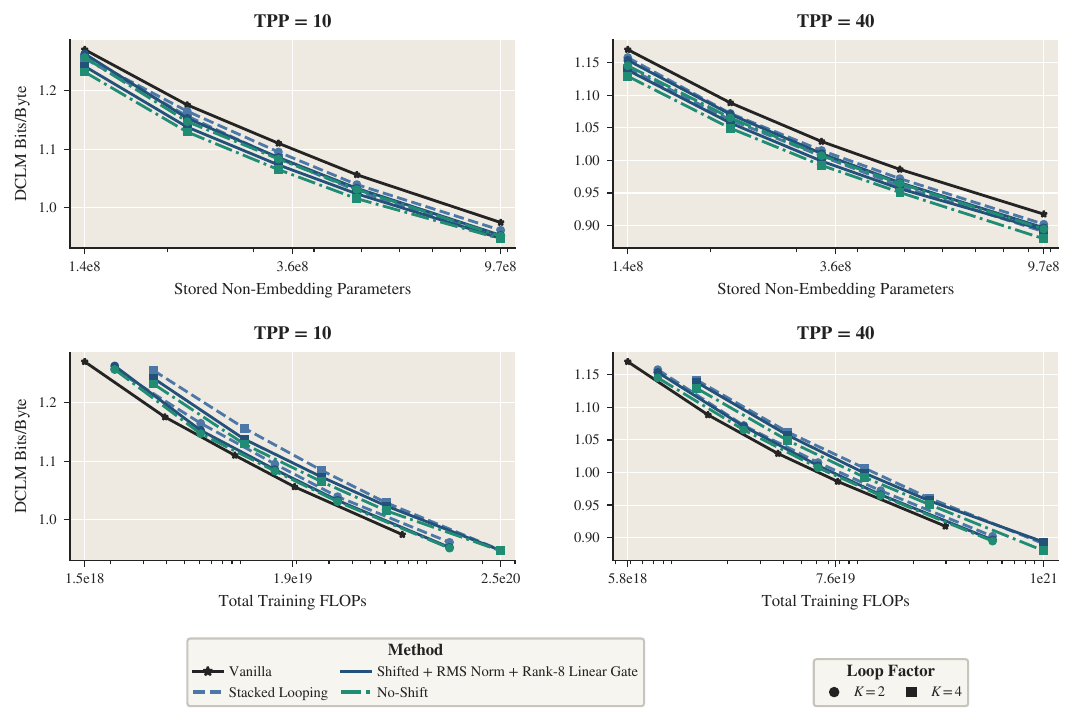
  }
  \caption{\textbf{Stored parameters and training FLOPs favor different loop
  counts.} DCLM BPB versus stored non-embedding parameters (top) and total
  training FLOPs (bottom). Columns fix TPP, and marker shape denotes loop
  count. Looping can improve quality with little additional parameter storage.
  When training FLOPs are the constraint and additional parameters can be
  stored, models made deeper with distinct Transformer layers generally give
  the stronger quality--compute trade-off. Lower BPB is better.}
  \label{fig:app-plt-params-flops}
\end{figure*}

\paragraph{At fixed TPP, parameters and training FLOPs favor different
choices.}
The parameter view in \autoref{fig:app-plt-params-flops} shows the main storage
benefit of looping. Increasing the number of loops can lower BPB while keeping
the number of stored parameters nearly unchanged because the same weights are
reused. If absolute quality is the main objective among the looped models,
  unshifted PLT with $L\in\{2,4\}$ is generally the strongest choice in the
evaluated range, although it must run its loops sequentially during decoding.
The training-FLOP view gives a different answer because every additional loop
requires another evaluation of the Transformer during training. If training
FLOPs are constrained, using fewer loops, typically $L=2$, therefore provides
a better return than $L=4$. When additional parameters can also be stored,
increasing physical depth by adding Transformer layers with their own weights
generally gives the strongest quality for the training FLOPs used. This
  reproduces the trade-off observed for ordinary looping in
  \autoref{sec:which-loops-work}: the best choice depends on whether the primary
constraint is stored parameters, training FLOPs, or inference efficiency.

\begin{figure*}[ht]
  \centering
  \includegraphics[width=0.94\textwidth]{
    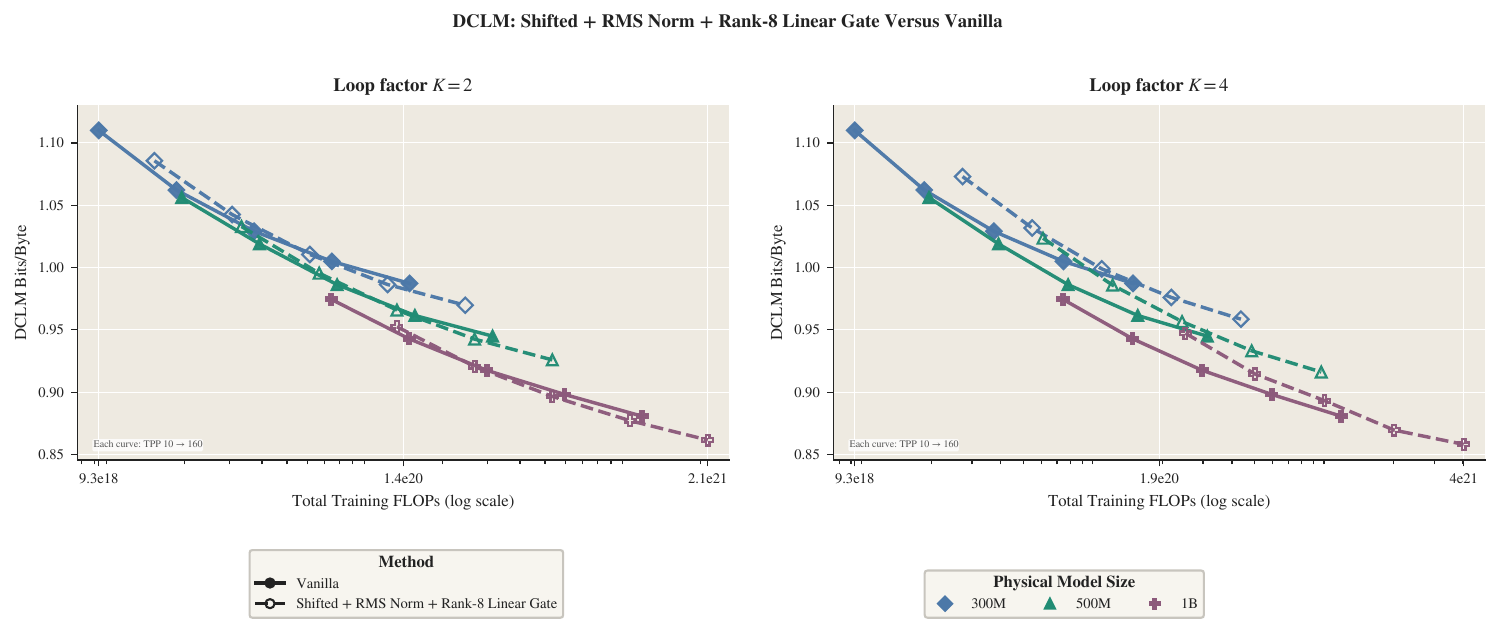
  }
  \caption{\textbf{Measured quality versus training FLOPs for shifted PLT at
  larger model sizes.} DCLM BPB for Vanilla (No Looping) and shifted PLT at
  300M, 500M, and 1B. Panels fix $L=2$ or $L=4$, and points along each curve
  increase from TPP 10 to 160. With $L=2$, the curves cross at some measured
  operating points: shifted PLT trained on fewer tokens can outperform
  Vanilla trained on more tokens. With $L=4$, the additional loop computation
  generally gives a weaker quality--compute trade-off. We do not extrapolate
  beyond the measured points. Lower BPB is better.}
  \label{fig:app-plt-compute-shifted}
\end{figure*}

\begin{figure*}[ht]
  \centering
  \includegraphics[width=0.90\textwidth]{
    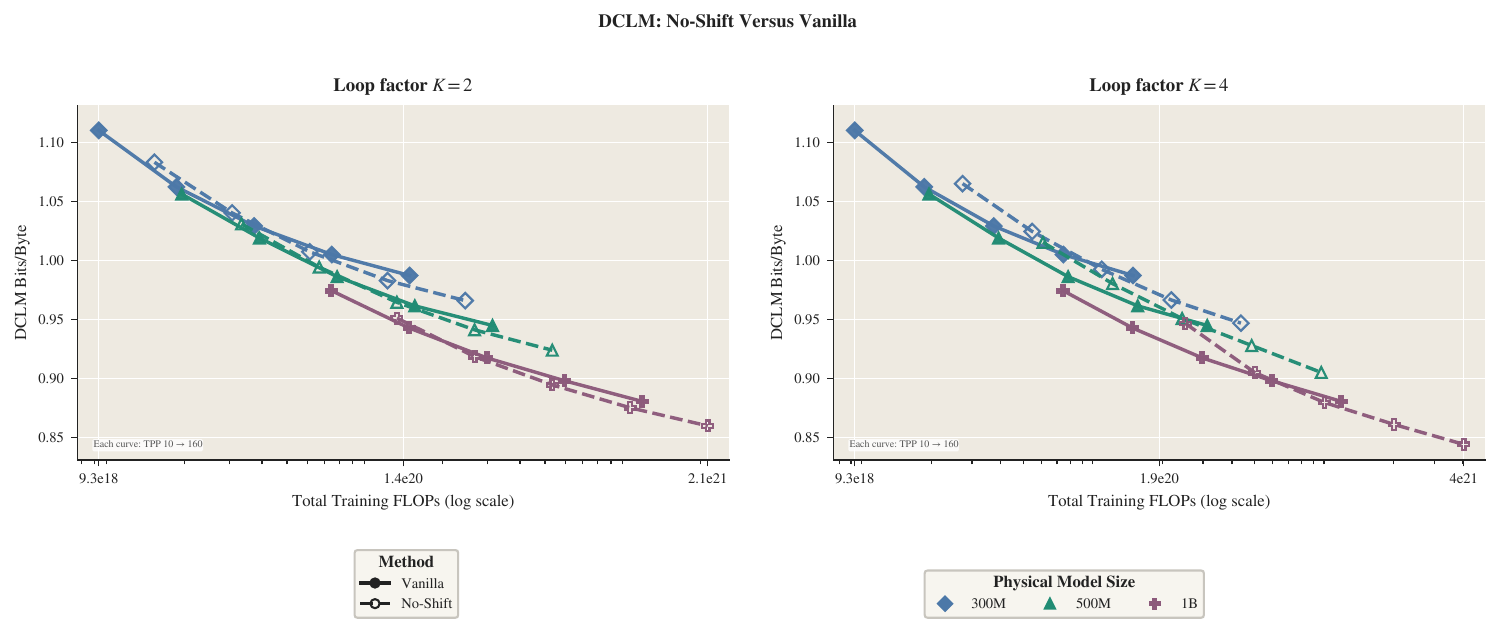
  }
  \caption{\textbf{Unshifted PLT provides a stronger quality reference but not
  parallel decoding.} DCLM BPB versus total training FLOPs for Vanilla (No
  Looping) and unshifted PLT at 300M, 500M, and 1B. Unshifted PLT passes each
  token's representation directly from one loop to the next, so its loops
  remain sequential during decoding. Lower BPB is better.}
  \label{fig:app-plt-compute-noshift}
\end{figure*}

\paragraph{Across TPP, two-loop PLT comes close to the measured compute
frontier.}
Allowing TPP to vary changes the comparison because each model size is now
evaluated under several allocations of training compute.
\autoref{fig:app-plt-compute-shifted} traces Vanilla (No Looping) and shifted
PLT from TPP 10 to 160. With $L=2$, shifted PLT appears to approach the best
quality--compute trade-off among the measured models as model size increases.
The unshifted version comes closer still
(\autoref{fig:app-plt-compute-noshift}), although it does not establish a new
quality--compute frontier and does not provide parallel decoding.

The curves also cross at some operating points. Within the same model scale, a
two-loop model trained at a lower TPP, and therefore on fewer tokens, can
outperform a Vanilla model trained at a higher TPP while using comparable
total training FLOPs. We do not interpret this as evidence that PLT is
uniformly more data-efficient. Instead, it shows that, for some model scales
and compute budgets, allocating part of the training compute to a second loop
can be more effective than allocating all of it to additional training tokens.
The same behavior appears in the headline comparison in
\autoref{fig:headline-looping-efficiency}. With $L=4$, the additional loop
computation generally moves the model farther from the best measured
quality--compute trade-off, suggesting that this benefit is specific to
relatively shallow looping over the range studied.

\autoref{fig:app-plt-compute-noshift} gives the corresponding comparison for
unshifted PLT. It achieves stronger quality than shifted PLT, but its loops
must run one after another during decoding. It therefore provides a useful
quality reference, but not PLT's inference-efficiency benefit.

\subsubsection{Training-FLOP Results for Each Model Size}

\begin{figure*}[ht]
  \centering
  \includegraphics[width=0.49\textwidth]{
    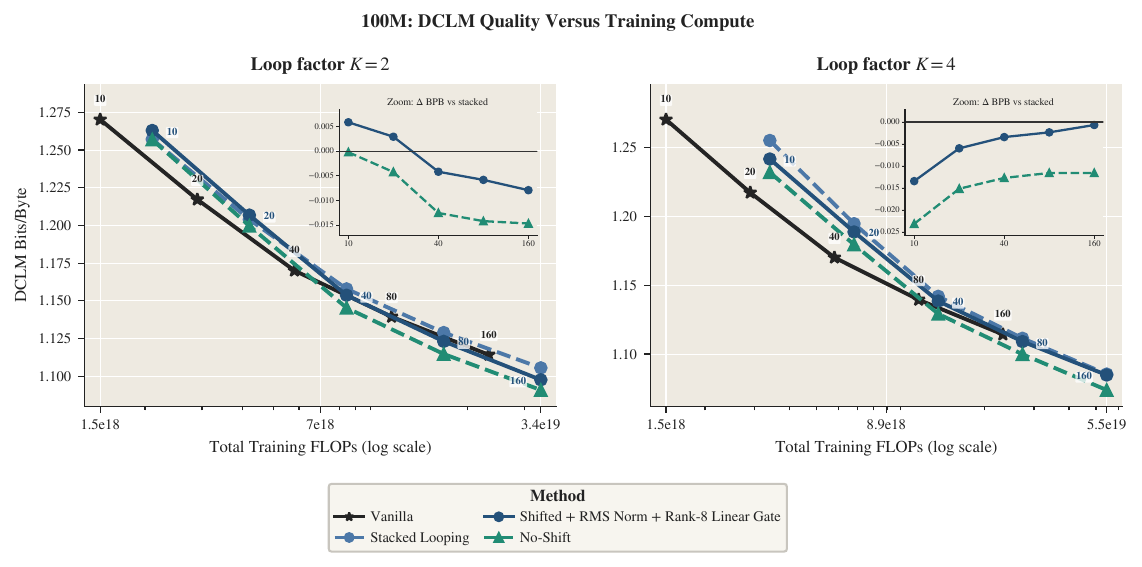
  }\hfill
  \includegraphics[width=0.49\textwidth]{
    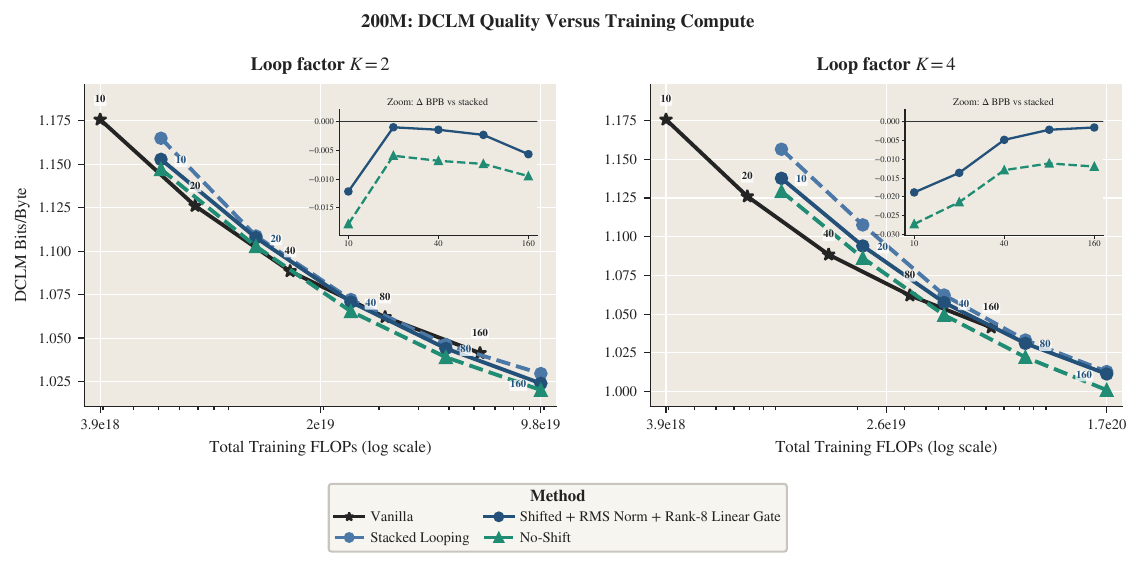
  }
  \caption{\textbf{DCLM quality versus total training FLOPs at 100M and
  200M.} The left plot shows 100M and the right plot shows 200M. Insets report
  BPB differences from the corresponding Full-Stack model. Lower BPB and
  negative inset values are better.}
  \label{fig:app-plt-flops-small}
\end{figure*}

\begin{figure*}[ht]
  \centering
  \includegraphics[width=0.49\textwidth]{
    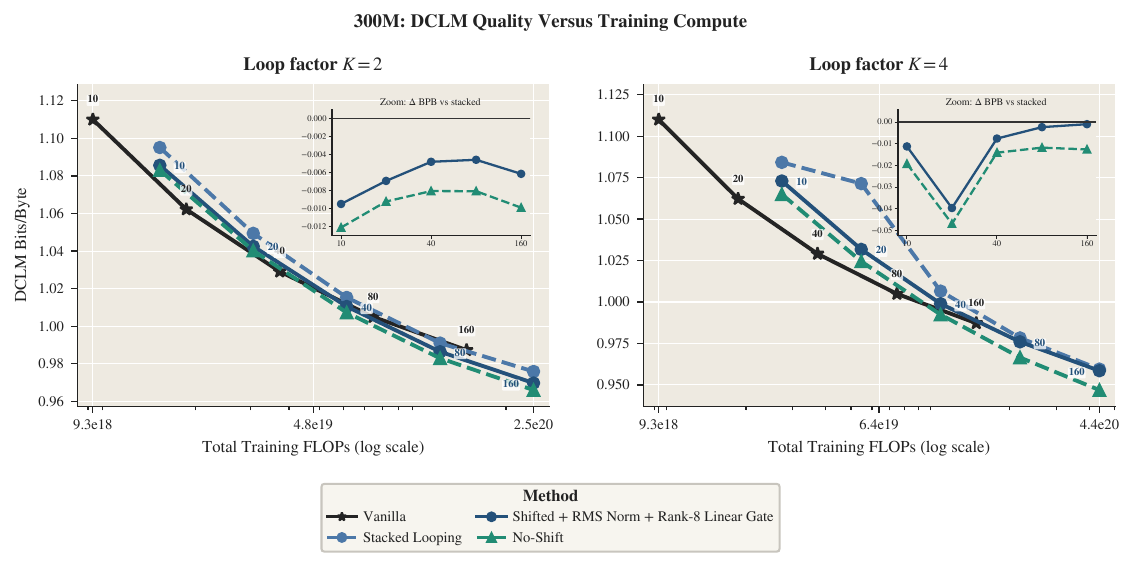
  }\hfill
  \includegraphics[width=0.49\textwidth]{
    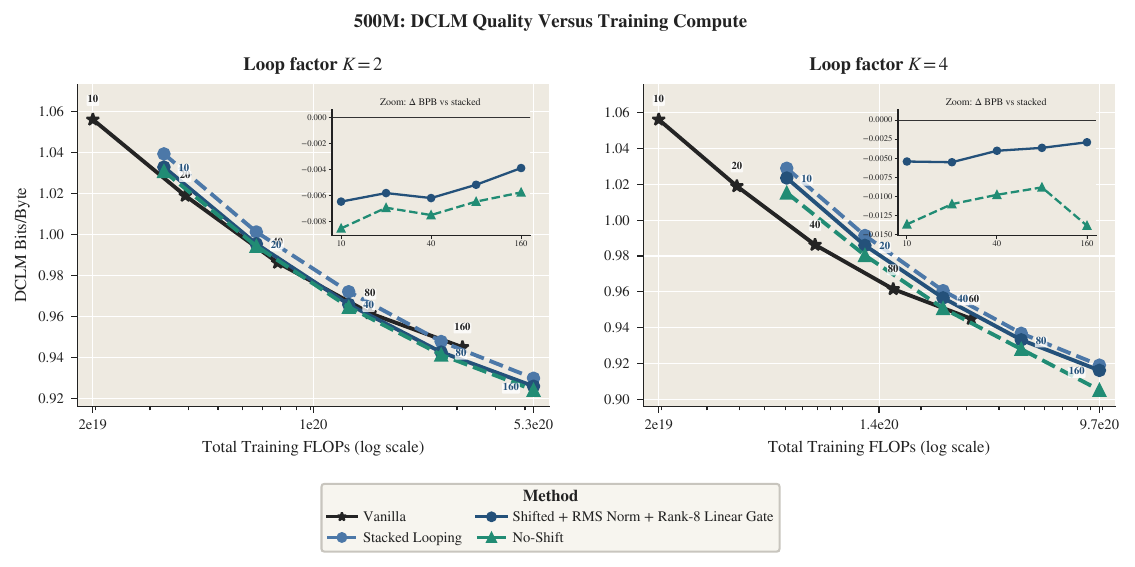
  }
  \caption{\textbf{DCLM quality versus total training FLOPs at 300M and
  500M.} The left plot shows 300M and the right plot shows 500M. Insets report
  BPB differences from the corresponding Full-Stack model. Lower BPB and
  negative inset values are better.}
  \label{fig:app-plt-flops-large}
\end{figure*}

\begin{figure*}[ht]
  \centering
  \includegraphics[width=0.6\textwidth]{
    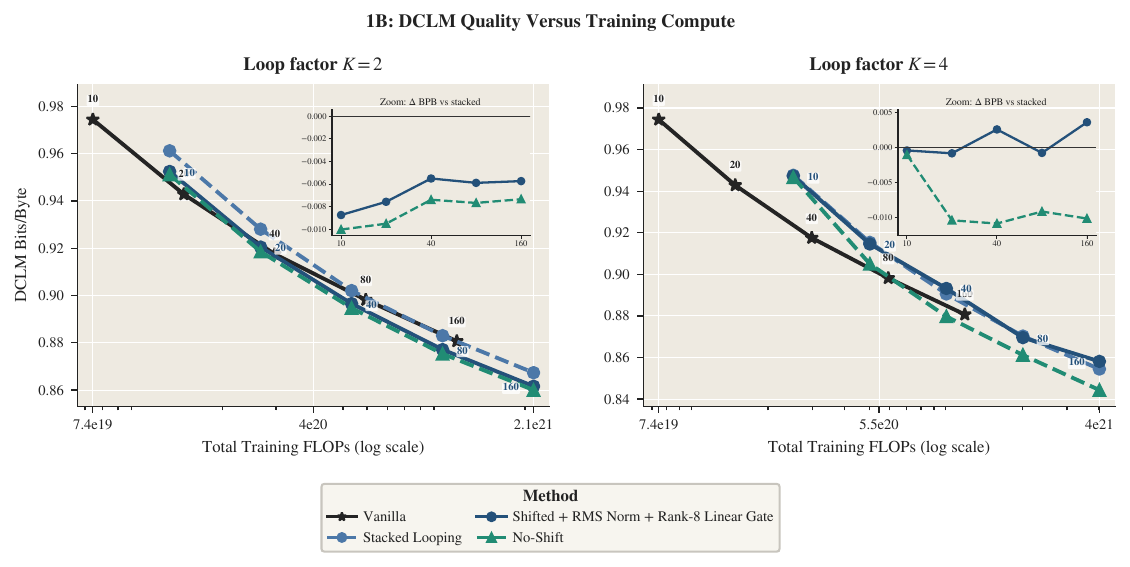
  }
  \caption{\textbf{DCLM quality versus total training FLOPs at 1B.} The
  $L=2$ and $L=4$ panels show all five measured training budgets. Insets
  report BPB differences from the corresponding Full-Stack model. Lower BPB
  and negative inset values are better.}
  \label{fig:app-plt-flops-1b}
\end{figure*}

For completeness, \autoref{fig:app-plt-flops-small} and
\autoref{fig:app-plt-flops-large} report the training-FLOP comparisons
separately for the 100M--500M models, while
\autoref{fig:app-plt-flops-1b} reports the corresponding 1B results. Each
figure includes every measured TPP value, and the insets report the BPB
difference from the corresponding Full-Stack model. These per-model results support the aggregate trends above. Unshifted PLT
generally achieves the best quality, while shifted PLT with $L=2$ becomes more
competitive as model size increases and often matches or improves upon
Full-Stack Looping. Using $L=4$ can lower BPB further, but generally provides a
weaker return per training FLOP.

\begin{takeawaybox}
For practitioners, the choice depends on the primary constraint. If the goal
is the strongest quality and sequential decoding is acceptable, unshifted PLT
with $L\in\{2,4\}$ is generally the strongest looped design. If decoding
efficiency matters, shifted PLT with separate input normalization and $L=2$ is
a strong default: it allows the loops to run concurrently, usually matches or
improves upon Full-Stack Looping, and approaches unshifted PLT as model size
increases. At some measured operating points, it also outperforms Vanilla
models trained on more tokens at comparable training FLOPs, showing that
shallow looping can improve token efficiency in some regimes. Increasing to
$L=4$ can further lower BPB when stored parameters are limited, but generally
provides a weaker return per training FLOP. If training FLOPs are the main
constraint and additional parameters can be stored, adding distinct
Transformer layers is typically preferable to increasing the number of loops.
\end{takeawaybox}

\clearpage

\clearpage
\section{Parallel Looping with Loop-Specific Specialization}
\label{app:plt-lora}

The preceding sections study loop-specific specialization and parallel
looping separately. Here, we provide a short supporting experiment asking
whether the two can be used together. We focus on \emph{Loop Matrix} because
it was generally the strongest of the specialization methods compared in
\autoref{sec:loop-lora}. We use the same method defined in
\autoref{def:loop-lora-mixing}, without introducing a new specialization
design. The figures label the same method as Matrix LoRA; throughout the text,
we use Loop Matrix. Its loop-specific
corrections are added to the attention and MLP projections and trained jointly
with the rest of the model. We compare three forms of looping. \emph{Full-Stack Looping} is ordinary looping, in
which each token passes through the complete Transformer repeatedly.
\emph{Shifted PLT} passes information across adjacent token positions and
allows the loops to run concurrently during autoregressive decoding.
\emph{Unshifted PLT} passes the current token's representation directly from
one loop to the next. It provides a useful quality reference, but its loops
remain sequential during decoding. We first examine DCLM in detail and then
check the same pattern across DCLM and five additional language-modeling
corpora.

\subsection{DCLM Results}
\label{app:plt-lora-dclm}

\begin{figure*}[ht]
  \centering
  \includegraphics[width=0.98\textwidth]{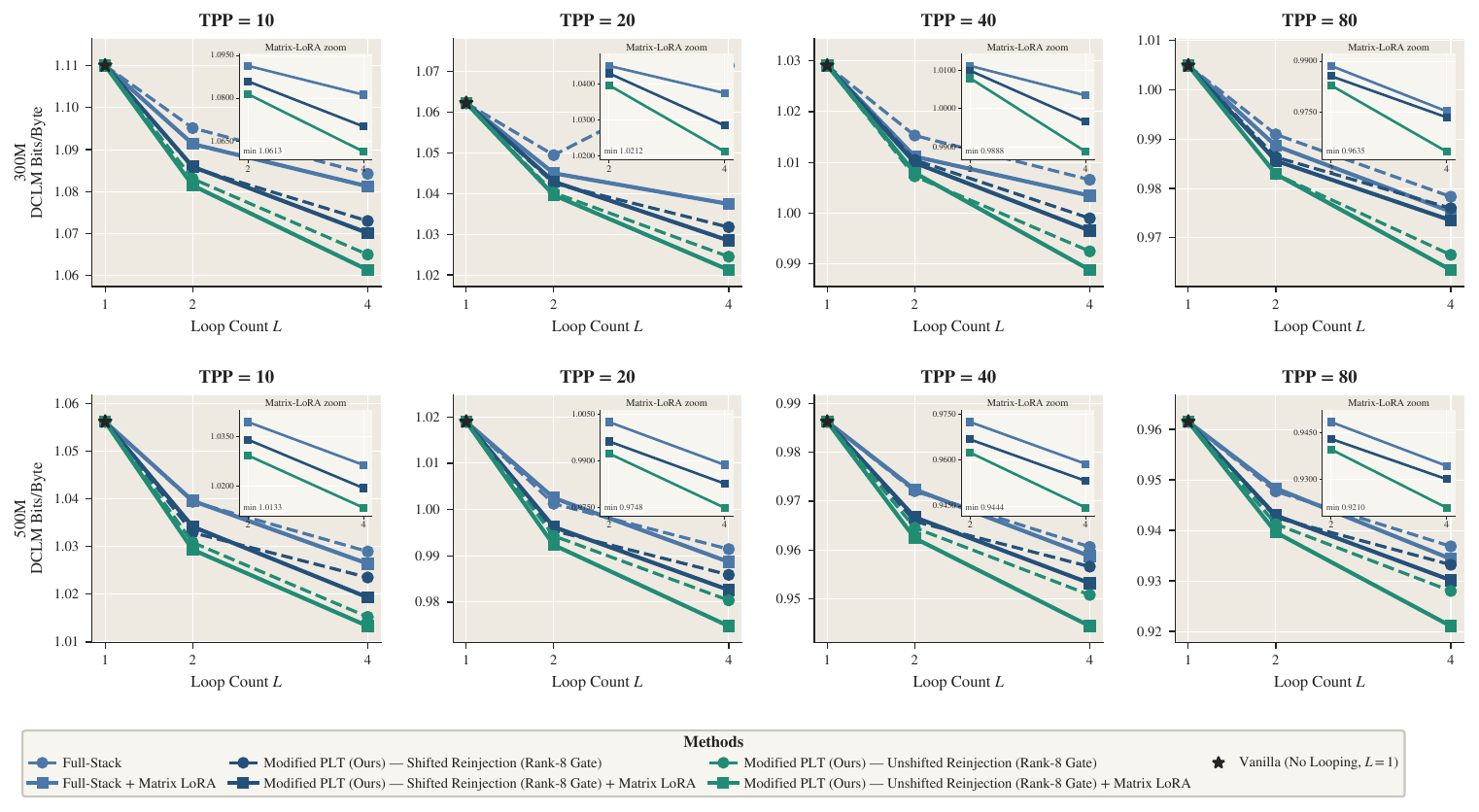}
  \caption{\textbf{Loop-specific specialization preserves the ordering among
  ordinary and parallel looping designs.} DCLM BPB for 300M and 500M models.
  The figure compares ordinary Full-Stack Looping, shifted PLT, and unshifted
  PLT with shared weights and with Loop Matrix. Unshifted PLT gives the
  strongest quality but remains sequential during decoding. Shifted PLT
  generally improves over Full-Stack Looping while retaining parallel
  decoding. Lower BPB is better.}
  \label{fig:app-plt-lora-shifted-unshifted}
\end{figure*}

\autoref{fig:app-plt-lora-shifted-unshifted} compares 300M and 500M models at
TPP 10--80 and $L\in\{2,4\}$. The shared-weight results are included for
context, while the Loop Matrix results show what happens when the loops are
allowed to specialize. The ordering is consistent with the preceding PLT
experiments. Unshifted PLT achieves the lowest BPB, but remains sequential
during decoding. Shifted PLT is generally stronger than ordinary Full-Stack
looping while retaining PLT's parallel decoding schedule. Adding Loop Matrix
does not change this overall ordering.

The effect of specialization also follows the pattern observed in
\autoref{sec:loop-lora}. Its benefit is small or mixed at $L=2$ and becomes
clearer at $L=4$, when the same Transformer weights are reused more times.
Thus, loop-specific specialization remains useful in PLT, but the central
comparison here is architectural: shifted PLT retains its advantage over
ordinary Full-Stack Looping after the same specialization is applied to both.

\subsection{Results Across Six Language-Modeling Corpora}
\label{app:plt-lora-corpora}

\begin{figure*}[ht]
  \centering
  \includegraphics[width=0.92\textwidth]{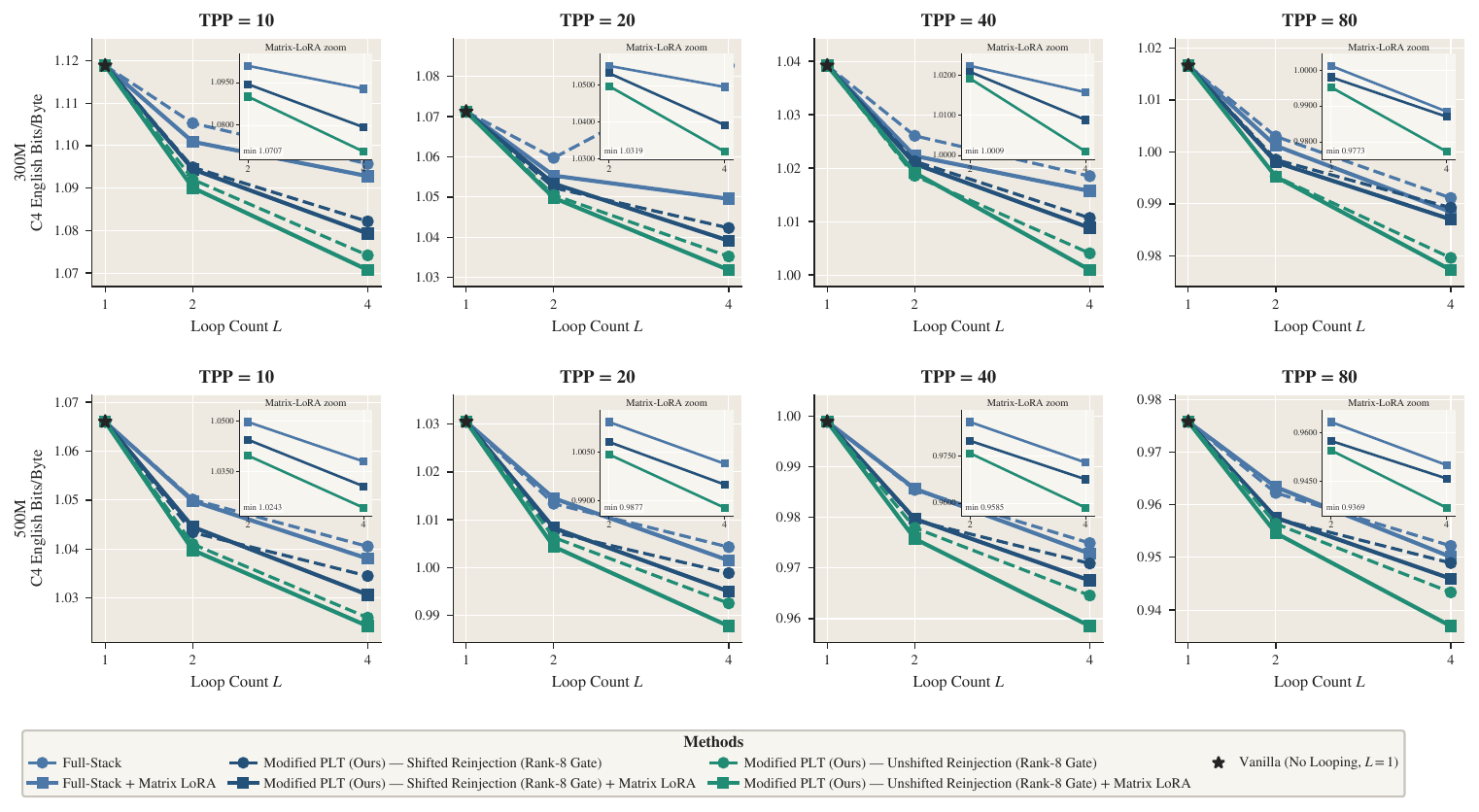}
  \caption{\textbf{Parallel looping with Loop Matrix on C4 English.}
  Absolute BPB for ordinary Full-Stack Looping, shifted PLT, and unshifted
  PLT with shared weights and with Loop Matrix. The comparison uses 300M and
  500M models, TPP 10--80, and $L\in\{2,4\}$. The ordering broadly matches
  DCLM: unshifted PLT is strongest, while shifted PLT generally improves over
  Full-Stack Looping and retains parallel decoding. Lower BPB is better.}
  \label{fig:app-plt-lora-c4}
\end{figure*}

\begin{figure*}[ht]
  \centering
  \includegraphics[width=0.92\textwidth]{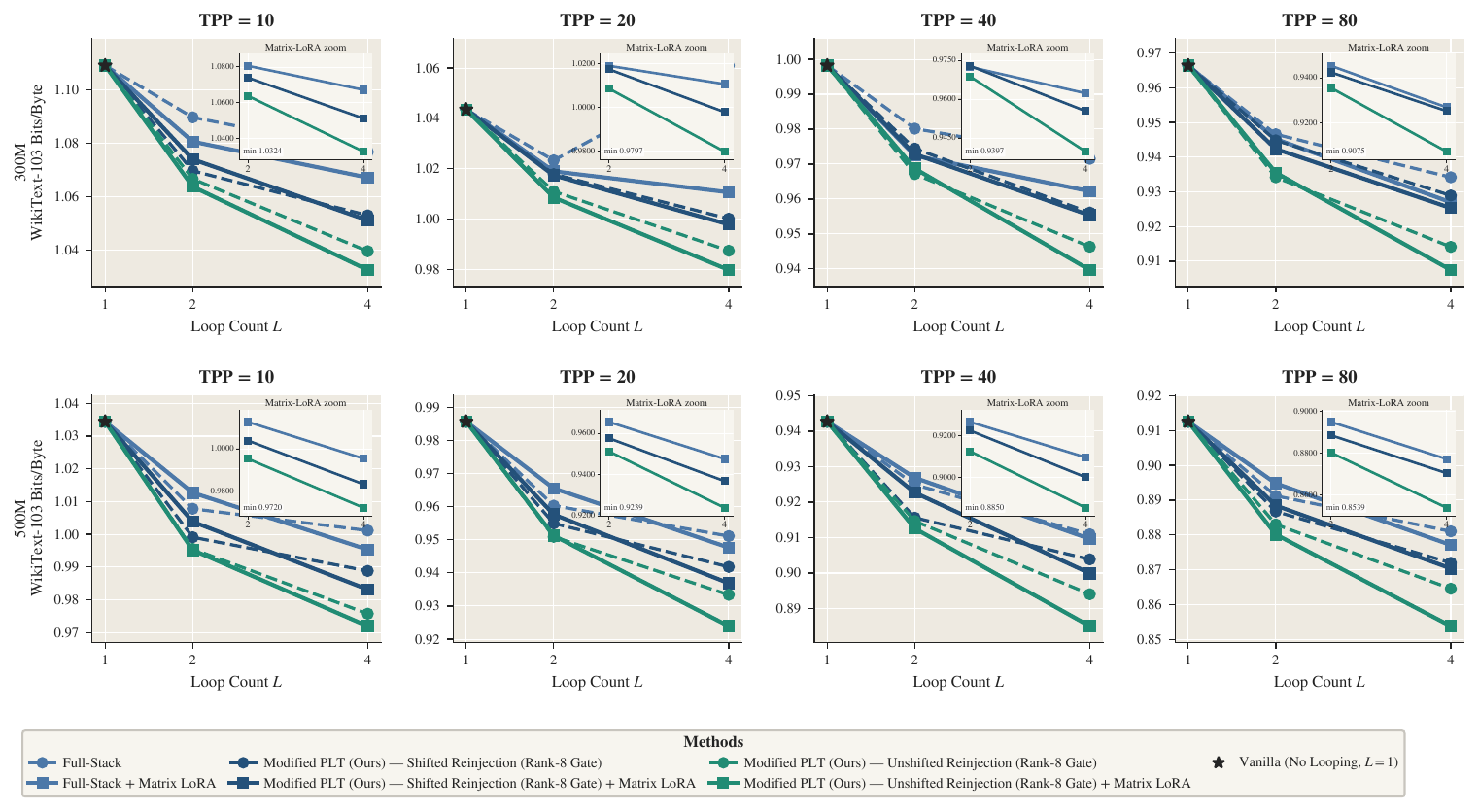}\par
  \vspace{0.5em}
  \includegraphics[width=0.92\textwidth]{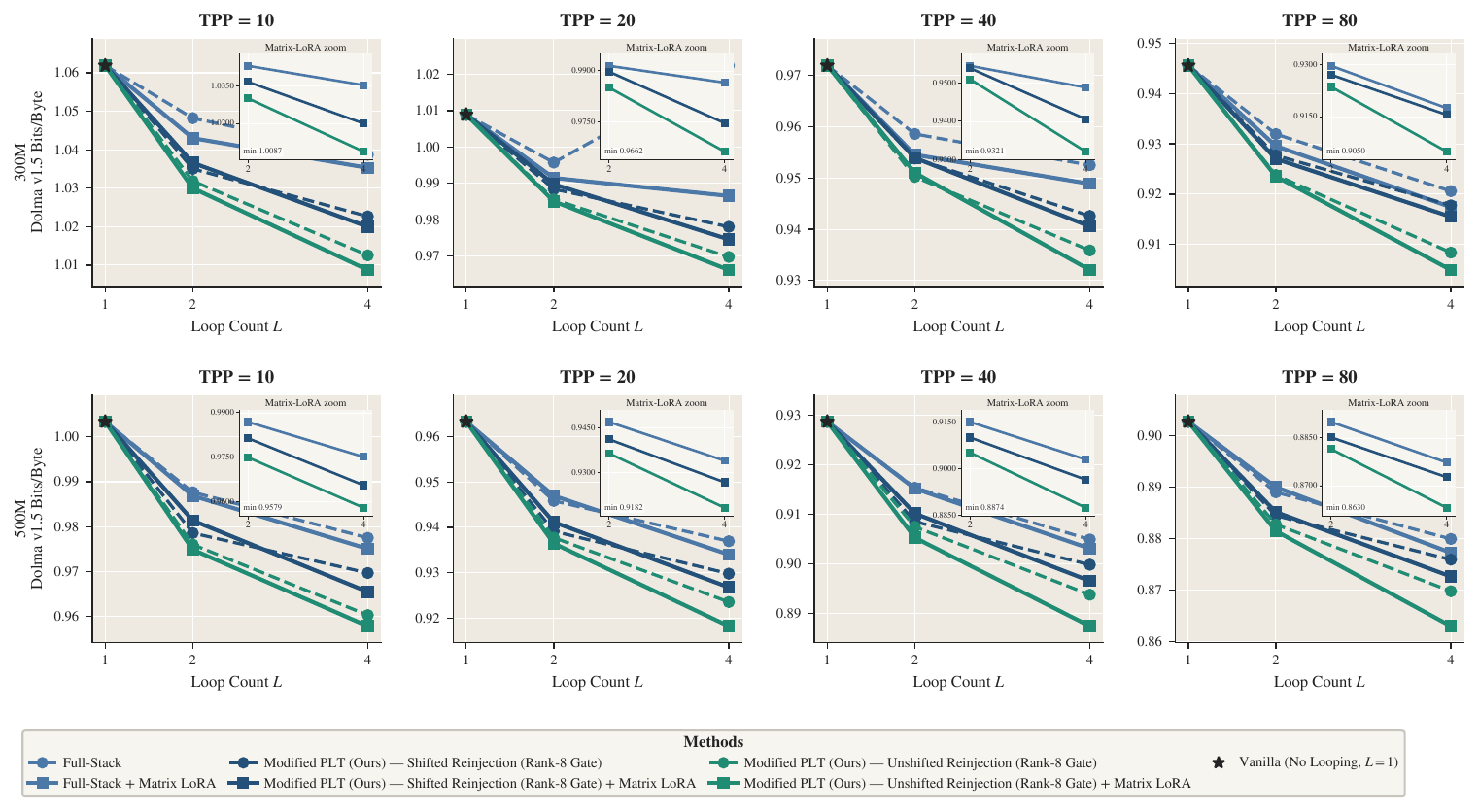}
  \caption{\textbf{Parallel looping with Loop Matrix on WikiText-103 and
  Dolma v1.5.} The top figure shows WikiText-103 and the bottom figure shows
  Dolma v1.5. Both compare ordinary Full-Stack Looping, shifted PLT, and
  unshifted PLT with shared weights and with Loop Matrix. Specialization is
  generally more useful at $L=4$ than at $L=2$. Lower BPB is better.}
  \label{fig:app-plt-lora-wiki-dolma}
\end{figure*}

\begin{figure*}[ht]
  \centering
  \includegraphics[width=0.92\textwidth]{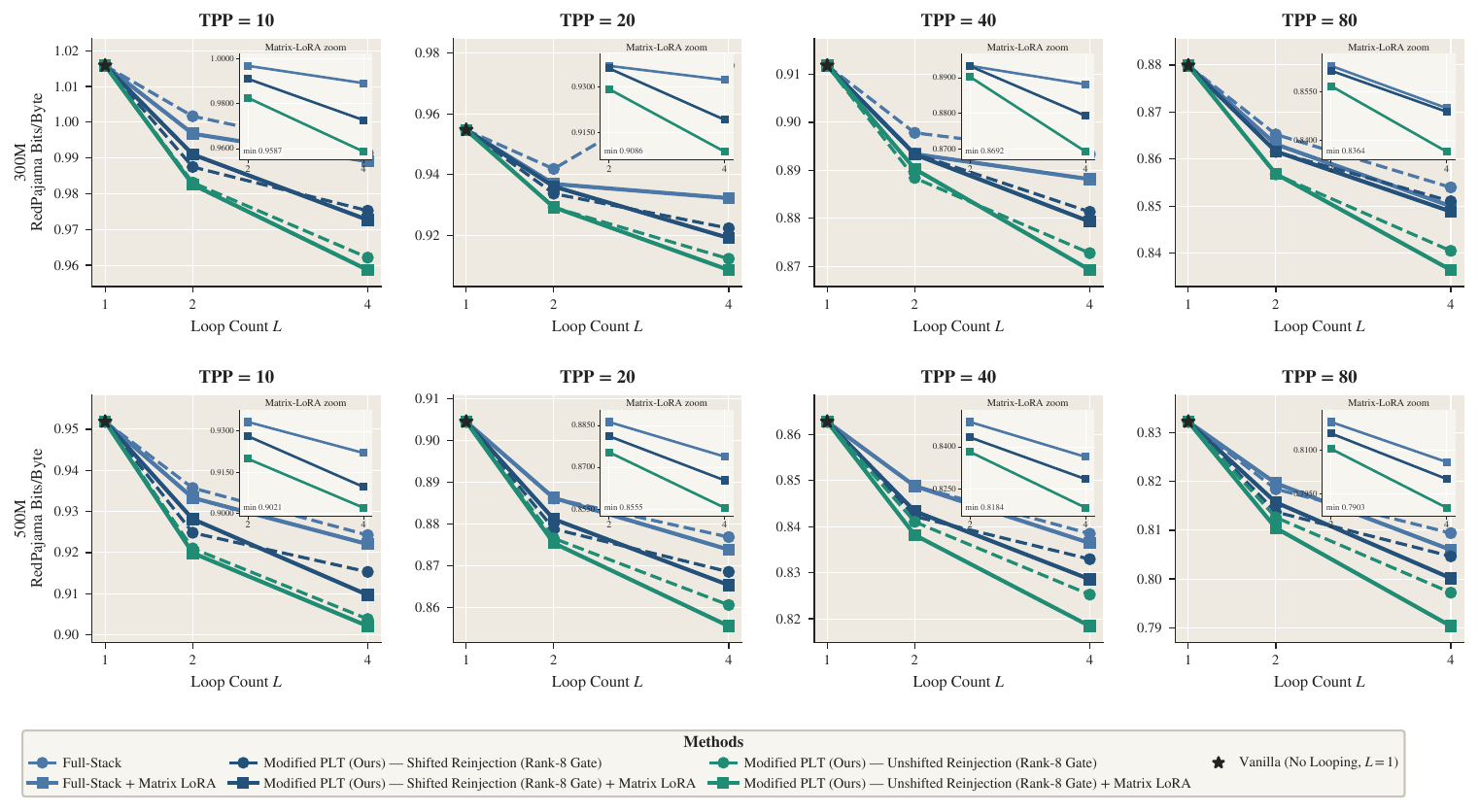}\par
  \vspace{0.5em}
  \includegraphics[width=0.92\textwidth]{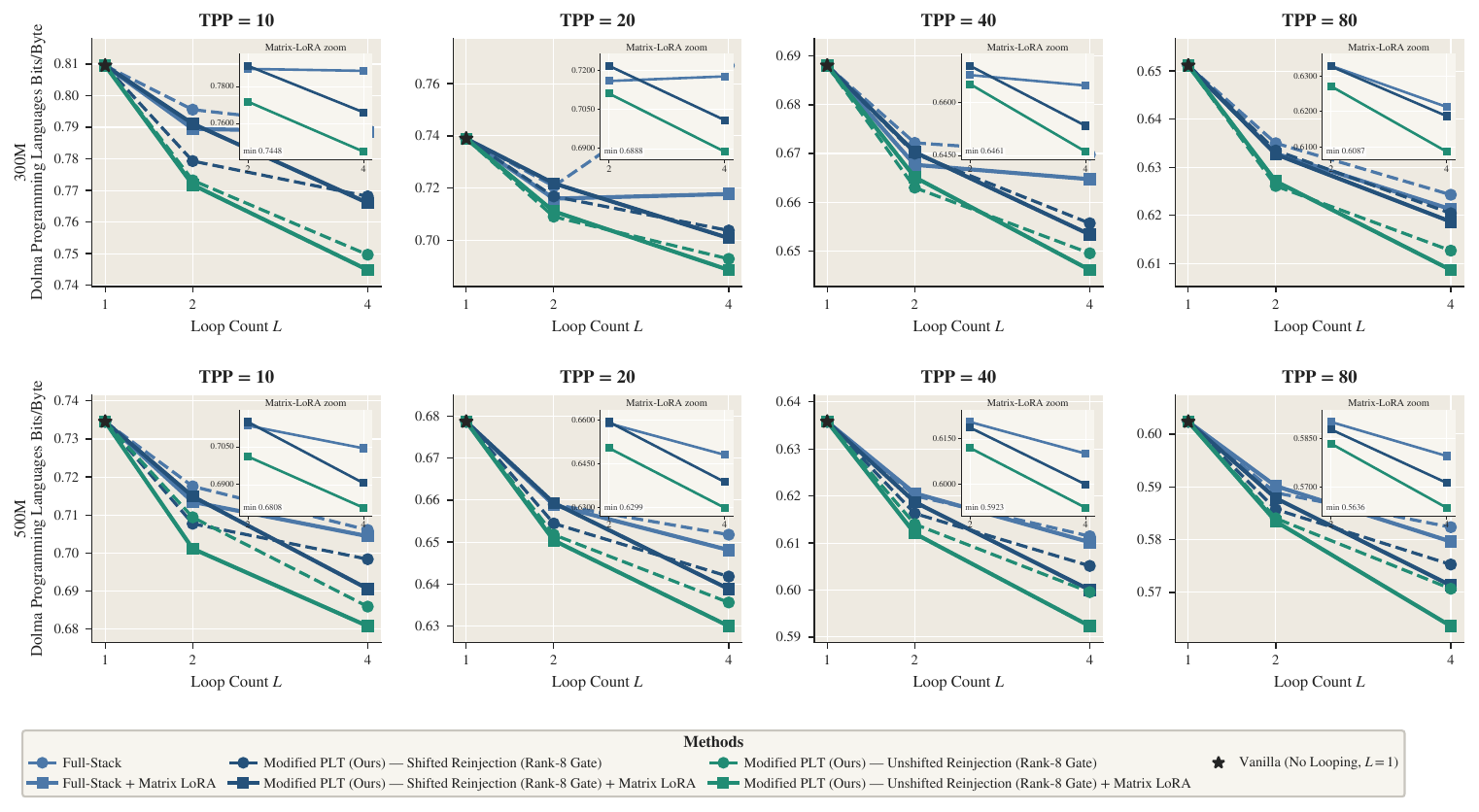}
  \caption{\textbf{Parallel looping with Loop Matrix on RedPajama and Dolma
  programming languages.} The top figure shows RedPajama and the bottom
  figure shows the code-heavy Dolma programming-languages split. The benefit
  of specialization is generally clearer at $L=4$; at $L=2$, the differences
  are smaller and less uniform, particularly on the programming-languages
  split. Lower BPB is better.}
  \label{fig:app-plt-lora-redpajama-code}
\end{figure*}

We next test whether the DCLM result in
\autoref{fig:app-plt-lora-shifted-unshifted} is specific to one evaluation
corpus. We report C4 English in \autoref{fig:app-plt-lora-c4}, WikiText-103
and Dolma v1.5 in \autoref{fig:app-plt-lora-wiki-dolma}, and RedPajama and
Dolma programming languages in \autoref{fig:app-plt-lora-redpajama-code}.
Each figure uses the same 300M and 500M models, TPP 10--80, and
$L\in\{2,4\}$ comparison as the DCLM figure. The figures show ordinary
Full-Stack Looping, shifted PLT, and unshifted PLT, both with shared weights
and with Loop Matrix. The same broad pattern appears across the six corpora. Unshifted PLT is
usually strongest but remains sequential during decoding. Shifted PLT
generally improves over ordinary Full-Stack Looping while retaining parallel
decoding. Loop Matrix also generally improves quality, with the clearest and
most consistent gains at $L=4$. At $L=2$, its effect is smaller and can be
mixed, particularly on the code-heavy Dolma programming-languages split
(\autoref{fig:app-plt-lora-redpajama-code}). Taken together,
\autoref{fig:app-plt-lora-c4}, \autoref{fig:app-plt-lora-wiki-dolma}, and
\autoref{fig:app-plt-lora-redpajama-code} show that the DCLM conclusion
transfers across evaluation corpora: specialization generally helps when the
weights are reused more deeply, and it does not remove the advantage of
shifted PLT over ordinary Full-Stack Looping.

Overall, the cross-corpus results show that loop-specific specialization and
parallel looping remain complementary, particularly when the same weights are
reused across more loops.

\begin{takeawaybox}
Across the six language-modeling corpora, Loop Matrix generally improves
quality, with clearer and more consistent gains at $L=4$ than at $L=2$.
Unshifted PLT usually gives the strongest quality but remains sequential during
decoding. When parallel decoding is required, shifted PLT with Loop Matrix is
the more useful choice: it generally outperforms ordinary Full-Stack Looping
while preserving PLT's parallel decoding schedule.
\end{takeawaybox}

\section{Limitations}
  \label{app:limitations}

  Our experiments establish consistent looping trends across dense decoder-only
  transformers trained from scratch at scales up to 1B parameters. We do not
  evaluate larger models, mixture-of-experts architectures, or post-trained
  models; future work can test how broadly these trends extend. We study a diverse set of looping strategies, low-rank specialization
  methods, training budgets, and evaluation datasets, but do not exhaust the
  space of possible designs. We also focus primarily on likelihood-based
  evaluation rather than free-form generation. Future work can broaden both the
  architectural and evaluation settings. Our normalized shifted PLT performs particularly well at \(L=2\) and remains
  competitive at \(L=4\). We do not establish whether this behavior persists at
  substantially larger loop counts and model scales. Closing the remaining gap
  to the unshifted control and developing a theoretical account of how
  normalization and current-token embedding reinjection affect stability are
  promising directions for future work.

\end{document}